\documentclass[a4paper,10pt]{article}
\usepackage[utf8]{inputenc}
\DeclareUnicodeCharacter{03C7}{$\chi$}
\usepackage[T1]{fontenc}

\usepackage{geometry}
\usepackage{url}
\usepackage{hyperref}
\usepackage{amsmath,mathtools}
\usepackage{libertine}
\usepackage[scale=.9]{sourcecodepro}
\usepackage{microtype}\DisableLigatures[f]{encoding=*,family=tt*}
\usepackage{fancyvrb}
\makeatletter\FV@AddToHook{\FV@ListParameterHook}{\topsep=0pt\partopsep=0pt}\makeatother
\usepackage{wrapfig}
\usepackage{relsize}
\usepackage[inline]{enumitem}\setlist{nolistsep,leftmargin=*,labelsep=.5ex}
\setlistdepth{8}\renewlist{itemize}{itemize}{8}
\setlist[itemize]{label=$\cdot$}
\setlist[itemize,1]{label=\textbullet}\setlist[itemize,2]{label=\textopenbullet}

\DeclareUrlCommand{\bfurl}{}
\newcommand{\urlCharBreak}{\penalty50\allowbreak\hspace{0pt plus 1em}}
\let\UrlSpecialsOld\UrlSpecials
\def\UrlSpecials{\UrlSpecialsOld%
\do\.{\penalty50\allowbreak\hspace{0pt plus 1em minus .2ex}\mathchar`.\penalty50\allowbreak\hspace{0pt plus 1em minus .2ex}}%
\do\A{\urlCharBreak\mathchar`A}%
\do\B{\urlCharBreak\mathchar`B}%
\do\C{\urlCharBreak\mathchar`C}%
\do\D{\urlCharBreak\mathchar`D}%
\do\E{\urlCharBreak\mathchar`E}%
\do\F{\urlCharBreak\mathchar`F}%
\do\G{\urlCharBreak\mathchar`G}%
\do\H{\urlCharBreak\mathchar`H}%
\do\I{\urlCharBreak\mathchar`I}%
\do\J{\urlCharBreak\mathchar`J}%
\do\K{\urlCharBreak\mathchar`K}%
\do\L{\urlCharBreak\mathchar`L}%
\do\M{\urlCharBreak\mathchar`M}%
\do\N{\urlCharBreak\mathchar`N}%
\do\O{\urlCharBreak\mathchar`O}%
\do\P{\urlCharBreak\mathchar`P}%
\do\Q{\urlCharBreak\mathchar`Q}%
\do\R{\urlCharBreak\mathchar`R}%
\do\S{\urlCharBreak\mathchar`S}%
\do\T{\urlCharBreak\mathchar`T}%
\do\U{\urlCharBreak\mathchar`U}%
\do\V{\urlCharBreak\mathchar`V}%
\do\W{\urlCharBreak\mathchar`W}%
\do\X{\urlCharBreak\mathchar`X}%
\do\Y{\urlCharBreak\mathchar`Y}%
\do\Z{\urlCharBreak\mathchar`Z}%
}%

\AtBeginDocument{
\addtolength{\textfloatsep}{-.8\textfloatsep}
\addtolength{\floatsep}{-.5\floatsep}
\addtolength{\abovedisplayskip}{-.5\abovedisplayskip}
\addtolength{\belowdisplayskip}{-.5\belowdisplayskip}
\addtolength{\abovedisplayshortskip}{-.5\abovedisplayshortskip}
\addtolength{\belowdisplayshortskip}{-.5\belowdisplayshortskip}
}

\newif\ifBiblatex\Biblatexfalse
\makeatletter
\@ifpackageloaded{biblatex}{\Biblatextrue}{}
\makeatother
\ifBiblatex
\newbibmacro{string+url}[1]{\iffieldundef{url}{\iffieldundef{doi}{#1}{\href{http://dx.doi.org/\thefield{doi}}{#1}}}{\href{\thefield{url}}{#1}}}
\DeclareFieldFormat{title}{\usebibmacro{string+url}{\mkbibemph{#1}}}
\DeclareFieldFormat[article,incollection,inproceedings]{title}{\usebibmacro{string+url}{\mkbibquote{#1}}}
\fi

\newcommand\rightmbox[1]{{%
      \unskip\nobreak\hfil\penalty50%
      \hskip2em\hbox{}\nobreak\hfil\mbox{#1}%
      \parfillskip=0pt \finalhyphendemerits=0 \par}}

\makeatletter
\@ifundefined{chapter}{
\newcommand{\pkginfo}[4]{%
\ifnum0#1<6\section{#4}\else%
\ifnum0#1<7\subsection{#4}\else%
\ifnum0#1<8\subsubsection{#4}\else%
\paragraph{#4}\fi\fi\fi
\label{pkg:#2#3}
\begingroup\raggedright\small\nolinkurl{#2}\penalty10\allowbreak\bfurl{#3} package:\endgroup
\nopagebreak}
\newcommand{\classinfo}[5]{%
\subparagraph{#5}\label{class:#1#2}\hfill
\rightmbox{%
\def\tmp{#4}\ifx\tmp\empty\else\cite{#4} \fi%
 (since~#3)}
\nopagebreak
\begingroup\raggedright\small\nolinkurl{#1}\penalty10\allowbreak\bfurl{#2}\par\endgroup
\nopagebreak
}
}{
\newcommand{\pkginfo}[4]{%
\ifnum0#1<6\chapter{#4}\else%
\ifnum0#1<7\section{#4}\else%
\ifnum0#1<8\subsection{#4}\else%
\subsubsection{#4}\fi\fi\fi
\label{pkg:#2#3}
\begingroup\raggedright\small\nolinkurl{#2}\penalty10\allowbreak\bfurl{#3} package:\endgroup
\nopagebreak}
\newcommand{\classinfo}[5]{%
\paragraph{#5}\label{class:#1#2}\hfill
\rightmbox{%
\def\tmp{#4}\ifx\tmp\empty\else\cite{#4} \fi%
 (since~#3)}
\nopagebreak
\begingroup\raggedright\small\nolinkurl{#1}\penalty10\allowbreak\bfurl{#2}\par\endgroup
\nopagebreak
}
}
\makeatother
\newcommand{\verbatimsize}{\relsize{-1}}

\usepackage[toc,page]{appendix}
\usepackage{minitoc}

\title{ELKI: A large open-source library for data analysis}
\date{\today}
\author{Erich Schubert \and Arthur Zimek}

\begin{document}
\begin{center}
\null\vskip 2em
\makeatletter
{\LARGE\bfseries\@title\par}
\makeatother

\vskip 1em
{\Large\bfseries
ELKI Release 0.7.5 ``Heidelberg''
\par}
\vskip .25em
{\large
\url{https://elki-project.github.io/}
\par}

\vskip 1.5em
\begin{minipage}{.4\linewidth}\centering
Erich Schubert
\\
Technische Universität Dortmund
\end{minipage}
\qquad
\begin{minipage}{.4\linewidth}\centering
Arthur Zimek
\\
University of Southern Denmark
\end{minipage}
\vskip 2em
\doparttoc\faketableofcontents

\end{center}

\section{Introduction}

This paper documents the release of the ELKI data mining framework, version 0.7.5.

\begin{wrapfigure}{r}{6cm}\vspace{-\baselineskip}%
\includegraphics[width=6cm]{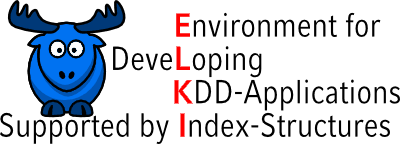}%
\vspace{-.5\baselineskip}%
\end{wrapfigure}
ELKI is an open source (AGPLv3) data mining software written in Java. The focus
of ELKI is research in algorithms, with an emphasis on unsupervised methods in
cluster analysis and outlier detection. In order to achieve high performance
and scalability, ELKI offers data index structures such as the R*-tree that can
provide major performance gains. ELKI is designed to be easy to extend for
researchers and students in this domain, and welcomes contributions of
additional methods. ELKI aims at providing a large collection of highly
parameterizable algorithms, in order to allow easy and fair evaluation and
benchmarking of algorithms.

\medskip
We will first outline the motivation for this release,
the plans for the future, and then give a brief overview over the new functionality in this version.
We also include an appendix presenting an overview on the overall implemented
functionality.

\section{ELKI 0.7.5: ``Heidelberg''}

The majority of the work in this release was done at Heidelberg University,
both by students (that for example contributed many unit tests as part of Java programming practicals)
and by Dr.~Erich Schubert.

The last official release 0.7.1 is now three years old.
Some changes in Java have broken this old release, and it does not run on Java 11 anymore
(trivial fixes are long available, but there has not been a released version) ---
three years where we have continuted to add functionality, improve APIs, fix bugs, \ldots
yet surprisingly many users still use the old version rather than building the latest
development version from git (albeit conveniently available at Github,
\url{https://github.com/elki-project/elki}).

There are a number of breaking changes queued up, waiting for a good moment to be pushed.
This includes an overhaul of the package names, migration to Java 11,
support for automatic indexing, some automatic memory management for automatically added indexes, etc.

Release 0.7.5 is intended as a snapshot to make all the new functionality of the last three years
easily available to everybody, and to open a new window where breaking changes are allowed
to get in, for the next version 0.8.0.

\section{ELKI Plans for the Future}

ELKI is an ongoing project. But because of new career responsibilities,
the former lead authors will likely be able to devote only much less time to this project
in the future.

If you want this project to continue being the largest open-source collection of clustering and outlier detection algorithms
-- and (not the least because of the support for data indexing) also one of the fastest~\cite{DBLP:journals/kais/KriegelSZ17} --
then you should consider contributing to this project yourself:
for example, by adding missing algorithms from the literature to our collection,
fixing smaller bugs, adding completely new functionality, and improving documentation. Thank you.

\section{New Functionality in ELKI 0.7.5}

\subsection{Clustering}
\begin{itemize}

\item Gaussian Mixture Modeling Improvements:\\
\begin{itemize*}
\item Additional models
\item Improved numerical stability \cite{DBLP:conf/ssdbm/SchubertG18}
\item MAP for robustness
\item Better initialization
\end{itemize*}

\item DBSCAN Improvements:

\begin{itemize*}
  \item {GDBSCAN with similarity functions} \cite{DBLP:conf/kdd/EsterKSX96}
  \item {Parallel Generalized DBSCAN} related to \cite{DBLP:conf/sc/PatwaryPALMC12}
\end{itemize*}

\item Hierarchical clustering additions:
\begin{itemize}
  \item {NN-Chain algorithm} \cite{DBLP:journals/cj/Murtagh83,DBLP:journals/corr/abs-1109-2378}
  \item {MiniMax clustering} \cite{DBLP:journals/bioinformatics/AoYNCFMS05,doi:10.1198/jasa.2011.tm10183}
  \item {Flexible Beta Linkage} \cite{doi:10.1093/comjnl/9.4.373}
  \item {Minimum Variance Linkage} \cite{books/misc/DidayLPT85,doi:10.1007/978-94-009-2432-1_5}
\end{itemize}

\item Cluster extraction from dendrograms
  {with handling of noise points and minimum cluster size} \cite{DBLP:journals/corr/abs-1708-03569}

\item {Basic BIRCH: clustering into leaves, with support for:} \cite{DBLP:journals/datamine/ZhangRL97,DBLP:conf/sigmod/ZhangRL96}

Distances:
\begin{itemize*}
  \item {Average Intercluster Distance} %
  \item {Average Intracluster Distance} %
  \item {Centroid Euclidean Distance} %
  \item {Centroid Manhattan Distance} %
  \item {Variance Increase Distance} %
\end{itemize*}

Merge Criterions:
\begin{itemize*}
  \item {Diameter Criterion} %
  \item {Euclidean Distance Criterion}
  \item {Radius Criterion} %
\end{itemize*}

\item PAM clustering additions:
\begin{itemize}
  \item {Reynolds' PAM} \cite{DBLP:journals/jmma/ReynoldsRIR06}
  and {FastPAM} \cite{DBLP:journals/corr/abs-1810-05691}
  \item Improvements to CLARA \cite{doi:10.1016/B978-0-444-87877-9.50039-X,doi:10.1002/9780470316801.ch3}
  and {FastCLARA} \cite{DBLP:journals/corr/abs-1810-05691}
  \item {CLARANS} \cite{DBLP:journals/tkde/NgH02}
  and {FastCLARANS} \cite{DBLP:journals/corr/abs-1810-05691}
\end{itemize}

\item $k$-Means clustering additions:
\begin{itemize}
  \item {Annulus algorithm} \cite{doi:10.1007/978-3-319-09259-1_2,mathesis/Drake13}
  \item {Exponion algorithm} \cite{DBLP:conf/icml/NewlingF16}
  \item {Simplified Elkan's algorithm} \cite{DBLP:conf/icml/NewlingF16}
  \item {$k$-Means-{}- (more robust to noise)} \cite{DBLP:conf/sdm/ChawlaG13}
\end{itemize}

\item $k$-Means and PAM initialization additions:
\begin{itemize}
  \item {Linear Approximative BUILD (LAB)} \cite{DBLP:journals/corr/abs-1810-05691}
  \item {Ostrovsky's initial means} \cite{DBLP:conf/focs/OstrovskyRSS062,DBLP:journals/jacm/OstrovskyRSS12}
  \item {Park's initial medoids} \cite{DBLP:journals/eswa/ParkJ09}
  \item {Generated from a normal distribution} \cite{doi:10.1071/BT9660127}
\end{itemize}

\item Leader Clustering \cite{books/wiley/Hartigan75/C3}
\item {FastDOC} subspace clustering \cite{DBLP:conf/sigmod/ProcopiucJAM02}
\end{itemize}

\subsection{Association Rule Mining}
\begin{itemize}
\item {Association Rule Generation} \cite{DBLP:books/cu/ZM2014}
\item Interestingness Measures:
\begin{itemize*}
  \item {Added Value} \cite{DBLP:conf/dmkdttt/SaharM99}
  \item {Certainty Factor} \cite{DBLP:journals/ida/GalianoBSM02}
  \item {Confidence} \cite{DBLP:conf/sigmod/AgrawalIS93}
  \item {Conviction} \cite{DBLP:conf/sigmod/BrinMUT97}
  \item {Cosine} \cite{tr/umn/TanK00}
  \item {Gini Index} \cite{tr/umn/TanK00,books/wa/BreimanFOS84}
  \item {J Measure} \cite{DBLP:books/mit/PF91/SmythG91}
  \item {Jaccard} \cite{DBLP:books/bu/Rijsbergen79,DBLP:journals/is/TanKS04}
  \item {Klosgen} \cite{DBLP:books/mit/fayyadPSU96/Klosgen96}
  \item {Leverage} \cite{DBLP:books/mit/PF91/Piatetsky91}
  \item {Lift} \cite{DBLP:conf/sigmod/BrinMS97}
\end{itemize*}
\end{itemize}

\subsection{Outlier Detection}
\begin{itemize}
\item {Cluster-Based Local Outlier Factor (CBLOF)} \cite{DBLP:journals/prl/HeXD03}
\item {KNN Data Descriptor (KNNDD)} \cite{conf/asci/deRidderTD98}
\item {Stochastic Outlier Selection (SOS)} \cite{DBLP:conf/sisap/SchubertG17,tr/tilburg/JanssensHPv12},
with {kNN approximation (KNNSOS)} \cite{DBLP:conf/sisap/SchubertG17,tr/tilburg/JanssensHPv12},
and {intrinsic dimensionality (ISOS)} \cite{DBLP:conf/sisap/SchubertG17}
\end{itemize}

\subsection{Projections and Embeddings}
\begin{itemize}
  \item {Stochastic Neighbor Embedding (SNE)} \cite{DBLP:conf/nips/HintonR02}
  \item {t-Stochastic Neighbor Embedding (t-SNE)} \cite{journals/jmlr/MaatenH08}
  \item {Barnes Hut approximation for tSNE} \cite{DBLP:journals/jmlr/Maaten14}
  \item {Intrinsic t-SNE} \cite{DBLP:conf/sisap/SchubertG17}
\end{itemize}

\pagebreak

\subsection{Change Point Detection in Time Series}
\begin{itemize}
  \item {Offline Change Point Detection} \cite{doi:10.2307/2333258,books/prentice/BassevilleN93/C2,doi:10.2307/1427090}
  \item {Signi-Trend-based Change Detection} \cite{DBLP:conf/kdd/SchubertWK14}
\end{itemize}

\subsection{Distance and Similarity Functions}
\begin{itemize*}
  \item Cosine distances optimized for unit-length vectors
  \item {Mahalanobis} \cite{journals/misc/Mahalanobis36}
  \item {Chi Distance} \cite{DBLP:journals/tit/EndresS03,DBLP:conf/iccv/PuzichaRTB99}
  \item {Fisher Rao Distance} \cite{doi:10.1007/978-3-642-00234-2,journals/bcalms/Rao45}
  \item {Triangular Discrimination} \cite{DBLP:journals/tit/Topsoe00}
  \item {Triangular Distance} \cite{DBLP:journals/corr/ConnorCVR16}
\end{itemize*}

\subsection{Evaluation}
\begin{itemize}
  \item {Density-Based Cluster Validation (DBCV)} \cite{DBLP:conf/sdm/MoulaviJCZS14}
  \item {Discounted Cumulative Gain (DCG)} \cite{DBLP:journals/tois/JarvelinK02}
  \item {Normalized Discounted Cumulative Gain (NDCG)} \cite{DBLP:journals/tois/JarvelinK02}
\end{itemize}

\subsection{Indexing Additions}
\begin{itemize}
\item Basic index support for similarities
\item {NN Descent} \cite{DBLP:conf/www/DongCL11}
\item M-Tree enhancements:

\begin{itemize*}
  \item {Farthest Points Split}
  \item {MST Split} \cite{DBLP:conf/edbt/TrainaTSF00}
  \item {Balanced Distribution} \cite{DBLP:conf/vldb/CiacciaPZ97}
  \item {Farthest Balanced Distribution}
  \item {Generalized Hyperplane Distribution} \cite{DBLP:conf/vldb/CiacciaPZ97}
\end{itemize*}
\item X-Tree \cite{DBLP:conf/vldb/BerchtoldKK96} (non-reviewed code)
\end{itemize}

\subsection{Statistics Layer}
\begin{itemize}
\item {Exp-Gamma distribution parameters using the method of moments}
\item {Log-Gamma distribution parameters using the method of moments}
\item {ALID estimator of intrinsic dimensionality} \cite{tr/nii/ChellyHK16}
\end{itemize}

\subsection{Other Improvements}
\begin{itemize}
\item ELKI Builder API (see Section~\ref{sec:elkibuilderapi})
\item Integer range parameters (e.g., \texttt{1,2,..,10,20,..,100,200,..,1000})
\item {Xoroshiro 128 fast random generator} \cite{blog/Lemire16,web/BlackmanV16}
\item Dendrogram visualization for hierarchical clusterings
\item Numerous unit tests
\item Many bug fixes found in testing and reported by users
\end{itemize}

\pagebreak
\section{ELKI Builder API}\label{sec:elkibuilderapi}

The ELKI Builder API is a simpler API to initialize classes from Java.

It is based on the standard ``builder'' pattern, and as it builds upon the
Parameterization API, it has support for \emph{default values} (a feature
unfortunately not in the Java API, nor are named parameters).

You can use standard Java constructors, but you will have to specify all parameters:
\begin{Verbatim}
new KMeansSort<>(
  /* distance= */ EuclideanDistanceFunction.STATIC,
  /* k= */ 10, /* maxiter= */ 0,
  /* init= */ new KMeansPlusPlusInitialMeans<NumberVector>(
    new RandomFactory(/* seed= */ 0L)));
\end{Verbatim}
For more complex classes such as the R$^*$-tree this can become quite complicated.

If we want to add R-tree indexes (with sort-tile-recursive bulk loading), we can initialize the factory
much easier with the ELKIBuilder API, and only specify the page size and bulk load:
\begin{Verbatim}
new ELKIBuilder<>(RStarTreeFactory.class)
    .with(AbstractPageFileFactory.Parameterizer.PAGE_SIZE_ID, 300)
    .with(RStarTreeFactory.Parameterizer.BULK_SPLIT_ID,
              SortTileRecursiveBulkSplit.class)
    .build();
\end{Verbatim}
Furthermore, the parameters directly correspond to the settings in the MiniGUI,
hence make it easier to first experiment in the MiniGUI, then transfer these settings
into Java code. While in the ELKI unit tests we always use the constants (to enable
automatic refactoring), you can also use string parameters here:
\begin{Verbatim}
new ELKIBuilder<>(RStarTreeFactory.class)
    .with("pagefile.pagesize", 300)
    .with("spatial.bulkstrategy", "str")
    .build();
\end{Verbatim}

\section{Availability}

ELKI Release 0.7.5 will be available

\begin{tabbing}
\enskip\textbullet\enskip
on the project website: \quad \=
\url{https://elki-project.github.io/releases/}
\\
\enskip\textbullet\enskip
on Github:
\>
\url{https://github.com/elki-project/elki/releases}
\\
\enskip\textbullet\enskip
on Maven central:
\>in the group \texttt{de.lmu.ifi.dbs.elki}\\
\>(0.7.5 will be the last version using this group id)
\\
\enskip\textbullet\enskip
via Jitpack: \>
\url{https://jitpack.io/p/elki-project/elki}
\\\>
(which also allows accessing development builds)
\end{tabbing}

\bigskip
An example project depending on ELKI is available at\\
\url{https://github.com/elki-project/example-elki-project}.

\pagebreak
\appendix
\addcontentsline{toc}{section}{Appendix} %
\part{Appendix}%

\null\vskip 2em\vskip -4ex %
{\LARGE\bfseries Functionality Available in ELKI 0.7.5\par}
\vskip 2em

In this Appendix, we give a structured overview of the functionality implemented in ELKI.

This is not a complete class list. We only include those classes that
are \emph{modular}, i.e., that can be substituted with other components to
configure the data analysis.

Where the classes are annotated with scientific references, we include them in this list,
as well as the version when this functionality was introduced
in ELKI (possibly with a different name). Note that a reference does not imply
this is based on the original code, nor a complete implementation of each paper; but the references can be used to study the underlying
ideas of the implementations.

This Appendix is, to a large extend, automatically generated from the ELKI JavaDoc documentation.

\bigskip
\mtcsetrules{*}{off}
\mtcsetoffset{parttoc}{0em}
\renewcommand{\ptcindent}{0em}
\setcounter{parttocdepth}{2}
\parttoc%
\pagebreak

\pkginfo{5}%
{de.lmu.ifi.dbs.elki.}{algorithm}%
{Algorithm}%

Algorithms suitable as a task for the {\begingroup\ttfamily{}KDDTask\endgroup} main routine.
 
 The {\begingroup\ttfamily{}KDDTask\endgroup} main routine expects an algorithm to implement
 the {\begingroup\ttfamily{}Algorithm\endgroup}-Interface.
 Basic functions are already provided within {\begingroup\ttfamily{}AbstractAlgorithm\endgroup},
 see there for basic instructions of how to implement an algorithm suitable to the framework.

\classinfo%
{de.lmu.ifi.dbs.elki.algorithm.}{DependencyDerivator}%
{0.1}%
{DBLP:conf/kdd/AchtertBKKZ06}%
{Dependency Derivator: Deriving numerical inter-dependencies on data}%

Dependency derivator computes quantitatively linear dependencies among
attributes of a given dataset based on a linear correlation PCA.

\classinfo%
{de.lmu.ifi.dbs.elki.algorithm.}{KNNDistancesSampler}%
{0.1}%
{DBLP:conf/kdd/EsterKSX96,DBLP:journals/tods/SchubertSEKX17}%
{KNN-Distance-Order}%

Provides an order of the kNN-distances for all objects within the database.

 This class can be used to estimate parameters for other algorithms, such as
estimating the epsilon parameter for DBSCAN: set k to minPts-1, and then
choose a percentile from the sample as epsilon, or plot the result as a graph
and look for a bend or knee in this plot.

\classinfo%
{de.lmu.ifi.dbs.elki.algorithm.}{KNNJoin}%
{0.1}%
{}%
{K-Nearest Neighbor Join}%

Joins in a given spatial database to each object its k-nearest neighbors.
This algorithm only supports spatial databases based on a spatial index
structure.

 Since this method compares the MBR of every single leaf with every other
leaf, it is essentially quadratic in the number of leaves, which may not be
appropriate for large trees. It does currently not yet use the tree structure
for pruning.

\pkginfo{6}%
{de.lmu.ifi.dbs.elki.algorithm.}{benchmark}%
{Benchmark}%

Benchmarking pseudo algorithms.
 
The algorithms in this package are meant to be used in run time benchmarks,
to evalute e.g. the performance of an index structure.

\classinfo%
{de.lmu.ifi.dbs.elki.algorithm.benchmark.}{KNNBenchmarkAlgorithm}%
{0.5.5}%
{}%
{KNN Benchmark Algorithm}%

Benchmarking algorithm that computes the k nearest neighbors for each query
point. The query points can either come from a separate data source, or from
the original database.

\classinfo%
{de.lmu.ifi.dbs.elki.algorithm.benchmark.}{RangeQueryBenchmarkAlgorithm}%
{0.5.5}%
{}%
{Range Query Benchmark Algorithm}%

Benchmarking algorithm that computes a range query for each point. The query
points can either come from a separate data source, or from the original
database. In the latter case, the database is expected to have an additional,
1-dimensional vector field. For the separate data source, the last dimension
will be cut off and used as query radius.

 The simplest data setup clearly is to have an input file:

\begin{Verbatim}[fontsize=\verbatimsize]
x y z label
1 2 3 Example1
4 5 6 Example2
7 8 9 Example3
\end{Verbatim}
 and a query file:

\begin{Verbatim}[fontsize=\verbatimsize]
x y z radius
1 2 3 1.2
4 5 6 3.3
7 8 9 4.1
\end{Verbatim}
 where the additional column is the radius.

 Alternatively, if you work with a single file, you need to use the filter
command \texttt{-dbc.filter SplitNumberVectorFilter -split.dims 1,2,3} to
split the relation into a 3-dimensional data vector, and 1 dimensional radius
vector.

\classinfo%
{de.lmu.ifi.dbs.elki.algorithm.benchmark.}{ValidateApproximativeKNNIndex}%
{0.6.0}%
{}%
{Validate Approximative KNN Index}%

Algorithm to validate the quality of an approximative kNN index, by
performing a number of queries and comparing them to the results obtained by
exact indexing (e.g. linear scanning).

\pkginfo{6}%
{de.lmu.ifi.dbs.elki.algorithm.}{classification}%
{Classification}%

Classification algorithms.

\classinfo%
{de.lmu.ifi.dbs.elki.algorithm.classification.}{KNNClassifier}%
{0.7.0}%
{}%
{kNN-classifier}%

KNNClassifier classifies instances based on the class distribution among the
k nearest neighbors in a database.

\classinfo%
{de.lmu.ifi.dbs.elki.algorithm.classification.}{PriorProbabilityClassifier}%
{0.7.0}%
{}%
{Prior Probability Classifier}%

Classifier to classify instances based on the prior probability of classes in
the database, without using the actual data values.

\pkginfo{6}%
{de.lmu.ifi.dbs.elki.algorithm.}{clustering}%
{Clustering}%

Clustering algorithms.
 
Clustering algorithms are supposed to implement the {\begingroup\ttfamily{}Algorithm\endgroup}-Interface.
The more specialized interface {\begingroup\ttfamily{}ClusteringAlgorithm\endgroup} requires an implementing algorithm to provide a special result class suitable as a partitioning of the database.
More relaxed clustering algorithms are allowed to provide a result that is a fuzzy clustering, does not
partition the database complete or is in any other sense a relaxed clustering result.

\classinfo%
{de.lmu.ifi.dbs.elki.algorithm.clustering.}{CanopyPreClustering}%
{0.6.0}%
{DBLP:conf/kdd/McCallumNU00}%
{Canopy Pre Clustering}%

Canopy pre-clustering is a simple preprocessing step for clustering.

\classinfo%
{de.lmu.ifi.dbs.elki.algorithm.clustering.}{DBSCAN}%
{0.1}%
{DBLP:conf/kdd/EsterKSX96,DBLP:journals/tods/SchubertSEKX17}%
{DBSCAN: Density-Based Clustering of Applications with Noise}%

Density-Based Clustering of Applications with Noise (DBSCAN), an algorithm to
find density-connected sets in a database.

\classinfo%
{de.lmu.ifi.dbs.elki.algorithm.clustering.}{GriDBSCAN}%
{0.7.1}%
{DBLP:conf/IEEEcit/MahranM08}%
{GriDBSCAN: Using Grid for Accelerating Density-Based Clustering}%

Using Grid for Accelerating Density-Based Clustering.

 An accelerated DBSCAN version for numerical data and Lp-norms only, by
partitioning the data set into overlapping grid cells. For best efficiency,
the overlap of the grid cells must be chosen well. The authors suggest a grid
width of 10 times epsilon.

 Because of partitioning the data, this version does not make use of indexes.

\classinfo%
{de.lmu.ifi.dbs.elki.algorithm.clustering.}{Leader}%
{0.7.5}%
{books/wiley/Hartigan75/C3}%
{Leader}%

Leader clustering algorithm.

\classinfo%
{de.lmu.ifi.dbs.elki.algorithm.clustering.}{NaiveMeanShiftClustering}%
{0.5.5}%
{DBLP:journals/pami/Cheng95}%
{Naive Mean Shift Clustering}%

Mean-shift based clustering algorithm. Naive implementation: there does not
seem to be "the" mean-shift clustering algorithm, but it is a general
concept. For the naive implementation, mean-shift is applied to all objects
until they converge to other. This implementation is quite naive, and various
optimizations can be made.

 It also is not really parameter-free: the kernel needs to be specified,
including a radius/bandwidth.

 By using range queries, the algorithm does benefit from index structures!

\classinfo%
{de.lmu.ifi.dbs.elki.algorithm.clustering.}{SNNClustering}%
{0.1}%
{DBLP:conf/sdm/ErtozSK03}%
{SNN: Shared Nearest Neighbor Clustering}%

Shared nearest neighbor clustering.

\pkginfo{7}%
{de.lmu.ifi.dbs.elki.algorithm.clustering.}{affinitypropagation}%
{Affinitypropagation}%

Affinity Propagation (AP) clustering.

\classinfo%
{de.lmu.ifi.dbs.elki.algorithm.clustering.affinitypropagation.}{AffinityPropagationClusteringAlgorithm}%
{0.6.0}%
{doi:10.1126/science.1136800}%
{Affinity Propagation: Clustering by Passing Messages Between Data Points}%

Cluster analysis by affinity propagation.

\classinfo%
{de.lmu.ifi.dbs.elki.algorithm.clustering.affinitypropagation.}{DistanceBasedInitializationWithMedian}%
{0.6.0}%
{}%
{Distance Based Initialization With Median}%

Distance based initialization.

\classinfo%
{de.lmu.ifi.dbs.elki.algorithm.clustering.affinitypropagation.}{SimilarityBasedInitializationWithMedian}%
{0.6.0}%
{}%
{Similarity Based Initialization With Median}%

Similarity based initialization.

\pkginfo{7}%
{de.lmu.ifi.dbs.elki.algorithm.clustering.}{biclustering}%
{Biclustering}%

Biclustering algorithms.

\classinfo%
{de.lmu.ifi.dbs.elki.algorithm.clustering.biclustering.}{ChengAndChurch}%
{0.6.0}%
{DBLP:conf/ismb/ChengC00}%
{Cheng And Church}%

Cheng and Church biclustering.

\pkginfo{7}%
{de.lmu.ifi.dbs.elki.algorithm.clustering.}{correlation}%
{Correlation}%

Correlation clustering algorithms

\classinfo%
{de.lmu.ifi.dbs.elki.algorithm.clustering.correlation.}{CASH}%
{0.1}%
{DBLP:conf/sdm/AchtertBDKZ08}%
{CASH: Robust clustering in arbitrarily oriented subspaces}%

The CASH algorithm is a subspace clustering algorithm based on the Hough
transform.

\classinfo%
{de.lmu.ifi.dbs.elki.algorithm.clustering.correlation.}{COPAC}%
{0.1}%
{DBLP:conf/sdm/AchtertBKKZ07}%
{COPAC: COrrelation PArtition Clustering}%

COPAC is an algorithm to partition a database according to the correlation
dimension of its objects and to then perform an arbitrary clustering
algorithm over the partitions.

\classinfo%
{de.lmu.ifi.dbs.elki.algorithm.clustering.correlation.}{ERiC}%
{0.1}%
{DBLP:conf/ssdbm/AchtertBKKZ07}%
{ERiC: Exploring Relationships among Correlation Clusters}%

Performs correlation clustering on the data partitioned according to local
correlation dimensionality and builds a hierarchy of correlation clusters
that allows multiple inheritance from the clustering result.

\classinfo%
{de.lmu.ifi.dbs.elki.algorithm.clustering.correlation.}{FourC}%
{0.1}%
{DBLP:conf/sigmod/BohmKKZ04}%
{4C: Computing Correlation Connected Clusters}%

4C identifies local subgroups of data objects sharing a uniform correlation.
The algorithm is based on a combination of PCA and density-based clustering
(DBSCAN).

\classinfo%
{de.lmu.ifi.dbs.elki.algorithm.clustering.correlation.}{HiCO}%
{0.1}%
{DBLP:conf/ssdbm/AchtertBKZ06}%
{Mining Hierarchies of Correlation Clusters}%

Implementation of the HiCO algorithm, an algorithm for detecting hierarchies
of correlation clusters.

\classinfo%
{de.lmu.ifi.dbs.elki.algorithm.clustering.correlation.}{LMCLUS}%
{0.5.0}%
{DBLP:journals/pr/HaralickH07}%
{LMCLUS}%

Linear manifold clustering in high dimensional spaces by stochastic search.

\classinfo%
{de.lmu.ifi.dbs.elki.algorithm.clustering.correlation.}{ORCLUS}%
{0.1}%
{DBLP:conf/sigmod/AggarwalY00}%
{ORCLUS: Arbitrarily ORiented projected CLUSter generation}%

ORCLUS: Arbitrarily ORiented projected CLUSter generation.

\pkginfo{7}%
{de.lmu.ifi.dbs.elki.algorithm.clustering.}{em}%
{EM}%

Expectation-Maximization clustering algorithm.

\classinfo%
{de.lmu.ifi.dbs.elki.algorithm.clustering.em.}{DiagonalGaussianModelFactory}%
{0.7.0}%
{}%
{Diagonal Gaussian Model Factory}%

Factory for EM with multivariate gaussian models using diagonal matrixes.

 These models have individual variances, but no covariance, so this
corresponds to the {\begingroup\ttfamily{}'VVI'\endgroup} model in Mclust (R).

\classinfo%
{de.lmu.ifi.dbs.elki.algorithm.clustering.em.}{EM}%
{0.1}%
{DBLP:journals/classification/FraleyR07,journals/jroyastatsocise2/DempsterLR77}%
{EM-Clustering: Clustering by Expectation Maximization}%

Clustering by expectation maximization (EM-Algorithm), also known as Gaussian
Mixture Modeling (GMM), with optional MAP regularization.

\classinfo%
{de.lmu.ifi.dbs.elki.algorithm.clustering.em.}{MultivariateGaussianModelFactory}%
{0.7.0}%
{}%
{Multivariate Gaussian Model Factory}%

Factory for EM with multivariate Gaussian models (with covariance; also known
as Gaussian Mixture Modeling, GMM).

These models have individual covariance matrixes, so this corresponds to the
{\begingroup\ttfamily{}'VVV'\endgroup} model in Mclust (R).

\classinfo%
{de.lmu.ifi.dbs.elki.algorithm.clustering.em.}{SphericalGaussianModelFactory}%
{0.7.0}%
{}%
{Spherical Gaussian Model Factory}%

Factory for EM with multivariate gaussian models using a single variance.

 These models have a single variances, no covariance, so this corresponds to
the {\begingroup\ttfamily{}'VII'\endgroup} model in Mclust (R).

\classinfo%
{de.lmu.ifi.dbs.elki.algorithm.clustering.em.}{TextbookMultivariateGaussianModelFactory}%
{0.7.5}%
{}%
{Textbook Multivariate Gaussian Model Factory}%

Factory for EM with multivariate Gaussian model, using the textbook
algorithm. There is no reason to use this in practice, it is only useful to
study the reliability of the textbook approach.

"Textbook" refers to the E[XY]-E[X]E[Y] equation for covariance, that is
numerically not reliable with floating point math, but popular in textbooks.

Again, do not use this. Always prefer {\begingroup\ttfamily{}\#MultivariateGaussianModelFactory\endgroup}.

\classinfo%
{de.lmu.ifi.dbs.elki.algorithm.clustering.em.}{TwoPassMultivariateGaussianModelFactory}%
{0.7.5}%
{}%
{Two Pass Multivariate Gaussian Model Factory}%

Factory for EM with multivariate Gaussian models (with covariance; also known
as Gaussian Mixture Modeling, GMM).

These models have individual covariance matrixes, so this corresponds to the
{\begingroup\ttfamily{}'VVV'\endgroup} model in Mclust (R).

\pkginfo{7}%
{de.lmu.ifi.dbs.elki.algorithm.clustering.}{gdbscan}%
{GDBSCAN}%

Generalized DBSCAN.
 Generalized DBSCAN is an abstraction of the original DBSCAN idea,
that allows the use of arbitrary "neighborhood" and "core point" predicates.

 For each object, the neighborhood as defined by the "neighborhood" predicate
is retrieved - in original DBSCAN, this is the objects within an epsilon sphere
around the query object. Then the core point predicate is evaluated to decide if
the object is considered dense. If so, a cluster is started (or extended) to
include the neighbors as well.

\classinfo%
{de.lmu.ifi.dbs.elki.algorithm.clustering.gdbscan.}{COPACNeighborPredicate}%
{0.7.0}%
{DBLP:conf/sdm/AchtertBKKZ07}%
{COPAC Neighbor Predicate}%

COPAC neighborhood predicate.

\classinfo%
{de.lmu.ifi.dbs.elki.algorithm.clustering.gdbscan.}{ERiCNeighborPredicate}%
{0.7.0}%
{DBLP:conf/ssdbm/AchtertBKKZ07}%
{ERiC Neighbor Predicate}%

ERiC neighborhood predicate.

\classinfo%
{de.lmu.ifi.dbs.elki.algorithm.clustering.gdbscan.}{EpsilonNeighborPredicate}%
{0.5.0}%
{DBLP:conf/kdd/EsterKSX96}%
{Epsilon Neighbor Predicate}%

The default DBSCAN and OPTICS neighbor predicate, using an
epsilon-neighborhood.

\classinfo%
{de.lmu.ifi.dbs.elki.algorithm.clustering.gdbscan.}{FourCCorePredicate}%
{0.7.0}%
{DBLP:conf/sigmod/BohmKKZ04}%
{Four C Core Predicate}%

The 4C core point predicate.

\classinfo%
{de.lmu.ifi.dbs.elki.algorithm.clustering.gdbscan.}{FourCNeighborPredicate}%
{0.7.0}%
{DBLP:conf/sigmod/BohmKKZ04}%
{Four C Neighbor Predicate}%

4C identifies local subgroups of data objects sharing a uniform correlation.
The algorithm is based on a combination of PCA and density-based clustering
(DBSCAN).

\classinfo%
{de.lmu.ifi.dbs.elki.algorithm.clustering.gdbscan.}{GeneralizedDBSCAN}%
{0.5.0}%
{DBLP:journals/datamine/SanderEKX98}%
{Generalized DBSCAN}%

Generalized DBSCAN, density-based clustering with noise.

\classinfo%
{de.lmu.ifi.dbs.elki.algorithm.clustering.gdbscan.}{LSDBC}%
{0.7.0}%
{DBLP:conf/icannga/BiciciY07}%
{LSDBC: Locally Scaled Density Based Clustering}%

Locally Scaled Density Based Clustering.

 This is a variant of DBSCAN which starts with the most dense point first,
then expands clusters until density has dropped below a threshold.

\classinfo%
{de.lmu.ifi.dbs.elki.algorithm.clustering.gdbscan.}{MinPtsCorePredicate}%
{0.5.0}%
{DBLP:conf/kdd/EsterKSX96}%
{MinPts Core Predicate}%

The DBSCAN default core point predicate -- having at least {\begingroup\ttfamily{}minpts\endgroup} neighbors.

\classinfo%
{de.lmu.ifi.dbs.elki.algorithm.clustering.gdbscan.}{PreDeConCorePredicate}%
{0.7.0}%
{DBLP:conf/icdm/BohmKKK04}%
{PreDeCon Core Predicate}%

The PreDeCon core point predicate -- having at least minpts. neighbors, and a
maximum preference dimensionality of lambda.

\classinfo%
{de.lmu.ifi.dbs.elki.algorithm.clustering.gdbscan.}{PreDeConNeighborPredicate}%
{0.7.0}%
{DBLP:conf/icdm/BohmKKK04}%
{PreDeCon Neighbor Predicate}%

Neighborhood predicate used by PreDeCon.

\classinfo%
{de.lmu.ifi.dbs.elki.algorithm.clustering.gdbscan.}{SimilarityNeighborPredicate}%
{0.7.5}%
{DBLP:conf/kdd/EsterKSX96}%
{Similarity Neighbor Predicate}%

The DBSCAN neighbor predicate for a {\begingroup\ttfamily{}SimilarityFunction\endgroup}, using all
neighbors with a minimum similarity.

\pkginfo{8}%
{de.lmu.ifi.dbs.elki.algorithm.clustering.gdbscan.}{parallel}%
{Gdbscan Parallel}%

Parallel versions of Generalized DBSCAN.

\classinfo%
{de.lmu.ifi.dbs.elki.algorithm.clustering.gdbscan.parallel.}{ParallelGeneralizedDBSCAN}%
{0.7.5}%
{DBLP:conf/sc/PatwaryPALMC12}%
{Parallel Generalized DBSCAN}%

Parallel version of DBSCAN clustering.

 This is the archetype of a non-linear shared-memory DBSCAN that does not
sequentially expand a cluster, but processes points in arbitrary order and
merges clusters when neighboring core points occur.

 Because of synchronization when labeling points, the speedup will only be
sublinear in the number of cores. But in particular without an index and on
large data, the majority of the work is finding the neighbors; not in
labeling the points.

\pkginfo{7}%
{de.lmu.ifi.dbs.elki.algorithm.clustering.}{hierarchical}%
{Hierarchical}%

Hierarchical agglomerative clustering (HAC).

\classinfo%
{de.lmu.ifi.dbs.elki.algorithm.clustering.hierarchical.}{AGNES}%
{0.6.0}%
{doi:10.2307/2344237,doi:10.1002/9780470316801.ch5,doi:10.1099/00221287-17-1-201}%
{AGNES}%

Hierarchical Agglomerative Clustering (HAC) or Agglomerative Nesting (AGNES)
is a classic hierarchical clustering algorithm. Initially, each element is
its own cluster; the closest clusters are merged at every step, until all the
data has become a single cluster.

 This is the naive O(n³) algorithm. See {\begingroup\ttfamily{}SLINK\endgroup} for a much faster
algorithm (however, only for single-linkage).

 This implementation uses the pointer-based representation used by SLINK, so
that the extraction algorithms we have can be used with either of them.

 The algorithm is believed to be first published (for single-linkage) by:

 P. H. Sneath\\
 The application of computers to taxonomy\\
 Journal of general microbiology, 17(1).

 This algorithm is also known as AGNES (Agglomerative Nesting), where the use
of alternative linkage criterions is discussed:

 L. Kaufman, P. J. Rousseeuw\\
 Agglomerative Nesting (Program AGNES),\\
 in Finding Groups in Data: An Introduction to Cluster Analysis

\classinfo%
{de.lmu.ifi.dbs.elki.algorithm.clustering.hierarchical.}{AbstractHDBSCAN}%
{0.7.0}%
{DBLP:conf/pakdd/CampelloMS13}%
{Abstract HDBSCAN}%

Abstract base class for HDBSCAN variations.

\classinfo%
{de.lmu.ifi.dbs.elki.algorithm.clustering.hierarchical.}{AnderbergHierarchicalClustering}%
{0.7.0}%
{books/academic/Anderberg73/Ch6}%
{Anderberg Hierarchical Clustering}%

This is a modification of the classic AGNES algorithm for hierarchical
clustering using a nearest-neighbor heuristic for acceleration.

 Instead of scanning the matrix (with cost O(n²)) to find the minimum, the
nearest neighbor of each object is remembered. On the downside, we need to
check these values at every merge, and it may now cost O(n²) to perform a
merge, so there is no worst-case advantage to this approach. The average case
however improves from O(n³) to O(n²), which yields a considerable
improvement in running time.

 This optimization is attributed to M. R. Anderberg.

\classinfo%
{de.lmu.ifi.dbs.elki.algorithm.clustering.hierarchical.}{CLINK}%
{0.7.0}%
{DBLP:journals/cj/Defays77}%
{CLINK}%

CLINK algorithm for complete linkage.

 This algorithm runs in O(n²) time, and needs only O(n) memory. The results
can differ from the standard algorithm in unfavorable ways, and are
order-dependent (Defays: "Modifications of the labeling permit us to obtain
different minimal superior ultrametric dissimilarities"). Unfortunately, the
results are usually perceived to be substantially worse than the more
expensive algorithms for complete linkage clustering. This arises from the
fact that this algorithm has to add the new object to the existing tree in
every step, instead of being able to always do the globally best merge.

\classinfo%
{de.lmu.ifi.dbs.elki.algorithm.clustering.hierarchical.}{HDBSCANLinearMemory}%
{0.7.0}%
{DBLP:conf/pakdd/CampelloMS13}%
{HDBSCAN: Hierarchical Density-Based Spatial Clustering of Applications with Noise}%

Linear memory implementation of HDBSCAN clustering.

 By not building a distance matrix, we can reduce memory usage to linear
memory only; but at the cost of roughly double the runtime (unless using
indexes) as we first need to compute all kNN distances (for core sizes), then
recompute distances when building the spanning tree.

 This implementation follows the HDBSCAN publication more closely than
{\begingroup\ttfamily{}SLINKHDBSCANLinearMemory\endgroup}, by computing the minimum spanning tree
using Prim's algorithm (instead of SLINK; although the two are remarkably
similar). In order to produce the preferred internal format of hierarchical
clusterings (the compact pointer representation introduced in {\begingroup\ttfamily{}SLINK\endgroup})
we have to perform a postprocessing conversion.

 This implementation does not include the cluster extraction
discussed as Step 4, which is provided in a separate step. For this reason,
we also do not include self-edges.

\classinfo%
{de.lmu.ifi.dbs.elki.algorithm.clustering.hierarchical.}{MiniMax}%
{0.7.5}%
{DBLP:journals/bioinformatics/AoYNCFMS05,doi:10.1198/jasa.2011.tm10183}%
{Mini Max}%

Minimax Linkage clustering.

\classinfo%
{de.lmu.ifi.dbs.elki.algorithm.clustering.hierarchical.}{MiniMaxAnderberg}%
{0.7.5}%
{books/academic/Anderberg73/Ch6}%
{Mini Max Anderberg}%

This is a modification of the classic MiniMax algorithm for hierarchical
clustering using a nearest-neighbor heuristic for acceleration.

 This optimization is attributed to M. R. Anderberg.

 This particular implementation is based on AnderbergHierarchicalClustering

\classinfo%
{de.lmu.ifi.dbs.elki.algorithm.clustering.hierarchical.}{MiniMaxNNChain}%
{0.7.5}%
{DBLP:journals/cj/Murtagh83,DBLP:journals/corr/abs-1109-2378}%
{Mini Max NN Chain}%

MiniMax hierarchical clustering using the NNchain algorithm.

\classinfo%
{de.lmu.ifi.dbs.elki.algorithm.clustering.hierarchical.}{NNChain}%
{0.7.5}%
{DBLP:journals/cj/Murtagh83,DBLP:journals/corr/abs-1109-2378}%
{NN Chain}%

NNchain clustering algorithm.

\classinfo%
{de.lmu.ifi.dbs.elki.algorithm.clustering.hierarchical.}{SLINK}%
{0.6.0}%
{DBLP:journals/cj/Sibson73}%
{SLINK: Single Link Clustering}%

Implementation of the efficient Single-Link Algorithm SLINK of R. Sibson.

This is probably the fastest exact single-link algorithm currently in use.

\classinfo%
{de.lmu.ifi.dbs.elki.algorithm.clustering.hierarchical.}{SLINKHDBSCANLinearMemory}%
{0.7.0}%
{DBLP:conf/pakdd/CampelloMS13}%
{SLINKHDBSCAN Linear Memory}%

Linear memory implementation of HDBSCAN clustering based on SLINK.

 By not building a distance matrix, we can reduce memory usage to linear
memory only; but at the cost of roughly double the runtime (unless using
indexes) as we first need to compute all kNN distances (for core sizes), then
recompute distances when building the spanning tree.

 This version uses the SLINK algorithm to directly produce the pointer
representation expected by the extraction methods. The SLINK algorithm is
closely related to Prim's minimum spanning tree, but produces the more
compact pointer representation instead of an edges list.

 This implementation does not include the cluster extraction
discussed as Step 4. This functionality should however already be provided by
{\begingroup\ttfamily{}HDBSCANHierarchyExtraction\endgroup} . For this reason, we also do not include self-edges.

\pkginfo{8}%
{de.lmu.ifi.dbs.elki.algorithm.clustering.hierarchical.}{birch}%
{Hierarchical Birch}%

BIRCH clustering.

\classinfo%
{de.lmu.ifi.dbs.elki.algorithm.clustering.hierarchical.birch.}{AverageInterclusterDistance}%
{0.7.5}%
{tr/wisc/Zhang97}%
{Average Intercluster Distance}%

Average intercluster distance.

\classinfo%
{de.lmu.ifi.dbs.elki.algorithm.clustering.hierarchical.birch.}{AverageIntraclusterDistance}%
{0.7.5}%
{tr/wisc/Zhang97}%
{Average Intracluster Distance}%

Average intracluster distance.

\classinfo%
{de.lmu.ifi.dbs.elki.algorithm.clustering.hierarchical.birch.}{BIRCHLeafClustering}%
{0.7.5}%
{DBLP:journals/datamine/ZhangRL97,DBLP:conf/sigmod/ZhangRL96}%
{BIRCH Leaf Clustering}%

BIRCH-based clustering algorithm that simply treats the leafs of the CFTree
as clusters.

\classinfo%
{de.lmu.ifi.dbs.elki.algorithm.clustering.hierarchical.birch.}{CFTree}%
{0.7.5}%
{DBLP:journals/datamine/ZhangRL97,DBLP:conf/sigmod/ZhangRL96}%
{CF Tree}%

Partial implementation of the CFTree as used by BIRCH.

 Important differences:

\begin{enumerate}
\item Leaf nodes and directory nodes have the same capacity
\item Condensing and memory limits are not implemented
\item Merging refinement (merge-resplit) is not implemented
\end{enumerate}
 Because we want to be able to track the cluster assignments of all data
points easily, we need to store the point IDs, and it is not possible to
implement the originally proposed page size management at the same time.

 Condensing and merging refinement are possible, and improvements to this code
are welcome - please send a pull request!

\classinfo%
{de.lmu.ifi.dbs.elki.algorithm.clustering.hierarchical.birch.}{CentroidEuclideanDistance}%
{0.7.5}%
{tr/wisc/Zhang97}%
{Centroid Euclidean Distance}%

Centroid Euclidean distance.

\classinfo%
{de.lmu.ifi.dbs.elki.algorithm.clustering.hierarchical.birch.}{CentroidManhattanDistance}%
{0.7.5}%
{tr/wisc/Zhang97}%
{Centroid Manhattan Distance}%

Centroid Manhattan Distance

\classinfo%
{de.lmu.ifi.dbs.elki.algorithm.clustering.hierarchical.birch.}{DiameterCriterion}%
{0.7.5}%
{DBLP:conf/sigmod/ZhangRL96}%
{Diameter Criterion}%

Average Radius (R) criterion.

\classinfo%
{de.lmu.ifi.dbs.elki.algorithm.clustering.hierarchical.birch.}{EuclideanDistanceCriterion}%
{0.7.5}%
{}%
{Euclidean Distance Criterion}%

Distance criterion.

This is not found in the original work, but used in many implementation
attempts: assign points if the Euclidean distances is below a threshold.

\classinfo%
{de.lmu.ifi.dbs.elki.algorithm.clustering.hierarchical.birch.}{RadiusCriterion}%
{0.7.5}%
{DBLP:conf/sigmod/ZhangRL96}%
{Radius Criterion}%

Average Radius (R) criterion.

\classinfo%
{de.lmu.ifi.dbs.elki.algorithm.clustering.hierarchical.birch.}{VarianceIncreaseDistance}%
{0.7.5}%
{tr/wisc/Zhang97}%
{Variance Increase Distance}%

Variance increase distance.

\pkginfo{8}%
{de.lmu.ifi.dbs.elki.algorithm.clustering.hierarchical.}{extraction}%
{Hierarchical Extraction}%

Extraction of partitional clusterings from hierarchical results.

\classinfo%
{de.lmu.ifi.dbs.elki.algorithm.clustering.hierarchical.extraction.}{ClustersWithNoiseExtraction}%
{0.7.5}%
{DBLP:journals/corr/abs-1708-03569}%
{Clusters With Noise Extraction}%

Extraction of a given number of clusters with a minimum size, and noise.

 This will execute the highest-most cut where we retain k clusters, each with
a minimum size, plus noise (single points that would only merge afterwards).
If no such cut can be found, it returns a result with a relaxed k.

 You need to specify: A) the minimum size of a cluster (it does not make much
sense to use 1 - then it will simply execute all but the last k merges) and
B) the desired number of clusters with at least minSize elements each.

\classinfo%
{de.lmu.ifi.dbs.elki.algorithm.clustering.hierarchical.extraction.}{CutDendrogramByHeight}%
{0.7.5}%
{}%
{Cut Dendrogram By Height}%

Extract a flat clustering from a full hierarchy, represented in pointer form.

\classinfo%
{de.lmu.ifi.dbs.elki.algorithm.clustering.hierarchical.extraction.}{CutDendrogramByNumberOfClusters}%
{0.7.5}%
{}%
{Cut Dendrogram By Number Of Clusters}%

Extract a flat clustering from a full hierarchy, represented in pointer form.

\classinfo%
{de.lmu.ifi.dbs.elki.algorithm.clustering.hierarchical.extraction.}{HDBSCANHierarchyExtraction}%
{0.7.0}%
{DBLP:conf/pakdd/CampelloMS13}%
{HDBSCAN Hierarchy Extraction}%

Extraction of simplified cluster hierarchies, as proposed in HDBSCAN.

 In contrast to the authors top-down approach, we use a bottom-up approach
based on the more efficient pointer representation introduced in SLINK.

 In particular, it can also be used to extract a hierarchy from a hierarchical
agglomerative clustering.

\classinfo%
{de.lmu.ifi.dbs.elki.algorithm.clustering.hierarchical.extraction.}{SimplifiedHierarchyExtraction}%
{0.7.0}%
{DBLP:conf/pakdd/CampelloMS13}%
{Simplified Hierarchy Extraction}%

Extraction of simplified cluster hierarchies, as proposed in HDBSCAN.

 In contrast to the authors top-down approach, we use a bottom-up approach
based on the more efficient pointer representation introduced in SLINK.

\pkginfo{8}%
{de.lmu.ifi.dbs.elki.algorithm.clustering.hierarchical.}{linkage}%
{Hierarchical Linkage}%

Linkages for hierarchical clustering.

\classinfo%
{de.lmu.ifi.dbs.elki.algorithm.clustering.hierarchical.linkage.}{CentroidLinkage}%
{0.6.0}%
{doi:10.2307/2528417}%
{Centroid Linkage}%

Centroid linkage — Unweighted Pair-Group Method using Centroids
(UPGMC).

 This is closely related to {\begingroup\ttfamily{}GroupAverageLinkage\endgroup} (UPGMA), but the
resulting distance corresponds to the distance of the cluster centroids when
used with squared Euclidean distance.

 For Lance-Williams, we can then obtain the following recursive definition:
\[d_{\text{UPGMC}}(A\cup B,C)=\tfrac{|A|}{|A|+|B|} d(A,C) +
\tfrac{|B|}{|A|+|B|} d(B,C) - \tfrac{|A|\cdot|B|}{(|A|+|B|)^2} d(A,B)\]

 With squared Euclidean distance, we then get the cluster distance:
\[d_{\text{UPGMC}}(A,B)=||\tfrac{1}{|A|}\sum\nolimits_{a\in A} a,
\tfrac{1}{|B|}\sum\nolimits_{b\in B} b||^2\]
but for other distances, this will not generally be true.

 Because the ELKI implementations use Lance-Williams, this linkage should only
be used with (squared) Euclidean distance.

 While titled "unweighted", this method does take cluster sizes into account
when merging clusters with Lance-Williams.

 While the idea of this method — at least for squared Euclidean —
is compelling (distance of cluster centers), it is not as well behaved as one
may think. It can yield so called "inversions", where a later merge has a
smaller distance than an early merge, because a cluster center can be closer to a neighboring cluster than any of the individual points. Because
of this, the {\begingroup\ttfamily{}GroupAverageLinkage\endgroup} (UPGMA) is usually preferable.

\classinfo%
{de.lmu.ifi.dbs.elki.algorithm.clustering.hierarchical.linkage.}{CompleteLinkage}%
{0.6.0}%
{doi:10.1093/comjnl/9.4.373,journals/misc/Sorensen48,journals/psychometrika/Johnson67,journals/misc/MacnaughtonSmith65}%
{Complete Linkage}%

Complete-linkage ("maximum linkage") clustering method.

 The distance of two clusters is simply the maximum of all pairwise distances
between the two clusters.

 The distance of two clusters is defined as:
\[d_{\max}(A,B):=\max_{a\in A}\max_{b\in B} d(a,b)\]

 This can be computed recursively using:
\[d_{\max}(A\cup B,C) = \max(d(A,C), d(B,C))\]

 Note that with similarity functions, one would need to use the minimum
instead to get the same effect.

 The algorithm {\begingroup\ttfamily{}CLINK\endgroup} is a faster algorithm to find such clusterings,
but it is very much order dependent and tends to find worse solutions.

\classinfo%
{de.lmu.ifi.dbs.elki.algorithm.clustering.hierarchical.linkage.}{FlexibleBetaLinkage}%
{0.7.5}%
{doi:10.1093/comjnl/9.4.373}%
{Flexible Beta Linkage}%

Flexible-beta linkage as proposed by Lance and Williams.

 Beta values larger than 0 cause chaining, and are thus not recommended.
Instead, choose a value between -1 and 0.

 The general form of the recursive definition is:
\[d_{\text{Flexible},\beta}(A\cup B, C) := \tfrac{1-\beta}{2} d(A,C)
+ \tfrac{1-\beta}{2} d(B,C) + \beta d(A,B) \]

\classinfo%
{de.lmu.ifi.dbs.elki.algorithm.clustering.hierarchical.linkage.}{GroupAverageLinkage}%
{0.6.0}%
{journals/kansas/SokalM1902}%
{Group Average Linkage}%

Group-average linkage clustering method (UPGMA).

 This is a good default linkage to use with hierarchical clustering, as it
neither exhibits the single-link chaining effect, nor has the strong tendency
of complete linkage to split large clusters. It is also easy to understand,
and it can be used with arbitrary distances and similarity functions.

 The distances of two clusters is defined as the between-group average
distance of two points $a$ and $b$, one from each cluster. It should be noted
that this is not the average distance within the resulting cluster, because
it does not take within-cluster distances into account.

 The distance of two clusters in this method is:
\[d_{\text{UPGMA}}(A,B)=\tfrac{1}{|A|\cdot|B|}
\sum\nolimits_{a\in A}\sum\nolimits_{b\in B} d(a,b)\]

 For Lance-Williams, we can then obtain the following recursive definition:
\[d_{\text{UPGMA}}(A\cup B,C)=\tfrac{|A|}{|A|+|B|} d(A,C) +
\tfrac{|B|}{|A|+|B|} d(B,C)\]

 While the method is also called "Unweighted Pair Group Method with Arithmetic
mean", it uses weights in the Lance-Williams formulation that account for the
cluster size. It is unweighted in the sense that every point keeps the same
weight, whereas in {\begingroup\ttfamily{}WeightedAverageLinkage\endgroup} (WPGMA), the weight of
points effectively depends on the depth in the cluster tree.

\classinfo%
{de.lmu.ifi.dbs.elki.algorithm.clustering.hierarchical.linkage.}{Linkage}%
{0.6.0}%
{doi:10.1093/comjnl/9.4.373}%
{Linkage}%

Abstract interface for implementing a new linkage method into hierarchical
clustering.

\classinfo%
{de.lmu.ifi.dbs.elki.algorithm.clustering.hierarchical.linkage.}{MedianLinkage}%
{0.6.0}%
{doi:10.2307/2528417}%
{Median Linkage}%

Median-linkage — weighted pair group method using centroids (WPGMC).

 Similar to {\begingroup\ttfamily{}WeightedAverageLinkage\endgroup} (WPGMA), the weight of points in
this method decreases with the depth of the tree. This yields to difficult to
understand semantics of the result, as it does not yield the distance of
medians. The method is best defined recursively:
\[d_{\text{WPGMC}}(A\cup B,C):=\tfrac{1}{2}d(A,C)+\tfrac{1}{2}d(B,C)
-\tfrac{1}{4}d(A,B)\]

\classinfo%
{de.lmu.ifi.dbs.elki.algorithm.clustering.hierarchical.linkage.}{MinimumVarianceLinkage}%
{0.7.5}%
{books/misc/DidayLPT85,doi:10.1007/978-94-009-2432-1_5}%
{Minimum Variance Linkage}%

Minimum increase in variance (MIVAR) linkage.

 This is subtly different from Ward's method ({\begingroup\ttfamily{}WardLinkage\endgroup}, MISSQ),
because variance is normalized by the cluster size; and Ward minimizes the
increase in sum of squares (without normalization).
\[d_{\text{MIVAR}}(A\cup B,C)=
\left(\tfrac{|A|+|C|}{|A|+|B|+|C|}\right)^2 d(A,C) +
\left(\tfrac{|B|+|C|}{|A|+|B|+|C|}\right)^2 d(B,C)
- \tfrac{|C|\cdot(|A|+|B|)}{(|A|+|B|+|C|)^2} d(A,B)\]
or equivalently:
\[d_{\text{MIVAR}}(A\cup B,C)=\tfrac{(|A|+|C|)^2 d(A,C)
+ (|B|+|C|)^2 d(B,C) - |C|\cdot(|A|+|B|) d(A,B)}{(|A|+|B|+|C|)^2}\]

\classinfo%
{de.lmu.ifi.dbs.elki.algorithm.clustering.hierarchical.linkage.}{SingleLinkage}%
{0.6.0}%
{journals/misc/FlorekLPSZ51}%
{Single Linkage}%

Single-linkage ("minimum") clustering method.

 The distance of two clusters is simply the minimum of all pairwise distances
between the two clusters.

 The distance of two clusters is defined as:
\[d_{\min}(A,B):=\min_{a\in A}\min_{b\in B} d(a,b)\]

 This can be computed recursively using:
\[d_{\min}(A\cup B,C) = \min(d(A,C), d(B,C))\]

 Note that with similarity functions, one would need to use the maximum
instead to get the same effect.

\classinfo%
{de.lmu.ifi.dbs.elki.algorithm.clustering.hierarchical.linkage.}{WardLinkage}%
{0.6.0}%
{doi:10.2307/2528688,doi:10.1080/01621459.1963.10500845}%
{Ward Linkage}%

Ward's method clustering method.

 This criterion minimizes the increase of squared errors, and should
be used with squared Euclidean distance. Usually, ELKI will try to
automatically square distances when you combine this with Euclidean distance.
For performance reasons, the direct use of squared distances is preferable!

 The distance of two clusters in this method is:
\[ d_{\text{Ward}}(A,B):=\text{SSE}(A\cup B)-\text{SSE}(A)-\text{SSE}(B) \]
where the sum of squared errors is defined as:
\[ \text{SSE}(X):=\sum\nolimits_{x\in X} (x-\mu_X)^2 \qquad \text{with }
\mu_X=\tfrac{1}{|X|}\sum\nolimits_{x\in X} X \]
This objective can be rewritten to
\[ d_{\text{Ward}}(A,B):=\tfrac{|A|\cdot|B|}{|A|+|B|} ||\mu_A-\mu_B||^2
= \tfrac{1}{1/|A|+1/|B|} ||\mu_A-\mu_B||^2 \]

 For Lance-Williams, we can then obtain the following recursive definition:
\[d_{\text{Ward}}(A\cup B,C)=\tfrac{|A|+|C|}{|A|+|B|+|C|} d(A,C) +
\tfrac{|B|+|C|}{|A|+|B|+|C|} d(B,C) - \tfrac{|C|}{|A|+|B|+|C|} d(A,B)\]

 These transformations rely on properties of the L2-norm, so they cannot be
used with arbitrary metrics, unless they are equivalent to the L2-norm in
some transformed space.

 Because the resulting distances are squared, when used with a non-squared
distance, ELKI implementations will apply the square root before returning
the final result. This is statistically somewhat questionable, but usually
yields more interpretable distances that — roughly — correspond
to the increase in standard deviation. With ELKI, you can get both behavior:
Either choose squared Euclidean distance, or regular Euclidean distance.

 This method is also referred to as "minimize increase of sum of squares"
(MISSQ) by Podani.

\classinfo%
{de.lmu.ifi.dbs.elki.algorithm.clustering.hierarchical.linkage.}{WeightedAverageLinkage}%
{0.6.0}%
{journals/kansas/SokalM1902}%
{Weighted Average Linkage}%

Weighted average linkage clustering method (WPGMA).

 This is somewhat a misnomer, as it actually ignores that the clusters should
likely be weighted differently according to their size when computing the
average linkage. See {\begingroup\ttfamily{}GroupAverageLinkage\endgroup} for the UPGMA method
that uses the group size to weight the objects the same way.
Because of this, it is sometimes also called "simple average".

 There does not appear to be a closed form distance for this clustering,
but it is only defined recursively on the previous clusters simply by
\[d_{\text{WPGMA}}(A\cup B,C):=\tfrac{1}{2}d(A,C)+\tfrac{1}{2}d(B,C)\]

 {\begingroup\ttfamily{}MedianLinkage\endgroup} (WPGMC) is similar in the sense that it is ignoring
the cluster sizes, and therefore the weight of points decreases with the
depth of the tree. The method is "weighted" in the sense that the new members
get the weight adjusted to match the old cluster members.

\pkginfo{7}%
{de.lmu.ifi.dbs.elki.algorithm.clustering.}{kmeans}%
{Kmeans}%

K-means clustering and variations.

\classinfo%
{de.lmu.ifi.dbs.elki.algorithm.clustering.kmeans.}{BestOfMultipleKMeans}%
{0.6.0}%
{}%
{Best Of Multiple K Means}%

Run K-Means multiple times, and keep the best run.

\classinfo%
{de.lmu.ifi.dbs.elki.algorithm.clustering.kmeans.}{CLARA}%
{0.7.0}%
{doi:10.1002/9780470316801.ch3,doi:10.1016/B978-0-444-87877-9.50039-X}%
{CLARA}%

Clustering Large Applications (CLARA) is a clustering method for large data
sets based on PAM, partitioning around medoids ({\begingroup\ttfamily{}KMedoidsPAM\endgroup}) based on
sampling.

\classinfo%
{de.lmu.ifi.dbs.elki.algorithm.clustering.kmeans.}{CLARANS}%
{0.7.5}%
{DBLP:journals/tkde/NgH02}%
{CLARANS}%

CLARANS: a method for clustering objects for spatial data mining
is inspired by PAM (partitioning around medoids, {\begingroup\ttfamily{}KMedoidsPAM\endgroup})
and CLARA and also based on sampling.

 This implementation tries to balance memory and computation time.
By caching the distances to the two nearest medoids, we usually only need
O(n) instead of O(nk) distance computations for one iteration, at
the cost of needing O(2n) memory to store them.

 The implementation is fairly ugly, because we have three solutions (the best
found so far, the current solution, and a neighbor candidate); and for each
point in each solution we need the best and second best assignments. But with
Java 11, we may be able to switch to value types that would clean this code
significantly, without the overhead of O(n) objects.

\classinfo%
{de.lmu.ifi.dbs.elki.algorithm.clustering.kmeans.}{FastCLARA}%
{0.7.5}%
{DBLP:journals/corr/abs-1810-05691}%
{Fast CLARA}%

Clustering Large Applications (CLARA) with the {\begingroup\ttfamily{}KMedoidsFastPAM\endgroup} improvements, to increase scalability in the number of clusters. This variant
will also default to twice the sample size, to improve quality.

\classinfo%
{de.lmu.ifi.dbs.elki.algorithm.clustering.kmeans.}{FastCLARANS}%
{0.7.5}%
{DBLP:journals/corr/abs-1810-05691}%
{Fast CLARANS}%

A faster variation of CLARANS, that can explore O(k) as many swaps at a
similar cost by considering all medoids for each candidate non-medoid. Since
this means sampling fewer non-medoids, we suggest to increase the subsampling
rate slightly to get higher quality than CLARANS, at better runtime.

\classinfo%
{de.lmu.ifi.dbs.elki.algorithm.clustering.kmeans.}{KMeansAnnulus}%
{0.7.5}%
{doi:10.1007/978-3-319-09259-1_2,mathesis/Drake13}%
{K Means Annulus}%

Annulus k-means algorithm. A variant of Hamerly with an additional bound,
based on comparing the norm of the mean and the norm of the points.

 This implementation could be further improved by precomputing and storing the
norms of all points (at the cost of O(n) memory additionally).

\classinfo%
{de.lmu.ifi.dbs.elki.algorithm.clustering.kmeans.}{KMeansBisecting}%
{0.6.0}%
{conf/kdd/SteinbachKK00}%
{K Means Bisecting}%

The bisecting k-means algorithm works by starting with an initial
partitioning into two clusters, then repeated splitting of the largest
cluster to get additional clusters.

\classinfo%
{de.lmu.ifi.dbs.elki.algorithm.clustering.kmeans.}{KMeansCompare}%
{0.7.1}%
{DBLP:conf/alenex/Phillips02}%
{Compare-Means}%

Compare-Means: Accelerated k-means by exploiting the triangle inequality and
pairwise distances of means to prune candidate means.

\classinfo%
{de.lmu.ifi.dbs.elki.algorithm.clustering.kmeans.}{KMeansElkan}%
{0.7.0}%
{DBLP:conf/icml/Elkan03}%
{K Means Elkan}%

Elkan's fast k-means by exploiting the triangle inequality.

 This variant needs O(n*k) additional memory to store bounds.

 See {\begingroup\ttfamily{}KMeansHamerly\endgroup} for a close variant that only uses O(n*2)
additional memory for bounds.

\classinfo%
{de.lmu.ifi.dbs.elki.algorithm.clustering.kmeans.}{KMeansExponion}%
{0.7.5}%
{DBLP:conf/icml/NewlingF16}%
{K Means Exponion}%

Newlings's exponion k-means algorithm, exploiting the triangle inequality.

 This is not a complete implementation, the approximative sorting part
is missing. We also had to guess on the paper how to make best use of F.

\classinfo%
{de.lmu.ifi.dbs.elki.algorithm.clustering.kmeans.}{KMeansHamerly}%
{0.7.0}%
{DBLP:conf/sdm/Hamerly10}%
{K Means Hamerly}%

Hamerly's fast k-means by exploiting the triangle inequality.

\classinfo%
{de.lmu.ifi.dbs.elki.algorithm.clustering.kmeans.}{KMeansLloyd}%
{0.5.0}%
{journals/biometrics/Forgy65,DBLP:journals/tit/Lloyd82}%
{k-Means (Lloyd/Forgy Algorithm)}%

The standard k-means algorithm, using bulk iterations and commonly attributed
to Lloyd and Forgy (independently).

\classinfo%
{de.lmu.ifi.dbs.elki.algorithm.clustering.kmeans.}{KMeansMacQueen}%
{0.1}%
{conf/bsmsp/MacQueen67}%
{k-Means (MacQueen Algorithm)}%

The original k-means algorithm, using MacQueen style incremental updates;
making this effectively an "online" (streaming) algorithm.

 This implementation will by default iterate over the data set until
convergence, although MacQueen likely only meant to do a single pass over the
data, but the result quality improves with multiple passes.

\classinfo%
{de.lmu.ifi.dbs.elki.algorithm.clustering.kmeans.}{KMeansMinusMinus}%
{0.7.5}%
{DBLP:conf/sdm/ChawlaG13}%
{K-Means--}%

k-means--: A Unified Approach to Clustering and Outlier Detection.

 Similar to Lloyds K-means algorithm, but ignores the farthest points when
updating the means, considering them to be outliers.

\classinfo%
{de.lmu.ifi.dbs.elki.algorithm.clustering.kmeans.}{KMeansSimplifiedElkan}%
{0.7.5}%
{DBLP:conf/icml/NewlingF16}%
{K Means Simplified Elkan}%

Simplified version of Elkan's k-means by exploiting the triangle inequality.

 Compared to {\begingroup\ttfamily{}KMeansElkan\endgroup}, this uses less pruning, but also does not
need to maintain a matrix of pairwise centroid separation.

\classinfo%
{de.lmu.ifi.dbs.elki.algorithm.clustering.kmeans.}{KMeansSort}%
{0.7.1}%
{DBLP:conf/alenex/Phillips02}%
{Sort-Means}%

Sort-Means: Accelerated k-means by exploiting the triangle inequality and
pairwise distances of means to prune candidate means (with sorting).

\classinfo%
{de.lmu.ifi.dbs.elki.algorithm.clustering.kmeans.}{KMediansLloyd}%
{0.5.0}%
{DBLP:conf/nips/BradleyMS96}%
{K Medians Lloyd}%

k-medians clustering algorithm, but using Lloyd-style bulk iterations instead
of the more complicated approach suggested by Kaufman and Rousseeuw (see
{\begingroup\ttfamily{}KMedoidsPAM\endgroup} instead).

\classinfo%
{de.lmu.ifi.dbs.elki.algorithm.clustering.kmeans.}{KMedoidsFastPAM}%
{0.7.5}%
{DBLP:journals/corr/abs-1810-05691}%
{K Medoids Fast PAM}%

FastPAM: An improved version of PAM, that is usually O(k) times faster. This
class incorporates the benefits of {\begingroup\ttfamily{}KMedoidsFastPAM1\endgroup}, but in addition
it tries to perform multiple swaps in each iteration (FastPAM2), which can
reduce the total number of iterations needed substantially for large k, if
some areas of the data are largely independent.

 There is a tolerance parameter, which controls how many additional swaps are
performed. When set to 0, it will only execute an additional swap if it
appears to be independent (i.e., the improvements resulting from the swap
have not decreased when the first swap was executed). We suggest to rather
leave it at the default of 1, which means to perform any additional swap
that gives an improvement. We could not observe a tendency to find worse
results when doing these additional swaps, but a reduced runtime.

 Because of the speed benefits, we also suggest to use a linear-time
initialization, such as the k-means++ initialization or the proposed
LAB (linear approximative BUILD, the third component of FastPAM)
initialization, and try multiple times if the runtime permits.

\classinfo%
{de.lmu.ifi.dbs.elki.algorithm.clustering.kmeans.}{KMedoidsFastPAM1}%
{0.7.5}%
{DBLP:journals/corr/abs-1810-05691}%
{K Medoids Fast PAM1}%

FastPAM1: A version of PAM that is O(k) times faster, i.e., now in O((n-k)²).
The change here feels pretty small - we handle all k medoids in parallel
using an array. But this means the innermost loop only gets executed in
O(1/k) of all iterations, and thus we benefit on average.

 This acceleration gives exactly (assuming perfect numerical
accuracy) the same results as the original PAM. For further improvements that
can affect the result, see also {\begingroup\ttfamily{}KMedoidsFastPAM\endgroup}, which is recommended
for usage in practice.

\classinfo%
{de.lmu.ifi.dbs.elki.algorithm.clustering.kmeans.}{KMedoidsPAM}%
{0.5.0}%
{books/misc/KauRou87,doi:10.1002/9780470316801.ch2}%
{Partioning Around Medoids}%

The original Partitioning Around Medoids (PAM) algorithm or k-medoids
clustering, as proposed by Kaufman and Rousseeuw in "Clustering by means of
Medoids".

\classinfo%
{de.lmu.ifi.dbs.elki.algorithm.clustering.kmeans.}{KMedoidsPAMReynolds}%
{0.7.5}%
{DBLP:journals/jmma/ReynoldsRIR06}%
{K Medoids PAM Reynolds}%

The Partitioning Around Medoids (PAM) algorithm with some additional
optimizations proposed by Reynolds et al.

 In our implementation, we could not observe a substantial improvement over
the original PAM algorithm. This may be because of modern CPU architectures,
where saving an addition may be neglibile compared to caching and pipelining.

\classinfo%
{de.lmu.ifi.dbs.elki.algorithm.clustering.kmeans.}{KMedoidsPark}%
{0.5.0}%
{DBLP:journals/jmma/ReynoldsRIR06,DBLP:journals/eswa/ParkJ09}%
{K Medoids Park}%

A k-medoids clustering algorithm, implemented as EM-style bulk algorithm.

 In contrast to PAM, which will in each iteration update one medoid with one
(arbitrary) non-medoid, this implementation follows the EM pattern. In the
expectation step, the best medoid from the cluster members is chosen; in the
M-step, the objects are reassigned to their nearest medoid.

 This implementation evolved naturally from EM and k-means algorithms, but
apparently a similar approach was published by Park and Jun, and also
Reynolds et al. discussed this kind of approach before as a side note.

 In our experiments, it tends to be much faster than PAM, but also find less
good solutions, as the medoids are only chosen from the cluster members. This
aligns with findings of Reynolds et al. and can be explained with the
requirement of the new medoid to cover the entire cluster.

\classinfo%
{de.lmu.ifi.dbs.elki.algorithm.clustering.kmeans.}{SingleAssignmentKMeans}%
{0.7.0}%
{}%
{Single Assignment K Means}%

Pseudo-k-Means variations, that assigns each object to the nearest center.

\classinfo%
{de.lmu.ifi.dbs.elki.algorithm.clustering.kmeans.}{XMeans}%
{0.7.0}%
{DBLP:conf/icml/PellegM00}%
{X Means}%

X-means: Extending K-means with Efficient Estimation on the Number of
Clusters.

 Note: this implementation does currently not use a k-d-tree for
acceleration. Also note that kmax is not a hard threshold - the algorithm
can return up to 2*kmax clusters!

\pkginfo{8}%
{de.lmu.ifi.dbs.elki.algorithm.clustering.kmeans.}{initialization}%
{Kmeans Initialization}%

Initialization strategies for k-means.

\classinfo%
{de.lmu.ifi.dbs.elki.algorithm.clustering.kmeans.initialization.}{FarthestPointsInitialMeans}%
{0.6.0}%
{}%
{Farthest Points Initial Means}%

K-Means initialization by repeatedly choosing the farthest point (by the
minimum distance to earlier points).

Note: this is less random than other initializations, so running multiple
times will be more likely to return the same local minima.

\classinfo%
{de.lmu.ifi.dbs.elki.algorithm.clustering.kmeans.initialization.}{FarthestSumPointsInitialMeans}%
{0.6.0}%
{}%
{Farthest Sum Points Initial Means}%

K-Means initialization by repeatedly choosing the farthest point (by the
sum of distances to previous objects).

Note: this is less random than other initializations, so running multiple
times will be more likely to return the same local minima.

\classinfo%
{de.lmu.ifi.dbs.elki.algorithm.clustering.kmeans.initialization.}{FirstKInitialMeans}%
{0.5.0}%
{conf/bsmsp/MacQueen67}%
{First K Initial Means}%

Initialize K-means by using the first k objects as initial means.

\classinfo%
{de.lmu.ifi.dbs.elki.algorithm.clustering.kmeans.initialization.}{KMeansPlusPlusInitialMeans}%
{0.5.0}%
{DBLP:conf/soda/ArthurV07}%
{K Means Plus Plus Initial Means}%

K-Means++ initialization for k-means.

\classinfo%
{de.lmu.ifi.dbs.elki.algorithm.clustering.kmeans.initialization.}{LABInitialMeans}%
{0.7.5}%
{DBLP:journals/corr/abs-1810-05691}%
{LAB Initial Means}%

Linear approximative BUILD (LAB) initialization for FastPAM (and k-means).

 This is a O(nk) aproximation of the original PAM BUILD. For performance, it
uses an O(sqrt(n)) sample to achieve linear run time. The results will be
worse than those of BUILD, but provide a good starting point for FastPAM
optimization.

\classinfo%
{de.lmu.ifi.dbs.elki.algorithm.clustering.kmeans.initialization.}{OstrovskyInitialMeans}%
{0.7.5}%
{DBLP:conf/focs/OstrovskyRSS062,DBLP:journals/jacm/OstrovskyRSS12}%
{Ostrovsky Initial Means}%

Ostrovsky initial means, a variant of k-means++ that is expected to give
slightly better results on average, but only works for k-means and not for,
e.g., PAM (k-medoids).

\classinfo%
{de.lmu.ifi.dbs.elki.algorithm.clustering.kmeans.initialization.}{PAMInitialMeans}%
{0.5.0}%
{books/misc/KauRou87,doi:10.1002/9780470316801.ch2}%
{PAM Initial Means}%

PAM initialization for k-means (and of course, for PAM).

\classinfo%
{de.lmu.ifi.dbs.elki.algorithm.clustering.kmeans.initialization.}{ParkInitialMeans}%
{0.7.5}%
{DBLP:journals/eswa/ParkJ09}%
{Park Initial Means}%

Initialization method proposed by Park and Jun.

 It is easy to imagine that this approach can become problematic, because it
does not take the distances between medoids into account. In the worst case,
it may choose k duplicates as initial centers, therefore we cannot recommend
this strategy, but it is provided for completeness.

\classinfo%
{de.lmu.ifi.dbs.elki.algorithm.clustering.kmeans.initialization.}{PredefinedInitialMeans}%
{0.7.0}%
{}%
{Predefined Initial Means}%

Run k-means with prespecified initial means.

\classinfo%
{de.lmu.ifi.dbs.elki.algorithm.clustering.kmeans.initialization.}{RandomNormalGeneratedInitialMeans}%
{0.7.5}%
{doi:10.1071/BT9660127}%
{Random Normal Generated Initial Means}%

Initialize k-means by generating random vectors (normal distributed
with \(N(\mu,\sigma)\) in each dimension).

 This is a different interpretation of the work of Jancey, who wrote little
more details but "introduced into known but arbitrary positions"; but
seemingly worked with standardized scores. In contrast to
{\begingroup\ttfamily{}RandomUniformGeneratedInitialMeans\endgroup} (which uses a uniform on the entire
value range), this class uses a normal distribution based on the estimated
parameters. The resulting means should be more central, and thus a bit less
likely to become empty (at least if you assume there is no correlation
amongst attributes... it is still not competitive with better methods).

 Warning: this still tends to produce empty clusters in many
situations, and is one of the least effective initialization strategies, not
recommended for use.

\classinfo%
{de.lmu.ifi.dbs.elki.algorithm.clustering.kmeans.initialization.}{RandomUniformGeneratedInitialMeans}%
{0.5.0}%
{doi:10.1071/BT9660127}%
{Random Uniform Generated Initial Means}%

Initialize k-means by generating random vectors (uniform, within the value
range of the data set).

 This is attributed to Jancey, but who wrote little more details but
"introduced into known but arbitrary positions". This class assumes this
refers to uniform positions within the value domain. For a normal distributed
variant, see {\begingroup\ttfamily{}RandomNormalGeneratedInitialMeans\endgroup}.

 Warning: this tends to produce empty clusters, and is one of the least
effective initialization strategies, not recommended for use.

\classinfo%
{de.lmu.ifi.dbs.elki.algorithm.clustering.kmeans.initialization.}{RandomlyChosenInitialMeans}%
{0.5.0}%
{journals/biometrics/Forgy65,journals/misc/McRae71,books/academic/Anderberg73/Ch7}%
{Randomly Chosen Initial Means}%

Initialize K-means by randomly choosing k existing elements as initial
cluster centers.

\classinfo%
{de.lmu.ifi.dbs.elki.algorithm.clustering.kmeans.initialization.}{SampleKMeansInitialization}%
{0.6.0}%
{DBLP:conf/icml/BradleyF98}%
{Sample K Means Initialization}%

Initialize k-means by running k-means on a sample of the data set only.

\pkginfo{8}%
{de.lmu.ifi.dbs.elki.algorithm.clustering.kmeans.}{parallel}%
{Kmeans Parallel}%

Parallelized implementations of k-means.

\classinfo%
{de.lmu.ifi.dbs.elki.algorithm.clustering.kmeans.parallel.}{ParallelLloydKMeans}%
{0.7.0}%
{}%
{Parallel Lloyd K Means}%

Parallel implementation of k-Means clustering.

\pkginfo{8}%
{de.lmu.ifi.dbs.elki.algorithm.clustering.kmeans.}{quality}%
{Kmeans Quality}%

Quality measures for k-Means results.

\classinfo%
{de.lmu.ifi.dbs.elki.algorithm.clustering.kmeans.quality.}{AbstractKMeansQualityMeasure}%
{0.7.0}%
{DBLP:conf/icml/PellegM00,DBLP:conf/ictai/ZhaoXF08}%
{Abstract K Means Quality Measure}%

Base class for evaluating clusterings by information criteria (such as AIC or
BIC). Provides helper functions (e.g. max likelihood calculation) to its
subclasses.

\classinfo%
{de.lmu.ifi.dbs.elki.algorithm.clustering.kmeans.quality.}{AkaikeInformationCriterion}%
{0.7.0}%
{DBLP:conf/icml/PellegM00,conf/isit/Akaike73}%
{Akaike Information Criterion}%

Akaike Information Criterion (AIC).

\classinfo%
{de.lmu.ifi.dbs.elki.algorithm.clustering.kmeans.quality.}{BayesianInformationCriterion}%
{0.7.0}%
{doi:10.1214/aos/1176344136}%
{Bayesian Information Criterion}%

Bayesian Information Criterion (BIC), also known as Schwarz criterion (SBC,
SBIC) for the use with evaluating k-means results.

\classinfo%
{de.lmu.ifi.dbs.elki.algorithm.clustering.kmeans.quality.}{BayesianInformationCriterionZhao}%
{0.7.0}%
{DBLP:conf/ictai/ZhaoXF08}%
{Bayesian Information Criterion Zhao}%

Different version of the BIC criterion.

\classinfo%
{de.lmu.ifi.dbs.elki.algorithm.clustering.kmeans.quality.}{WithinClusterMeanDistanceQualityMeasure}%
{0.6.0}%
{}%
{Within Cluster Mean Distance Quality Measure}%

Class for computing the average overall distance.

The average of all average pairwise distances in a cluster.

\classinfo%
{de.lmu.ifi.dbs.elki.algorithm.clustering.kmeans.quality.}{WithinClusterVarianceQualityMeasure}%
{0.6.0}%
{}%
{Within Cluster Variance Quality Measure}%

Class for computing the variance in a clustering result (sum-of-squares).

\pkginfo{7}%
{de.lmu.ifi.dbs.elki.algorithm.clustering.}{meta}%
{Meta}%

Meta clustering algorithms, that get their result from other clusterings or external sources.

\classinfo%
{de.lmu.ifi.dbs.elki.algorithm.clustering.meta.}{ExternalClustering}%
{0.7.0}%
{}%
{External Clustering}%

Read an external clustering result from a file, such as produced by
{\begingroup\ttfamily{}ClusteringVectorDumper\endgroup}.

 The input format of this parser is text-based:

\begin{Verbatim}[fontsize=\verbatimsize]
# Optional comment
1 1 1 2 2 2 -1 Example label
\end{Verbatim}
 Where non-negative numbers are cluster assignments, negative numbers are
considered noise clusters.

\pkginfo{7}%
{de.lmu.ifi.dbs.elki.algorithm.clustering.}{onedimensional}%
{Onedimensional}%

Clustering algorithms for one-dimensional data.

\classinfo%
{de.lmu.ifi.dbs.elki.algorithm.clustering.onedimensional.}{KNNKernelDensityMinimaClustering}%
{0.6.0}%
{}%
{KNN Kernel Density Minima Clustering}%

Cluster one-dimensional data by splitting the data set on local minima after
performing kernel density estimation.

\pkginfo{7}%
{de.lmu.ifi.dbs.elki.algorithm.clustering.}{optics}%
{Optics}%

OPTICS family of clustering algorithms.

 Note that some OPTICS based algorithms (HiCO, DiSH) are in the
subspace and correlation packages, which better describes their use case.

\classinfo%
{de.lmu.ifi.dbs.elki.algorithm.clustering.optics.}{AbstractOPTICS}%
{0.7.0}%
{DBLP:conf/sigmod/AnkerstBKS99}%
{Abstract OPTICS}%

The OPTICS algorithm for density-based hierarchical clustering.

 This is the abstract base class, providing the shared parameters only.

\classinfo%
{de.lmu.ifi.dbs.elki.algorithm.clustering.optics.}{DeLiClu}%
{0.1}%
{DBLP:conf/pakdd/AchtertBK06}%
{DeliClu: Density-Based Hierarchical Clustering}%

DeliClu: Density-Based Hierarchical Clustering

 A hierarchical algorithm to find density-connected sets in a database,
closely related to OPTICS but exploiting the structure of a R-tree for
acceleration.

\classinfo%
{de.lmu.ifi.dbs.elki.algorithm.clustering.optics.}{FastOPTICS}%
{0.7.0}%
{DBLP:conf/cikm/SchneiderV13}%
{Fast OPTICS}%

FastOPTICS algorithm (Fast approximation of OPTICS)

 Note that this is not FOPTICS as in "Fuzzy OPTICS"!

\classinfo%
{de.lmu.ifi.dbs.elki.algorithm.clustering.optics.}{OPTICSHeap}%
{0.1}%
{DBLP:conf/sigmod/AnkerstBKS99}%
{OPTICS: Density-Based Hierarchical Clustering (implementation using a heap)}%

The OPTICS algorithm for density-based hierarchical clustering.

 Algorithm to find density-connected sets in a database based on the
parameters 'minPts' and 'epsilon' (specifying a volume). These two parameters
determine a density threshold for clustering.

 This implementation uses a heap.

\classinfo%
{de.lmu.ifi.dbs.elki.algorithm.clustering.optics.}{OPTICSList}%
{0.7.0}%
{DBLP:conf/sigmod/AnkerstBKS99}%
{OPTICS: Density-Based Hierarchical Clustering (implementation using a list)}%

The OPTICS algorithm for density-based hierarchical clustering.

 Algorithm to find density-connected sets in a database based on the
parameters 'minPts' and 'epsilon' (specifying a volume). These two parameters
determine a density threshold for clustering.

 This version is implemented using a list, always scanning the list for the
maximum. While this could be cheaper than the complex heap updates,
benchmarks indicate the heap version is usually still preferable.

\classinfo%
{de.lmu.ifi.dbs.elki.algorithm.clustering.optics.}{OPTICSXi}%
{0.7.0}%
{DBLP:conf/sigmod/AnkerstBKS99,DBLP:conf/lwa/SchubertG18}%
{OPTICS Xi Cluster Extraction}%

Extract clusters from OPTICS Plots using the original Xi extraction.

 Note: this implementation includes an additional filter step that prunes
elements from a steep up area that don't have the predecessor in the cluster.
This removes a popular type of artifacts.

\pkginfo{7}%
{de.lmu.ifi.dbs.elki.algorithm.clustering.}{subspace}%
{Subspace}%

Axis-parallel subspace clustering algorithms

 The clustering algorithms in this package are instances of both, projected
clustering algorithms or subspace clustering algorithms according to the
classical but somewhat obsolete classification schema of clustering
algorithms for axis-parallel subspaces.

\classinfo%
{de.lmu.ifi.dbs.elki.algorithm.clustering.subspace.}{CLIQUE}%
{0.1}%
{DBLP:conf/sigmod/AgrawalGGR98}%
{CLIQUE: Automatic Subspace Clustering of High Dimensional Data for Data Mining Applications}%

Implementation of the CLIQUE algorithm, a grid-based algorithm to identify
dense clusters in subspaces of maximum dimensionality.

 The implementation consists of two steps:\\
 1. Identification of subspaces that contain clusters \\
 2. Identification of clusters

 The third step of the original algorithm (Generation of minimal description
for the clusters) is not (yet) implemented.

 Note: this is fairly old code, and not well optimized.
Do not use this for runtime benchmarking!

\classinfo%
{de.lmu.ifi.dbs.elki.algorithm.clustering.subspace.}{DOC}%
{0.6.0}%
{DBLP:conf/sigmod/ProcopiucJAM02}%
{DOC: Density-based Optimal projective Clustering}%

DOC is a sampling based subspace clustering algorithm.

\classinfo%
{de.lmu.ifi.dbs.elki.algorithm.clustering.subspace.}{DiSH}%
{0.1}%
{DBLP:conf/dasfaa/AchtertBKKMZ07}%
{DiSH: Detecting Subspace cluster Hierarchies}%

Algorithm for detecting subspace hierarchies.

\classinfo%
{de.lmu.ifi.dbs.elki.algorithm.clustering.subspace.}{FastDOC}%
{0.7.5}%
{DBLP:conf/sigmod/ProcopiucJAM02}%
{FastDOC: Density-based Optimal projective Clustering}%

The heuristic variant of the DOC algorithm, FastDOC

\classinfo%
{de.lmu.ifi.dbs.elki.algorithm.clustering.subspace.}{HiSC}%
{0.1}%
{DBLP:conf/pkdd/AchtertBKKMZ06}%
{Finding Hierarchies of Subspace Clusters}%

Implementation of the HiSC algorithm, an algorithm for detecting hierarchies
of subspace clusters.

\classinfo%
{de.lmu.ifi.dbs.elki.algorithm.clustering.subspace.}{P3C}%
{0.6.0}%
{DBLP:conf/icdm/MoiseSE06}%
{P3C: A Robust Projected Clustering Algorithm.}%

P3C: A Robust Projected Clustering Algorithm.

\classinfo%
{de.lmu.ifi.dbs.elki.algorithm.clustering.subspace.}{PROCLUS}%
{0.1}%
{doi:10.1145/304181.304188}%
{PROCLUS: PROjected CLUStering}%

The PROCLUS algorithm, an algorithm to find subspace clusters in high
dimensional spaces.

\classinfo%
{de.lmu.ifi.dbs.elki.algorithm.clustering.subspace.}{PreDeCon}%
{0.1}%
{DBLP:conf/icdm/BohmKKK04}%
{PreDeCon: Subspace Preference weighted Density Connected Clustering}%

PreDeCon computes clusters of subspace preference weighted connected points.
The algorithm searches for local subgroups of a set of feature vectors having
a low variance along one or more (but not all) attributes.

\classinfo%
{de.lmu.ifi.dbs.elki.algorithm.clustering.subspace.}{SUBCLU}%
{0.3}%
{DBLP:conf/sdm/KroegerKK04}%
{SUBCLU: Density connected Subspace Clustering}%

Implementation of the SUBCLU algorithm, an algorithm to detect arbitrarily
shaped and positioned clusters in subspaces. SUBCLU delivers for each
subspace the same clusters DBSCAN would have found, when applied to this
subspace separately.

\pkginfo{7}%
{de.lmu.ifi.dbs.elki.algorithm.clustering.}{uncertain}%
{Uncertain}%

Clustering algorithms for uncertain data.

\classinfo%
{de.lmu.ifi.dbs.elki.algorithm.clustering.uncertain.}{CKMeans}%
{0.7.0}%
{DBLP:conf/icdm/LeeKC07}%
{CK Means}%

Run k-means on the centers of each uncertain object.

 This is a baseline reference method, that computes the center of mass
(centroid) of each object, then runs k-means on this.

\classinfo%
{de.lmu.ifi.dbs.elki.algorithm.clustering.uncertain.}{CenterOfMassMetaClustering}%
{0.7.0}%
{DBLP:journals/pvldb/SchubertKEZSZ15}%
{Center Of Mass Meta Clustering}%

Center-of-mass meta clustering reduces uncertain objects to their center of
mass, then runs a vector-oriented clustering algorithm on this data set.

\classinfo%
{de.lmu.ifi.dbs.elki.algorithm.clustering.uncertain.}{FDBSCAN}%
{0.7.0}%
{DBLP:conf/kdd/KriegelP05}%
{FDBSCAN: Density-based Clustering of Applications with Noise on fuzzy objects}%

FDBSCAN is an adaption of DBSCAN for fuzzy (uncertain) objects.

 This implementation is based on GeneralizedDBSCAN. All implementation of
FDBSCAN functionality is located in the neighbor predicate
{\begingroup\ttfamily{}FDBSCANNeighborPredicate\endgroup}.

\classinfo%
{de.lmu.ifi.dbs.elki.algorithm.clustering.uncertain.}{FDBSCANNeighborPredicate}%
{0.7.0}%
{DBLP:conf/kdd/KriegelP05}%
{FDBSCAN Neighbor Predicate}%

Density-based Clustering of Applications with Noise and Fuzzy objects
(FDBSCAN) is an Algorithm to find sets in a fuzzy database that are
density-connected with minimum probability.

\classinfo%
{de.lmu.ifi.dbs.elki.algorithm.clustering.uncertain.}{RepresentativeUncertainClustering}%
{0.7.0}%
{DBLP:conf/kdd/ZufleESMZR14}%
{Representative Uncertain Clustering}%

Representative clustering of uncertain data.

 This algorithm clusters uncertain data by repeatedly sampling a possible
world, then running a traditional clustering algorithm on this sample.

 The resulting "possible" clusterings are then clustered themselves, using a
clustering similarity measure. This yields a number of representatives for
the set of all possible worlds.

\classinfo%
{de.lmu.ifi.dbs.elki.algorithm.clustering.uncertain.}{UKMeans}%
{0.7.0}%
{DBLP:conf/pakdd/ChauCKN06}%
{UK Means}%

Uncertain K-Means clustering, using the average deviation from the center.

 Note: this method is, essentially, superficial. It was shown to be equivalent
to doing regular K-means on the object centroids instead (see {\begingroup\ttfamily{}CKMeans\endgroup} for the reference and an implementation). This is only for completeness.

\pkginfo{6}%
{de.lmu.ifi.dbs.elki.algorithm.}{itemsetmining}%
{Itemsetmining}%

Algorithms for frequent itemset mining such as APRIORI.

\classinfo%
{de.lmu.ifi.dbs.elki.algorithm.itemsetmining.}{APRIORI}%
{0.1}%
{DBLP:conf/vldb/AgrawalS94}%
{APRIORI: Algorithm for Mining Association Rules}%

The APRIORI algorithm for Mining Association Rules.

\classinfo%
{de.lmu.ifi.dbs.elki.algorithm.itemsetmining.}{Eclat}%
{0.7.0}%
{DBLP:conf/kdd/ZakiPOL97}%
{Eclat}%

Eclat is a depth-first discovery algorithm for mining frequent itemsets.

 Eclat discovers frequent itemsets by first transforming the data into a
(sparse) column-oriented form, then performing a depth-first traversal of the
prefix lattice, stopping traversal when the minimum support is no longer
satisfied.

 This implementation is the basic algorithm only, and does not use diffsets.
Columns are represented using a sparse representation, which theoretically is
beneficial when the density is less than 1/31. This corresponds roughly to a
minimum support of 3\% for 1-itemsets. When searching for itemsets with a
larger minimum support, it may be desirable to use a dense bitset
representation instead and/or implement an automatic switching technique!

 Performance of this implementation is probably surpassed with a low-level C
implementation based on SIMD bitset operations as long as support of an
itemset is high, which are not easily accessible in Java.

\classinfo%
{de.lmu.ifi.dbs.elki.algorithm.itemsetmining.}{FPGrowth}%
{0.7.0}%
{DBLP:conf/sigmod/HanPY00}%
{FP Growth}%

FP-Growth is an algorithm for mining the frequent itemsets by using a
compressed representation of the database called {\begingroup\ttfamily{}FPGrowth.FPTree\endgroup}.

 FP-Growth first sorts items by the overall frequency, since having high
frequent items appear first in the tree leads to a much smaller tree since
frequent subsets will likely share the same path in the tree. FP-Growth is
beneficial when you have a lot of (near-) duplicate transactions, and are
using a not too high support threshold, as it only prunes single items, not
item combinations.

 This implementation is in-memory only, and has not yet been carefully
optimized.

 The worst case memory use probably is \(O(\min(n\cdot l,i^l))\) where i is the
number of items, l the average itemset length, and n the number of items. The
worst case scenario is when every item is frequent, and every transaction is
unique. The resulting tree will then be larger than the original data.

\pkginfo{7}%
{de.lmu.ifi.dbs.elki.algorithm.itemsetmining.}{associationrules}%
{Associationrules}%

Association rule mining.

\classinfo%
{de.lmu.ifi.dbs.elki.algorithm.itemsetmining.associationrules.}{AssociationRuleGeneration}%
{0.7.5}%
{DBLP:books/cu/ZM2014}%
{Association Rule Generation}%

Association rule generation from frequent itemsets

 This algorithm calls a specified frequent itemset algorithm
and calculates all association rules, having a interest value between
then the specified boundaries form the obtained frequent itemsets

\pkginfo{8}%
{de.lmu.ifi.dbs.elki.algorithm.itemsetmining.associationrules.}{interest}%
{Associationrules Interest}%

Association rule interestingness measures.

 Much of the confusion with these measures arises from the anti-monotonicity
of itemsets, which are omnipresent in the literature.

 In the itemset notation, the itemset \(X\) denotes the set of matching
transactions \(\{T|X\subseteq T\}\) that contain the itemset \(X\).
If we enlarge \(Z=X\cup Y\), the resulting set shrinks:
\(\{T|Z\subseteq T\}=\{T|X\subseteq T\}\cap\{T|Y\subseteq T\}\).

 Because of this: \(\text{support}(X\cup Y) = P(X \cap Y)\)
and \(\text{support}(X\cap Y) = P(X \cup Y)\). With "support" and
"confidence", it is common to see the reversed semantics (the union on the
constraints is the intersection on the matches, and conversely); with
probabilities it is common to use "events" as in frequentist inference.

 To make things worse, the "support" is sometimes in absolute (integer)
counts, and sometimes used in a relative share.

\classinfo%
{de.lmu.ifi.dbs.elki.algorithm.itemsetmining.associationrules.interest.}{AddedValue}%
{0.7.5}%
{DBLP:conf/dmkdttt/SaharM99}%
{Added Value}%

Added value (AV) interestingness measure:
\( \text{confidence}(X \rightarrow Y) - \text{support}(Y) = P(Y|X)-P(Y) \).

\classinfo%
{de.lmu.ifi.dbs.elki.algorithm.itemsetmining.associationrules.interest.}{CertaintyFactor}%
{0.7.5}%
{DBLP:journals/ida/GalianoBSM02}%
{Certainty Factor}%

Certainty factor (CF; Loevinger) interestingness measure.
\( \tfrac{\text{confidence}(X \rightarrow Y) -
\text{support}(Y)}{\text{support}(\neg Y)} \).

\classinfo%
{de.lmu.ifi.dbs.elki.algorithm.itemsetmining.associationrules.interest.}{Confidence}%
{0.7.5}%
{DBLP:conf/sigmod/AgrawalIS93}%
{Confidence}%

Confidence interestingness measure,
\( \tfrac{\text{support}(X \cup Y)}{\text{support}(X)}
= \tfrac{P(X \cap Y)}{P(X)}=P(Y|X) \).

\classinfo%
{de.lmu.ifi.dbs.elki.algorithm.itemsetmining.associationrules.interest.}{Conviction}%
{0.7.5}%
{DBLP:conf/sigmod/BrinMUT97}%
{Conviction}%

Conviction interestingness measure:
\(\frac{P(X) P(\neg Y)}{P(X\cap\neg Y)}\).

\classinfo%
{de.lmu.ifi.dbs.elki.algorithm.itemsetmining.associationrules.interest.}{Cosine}%
{0.7.5}%
{tr/umn/TanK00}%
{Cosine}%

Cosine interestingness measure,
\(\tfrac{\text{support}(A\cup B)}{\sqrt{\text{support}(A)\text{support}(B)}}
=\tfrac{P(A\cap B)}{\sqrt{P(A)P(B)}}\).

 The interestingness measure called IS by Tan and Kumar.

\classinfo%
{de.lmu.ifi.dbs.elki.algorithm.itemsetmining.associationrules.interest.}{GiniIndex}%
{0.7.5}%
{tr/umn/TanK00,books/wa/BreimanFOS84}%
{Gini Index}%

Gini-index based interestingness measure, using the weighted squared
conditional probabilities compared to the non-conditional priors.
 \[ P(X)\left(P(Y|X)^2+P(\neg Y|X)^2\right)
+ P(\neg X)\left(P(Y|\neg X)^2+P(\neg Y|\neg X)^2\right)
- P(Y)^2 - P(\neg Y)^2 \]

\classinfo%
{de.lmu.ifi.dbs.elki.algorithm.itemsetmining.associationrules.interest.}{JMeasure}%
{0.7.5}%
{DBLP:books/mit/PF91/SmythG91}%
{J Measure}%

J-Measure interestingness measure.
\(P(X\cap Y)\log\tfrac{P(Y|X)}{P(Y)}
+ P(X\cap \neg Y)\log\tfrac{P(\neg Y|X)}{P(\neg Y)}\).

\classinfo%
{de.lmu.ifi.dbs.elki.algorithm.itemsetmining.associationrules.interest.}{Jaccard}%
{0.7.5}%
{DBLP:books/bu/Rijsbergen79,DBLP:journals/is/TanKS04}%
{Jaccard}%

Jaccard interestingness measure:
 \[\tfrac{\text{support}(A \cup B)}{\text{support}(A \cap B)}
=\tfrac{P(A \cap B)}{P(A)+P(B)-P(A \cap B)}
=\tfrac{P(A \cap B)}{P(A \cup B)}\]

\classinfo%
{de.lmu.ifi.dbs.elki.algorithm.itemsetmining.associationrules.interest.}{Klosgen}%
{0.7.5}%
{DBLP:books/mit/fayyadPSU96/Klosgen96}%
{Klosgen}%

Klösgen interestingness measure.
 \[ \sqrt{\text{support}(X\cup Y)}
\left(\text{confidence}(X \rightarrow Y) - \text{support}(Y)\right)
= \sqrt{P(X\cap Y)}\left(P(Y|X)-P(Y)\right) \]

\classinfo%
{de.lmu.ifi.dbs.elki.algorithm.itemsetmining.associationrules.interest.}{Leverage}%
{0.7.5}%
{DBLP:books/mit/PF91/Piatetsky91}%
{Leverage}%

Leverage interestingness measure.
 \[ \text{support}(X\Rightarrow Y)-\text{support}(X)\text{support}(Y)
=P(X\cap Y)-P(X)P(Y) \]

\classinfo%
{de.lmu.ifi.dbs.elki.algorithm.itemsetmining.associationrules.interest.}{Lift}%
{0.7.5}%
{DBLP:conf/sigmod/BrinMS97}%
{Lift}%

Lift interestingness measure.
 \[ \tfrac{\text{confidence}(X\rightarrow Y)}{\text{support}(Y)}
= \tfrac{\text{confidence}(Y\rightarrow X)}{\text{support}(X)}
= \tfrac{P(X\cap Y)}{P(X)P(Y)} \]

\pkginfo{6}%
{de.lmu.ifi.dbs.elki.algorithm.}{outlier}%
{Outlier}%

Outlier detection algorithms

\classinfo%
{de.lmu.ifi.dbs.elki.algorithm.outlier.}{COP}%
{0.5.5}%
{DBLP:conf/icdm/KriegelKSZ12}%
{COP: Correlation Outlier Probability}%

Correlation outlier probability: Outlier Detection in Arbitrarily Oriented
Subspaces

\classinfo%
{de.lmu.ifi.dbs.elki.algorithm.outlier.}{DWOF}%
{0.6.0}%
{DBLP:conf/ibpria/MomtazMG13}%
{DWOF: Dynamic Window Outlier Factor}%

Algorithm to compute dynamic-window outlier factors in a database based on a
specified parameter k, which specifies the number of the neighbors to be
considered during the calculation of the DWOF score.

\classinfo%
{de.lmu.ifi.dbs.elki.algorithm.outlier.}{GaussianModel}%
{0.3}%
{}%
{Gaussian Model Outlier Detection}%

Outlier detection based on the probability density of the single normal
distribution.

\classinfo%
{de.lmu.ifi.dbs.elki.algorithm.outlier.}{GaussianUniformMixture}%
{0.3}%
{DBLP:conf/icml/Eskin00}%
{Gaussian-Uniform Mixture Model Outlier Detection}%

Outlier detection algorithm using a mixture model approach. The data is
modeled as a mixture of two distributions, a Gaussian distribution for
ordinary data and a uniform distribution for outliers. At first all Objects
are in the set of normal objects and the set of anomalous objects is empty.
An iterative procedure then transfers objects from the ordinary set to the
anomalous set if the transfer increases the overall likelihood of the data.

\classinfo%
{de.lmu.ifi.dbs.elki.algorithm.outlier.}{OPTICSOF}%
{0.3}%
{DBLP:conf/pkdd/BreunigKNS99}%
{OPTICS-OF: Identifying Local Outliers}%

OPTICS-OF outlier detection algorithm, an algorithm to find Local Outliers in
a database based on ideas from {\begingroup\ttfamily{}OPTICSTypeAlgorithm\endgroup} clustering.

\classinfo%
{de.lmu.ifi.dbs.elki.algorithm.outlier.}{SimpleCOP}%
{0.5.5}%
{phd/dnb/Zimek08/Ch18}%
{Simple COP: Correlation Outlier Probability}%

Algorithm to compute local correlation outlier probability.

 This is the simpler, original version of COP, as published in

\pkginfo{7}%
{de.lmu.ifi.dbs.elki.algorithm.outlier.}{anglebased}%
{Anglebased}%

Angle-based outlier detection algorithms.

\classinfo%
{de.lmu.ifi.dbs.elki.algorithm.outlier.anglebased.}{ABOD}%
{0.2}%
{DBLP:conf/kdd/KriegelSZ08}%
{ABOD: Angle-Based Outlier Detection}%

Angle-Based Outlier Detection / Angle-Based Outlier Factor.

 Outlier detection using variance analysis on angles, especially for high
dimensional data sets. Exact version, which has cubic runtime (see also
{\begingroup\ttfamily{}FastABOD\endgroup} and {\begingroup\ttfamily{}LBABOD\endgroup} for faster versions).

\classinfo%
{de.lmu.ifi.dbs.elki.algorithm.outlier.anglebased.}{FastABOD}%
{0.6.0}%
{DBLP:conf/kdd/KriegelSZ08}%
{Approximate ABOD: Angle-Based Outlier Detection}%

Fast-ABOD (approximateABOF) version of
Angle-Based Outlier Detection / Angle-Based Outlier Factor.

 Note: the minimum k is 3. The 2 nearest neighbors yields one 1 angle, which
implies a constant 0 variance everywhere.

\classinfo%
{de.lmu.ifi.dbs.elki.algorithm.outlier.anglebased.}{LBABOD}%
{0.6.0}%
{DBLP:conf/kdd/KriegelSZ08}%
{LB-ABOD: Lower Bounded Angle-Based Outlier Detection}%

LB-ABOD (lower-bound) version of
Angle-Based Outlier Detection / Angle-Based Outlier Factor.

 Exact on the top k outliers, approximate on the remaining.

 Outlier detection using variance analysis on angles, especially for high
dimensional data sets.

\pkginfo{7}%
{de.lmu.ifi.dbs.elki.algorithm.outlier.}{clustering}%
{Clustering}%

Clustering based outlier detection.

\classinfo%
{de.lmu.ifi.dbs.elki.algorithm.outlier.clustering.}{CBLOF}%
{0.7.5}%
{DBLP:journals/prl/HeXD03}%
{Discovering cluster-based local outliers}%

Cluster-based local outlier factor (CBLOF).

\classinfo%
{de.lmu.ifi.dbs.elki.algorithm.outlier.clustering.}{EMOutlier}%
{0.3}%
{}%
{EM Outlier: Outlier Detection based on the generic EM clustering}%

Outlier detection algorithm using EM Clustering.

 If an object does not belong to any cluster it is supposed to be an outlier.
If the probability for an object to belong to the most probable cluster is
still relatively low this object is an outlier.

\classinfo%
{de.lmu.ifi.dbs.elki.algorithm.outlier.clustering.}{KMeansOutlierDetection}%
{0.7.0}%
{}%
{K Means Outlier Detection}%

Outlier detection by using k-means clustering.

 The scores are assigned by the objects distance to the nearest center.

 We don't have a clear reference for this approach, but it seems to be a best
practise in some areas to remove objects that have the largest distance from
their center. If you need to cite this approach, please cite the ELKI version
you used (use the \href{https://elki-project.github.io/publications}{ELKI
publication list} for citation information and BibTeX templates).

\classinfo%
{de.lmu.ifi.dbs.elki.algorithm.outlier.clustering.}{SilhouetteOutlierDetection}%
{0.7.0}%
{doi:10.1016/0377-04278790125-7}%
{Silhouette Outlier Detection}%

Outlier detection by using the Silhouette Coefficients.

 Silhouette values are computed as by Rousseeuw and then used as outlier
scores. To cite this outlier detection approach,
please cite the ELKI version you used (use the
\href{https://elki-project.github.io/publications}{ELKI publication
list} for citation information and BibTeX templates).

\pkginfo{7}%
{de.lmu.ifi.dbs.elki.algorithm.outlier.}{distance}%
{Distance}%

Distance-based outlier detection algorithms, such as DBOutlier and kNN.

For methods based on local density, see package
{\begingroup\ttfamily{}de.lmu.ifi.dbs.elki.algorithm.outlier.lof\endgroup} instead.

\classinfo%
{de.lmu.ifi.dbs.elki.algorithm.outlier.distance.}{AbstractDBOutlier}%
{0.3}%
{DBLP:conf/vldb/KnorrN98}%
{Abstract DB Outlier}%

Simple distance based outlier detection algorithms.

\classinfo%
{de.lmu.ifi.dbs.elki.algorithm.outlier.distance.}{DBOutlierDetection}%
{0.3}%
{DBLP:conf/vldb/KnorrN98}%
{DBOD: Distance Based Outlier Detection}%

Simple distanced based outlier detection algorithm. User has to specify two
parameters An object is flagged as an outlier if at least a fraction p of all
data objects has a distance above d from c.

\classinfo%
{de.lmu.ifi.dbs.elki.algorithm.outlier.distance.}{DBOutlierScore}%
{0.3}%
{DBLP:conf/vldb/KnorrN98}%
{Distance Based Outlier Score}%

Compute percentage of neighbors in the given neighborhood with size d.

 Generalization of the DB Outlier Detection by using the fraction as outlier
score thus eliminating this parameter and turning the method into a ranking
method instead of a labelling one.

\classinfo%
{de.lmu.ifi.dbs.elki.algorithm.outlier.distance.}{HilOut}%
{0.5.0}%
{DBLP:conf/pkdd/AngiulliP02}%
{Fast Outlier Detection in High Dimensional Spaces}%

Fast Outlier Detection in High Dimensional Spaces

 Outlier Detection using Hilbert space filling curves

\classinfo%
{de.lmu.ifi.dbs.elki.algorithm.outlier.distance.}{KNNDD}%
{0.7.5}%
{conf/asci/deRidderTD98}%
{KNNDD: k-Nearest Neighbor Data Description}%

Nearest Neighbor Data Description.

 A variation inbetween of KNN outlier and LOF, comparing the nearest neighbor
distance of a point to the nearest neighbor distance of the nearest neighbor.

 The initial description used k=1, where this equation makes most sense.
For k > 1, one may want to use averaging similar to LOF.

\classinfo%
{de.lmu.ifi.dbs.elki.algorithm.outlier.distance.}{KNNOutlier}%
{0.3}%
{DBLP:conf/sigmod/RamaswamyRS00}%
{KNN outlier: Efficient Algorithms for Mining Outliers from Large Data Sets}%

Outlier Detection based on the distance of an object to its k nearest
neighbor.

 This implementation differs from the original pseudocode: the k nearest
neighbors do not exclude the point that is currently evaluated. I.e. for k=1
the resulting score is the distance to the 1-nearest neighbor that is not the
query point and therefore should match k=2 in the exact pseudocode - a value
of k=1 in the original code does not make sense, as the 1NN distance will be
0 for every point in the database. If you for any reason want to use the
original algorithm, subtract 1 from the k parameter.

\classinfo%
{de.lmu.ifi.dbs.elki.algorithm.outlier.distance.}{KNNSOS}%
{0.7.5}%
{DBLP:conf/sisap/SchubertG17,tr/tilburg/JanssensHPv12}%
{KNNSOS: k-Nearest-Neighbor Stochastic Outlier Selection}%

kNN-based adaption of Stochastic Outlier Selection.

 This is a trivial variation of Stochastic Outlier Selection to benefit from
KNN indexes, but not discussed in the original publication. Instead of
setting perplexity, we choose the number of neighbors k, and set perplexity
simply to k/3. Objects outside of the kNN are not considered anymore.

\classinfo%
{de.lmu.ifi.dbs.elki.algorithm.outlier.distance.}{KNNWeightOutlier}%
{0.3}%
{DBLP:conf/pkdd/AngiulliP02}%
{KNNWeight outlier detection}%

Outlier Detection based on the accumulated distances of a point to its k
nearest neighbors.

 As in the original publication (as far as we could tell from the pseudocode
included), the current point is not included in the nearest neighbors (see
figures in the publication). This matches the intuition common in nearest
neighbor classification, where the evaluated instances are not part of the
training set; but it contrasts to the pseudocode of the kNN outlier method
and the database interpretation (which returns all objects stored in the
database).

 Furthermore, we report the sum of the k distances (called "weight" in the
original publication). Other implementations may return the average distance
instead, and therefore yield different results.

\classinfo%
{de.lmu.ifi.dbs.elki.algorithm.outlier.distance.}{LocalIsolationCoefficient}%
{0.7.0}%
{doi:10.1109/ITCS.2009.230}%
{Local Isolation Coefficient}%

The Local Isolation Coefficient is the sum of the kNN distance and the
average distance to its k nearest neighbors.

 The algorithm originally used a normalized Manhattan distance on numerical
attributes, and Hamming distance on categorial attributes.

\classinfo%
{de.lmu.ifi.dbs.elki.algorithm.outlier.distance.}{ODIN}%
{0.6.0}%
{DBLP:conf/icpr/HautamakiKF04}%
{ODIN: Outlier Detection Using k-Nearest Neighbour Graph}%

Outlier detection based on the in-degree of the kNN graph.

 This is a curried version: instead of using a threshold T to obtain a binary
decision, we use the computed value as outlier score; normalized by k to make
the numbers more comparable across different parameterizations.

\classinfo%
{de.lmu.ifi.dbs.elki.algorithm.outlier.distance.}{ReferenceBasedOutlierDetection}%
{0.3}%
{DBLP:conf/icdm/PeiZG06}%
{An Efficient Reference-based Approach to Outlier Detection in Large Datasets}%

Reference-Based Outlier Detection algorithm, an algorithm that computes kNN
distances approximately, using reference points.

 kNN distances are approximated by the difference in distance from a reference
point. For this approximation to be of high quality, triangle inequality is
required; but the algorithm can also process non-metric distances.

\classinfo%
{de.lmu.ifi.dbs.elki.algorithm.outlier.distance.}{SOS}%
{0.7.5}%
{DBLP:conf/sisap/SchubertG17,tr/tilburg/JanssensHPv12}%
{SOS: Stochastic Outlier Selection}%

Stochastic Outlier Selection.

\pkginfo{8}%
{de.lmu.ifi.dbs.elki.algorithm.outlier.distance.}{parallel}%
{Distance Parallel}%

Parallel implementations of distance-based outlier detectors.

\classinfo%
{de.lmu.ifi.dbs.elki.algorithm.outlier.distance.parallel.}{ParallelKNNOutlier}%
{0.7.0}%
{DBLP:journals/datamine/SchubertZK14}%
{Parallel KNN Outlier}%

Parallel implementation of KNN Outlier detection.

\classinfo%
{de.lmu.ifi.dbs.elki.algorithm.outlier.distance.parallel.}{ParallelKNNWeightOutlier}%
{0.7.0}%
{DBLP:journals/datamine/SchubertZK14}%
{Parallel KNN Weight Outlier}%

Parallel implementation of KNN Weight Outlier detection.

\pkginfo{7}%
{de.lmu.ifi.dbs.elki.algorithm.outlier.}{intrinsic}%
{Intrinsic}%

Outlier detection algorithms based on intrinsic dimensionality.

\classinfo%
{de.lmu.ifi.dbs.elki.algorithm.outlier.intrinsic.}{IDOS}%
{0.7.0}%
{tr/nii/BrunkenHZ15}%
{IDOS: Intrinsic Dimensional Outlier Score}%

Intrinsic Dimensional Outlier Detection in High-Dimensional Data.

\classinfo%
{de.lmu.ifi.dbs.elki.algorithm.outlier.intrinsic.}{ISOS}%
{0.7.5}%
{DBLP:conf/sisap/SchubertG17}%
{ISOS: Intrinsic Stochastic Outlier Selection}%

Intrinsic Stochastic Outlier Selection.

\classinfo%
{de.lmu.ifi.dbs.elki.algorithm.outlier.intrinsic.}{IntrinsicDimensionalityOutlier}%
{0.7.0}%
{DBLP:conf/sisap/HouleSZ18}%
{Intrinsic Dimensionality Outlier}%

Use intrinsic dimensionality for outlier detection.

\pkginfo{7}%
{de.lmu.ifi.dbs.elki.algorithm.outlier.}{lof}%
{Lof}%

LOF family of outlier detection algorithms.

\classinfo%
{de.lmu.ifi.dbs.elki.algorithm.outlier.lof.}{ALOCI}%
{0.5.0}%
{DBLP:conf/icde/PapadimitriouKGF03}%
{Approximate LOCI: Fast Outlier Detection Using the Local Correlation Integral}%

Fast Outlier Detection Using the "approximate Local Correlation Integral".

 Outlier detection using multiple epsilon neighborhoods.

\classinfo%
{de.lmu.ifi.dbs.elki.algorithm.outlier.lof.}{COF}%
{0.7.0}%
{DBLP:conf/pakdd/TangCFC02}%
{COF: Connectivity-based Outlier Factor}%

Connectivity-based Outlier Factor (COF).

\classinfo%
{de.lmu.ifi.dbs.elki.algorithm.outlier.lof.}{FlexibleLOF}%
{0.2}%
{DBLP:conf/sigmod/BreunigKNS00}%
{FlexibleLOF: Local Outlier Factor with additional options}%

Flexible variant of the "Local Outlier Factor" algorithm.

 This implementation diverts from the original LOF publication in that it
allows the user to use a different distance function for the reachability
distance and neighborhood determination (although the default is to use the
same value.)

 The k nearest neighbors are determined using the standard distance function,
while the reference set used in reachability distance computation is
configured using a separate reachability distance function.

 The original LOF parameter was called "minPts". For consistency
with the name "kNN query", we chose to rename the parameter to {\begingroup\ttfamily{}k\endgroup}.
Flexible LOF allows you to set the two values different, which yields the
parameters {\begingroup\ttfamily{}-lof.krefer\endgroup} and {\begingroup\ttfamily{}-lof.kreach\endgroup}.

\classinfo%
{de.lmu.ifi.dbs.elki.algorithm.outlier.lof.}{INFLO}%
{0.3}%
{DBLP:conf/pakdd/JinTHW06}%
{INFLO: Influenced Outlierness Factor}%

Influence Outliers using Symmetric Relationship (INFLO) using two-way search,
is an outlier detection method based on LOF; but also using the reverse kNN.

\classinfo%
{de.lmu.ifi.dbs.elki.algorithm.outlier.lof.}{KDEOS}%
{0.7.0}%
{DBLP:conf/sdm/SchubertZK14}%
{KDEOS: Kernel Density Estimator Outlier Score}%

Generalized Outlier Detection with Flexible Kernel Density Estimates.

 This is an outlier detection inspired by LOF, but using kernel density
estimation (KDE) from statistics. Unfortunately, for higher dimensional data,
kernel density estimation itself becomes difficult. At this point, the
\texttt{kdeos.idim} parameter can become useful, which allows to either
disable dimensionality adjustment completely (\texttt{0}) or to set it to a
lower dimensionality than the data representation. This may sound like a hack
at first, but real data is often of lower intrinsic dimensionality, and
embedded into a higher data representation. Adjusting the kernel to account
for the representation seems to yield worse results than using a lower,
intrinsic, dimensionality.

 If your data set has many duplicates, the \texttt{kdeos.kernel.minbw} parameter sets a minimum kernel bandwidth, which may improve results in these
cases, as it prevents kernels from degenerating to single points.

\classinfo%
{de.lmu.ifi.dbs.elki.algorithm.outlier.lof.}{LDF}%
{0.5.5}%
{DBLP:conf/mldm/LateckiLP07}%
{LDF: Outlier Detection with Kernel Density Functions}%

Outlier Detection with Kernel Density Functions.

 A variation of LOF which uses kernel density estimation, but in contrast to
{\begingroup\ttfamily{}SimpleKernelDensityLOF\endgroup} also uses the reachability concept of LOF.

\classinfo%
{de.lmu.ifi.dbs.elki.algorithm.outlier.lof.}{LDOF}%
{0.3}%
{DBLP:conf/pakdd/ZhangHJ09}%
{LDOF: Local Distance-Based Outlier Factor}%

Computes the LDOF (Local Distance-Based Outlier Factor) for all objects of a
Database.

\classinfo%
{de.lmu.ifi.dbs.elki.algorithm.outlier.lof.}{LOCI}%
{0.2}%
{DBLP:conf/icde/PapadimitriouKGF03}%
{LOCI: Fast Outlier Detection Using the Local Correlation Integral}%

Fast Outlier Detection Using the "Local Correlation Integral".

 Exact implementation only, not aLOCI. See {\begingroup\ttfamily{}ALOCI\endgroup}.

 Outlier detection using multiple epsilon neighborhoods.

 This implementation has O(n$^{3}$ log n) runtime complexity!

\classinfo%
{de.lmu.ifi.dbs.elki.algorithm.outlier.lof.}{LOF}%
{0.2}%
{DBLP:conf/sigmod/BreunigKNS00}%
{LOF: Local Outlier Factor}%

Algorithm to compute density-based local outlier factors in a database based
on a specified parameter {\begingroup\ttfamily{}-lof.k\endgroup}.

 The original LOF parameter was called "minPts", but for consistency
within ELKI we have renamed this parameter to "k".

 Compatibility note: as of ELKI 0.7.0, we no longer include the query point,
for consistency with other methods.

\classinfo%
{de.lmu.ifi.dbs.elki.algorithm.outlier.lof.}{LoOP}%
{0.3}%
{DBLP:conf/cikm/KriegelKSZ09}%
{LoOP: Local Outlier Probabilities}%

LoOP: Local Outlier Probabilities

 Distance/density based algorithm similar to LOF to detect outliers, but with
statistical methods to achieve better result stability.

\classinfo%
{de.lmu.ifi.dbs.elki.algorithm.outlier.lof.}{OnlineLOF}%
{0.4.0}%
{}%
{Online LOF}%

Incremental version of the {\begingroup\ttfamily{}LOF\endgroup} Algorithm, supports insertions and
removals.

\classinfo%
{de.lmu.ifi.dbs.elki.algorithm.outlier.lof.}{SimpleKernelDensityLOF}%
{0.5.5}%
{}%
{Simple Kernel Density LOF}%

A simple variant of the LOF algorithm, which uses a simple kernel density
estimation instead of the local reachability density.

\classinfo%
{de.lmu.ifi.dbs.elki.algorithm.outlier.lof.}{SimplifiedLOF}%
{0.5.5}%
{DBLP:journals/datamine/SchubertZK14}%
{Simplified LOF}%

A simplified version of the original LOF algorithm, which does not use the
reachability distance, yielding less stable results on inliers.

\classinfo%
{de.lmu.ifi.dbs.elki.algorithm.outlier.lof.}{VarianceOfVolume}%
{0.7.0}%
{DBLP:journals/prl/HuS03}%
{Variance Of Volume}%

Variance of Volume for outlier detection.

 The volume is estimated by the distance to the k-nearest neighbor, then the
variance of volume is computed.

 Unfortunately, this approach needs an enormous numerical precision, and may
not work for high-dimensional, non-normalized data. We therefore divide each
volume by the average across the data set. This means values are even less
comparable across data sets, but this avoids some of the numerical problems
of this method.

\pkginfo{8}%
{de.lmu.ifi.dbs.elki.algorithm.outlier.lof.}{parallel}%
{Lof Parallel}%

Parallelized variants of LOF.

 This parallelization is based on the generalization of outlier detection
published in:

\classinfo%
{de.lmu.ifi.dbs.elki.algorithm.outlier.lof.parallel.}{ParallelLOF}%
{0.7.0}%
{DBLP:journals/datamine/SchubertZK14}%
{Parallel LOF}%

Parallel implementation of Local Outlier Factor using processors.

 This parallelized implementation is based on the easy-to-parallelize
generalized pattern discussed in

 Erich Schubert, Arthur Zimek, Hans-Peter Kriegel\\
 Local Outlier Detection Reconsidered: a Generalized View on Locality with
Applications to Spatial, Video, and Network Outlier Detection\\
 Data Mining and Knowledge Discovery 28(1)

\classinfo%
{de.lmu.ifi.dbs.elki.algorithm.outlier.lof.parallel.}{ParallelSimplifiedLOF}%
{0.7.0}%
{DBLP:journals/datamine/SchubertZK14}%
{Parallel Simplified LOF}%

Parallel implementation of Simplified-LOF Outlier detection using processors.

 This parallelized implementation is based on the easy-to-parallelize
generalized pattern discussed in

 Erich Schubert, Arthur Zimek, Hans-Peter Kriegel\\
 Local Outlier Detection Reconsidered: a Generalized View on Locality with
Applications to Spatial, Video, and Network Outlier Detection\\
 Data Mining and Knowledge Discovery 28(1)

\pkginfo{7}%
{de.lmu.ifi.dbs.elki.algorithm.outlier.}{meta}%
{Meta}%

Meta outlier detection algorithms: external scores, score rescaling.

\classinfo%
{de.lmu.ifi.dbs.elki.algorithm.outlier.meta.}{ExternalDoubleOutlierScore}%
{0.4.0}%
{}%
{External Double Outlier Score}%

External outlier detection scores, loading outlier scores from an external
file. This class is meant to be able to read the default output of ELKI, i.e.
one object per line, with the DBID specified as \texttt{ID=} and the outlier
score specified with an algorithm-specific prefix.

\classinfo%
{de.lmu.ifi.dbs.elki.algorithm.outlier.meta.}{FeatureBagging}%
{0.4.0}%
{DBLP:conf/kdd/LazarevicK05}%
{Feature Bagging for Outlier Detection}%

A simple ensemble method called "Feature bagging" for outlier detection.

 Since the proposed method is only sensible to run on multiple instances of
the same algorithm (due to incompatible score ranges), we do not allow using
arbitrary algorithms.

\classinfo%
{de.lmu.ifi.dbs.elki.algorithm.outlier.meta.}{HiCS}%
{0.5.0}%
{DBLP:conf/icde/KellerMB12}%
{HiCS: High Contrast Subspaces for Density-Based Outlier Ranking}%

Algorithm to compute High Contrast Subspaces for Density-Based Outlier
Ranking.

\classinfo%
{de.lmu.ifi.dbs.elki.algorithm.outlier.meta.}{RescaleMetaOutlierAlgorithm}%
{0.4.0}%
{}%
{Rescale Meta Outlier Algorithm}%

Scale another outlier score using the given scaling function.

\classinfo%
{de.lmu.ifi.dbs.elki.algorithm.outlier.meta.}{SimpleOutlierEnsemble}%
{0.5.5}%
{}%
{Simple Outlier Ensemble}%

Simple outlier ensemble method.

\pkginfo{7}%
{de.lmu.ifi.dbs.elki.algorithm.outlier.}{spatial}%
{Spatial}%

Spatial outlier detection algorithms

\classinfo%
{de.lmu.ifi.dbs.elki.algorithm.outlier.spatial.}{CTLuGLSBackwardSearchAlgorithm}%
{0.4.0}%
{DBLP:conf/kdd/ChenLB10}%
{GLS-Backward Search}%

GLS-Backward Search is a statistical approach to detecting spatial outliers.

 Implementation note: this is just the most basic version of this algorithm.
The spatial relation must be two dimensional, the set of spatial basis
functions is hard-coded (but trivial to enhance) to \(\{1,x,y,x^2,y^2,xy\}\),
and we assume the neighborhood is large enough for the simpler formulas to
work that make the optimization problem convex.

\classinfo%
{de.lmu.ifi.dbs.elki.algorithm.outlier.spatial.}{CTLuMeanMultipleAttributes}%
{0.4.0}%
{DBLP:conf/ictai/LuCK03}%
{CTLu Mean Multiple Attributes}%

Mean Approach is used to discover spatial outliers with multiple attributes.

\classinfo%
{de.lmu.ifi.dbs.elki.algorithm.outlier.spatial.}{CTLuMedianAlgorithm}%
{0.4.0}%
{DBLP:conf/icdm/LuCK03}%
{Median Algorithm for Spatial Outlier Detection}%

Median Algorithm of C.-T. Lu

\classinfo%
{de.lmu.ifi.dbs.elki.algorithm.outlier.spatial.}{CTLuMedianMultipleAttributes}%
{0.4.0}%
{DBLP:conf/ictai/LuCK03}%
{CTLu Median Multiple Attributes}%

Median Approach is used to discover spatial outliers with multiple
attributes.

\classinfo%
{de.lmu.ifi.dbs.elki.algorithm.outlier.spatial.}{CTLuMoranScatterplotOutlier}%
{0.4.0}%
{DBLP:journals/geoinformatica/ShekharLZ03}%
{Moran Scatterplot Outlier}%

Moran scatterplot outliers, based on the standardized deviation from the
local and global means. In contrast to the definition given in the reference,
we use this as a ranking outlier detection by not applying the signedness
test,
but by using the score (- localZ) * (Average localZ of Neighborhood)
directly.
This allows us to differentiate a bit between stronger and weaker outliers.

\classinfo%
{de.lmu.ifi.dbs.elki.algorithm.outlier.spatial.}{CTLuRandomWalkEC}%
{0.4.0}%
{DBLP:conf/gis/LiuLC10}%
{Random Walk on Exhaustive Combination}%

Spatial outlier detection based on random walks.

 Note: this method can only handle one-dimensional data, but could probably be
easily extended to higher dimensional data by using an distance function
instead of the absolute difference.

\classinfo%
{de.lmu.ifi.dbs.elki.algorithm.outlier.spatial.}{CTLuScatterplotOutlier}%
{0.4.0}%
{DBLP:journals/geoinformatica/ShekharLZ03}%
{Scatterplot Spatial Outlier}%

Scatterplot-outlier is a spatial outlier detection method that performs a
linear regression of object attributes and their neighbors average value.

\classinfo%
{de.lmu.ifi.dbs.elki.algorithm.outlier.spatial.}{CTLuZTestOutlier}%
{0.4.0}%
{DBLP:journals/geoinformatica/ShekharLZ03}%
{Z-Test Outlier Detection}%

Detect outliers by comparing their attribute value to the mean and standard
deviation of their neighborhood.

\classinfo%
{de.lmu.ifi.dbs.elki.algorithm.outlier.spatial.}{SLOM}%
{0.4.0}%
{DBLP:journals/kais/ChawlaS06}%
{SLOM: a new measure for local spatial outliers}%

SLOM: a new measure for local spatial outliers

\classinfo%
{de.lmu.ifi.dbs.elki.algorithm.outlier.spatial.}{SOF}%
{0.4.0}%
{DBLP:conf/icig/HuangQ04}%
{Spatial Outlier Factor}%

The Spatial Outlier Factor (SOF) is a spatial
{\begingroup\ttfamily{}LOF\endgroup} variation.

 Since the "reachability distance" of LOF cannot be used canonically in the
bichromatic case, this part of LOF is dropped and the exact distance is used
instead.

\classinfo%
{de.lmu.ifi.dbs.elki.algorithm.outlier.spatial.}{TrimmedMeanApproach}%
{0.4.0}%
{DBLP:journals/ida/HuS04}%
{A Trimmed Mean Approach to Finding Spatial Outliers}%

A Trimmed Mean Approach to Finding Spatial Outliers.

 Outliers are defined by their value deviation from a trimmed mean of the
neighbors.

\pkginfo{8}%
{de.lmu.ifi.dbs.elki.algorithm.outlier.spatial.}{neighborhood}%
{Spatial Neighborhood}%

Spatial outlier neighborhood classes

\classinfo%
{de.lmu.ifi.dbs.elki.algorithm.outlier.spatial.neighborhood.}{ExtendedNeighborhood}%
{0.4.0}%
{}%
{Extended Neighborhood}%

Neighborhood obtained by computing the k-fold closure of an existing
neighborhood.

\classinfo%
{de.lmu.ifi.dbs.elki.algorithm.outlier.spatial.neighborhood.}{ExternalNeighborhood}%
{0.4.0}%
{}%
{External Neighborhood}%

A precomputed neighborhood, loaded from an external file.

\classinfo%
{de.lmu.ifi.dbs.elki.algorithm.outlier.spatial.neighborhood.}{PrecomputedKNearestNeighborNeighborhood}%
{0.4.0}%
{}%
{Precomputed K Nearest Neighbor Neighborhood}%

Neighborhoods based on k nearest neighbors.

\pkginfo{7}%
{de.lmu.ifi.dbs.elki.algorithm.outlier.}{subspace}%
{Subspace}%

Subspace outlier detection methods.
 
Methods that detect outliers in subspaces (projections) of the data set.

\classinfo%
{de.lmu.ifi.dbs.elki.algorithm.outlier.subspace.}{AbstractAggarwalYuOutlier}%
{0.4.0}%
{DBLP:conf/sigmod/AggarwalY01}%
{Abstract Aggarwal Yu Outlier}%

Abstract base class for the sparse-grid-cell based outlier detection of
Aggarwal and Yu.

\classinfo%
{de.lmu.ifi.dbs.elki.algorithm.outlier.subspace.}{AggarwalYuEvolutionary}%
{0.4.0}%
{DBLP:conf/sigmod/AggarwalY01}%
{EAFOD: the evolutionary outlier detection algorithm}%

Evolutionary variant (EAFOD) of the high-dimensional outlier detection
algorithm by Aggarwal and Yu.

\classinfo%
{de.lmu.ifi.dbs.elki.algorithm.outlier.subspace.}{AggarwalYuNaive}%
{0.4.0}%
{DBLP:conf/sigmod/AggarwalY01}%
{BruteForce: Outlier detection for high dimensional data}%

BruteForce variant of the high-dimensional outlier detection algorithm by
Aggarwal and Yu.

 The evolutionary approach is implemented as
{\begingroup\ttfamily{}AggarwalYuEvolutionary\endgroup}

\classinfo%
{de.lmu.ifi.dbs.elki.algorithm.outlier.subspace.}{OUTRES}%
{0.5.0}%
{DBLP:conf/cikm/MullerSS10}%
{OUTRES}%

Adaptive outlierness for subspace outlier ranking (OUTRES).

 Note: this algorithm seems to have a O(n³d!) complexity with no obvious way
to accelerate it with usual index structures for range queries: each object
in each tested subspace will need to know the mean and standard deviation of
the density of the neighbors, which in turn needs another range query; except
if we precomputed the densities for each of O(d!) possible subsets of
dimensions.

\classinfo%
{de.lmu.ifi.dbs.elki.algorithm.outlier.subspace.}{OutRankS1}%
{0.5.0}%
{DBLP:conf/icde/MullerASS08}%
{OutRank: ranking outliers in high dimensional data}%

OutRank: ranking outliers in high dimensional data.

 Algorithm to score outliers based on a subspace clustering result. This class
implements score 1 of the OutRank publication, which is a score based on
cluster sizes and cluster dimensionality.

\classinfo%
{de.lmu.ifi.dbs.elki.algorithm.outlier.subspace.}{SOD}%
{0.2}%
{DBLP:conf/pakdd/KriegelKSZ09}%
{SOD: Subspace outlier degree}%

Subspace Outlier Degree. Outlier detection method for axis-parallel
subspaces.

\pkginfo{7}%
{de.lmu.ifi.dbs.elki.algorithm.outlier.}{svm}%
{Svm}%

Support-Vector-Machines for outlier detection.

\classinfo%
{de.lmu.ifi.dbs.elki.algorithm.outlier.svm.}{LibSVMOneClassOutlierDetection}%
{0.7.0}%
{DBLP:journals/neco/ScholkopfPSSW01}%
{LibSVMOne Class Outlier Detection}%

Outlier-detection using one-class support vector machines.

 Important note: from literature, the one-class SVM is trained as if 0 was the
only counterexample. Outliers will only be detected when they are close to
the origin in kernel space! In our experience, results from this method are
rather mixed, in particular as you would likely need to tune hyperparameters.
Results may be better if you have a training data set with positive examples
only, then apply it only to new data (which is currently not supported in
this implementation, it assumes a single-dataset scenario).

\pkginfo{6}%
{de.lmu.ifi.dbs.elki.algorithm.}{projection}%
{Projection}%

Data projections (see also preprocessing filters for basic projections).

\classinfo%
{de.lmu.ifi.dbs.elki.algorithm.projection.}{BarnesHutTSNE}%
{0.7.5}%
{DBLP:journals/jmlr/Maaten14}%
{Barnes Hut TSNE}%

tSNE using Barnes-Hut-Approximation.

 For larger data sets, use an index to make finding the nearest neighbors
faster, e.g. cover tree or k-d-tree.

\classinfo%
{de.lmu.ifi.dbs.elki.algorithm.projection.}{GaussianAffinityMatrixBuilder}%
{0.7.5}%
{DBLP:conf/nips/HintonR02}%
{Gaussian Affinity Matrix Builder}%

Compute the affinity matrix for SNE and tSNE using a Gaussian distribution
with a constant sigma.

\classinfo%
{de.lmu.ifi.dbs.elki.algorithm.projection.}{IntrinsicNearestNeighborAffinityMatrixBuilder}%
{0.7.5}%
{DBLP:conf/sisap/SchubertG17}%
{Intrinsic t-Stochastic Neighbor Embedding}%

Build sparse affinity matrix using the nearest neighbors only, adjusting for
intrinsic dimensionality. On data sets with high intrinsic dimensionality,
this can give better results.

 Furthermore, this approach uses a different rule to combine affinities:
rather than taking the arithmetic average of \(p_{ij}\) and \(p_{ji}\), we
use \(\sqrt{p_{ij} \cdot p_{ji}}\), which prevents outliers from attaching
closely to nearby clusters.

\classinfo%
{de.lmu.ifi.dbs.elki.algorithm.projection.}{NearestNeighborAffinityMatrixBuilder}%
{0.7.5}%
{DBLP:journals/jmlr/Maaten14}%
{Nearest Neighbor Affinity Matrix Builder}%

Build sparse affinity matrix using the nearest neighbors only.

\classinfo%
{de.lmu.ifi.dbs.elki.algorithm.projection.}{PerplexityAffinityMatrixBuilder}%
{0.7.5}%
{DBLP:conf/nips/HintonR02}%
{Perplexity Affinity Matrix Builder}%

Compute the affinity matrix for SNE and tSNE.

\classinfo%
{de.lmu.ifi.dbs.elki.algorithm.projection.}{SNE}%
{0.7.5}%
{DBLP:conf/nips/HintonR02}%
{SNE}%

Stochastic Neighbor Embedding is a projection technique designed for
visualization that tries to preserve the nearest neighbor structure.

\classinfo%
{de.lmu.ifi.dbs.elki.algorithm.projection.}{TSNE}%
{0.7.5}%
{journals/jmlr/MaatenH08}%
{t-SNE}%

t-Stochastic Neighbor Embedding is a projection technique designed for
visualization that tries to preserve the nearest neighbor structure.

\pkginfo{6}%
{de.lmu.ifi.dbs.elki.algorithm.}{statistics}%
{Statistics}%

Statistical analysis algorithms.
 
 The algorithms in this package perform statistical analysis of the data
 (e.g. compute distributions, distance distributions etc.)

\classinfo%
{de.lmu.ifi.dbs.elki.algorithm.statistics.}{AddSingleScale}%
{0.5.0}%
{}%
{Add Single Scale}%

Pseudo "algorithm" that computes the global min/max for a relation across all
attributes.

\classinfo%
{de.lmu.ifi.dbs.elki.algorithm.statistics.}{AddUniformScale}%
{0.7.5}%
{}%
{Add Uniform Scale}%

Pseudo "algorithm" that computes the global min/max for a relation across all
attributes.

\classinfo%
{de.lmu.ifi.dbs.elki.algorithm.statistics.}{AveragePrecisionAtK}%
{0.5.0}%
{}%
{Average Precision At K}%

Evaluate a distance functions performance by computing the average precision
at k, when ranking the objects by distance.

\classinfo%
{de.lmu.ifi.dbs.elki.algorithm.statistics.}{DistanceQuantileSampler}%
{0.7.0}%
{}%
{Distance Quantile Sampler}%

Compute a quantile of a distance sample, useful for choosing parameters for
algorithms.

\classinfo%
{de.lmu.ifi.dbs.elki.algorithm.statistics.}{DistanceStatisticsWithClasses}%
{0.2}%
{}%
{Distance Histogram}%

Algorithm to gather statistics over the distance distribution in the data
set.

\classinfo%
{de.lmu.ifi.dbs.elki.algorithm.statistics.}{EstimateIntrinsicDimensionality}%
{0.7.0}%
{}%
{Estimate Intrinsic Dimensionality}%

Estimate global average intrinsic dimensionality of a data set.

Note: this algorithm does not produce a result, but only logs statistics.

\classinfo%
{de.lmu.ifi.dbs.elki.algorithm.statistics.}{EvaluateRankingQuality}%
{0.2}%
{}%
{Evaluate Ranking Quality}%

Evaluate a distance function with respect to kNN queries. For each point, the
neighbors are sorted by distance, then the ROC AUC is computed. A score of 1
means that the distance function provides a perfect ordering of relevant
neighbors first, then irrelevant neighbors. A value of 0.5 can be obtained by
random sorting. A value of 0 means the distance function is inverted, i.e. a
similarity.

In contrast to {\begingroup\ttfamily{}RankingQualityHistogram\endgroup}, this method uses a binning
based on the centrality of objects. This allows analyzing whether or not a
particular distance degrades for the outer parts of a cluster.

\classinfo%
{de.lmu.ifi.dbs.elki.algorithm.statistics.}{EvaluateRetrievalPerformance}%
{0.7.0}%
{}%
{Evaluate Retrieval Performance}%

Evaluate a distance functions performance by computing the mean average
precision, ROC, and NN classification performance when ranking the objects by
distance.

\classinfo%
{de.lmu.ifi.dbs.elki.algorithm.statistics.}{HopkinsStatisticClusteringTendency}%
{0.7.0}%
{doi:10.1093/oxfordjournals.aob.a083391}%
{Hopkins Statistic Clustering Tendency}%

The Hopkins Statistic of Clustering Tendency measures the probability that a
data set is generated by a uniform data distribution.

 The statistic compares the ratio of the 1NN distance for objects from the
data set compared to the 1NN distances of uniform distributed objects.

\classinfo%
{de.lmu.ifi.dbs.elki.algorithm.statistics.}{RangeQuerySelectivity}%
{0.7.0}%
{}%
{Range Query Selectivity}%

Evaluate the range query selectivity.

\classinfo%
{de.lmu.ifi.dbs.elki.algorithm.statistics.}{RankingQualityHistogram}%
{0.2}%
{}%
{Ranking Quality Histogram}%

Evaluate a distance function with respect to kNN queries. For each point, the
neighbors are sorted by distance, then the ROC AUC is computed. A score of 1
means that the distance function provides a perfect ordering of relevant
neighbors first, then irrelevant neighbors. A value of 0.5 can be obtained by
random sorting. A value of 0 means the distance function is inverted, i.e. a
similarity.

\pkginfo{6}%
{de.lmu.ifi.dbs.elki.algorithm.}{timeseries}%
{Timeseries}%

\classinfo%
{de.lmu.ifi.dbs.elki.algorithm.timeseries.}{OfflineChangePointDetectionAlgorithm}%
{0.7.5}%
{doi:10.2307/2333258,books/prentice/BassevilleN93/C2,doi:10.2307/1427090}%
{Off-line Change Point Detection}%

Off-line change point detection algorithm detecting a change in mean, based
on the cumulative sum (CUSUM), same-variance assumption, and using bootstrap
sampling for significance estimation.

\classinfo%
{de.lmu.ifi.dbs.elki.algorithm.timeseries.}{SigniTrendChangeDetection}%
{0.7.5}%
{DBLP:conf/kdd/SchubertWK14}%
{Signi-Trend: scalable detection of emerging topics in textual streams by hashed significance thresholds}%

Signi-Trend detection algorithm applies to a single time-series.

This is not a complete implementation of the method, but a modified
(two-sided) version of the significance score use in Signi-Trend for change
detection. The hashing and scalability parts of Signi-Trend are not
applicable here.

This implementation currently does not use timestamps, and thus only works
for fixed-interval measurements. It could be extended to allow dynamic data
windows by adjusting the alpha parameter based on time deltas.

\pkginfo{5}%
{de.lmu.ifi.dbs.elki.}{application}%
{Application}%

Base classes for stand alone applications.

\classinfo%
{de.lmu.ifi.dbs.elki.application.}{AbstractApplication}%
{0.2}%
{DBLP:journals/pvldb/SchubertKEZSZ15}%
{Abstract Application}%

AbstractApplication sets the values for flags verbose and help.

 Any Wrapper class that makes use of these flags may extend this class. Beware
to make correct use of parameter settings via optionHandler as commented with
constructor and methods.

\classinfo%
{de.lmu.ifi.dbs.elki.application.}{ClassifierHoldoutEvaluationTask}%
{0.7.0}%
{}%
{Classifier Holdout Evaluation Task}%

Evaluate a classifier.

\classinfo%
{de.lmu.ifi.dbs.elki.application.}{ConvertToBundleApplication}%
{0.5.5}%
{}%
{Convert To Bundle Application}%

Convert an input file to the more efficient ELKI bundle format.

\classinfo%
{de.lmu.ifi.dbs.elki.application.}{GeneratorXMLSpec}%
{0.2}%
{}%
{Generator XML Spec}%

Generate a data set based on a specified model (using an XML specification)

\classinfo%
{de.lmu.ifi.dbs.elki.application.}{KDDCLIApplication}%
{0.3}%
{}%
{KDDCLI Application}%

Basic command line application for Knowledge Discovery in Databases use
cases. It allows running unsupervised {\begingroup\ttfamily{}Algorithm\endgroup}s to run on any
{\begingroup\ttfamily{}DatabaseConnection\endgroup}.

\pkginfo{6}%
{de.lmu.ifi.dbs.elki.application.}{cache}%
{Cache}%

Utility applications for the persistence layer such as distance cache builders.

{\begingroup\ttfamily{}CacheDoubleDistanceInOnDiskMatrix\endgroup} and
{\begingroup\ttfamily{}CacheFloatDistanceInOnDiskMatrix\endgroup} are conversion utilities that materialize an arbitrary distance into a binary distance
cache on the harddisk (using {\begingroup\ttfamily{}OnDiskUpperTriangleMatrix\endgroup})

\classinfo%
{de.lmu.ifi.dbs.elki.application.cache.}{CacheDoubleDistanceInOnDiskMatrix}%
{0.2}%
{}%
{Cache Double Distance In On Disk Matrix}%

Precompute an on-disk distance matrix, using double precision.

\classinfo%
{de.lmu.ifi.dbs.elki.application.cache.}{CacheDoubleDistanceKNNLists}%
{0.6.0}%
{}%
{Cache Double Distance KNN Lists}%

Precompute the k nearest neighbors in a disk cache.

\classinfo%
{de.lmu.ifi.dbs.elki.application.cache.}{CacheDoubleDistanceRangeQueries}%
{0.6.0}%
{}%
{Cache Double Distance Range Queries}%

Precompute the k nearest neighbors in a disk cache.

\classinfo%
{de.lmu.ifi.dbs.elki.application.cache.}{CacheFloatDistanceInOnDiskMatrix}%
{0.2}%
{}%
{Cache Float Distance In On Disk Matrix}%

Precompute an on-disk distance matrix, using float precision.

\classinfo%
{de.lmu.ifi.dbs.elki.application.cache.}{PrecomputeDistancesAsciiApplication}%
{0.2}%
{}%
{Precompute Distances Ascii Application}%

Application to precompute pairwise distances into an ascii file.

IDs in the output file will always begin at 0.

The result can then be used with the DoubleDistanceParse.

Symmetry is assumed.

\pkginfo{6}%
{de.lmu.ifi.dbs.elki.application.}{experiments}%
{Experiments}%

Packaged experiments to make them easy to reproduce.

\classinfo%
{de.lmu.ifi.dbs.elki.application.experiments.}{EvaluateIntrinsicDimensionalityEstimators}%
{0.7.0}%
{}%
{Evaluate Intrinsic Dimensionality Estimators}%

Class for testing the estimation quality of intrinsic dimensionality
estimators.

\classinfo%
{de.lmu.ifi.dbs.elki.application.experiments.}{VisualizeGeodesicDistances}%
{0.5.5}%
{DBLP:conf/ssd/SchubertZK13}%
{Visualize Geodesic Distances}%

Visualization function for Cross-track, Along-track, and minimum distance
function.

\pkginfo{6}%
{de.lmu.ifi.dbs.elki.application.}{greedyensemble}%
{Greedyensemble}%

 Greedy ensembles for outlier detection.

 This package contains code used for the greedy ensemble experiment in
 Erich Schubert, Remigius Wojdanowski, Arthur Zimek, Hans-Peter Kriegel\\
 On Evaluation of Outlier Rankings and Outlier Scores\\
 Proc. 12th SIAM Int. Conf. on Data Mining (SDM 2012)

\classinfo%
{de.lmu.ifi.dbs.elki.application.greedyensemble.}{ComputeKNNOutlierScores}%
{0.5.0}%
{DBLP:conf/sdm/SchubertWZK12}%
{Compute KNN Outlier Scores}%

Application that runs a series of kNN-based algorithms on a data set, for
building an ensemble in a second step. The output file consists of a label
and one score value for each object.

 Since some algorithms can be too slow to run on large data sets and for large
values of k, they can be disabled. For example
\texttt{-disable '(LDOF|DWOF|COF|FastABOD)'} disables these two methods
completely. Alternatively, you can use the parameter \texttt{-ksquaremax} to control the maximum k for these four methods separately.

 For methods where k=1 does not make sense, this value will be skipped, and
the procedure will commence at 1+stepsize.

\classinfo%
{de.lmu.ifi.dbs.elki.application.greedyensemble.}{EvaluatePrecomputedOutlierScores}%
{0.7.0}%
{}%
{Evaluate Precomputed Outlier Scores}%

Class to load an outlier detection summary file, as produced by
{\begingroup\ttfamily{}ComputeKNNOutlierScores\endgroup}, and compute popular evaluation metrics for
it.

 File format description:

\begin{itemize}
\item Each column is one object in the data set
\item Each line is a different algorithm
\item There is a mandatory label column, containing the method name
\item The first line must contain the ground-truth, titled
\texttt{bylabel}, where \texttt{0} indicates an inlier and \texttt{1} indicates an outlier
\end{itemize}
 The evaluation assumes that high scores correspond to outliers, unless the
method name matches the pattern given using {\begingroup\ttfamily{}-reversed\endgroup}.
The default value matches several scores known to use reversed values.

\classinfo%
{de.lmu.ifi.dbs.elki.application.greedyensemble.}{GreedyEnsembleExperiment}%
{0.5.0}%
{DBLP:conf/sdm/SchubertWZK12}%
{Greedy Ensemble Experiment}%

Class to load an outlier detection summary file, as produced by
{\begingroup\ttfamily{}ComputeKNNOutlierScores\endgroup}, and compute a naive ensemble for it. Based
on this initial estimation, and optimized ensemble is built using a greedy
strategy. Starting with the best candidate only as initial ensemble, the most
diverse candidate is investigated at each step. If it improves towards the
(estimated) target vector, it is added, otherwise it is discarded.

 This approach is naive, and it may be surprising that it can improve results.
The reason is probably that diversity will result in a comparable ensemble,
while the reduced ensemble size is actually responsible for the improvements,
by being more decisive and less noisy due to dropping "unhelpful" members.

 This still leaves quite a bit of room for improvement. If you build upon this
basic approach, please acknowledge our proof of concept work.

\classinfo%
{de.lmu.ifi.dbs.elki.application.greedyensemble.}{VisualizePairwiseGainMatrix}%
{0.5.0}%
{DBLP:conf/sdm/SchubertWZK12}%
{Visualize Pairwise Gain Matrix}%

Class to load an outlier detection summary file, as produced by
{\begingroup\ttfamily{}ComputeKNNOutlierScores\endgroup}, and compute a matrix with the pairwise
gains. It will have one column / row obtained for each combination.

 The gain is always computed in relation to the better of the two input
methods. Green colors indicate the result has improved, red indicate it
became worse.

\pkginfo{5}%
{de.lmu.ifi.dbs.elki.}{data}%
{Data}%

Basic classes for different data types, database object types and label types.

\classinfo%
{de.lmu.ifi.dbs.elki.data.}{HierarchicalClassLabel}%
{0.1}%
{}%
{Hierarchical Class Label}%

A HierarchicalClassLabel is a ClassLabel to reflect a hierarchical structure
of classes.

\classinfo%
{de.lmu.ifi.dbs.elki.data.}{SimpleClassLabel}%
{0.1}%
{}%
{Simple Class Label}%

A simple class label casting a String as it is as label.

\pkginfo{6}%
{de.lmu.ifi.dbs.elki.data.}{projection}%
{Projection}%

Data projections.

\classinfo%
{de.lmu.ifi.dbs.elki.data.projection.}{FeatureSelection}%
{0.5.0}%
{}%
{Feature Selection}%

Projection class for number vectors.

\classinfo%
{de.lmu.ifi.dbs.elki.data.projection.}{LatLngToECEFProjection}%
{0.6.0}%
{}%
{LatLngToECEF Projection}%

Project (Latitude, Longitude) vectors to (X, Y, Z), from spherical
coordinates to ECEF (earth-centered earth-fixed).

\classinfo%
{de.lmu.ifi.dbs.elki.data.projection.}{LngLatToECEFProjection}%
{0.6.0}%
{}%
{LngLatToECEF Projection}%

Project (Longitude, Latitude) vectors to (X, Y, Z), from spherical
coordinates to ECEF (earth-centered earth-fixed).

\classinfo%
{de.lmu.ifi.dbs.elki.data.projection.}{NumericalFeatureSelection}%
{0.5.0}%
{}%
{Numerical Feature Selection}%

Projection class for number vectors.

\classinfo%
{de.lmu.ifi.dbs.elki.data.projection.}{RandomProjection}%
{0.6.0}%
{}%
{Random Projection}%

Randomized projections of the data.

This class allows projecting the data with different types of random
projections, in particular database friendly projections (as suggested by
Achlioptas, see {\begingroup\ttfamily{}AchlioptasRandomProjectionFamily\endgroup}), but also as
suggested for locality sensitive hashing (LSH).

\pkginfo{7}%
{de.lmu.ifi.dbs.elki.data.projection.}{random}%
{Random}%

Random projection families.

\classinfo%
{de.lmu.ifi.dbs.elki.data.projection.random.}{AchlioptasRandomProjectionFamily}%
{0.6.0}%
{DBLP:conf/pods/Achlioptas01}%
{Achlioptas Random Projection Family}%

Random projections as suggested by Dimitris Achlioptas.

\classinfo%
{de.lmu.ifi.dbs.elki.data.projection.random.}{CauchyRandomProjectionFamily}%
{0.6.0}%
{DBLP:conf/compgeom/DatarIIM04}%
{Cauchy Random Projection Family}%

Random projections using Cauchy distributions (1-stable).

\classinfo%
{de.lmu.ifi.dbs.elki.data.projection.random.}{GaussianRandomProjectionFamily}%
{0.6.0}%
{DBLP:conf/compgeom/DatarIIM04}%
{Gaussian Random Projection Family}%

Random projections using Cauchy distributions (1-stable).

\classinfo%
{de.lmu.ifi.dbs.elki.data.projection.random.}{RandomSubsetProjectionFamily}%
{0.6.0}%
{DBLP:journals/ml/Breiman96b}%
{Random Subset Projection Family}%

Random projection family based on selecting random features.

 The basic idea of using this for data mining should probably be attributed to
L. Breiman, who used it to improve the performance of predictors in an
ensemble.

\classinfo%
{de.lmu.ifi.dbs.elki.data.projection.random.}{SimplifiedRandomHyperplaneProjectionFamily}%
{0.7.0}%
{DBLP:conf/sigir/Henzinger06}%
{Simplified Random Hyperplane Projection Family}%

Random hyperplane projection family.

\pkginfo{6}%
{de.lmu.ifi.dbs.elki.data.}{uncertain}%
{Uncertain}%

Uncertain data objects.

\classinfo%
{de.lmu.ifi.dbs.elki.data.uncertain.}{UnweightedDiscreteUncertainObject}%
{0.7.0}%
{DBLP:journals/cacm/DalviRS09,DBLP:conf/vldb/BenjellounSHW06}%
{Unweighted Discrete Uncertain Object}%

Unweighted implementation of discrete uncertain objects.
\begin{itemize}
\item Every object is represented by a finite number of discrete samples.
\item Every sample has the same weight.
\item Every sample is equally likely to be returned by {\begingroup\ttfamily{}drawSample(java.util.Random)\endgroup}.

\end{itemize}

\classinfo%
{de.lmu.ifi.dbs.elki.data.uncertain.}{WeightedDiscreteUncertainObject}%
{0.7.0}%
{DBLP:journals/cacm/DalviRS09,DBLP:conf/vldb/BenjellounSHW06,DBLP:conf/kdd/BerneckerKRVZ09}%
{Weighted Discrete Uncertain Object}%

Weighted version of discrete uncertain objects.
\begin{itemize}
\item Every object is represented by a finite number of discrete samples.
\item Every sample has a weight associated with it.
\item Samples with higher weight are more likely to be returned by
{\begingroup\ttfamily{}drawSample(java.util.Random)\endgroup}.

\end{itemize}

\pkginfo{7}%
{de.lmu.ifi.dbs.elki.data.uncertain.}{uncertainifier}%
{Uncertainifier}%

Classes to generate uncertain objects from existing certain data.

\classinfo%
{de.lmu.ifi.dbs.elki.data.uncertain.uncertainifier.}{SimpleGaussianUncertainifier}%
{0.7.0}%
{}%
{Simple Gaussian Uncertainifier}%

Vector factory

\classinfo%
{de.lmu.ifi.dbs.elki.data.uncertain.uncertainifier.}{UniformUncertainifier}%
{0.7.0}%
{}%
{Uniform Uncertainifier}%

Factory class.

\classinfo%
{de.lmu.ifi.dbs.elki.data.uncertain.uncertainifier.}{UnweightedDiscreteUncertainifier}%
{0.7.0}%
{}%
{Unweighted Discrete Uncertainifier}%

Class to generate unweighted discrete uncertain objects.

This is a second-order generator: it requires the use of another generator to
sample from (e.g. {\begingroup\ttfamily{}UniformUncertainifier\endgroup} or
{\begingroup\ttfamily{}SimpleGaussianUncertainifier\endgroup}).

\classinfo%
{de.lmu.ifi.dbs.elki.data.uncertain.uncertainifier.}{WeightedDiscreteUncertainifier}%
{0.7.0}%
{}%
{Weighted Discrete Uncertainifier}%

Class to generate weighted discrete uncertain objects.

This is a second-order generator: it requires the use of another generator to
sample from (e.g. {\begingroup\ttfamily{}UniformUncertainifier\endgroup} or
{\begingroup\ttfamily{}SimpleGaussianUncertainifier\endgroup}).

\pkginfo{5}%
{de.lmu.ifi.dbs.elki.}{database}%
{Database}%

ELKI database layer - loading, storing, indexing and accessing data

\pkginfo{6}%
{de.lmu.ifi.dbs.elki.database.}{ids}%
{Ids}%

Database object identification and ID group handling API.

 Database IDs (short: DBID) in ELKI are based on the factory pattern, to allow replacing
the simple Integer-based DBIDs with more complex implementations, e.g. for use with external
databases or to add tracking for debugging purposes. This also allows adding of more efficient
implementations later on in a single place.

\paragraph{DBID interface:}\hfill\\

 The {\begingroup\ttfamily{}DBID\endgroup} object identifies a single object.

 The {\begingroup\ttfamily{}DBIDs\endgroup} hierarchy contains classes for handling groups (sets, arrays) of IDs, that can
be seen as a two-dimensional matrix consisting

    {\begingroup\ttfamily{}ArrayDBIDs\endgroup}  {\begingroup\ttfamily{}HashSetDBIDs\endgroup} 

  {\begingroup\ttfamily{}ModifiableDBIDs\endgroup}  {\begingroup\ttfamily{}ArrayModifiableDBIDs\endgroup}  {\begingroup\ttfamily{}HashSetModifiableDBIDs\endgroup} 

  {\begingroup\ttfamily{}StaticDBIDs\endgroup}  {\begingroup\ttfamily{}ArrayStaticDBIDs\endgroup}  n/a 

 {\begingroup\ttfamily{}StaticDBIDs\endgroup} are structures that cannot support
modifications, but thus can be implemented more efficiently, for example as Interval. They are
mostly used by the data sources.

 These interfaces cannot be instantiated, obviously. Instead, use the static
{\begingroup\ttfamily{}DBIDFactory.FACTORY\endgroup}, which is also wrapped in the {\begingroup\ttfamily{}DBIDUtil\endgroup} class.

\paragraph{Examples:}\hfill\\
\begin{Verbatim}[fontsize=\verbatimsize]
DBIDs allids = database.getIDs();
// preallocate an array of initial capacity 123 
ArrayModifiableDBIDs array = DBIDUtil.newArraySet(123);
// new DBID hash set with minimum initial capacity
ModifiableDBIDs hash = DBIDUtil.newHashSet();

// add all DBIDs from the hash
tree.addDBIDs(hash)
\end{Verbatim}

\paragraph{Utility functions:}\hfill\\
\begin{itemize}
\item {\begingroup\ttfamily{}DBIDUtil.ensureArray\endgroup} to ensure {\begingroup\ttfamily{}ArrayDBIDs\endgroup}
\item {\begingroup\ttfamily{}DBIDUtil.ensureModifiable\endgroup} to ensure {\begingroup\ttfamily{}ModifiableDBIDS\endgroup}
\item {\begingroup\ttfamily{}DBIDUtil.makeUnmodifiable\endgroup} to wrap DBIDs unmodifiable
\end{itemize}

\paragraph{Generic utility classes:}\hfill\\

 {\begingroup\ttfamily{}MaskedDBIDs\endgroup} allows masking an ArrayDBIDs with a BitSet.

\pkginfo{7}%
{de.lmu.ifi.dbs.elki.database.ids.}{integer}%
{Integer}%

Integer-based DBID implementation --
do not use directly - always use {\begingroup\ttfamily{}DBIDUtil\endgroup}.

\classinfo%
{de.lmu.ifi.dbs.elki.database.ids.integer.}{IntegerDBIDArrayQuickSort}%
{0.5.5}%
{web/Yaroslavskiy09}%
{Integer DBID Array Quick Sort}%

Class to sort an integer DBID array, using a modified quicksort.

 Two array iterators will be used to seek to the elements to compare, while
the backing storage is a plain integer array.

\classinfo%
{de.lmu.ifi.dbs.elki.database.ids.integer.}{ReusingDBIDFactory}%
{0.4.0}%
{}%
{Reusing DBID Factory}%

Slightly more complex DBID management, that allows reuse of DBIDs.

NOT tested a lot yet. Not reusing is much simpler!

\classinfo%
{de.lmu.ifi.dbs.elki.database.ids.integer.}{SimpleDBIDFactory}%
{0.4.0}%
{}%
{Simple DBID Factory}%

Simple DBID management, that never reuses IDs. Statically allocated DBID
ranges are given positive values, Dynamically allocated DBIDs are given
negative values.

\classinfo%
{de.lmu.ifi.dbs.elki.database.ids.integer.}{TrivialDBIDFactory}%
{0.4.0}%
{}%
{Trivial DBID Factory}%

Trivial DBID management, that never reuses IDs and just gives them out in
sequence. All IDs will be positive.

\pkginfo{5}%
{de.lmu.ifi.dbs.elki.}{datasource}%
{Datasource}%

Data normalization (and reconstitution) of data sets.

\pkginfo{6}%
{de.lmu.ifi.dbs.elki.datasource.}{filter}%
{Filter}%

Data filtering, in particular for normalization and projection.

\classinfo%
{de.lmu.ifi.dbs.elki.datasource.filter.}{FixedDBIDsFilter}%
{0.4.0}%
{}%
{Fixed DBI Ds Filter}%

This filter assigns static DBIDs, based on the sequence the objects appear in
the bundle by adding a column of DBID type to the bundle.

\pkginfo{7}%
{de.lmu.ifi.dbs.elki.datasource.filter.}{cleaning}%
{Cleaning}%

Filters for data cleaning.

\classinfo%
{de.lmu.ifi.dbs.elki.datasource.filter.cleaning.}{DropNaNFilter}%
{0.6.0}%
{}%
{Drop Na N Filter}%

A filter to drop all records that contain NaN values.

Note: currently, only dense vector columns are supported.

\classinfo%
{de.lmu.ifi.dbs.elki.datasource.filter.cleaning.}{NoMissingValuesFilter}%
{0.4.0}%
{}%
{No Missing Values Filter}%

A filter to remove entries that have missing values.

\classinfo%
{de.lmu.ifi.dbs.elki.datasource.filter.cleaning.}{ReplaceNaNWithRandomFilter}%
{0.6.0}%
{}%
{Replace Na N With Random Filter}%

A filter to replace all NaN values with random values.

Note: currently, only dense vector columns are supported.

\classinfo%
{de.lmu.ifi.dbs.elki.datasource.filter.cleaning.}{VectorDimensionalityFilter}%
{0.7.0}%
{}%
{Vector Dimensionality Filter}%

Filter to remove all vectors that do not have the desired dimensionality.

\pkginfo{7}%
{de.lmu.ifi.dbs.elki.datasource.filter.}{normalization}%
{Normalization}%

Data normalization.

\pkginfo{8}%
{de.lmu.ifi.dbs.elki.datasource.filter.normalization.}{columnwise}%
{Normalization Columnwise}%

Normalizations operating on columns / variates; where each column is treated independently.

\classinfo%
{de.lmu.ifi.dbs.elki.datasource.filter.normalization.columnwise.}{AttributeWiseBetaNormalization}%
{0.7.0}%
{}%
{Attribute Wise Beta Normalization}%

Project the data using a Beta distribution.

This is a crude heuristic, that may or may not work for your data set. There
currently is no theoretical foundation of why it may be sensible or not to do
this.

\classinfo%
{de.lmu.ifi.dbs.elki.datasource.filter.normalization.columnwise.}{AttributeWiseCDFNormalization}%
{0.6.0}%
{}%
{Attribute Wise CDF Normalization}%

Class to perform and undo a normalization on real vectors by estimating the
distribution of values along each dimension independently, then rescaling
objects to the cumulative density function (CDF) value at the original
coordinate.

 This process is for example also mentioned in section 3.4 of

 Effects of Feature Normalization on Image Retrieval\\
 S. Aksoy, R. M. Haralick

 but they do not detail how to obtain an appropriate function `F`.

\classinfo%
{de.lmu.ifi.dbs.elki.datasource.filter.normalization.columnwise.}{AttributeWiseMADNormalization}%
{0.6.0}%
{}%
{Attribute Wise MAD Normalization}%

Median Absolute Deviation is used for scaling the data set as follows:

First, the median, and median absolute deviation are computed in each axis.
Then, each value is projected to (x - median(X)) / MAD(X).

This is similar to z-standardization of data sets, except that it is more
robust towards outliers, and only slightly more expensive to compute.

\classinfo%
{de.lmu.ifi.dbs.elki.datasource.filter.normalization.columnwise.}{AttributeWiseMeanNormalization}%
{0.4.0}%
{}%
{Attribute Wise Mean Normalization}%

Normalization designed for data with a meaningful zero:\\
 The 0 is retained, and the data is linearly scaled to have a mean of 1,
by projection with f(x) = x / mean(X).

Each attribute is processed separately.

\classinfo%
{de.lmu.ifi.dbs.elki.datasource.filter.normalization.columnwise.}{AttributeWiseMinMaxNormalization}%
{0.4.0}%
{}%
{Attribute Wise Min Max Normalization}%

Class to perform and undo a normalization on real vectors with respect to
a given minimum and maximum in each dimension. This class performs a linear
scaling on the data.

\classinfo%
{de.lmu.ifi.dbs.elki.datasource.filter.normalization.columnwise.}{AttributeWiseVarianceNormalization}%
{0.4.0}%
{}%
{Attribute Wise Variance Normalization}%

Class to perform and undo a normalization on real vectors with respect to
given mean and standard deviation in each dimension.

We use the biased variance ({\begingroup\ttfamily{}MeanVariance.getNaiveStddev()\endgroup}), because
this produces that with exactly standard deviation 1. While often the
unbiased estimate ({\begingroup\ttfamily{}MeanVariance.getSampleStddev()\endgroup}) is more
appropriate, it will not ensure this interesting property. For large data,
the difference will be small anyway.

\classinfo%
{de.lmu.ifi.dbs.elki.datasource.filter.normalization.columnwise.}{IntegerRankTieNormalization}%
{0.5.0}%
{}%
{Integer Rank Tie Normalization}%

Normalize vectors according to their rank in the attributes.

Note: ranks are multiplied by 2, to be able to give ties an integer
rank. (e.g. when the first two records are tied, they both have rank "1"
then, followed by the next on "4")

\classinfo%
{de.lmu.ifi.dbs.elki.datasource.filter.normalization.columnwise.}{InverseDocumentFrequencyNormalization}%
{0.4.0}%
{}%
{Inverse Document Frequency Normalization}%

Normalization for text frequency (TF) vectors, using the inverse document
frequency (IDF). See also: TF-IDF for text analysis.

\pkginfo{8}%
{de.lmu.ifi.dbs.elki.datasource.filter.normalization.}{instancewise}%
{Normalization Instancewise}%

Instancewise normalization, where each instance is normalized independently.

\classinfo%
{de.lmu.ifi.dbs.elki.datasource.filter.normalization.instancewise.}{HellingerHistogramNormalization}%
{0.7.0}%
{}%
{Hellinger Histogram Normalization}%

Normalize histograms by scaling them to unit absolute sum, then taking the
square root of the absolute value in each attribute, times the normalization
constant \(1/\sqrt{2}\).
 \[ H(x_i)=\tfrac{\sqrt{|x_i|/\Sigma}}{\sqrt{2}}
\quad\text{ with } \Sigma=\sum\nolimits_i |x_i| \]

 Using Euclidean distance (linear kernel) and this transformation is the same
as using Hellinger distance:
{\begingroup\ttfamily{}HellingerDistanceFunction\endgroup}

\classinfo%
{de.lmu.ifi.dbs.elki.datasource.filter.normalization.instancewise.}{InstanceLogRankNormalization}%
{0.7.0}%
{}%
{Instance Log Rank Normalization}%

Normalize vectors such that the smallest value of each instance is 0, the
largest is 1, but using \( \log_2(1+x) \).

\classinfo%
{de.lmu.ifi.dbs.elki.datasource.filter.normalization.instancewise.}{InstanceMeanVarianceNormalization}%
{0.7.0}%
{}%
{Instance Mean Variance Normalization}%

Normalize vectors such that they have zero mean and unit variance.

\classinfo%
{de.lmu.ifi.dbs.elki.datasource.filter.normalization.instancewise.}{InstanceMinMaxNormalization}%
{0.7.0}%
{}%
{Instance Min Max Normalization}%

Normalize vectors with respect to a given minimum and maximum in each
dimension. By default, minimum 0 and maximum 1 is used. This class
performs a linear scaling on the data.

\classinfo%
{de.lmu.ifi.dbs.elki.datasource.filter.normalization.instancewise.}{InstanceRankNormalization}%
{0.7.0}%
{}%
{Instance Rank Normalization}%

Normalize vectors such that the smallest value of each instance is 0, the
largest is 1.

\classinfo%
{de.lmu.ifi.dbs.elki.datasource.filter.normalization.instancewise.}{LengthNormalization}%
{0.5.0}%
{}%
{Length Normalization}%

Class to perform a normalization on vectors to norm 1.

\classinfo%
{de.lmu.ifi.dbs.elki.datasource.filter.normalization.instancewise.}{Log1PlusNormalization}%
{0.7.0}%
{}%
{Log1 Plus Normalization}%

Normalize the data set by applying \( \frac{\log(1+|x|b)}{\log 1+b} \) to any
value. If the input data was in [0;1], then the resulting values will be in
[0;1], too.

 By default b=1, and thus the transformation is \(\log_2(1+|x|)\).

\pkginfo{7}%
{de.lmu.ifi.dbs.elki.datasource.filter.}{selection}%
{Selection}%

Filters for selecting and sorting data to process.

\classinfo%
{de.lmu.ifi.dbs.elki.datasource.filter.selection.}{ByLabelFilter}%
{0.4.0}%
{}%
{By Label Filter}%

A filter to select data set by their label.

\classinfo%
{de.lmu.ifi.dbs.elki.datasource.filter.selection.}{RandomSamplingStreamFilter}%
{0.5.0}%
{}%
{Random Sampling Stream Filter}%

Subsampling stream filter.

\classinfo%
{de.lmu.ifi.dbs.elki.datasource.filter.selection.}{ShuffleObjectsFilter}%
{0.4.0}%
{}%
{Shuffle Objects Filter}%

A filter to shuffle the dataset.

\classinfo%
{de.lmu.ifi.dbs.elki.datasource.filter.selection.}{SortByLabelFilter}%
{0.4.0}%
{}%
{Sort By Label Filter}%

A filter to sort the data set by some label. The filter sorts the
labels in alphabetical order.

\pkginfo{7}%
{de.lmu.ifi.dbs.elki.datasource.filter.}{transform}%
{Transform}%

Data space transformations.

\classinfo%
{de.lmu.ifi.dbs.elki.datasource.filter.transform.}{ClassicMultidimensionalScalingTransform}%
{0.6.0}%
{}%
{Classic Multidimensional Scaling Transform}%

Rescale the data set using multidimensional scaling, MDS.

Note: the current implementation is rather expensive, both memory- and
runtime wise. Don't use for large data sets! Instead, have a look at
{\begingroup\ttfamily{}FastMultidimensionalScalingTransform\endgroup} which uses power iterations
instead.

\classinfo%
{de.lmu.ifi.dbs.elki.datasource.filter.transform.}{FastMultidimensionalScalingTransform}%
{0.7.0}%
{}%
{Fast Multidimensional Scaling Transform}%

Rescale the data set using multidimensional scaling, MDS.

 This implementation uses power iterations, which is faster when the number of
data points is much larger than the desired number of dimensions.

 This implementation is O(n²), and uses O(n²) memory.

\classinfo%
{de.lmu.ifi.dbs.elki.datasource.filter.transform.}{GlobalPrincipalComponentAnalysisTransform}%
{0.5.0}%
{}%
{Global Principal Component Analysis Transform}%

Apply Principal Component Analysis (PCA) to the data set.

 This is also popular form of "Whitening transformation", and will project the
data to have a unit covariance matrix.

 If you want to also reduce dimensionality, set the {\begingroup\ttfamily{}-pca.filter\endgroup} parameter! Note that this implementation currently will always perform a full
matrix inversion. For very high dimensional data, this can take an excessive
amount of time O(d³) and memory O(d²). Please contribute a better
implementation to ELKI that only computes the requiried dimensions, yet
allows for the same filtering flexibility.

\classinfo%
{de.lmu.ifi.dbs.elki.datasource.filter.transform.}{HistogramJitterFilter}%
{0.5.5}%
{}%
{Histogram Jitter Filter}%

Add Jitter, preserving the histogram properties (same sum, nonnegative).

For each vector, the total sum of all dimensions is computed.\\
 Then a random vector of the average length {\begingroup\ttfamily{}jitter * scale\endgroup} is
added and the result normalized to the original vectors sum. The individual
dimensions are drawn from an exponential distribution with scale
{\begingroup\ttfamily{}jitter / dimensionality\endgroup}, so it is expected that the error in
most dimensions will be low, and higher in few.

This is designed to degrade the quality of a histogram, while preserving the
total sum (e.g. to keep the normalization). The factor "jitter" can be used
to control the degradation amount.

\classinfo%
{de.lmu.ifi.dbs.elki.datasource.filter.transform.}{LatLngToECEFFilter}%
{0.6.0}%
{}%
{LatLngToECEF Filter}%

Project a 2D data set (latitude, longitude) to a 3D coordinate system (X, Y,
Z), such that Euclidean distance is line-of-sight.

\classinfo%
{de.lmu.ifi.dbs.elki.datasource.filter.transform.}{LinearDiscriminantAnalysisFilter}%
{0.6.0}%
{doi:10.1111/j.1469-1809.1936.tb02137.x}%
{Linear Discriminant Analysis Filter}%

Linear Discriminant Analysis (LDA) / Fisher's linear discriminant.

\classinfo%
{de.lmu.ifi.dbs.elki.datasource.filter.transform.}{LngLatToECEFFilter}%
{0.6.0}%
{}%
{LngLatToECEF Filter}%

Project a 2D data set (longitude, latitude) to a 3D coordinate system (X, Y,
Z), such that Euclidean distance is line-of-sight.

\classinfo%
{de.lmu.ifi.dbs.elki.datasource.filter.transform.}{NumberVectorFeatureSelectionFilter}%
{0.5.5}%
{}%
{Number Vector Feature Selection Filter}%

 Parser to project the ParsingResult obtained by a suitable base parser onto a
selected subset of attributes.

\classinfo%
{de.lmu.ifi.dbs.elki.datasource.filter.transform.}{NumberVectorRandomFeatureSelectionFilter}%
{0.4.0}%
{}%
{Number Vector Random Feature Selection Filter}%

Parser to project the ParsingResult obtained by a suitable base parser onto a
randomly selected subset of attributes.

\classinfo%
{de.lmu.ifi.dbs.elki.datasource.filter.transform.}{PerturbationFilter}%
{0.7.0}%
{DBLP:conf/ssdbm/ZimekCS14}%
{Data Perturbation for Outlier Detection Ensembles}%

A filter to perturb the values by adding micro-noise.

The added noise is generated, attribute-wise, by a Gaussian with mean=0 and a
specified standard deviation or by a uniform distribution with a specified
range. The standard deviation or the range can be scaled, attribute-wise, to
a given percentage of the original standard deviation in the data
distribution (assuming a Gaussian distribution there), or to a percentage of
the extension in each attribute ({\begingroup\ttfamily{}maximumValue - minimumValue\endgroup}).

This filter has a potentially wide use but has been implemented for the
following publication:

\classinfo%
{de.lmu.ifi.dbs.elki.datasource.filter.transform.}{ProjectionFilter}%
{0.6.0}%
{}%
{Projection Filter}%

Apply a projection to the data.

\pkginfo{7}%
{de.lmu.ifi.dbs.elki.datasource.filter.}{typeconversions}%
{Typeconversions}%

Filters to perform data type conversions.

\classinfo%
{de.lmu.ifi.dbs.elki.datasource.filter.typeconversions.}{ClassLabelFilter}%
{0.4.0}%
{}%
{Class Label Filter}%

Class that turns a label column into a class label column.

\classinfo%
{de.lmu.ifi.dbs.elki.datasource.filter.typeconversions.}{ClassLabelFromPatternFilter}%
{0.6.0}%
{}%
{Class Label From Pattern Filter}%

Streaming filter to derive an outlier class label.

\classinfo%
{de.lmu.ifi.dbs.elki.datasource.filter.typeconversions.}{ExternalIDFilter}%
{0.4.0}%
{}%
{External ID Filter}%

Class that turns a label column into an external ID column.

\classinfo%
{de.lmu.ifi.dbs.elki.datasource.filter.typeconversions.}{MultivariateTimeSeriesFilter}%
{0.7.0}%
{}%
{Multivariate Time Series Filter}%

Class to "fold" a flat number vector into a multivariate time series.

\classinfo%
{de.lmu.ifi.dbs.elki.datasource.filter.typeconversions.}{SparseVectorFieldFilter}%
{0.5.0}%
{}%
{Sparse Vector Field Filter}%

Class that turns sparse float vectors into a proper vector field, by setting
the maximum dimensionality for each vector.

\classinfo%
{de.lmu.ifi.dbs.elki.datasource.filter.typeconversions.}{SplitNumberVectorFilter}%
{0.4.0}%
{}%
{Split Number Vector Filter}%

Split an existing column into two types.

\classinfo%
{de.lmu.ifi.dbs.elki.datasource.filter.typeconversions.}{UncertainSplitFilter}%
{0.7.0}%
{}%
{Uncertain Split Filter}%

Filter to transform a single vector into a set of samples to interpret as
uncertain observation.

\classinfo%
{de.lmu.ifi.dbs.elki.datasource.filter.typeconversions.}{UncertainifyFilter}%
{0.7.0}%
{}%
{Uncertainify Filter}%

Filter class to transform a database containing vector fields into a database
containing {\begingroup\ttfamily{}UncertainObject\endgroup} fields by invoking a
{\begingroup\ttfamily{}Uncertainifier\endgroup} on each vector.

 The purpose for that is to use those transformed databases in experiments
regarding uncertain data in some way.

\classinfo%
{de.lmu.ifi.dbs.elki.datasource.filter.typeconversions.}{WeightedUncertainSplitFilter}%
{0.7.0}%
{}%
{Weighted Uncertain Split Filter}%

Filter to transform a single vector into a set of samples and weights to
interpret as uncertain observation.

\pkginfo{6}%
{de.lmu.ifi.dbs.elki.datasource.}{parser}%
{Parser}%

Parsers for different file formats and data types.

The general use-case for any parser is to create objects out of an
{\begingroup\ttfamily{}InputStream\endgroup} (e.g. by reading a data file).
The objects are packed in a
{\begingroup\ttfamily{}MultipleObjectsBundle\endgroup} which,
in turn, is used by a {\begingroup\ttfamily{}DatabaseConnection\endgroup}-Object
to fill a {\begingroup\ttfamily{}Database\endgroup} containing the corresponding objects.

By default (i.e., if the user does not specify any specific requests),
any {\begingroup\ttfamily{}KDDTask\endgroup} will
use the {\begingroup\ttfamily{}StaticArrayDatabase\endgroup} which,
in turn, will use a {\begingroup\ttfamily{}FileBasedDatabaseConnection\endgroup} and a {\begingroup\ttfamily{}NumberVectorLabelParser\endgroup} to parse a specified data file creating
a {\begingroup\ttfamily{}StaticArrayDatabase\endgroup} containing {\begingroup\ttfamily{}DoubleVector\endgroup}-Objects.

Thus, the standard procedure to use a data set of a real-valued vector space
is to prepare the data set in a file of the following format
(as suitable to {\begingroup\ttfamily{}NumberVectorLabelParser\endgroup}):

\begin{itemize}
\item One point per line, attributes separated by whitespace.
\item Several labels may be given per point. A label must not be parseable as double.
\item Lines starting with "\#" will be ignored.
\item An index can be specified to identify an entry to be treated as class label.
     This index counts all entries (numeric and labels as well) starting with 0.
\item Files can be gzip compressed.
\end{itemize}
 This file format is e.g. also suitable to gnuplot.

As an example file following these requirements consider e.g.:
\href{http://www.dbs.ifi.lmu.de/research/KDD/ELKI/datasets/example/exampledata.txt}{exampledata.txt}

\classinfo%
{de.lmu.ifi.dbs.elki.datasource.parser.}{ArffParser}%
{0.4.0}%
{}%
{ARFF File Format Parser}%

Parser to load WEKA .arff files into ELKI.

 This parser is quite hackish, and contains lots of not yet configurable
magic.

\classinfo%
{de.lmu.ifi.dbs.elki.datasource.parser.}{BitVectorLabelParser}%
{0.1}%
{}%
{Bit Vector Label Parser}%

Parser for parsing one BitVector per line, bits separated by whitespace.

 Several labels may be given per BitVector. A label must not be parseable as
Bit. Lines starting with "\#" will be ignored.

\classinfo%
{de.lmu.ifi.dbs.elki.datasource.parser.}{CSVReaderFormat}%
{0.1}%
{}%
{CSV Reader Format}%

Basic format factory for parsing CSV-like formats.

To read CSV files into ELKI, see {\begingroup\ttfamily{}NumberVectorLabelParser\endgroup}.

This class encapsulates csv format settings, that need to be parsed in
multiple places from ELKI, not only on input vector files.

\classinfo%
{de.lmu.ifi.dbs.elki.datasource.parser.}{CategorialDataAsNumberVectorParser}%
{0.6.0}%
{}%
{Categorial Data As Number Vector Parser}%

A very simple parser for categorial data, which will then be encoded as
numbers. This is closely modeled after the number vector parser.

\classinfo%
{de.lmu.ifi.dbs.elki.datasource.parser.}{ClusteringVectorParser}%
{0.7.0}%
{}%
{Clustering Vector Parser}%

Parser for simple clustering results in vector form, as written by
{\begingroup\ttfamily{}ClusteringVectorDumper\endgroup}.

This allows reading the output of multiple clustering runs, and
analyze the results using ELKI algorithm.

The input format is very simple, each line containing a sequence of cluster
assignments in integer form, and an optional label:

\begin{Verbatim}[fontsize=\verbatimsize]
0 0 1 1 0 First
0 0 0 1 2 Second
\end{Verbatim}
 represents two clusterings for 5 objects. The first clustering has two
clusters, the second contains three clusters.

\classinfo%
{de.lmu.ifi.dbs.elki.datasource.parser.}{LibSVMFormatParser}%
{0.7.0}%
{}%
{libSVM Format Parser}%

Parser to read libSVM format files.

 The format of libSVM is roughly specified in the README given:

\begin{Verbatim}[fontsize=\verbatimsize]
<label> <index1>:<value1> <index2>:<value2> ...
\end{Verbatim}
 
i.e. a mandatory integer class label in the beginning followed by a classic
sparse vector representation of the data. indexes are integers, starting at 1
(Note that ELKI uses 0-based indexing, so we will map these to index-1) to
not always have a constant-0 dimension 0.

 The libSVM FAQ states that you can also put comments into the file, separated
by a hash: \texttt{\#}, but they must not contain colons and are not
officially supported.\\
 ELKI will simply stop parsing a line when encountering a \texttt{\#}.

\classinfo%
{de.lmu.ifi.dbs.elki.datasource.parser.}{NumberVectorLabelParser}%
{0.1}%
{}%
{Number Vector Label Parser}%

Parser for a simple CSV type of format, with columns separated by the given
pattern (default: whitespace).

Several labels may be given per point. A label must not be parseable as
double. Lines starting with "\#" will be ignored.

An index can be specified to identify an entry to be treated as class label.
This index counts all entries (numeric and labels as well) starting with 0.

\classinfo%
{de.lmu.ifi.dbs.elki.datasource.parser.}{SimplePolygonParser}%
{0.4.0}%
{}%
{Simple Polygon Parser}%

Parser to load polygon data (2D and 3D only) from a simple format. One record
per line, points separated by whitespace, numbers separated by colons.
Multiple polygons components can be separated using {\begingroup\ttfamily{}--\endgroup}.

 Unparseable parts will be treated as labels.

\classinfo%
{de.lmu.ifi.dbs.elki.datasource.parser.}{SimpleTransactionParser}%
{0.7.0}%
{}%
{Simple Transaction Parser}%

Simple parser for transactional data, such as market baskets.

To keep the input format simple and readable, all tokens are assumed to be of
text and separated by whitespace, and each transaction is on a separate line.

An example file containing two transactions looks like this

\begin{Verbatim}[fontsize=\verbatimsize]
bread butter milk
paste tomato basil
\end{Verbatim}

\classinfo%
{de.lmu.ifi.dbs.elki.datasource.parser.}{SparseNumberVectorLabelParser}%
{0.2}%
{}%
{Sparse Vector Label Parser}%

 Parser for parsing one point per line, attributes separated by whitespace.

 Several labels may be given per point. A label must not be parseable as
double. Lines starting with "\#" will be ignored.

 A line is expected in the following format: The first entry of each line is
the number of attributes with coordinate value not zero. Subsequent entries
are of the form {\begingroup\ttfamily{}index value \endgroup} each, where index is the number of
the corresponding dimension, and value is the value of the corresponding
attribute. A complete line then could look like this:

\begin{Verbatim}[fontsize=\verbatimsize]
3 7 12.34 8 56.78 11 1.234 objectlabel
\end{Verbatim}
 where {\begingroup\ttfamily{}3\endgroup} indicates there are three attributes set,
{\begingroup\ttfamily{}7,8,11\endgroup} are the attributes indexes and there is a non-numerical
object label.

 An index can be specified to identify an entry to be treated as class label.
This index counts all entries (numeric and labels as well) starting with 0.

\classinfo%
{de.lmu.ifi.dbs.elki.datasource.parser.}{StringParser}%
{0.6.0}%
{}%
{String Parser}%

Parser that loads a text file for use with string similarity measures.

The parser produces two relations: the first of type String, the second of
type label list, which contains the same data for convenience.

\classinfo%
{de.lmu.ifi.dbs.elki.datasource.parser.}{TermFrequencyParser}%
{0.4.0}%
{}%
{Term Frequency Parser}%

A parser to load term frequency data, which essentially are sparse vectors
with text keys.

 Parse a file containing term frequencies. The expected format is:

\begin{Verbatim}[fontsize=\verbatimsize]
rowlabel1 term1 <freq> term2 <freq> ...
rowlabel2 term1 <freq> term3 <freq> ...
\end{Verbatim}
 
Terms must not contain the separator character!

 If your data does not contain frequencies, you can maybe use
{\begingroup\ttfamily{}SimpleTransactionParser\endgroup} instead.

\pkginfo{5}%
{de.lmu.ifi.dbs.elki.}{distance}%
{Distance}%

\pkginfo{6}%
{de.lmu.ifi.dbs.elki.distance.}{distancefunction}%
{Distancefunction}%

Distance functions for use within ELKI.

\paragraph{Distance functions}\hfill\\

There are three basic types of distance functions:

\begin{itemize}
\item {\begingroup\ttfamily{}Primitive Distance Function\endgroup}s that can be computed for any two objects.
\item {\begingroup\ttfamily{}DBID Distance Function\endgroup}s, that are only defined for object IDs, e.g. an external distance matrix
\item {\begingroup\ttfamily{}Index-Based Distance Function\endgroup}s, that require an indexing/preprocessing step, and are then valid for existing database objects.
\end{itemize}
 These types differ significantly in both implementation and use.

\paragraph{Using distance functions}\hfill\\

 As a 'consumer' of distances, you usually do not care about the type of distance function you
want to use. To facilitate this, a distance function can be bound to a database by calling
the 'instantiate' method to obtain a {\begingroup\ttfamily{}DistanceQuery\endgroup} object.
A distance query is a best-effort adapter for the given distance function. Usually, you pass it
two DBIDs and get the distance value back. When required, the adapter will get the appropriate
records from the database needed to compute the distance.

 Note: instantiating a preprocessor based distance will invoke the preprocessing step.
It is recommended to do this as soon as possible, and only instantiate the query once,
then pass the query object through the various methods.

\paragraph{Code example}\hfill\\
{\begingroup\ttfamily{}DistanceQuery<V> distanceQuery = database.getDistanceQuery(EuclideanDistanceFunction.STATIC);
\endgroup}

\classinfo%
{de.lmu.ifi.dbs.elki.distance.distancefunction.}{ArcCosineDistanceFunction}%
{0.2}%
{}%
{Arc Cosine Distance Function}%

Arcus cosine distance function for feature vectors.

 The arc cosine distance is computed as the arcus from the cosine similarity
value, i.e., {\begingroup\ttfamily{}arccos(<v1,v2>)\endgroup}.

 Cosine similarity is defined as
\[ \tfrac{\vec{x}\cdot\vec{y}}{||a||\cdot||b||} \]
Arcus cosine distance then is
\[ \text{arccos} \tfrac{\vec{x}\cdot\vec{y}}{||a||\cdot||b||} \in [0;\pi] \]

 {\begingroup\ttfamily{}CosineDistanceFunction\endgroup} is a bit less expensive, and will yield the
same ranking of neighbors.

\classinfo%
{de.lmu.ifi.dbs.elki.distance.distancefunction.}{ArcCosineUnitlengthDistanceFunction}%
{0.7.5}%
{}%
{Arc Cosine Unitlength Distance Function}%

Arcus cosine distance function for feature vectors.

 The arc cosine distance is computed as the arcus from the cosine similarity
value, i.e., {\begingroup\ttfamily{}arccos(<v1,v2>)\endgroup}.

 Cosine similarity is defined as
\[ \tfrac{\vec{x}\cdot\vec{y}}{||a||\cdot||b||}
=_{||a||=||b||=1} \vec{x}\cdot\vec{y} \]
Arcus cosine distance then is
\[ \operatorname{arccos} \tfrac{\vec{x}\cdot\vec{y}}{||a||\cdot||b||}
=_{||a||=||b||=1} \operatorname{arccos} \vec{x}\cdot\vec{y} \in [0;\pi] \]

 This implementation assumes that \(||a||=||b||=1\). If this does not
hold for your data, use {\begingroup\ttfamily{}ArcCosineDistanceFunction\endgroup} instead!

 {\begingroup\ttfamily{}CosineUnitlengthDistanceFunction\endgroup} is a bit less expensive, and will
yield the same ranking of neighbors.

\classinfo%
{de.lmu.ifi.dbs.elki.distance.distancefunction.}{BrayCurtisDistanceFunction}%
{0.6.0}%
{journals/misc/Sorensen48,doi:10.2307/1942268,doi:10.2307/1932409}%
{Bray Curtis Distance Function}%

Bray-Curtis distance function / Sørensen–Dice coefficient for continuous
vector spaces (not only binary data).

\classinfo%
{de.lmu.ifi.dbs.elki.distance.distancefunction.}{CanberraDistanceFunction}%
{0.5.0}%
{doi:10.1093/comjnl/9.1.60}%
{Canberra Distance Function}%

Canberra distance function, a variation of Manhattan distance.

 Canberra distance is defined as:
\[ \text{Canberra}(\vec{x},\vec{y}) :=
\sum\nolimits_i \tfrac{|x_i-y_i|}{|x_i|+|y_i|} \]

\classinfo%
{de.lmu.ifi.dbs.elki.distance.distancefunction.}{ClarkDistanceFunction}%
{0.6.0}%
{doi:10.1007/978-3-642-00234-2}%
{Clark Distance Function}%

Clark distance function for vector spaces.

 Clark distance is defined as:
\[ \text{Clark}(\vec{x},\vec{y}) :=
\sqrt{\tfrac{1}{d}\sum\nolimits_i \left(\tfrac{|x_i-y_i|}{|x_i|+|y_i|}\right)^2} \]

\classinfo%
{de.lmu.ifi.dbs.elki.distance.distancefunction.}{CosineDistanceFunction}%
{0.1}%
{}%
{Cosine Distance Function}%

Cosine distance function for feature vectors.

 The cosine distance is computed from the cosine similarity by
{\begingroup\ttfamily{}1-(cosine similarity)\endgroup}.

 Cosine similarity is defined as
\[ \tfrac{\vec{x}\cdot\vec{y}}{||a||\cdot||b||} \]
Cosine distance then is defined as
\[ 1 - \tfrac{\vec{x}\cdot\vec{y}}{||a||\cdot||b||} \in [0;2] \]

 {\begingroup\ttfamily{}ArcCosineDistanceFunction\endgroup} may sometimes be more appropriate, but also
more computationally expensive.

\classinfo%
{de.lmu.ifi.dbs.elki.distance.distancefunction.}{CosineUnitlengthDistanceFunction}%
{0.7.5}%
{}%
{Cosine Unitlength Distance Function}%

Cosine distance function for unit length feature vectors.

 The cosine distance is computed from the cosine similarity by
{\begingroup\ttfamily{}1-(cosine similarity)\endgroup}.

 Cosine similarity is defined as
\[ \tfrac{\vec{x}\cdot\vec{y}}{||a||\cdot||b||}
=_{||a||=||b||=1} \vec{x}\cdot\vec{y}
\]
Cosine distance then is defined as
\[ 1 - \tfrac{\vec{x}\cdot\vec{y}}{||a||\cdot||b||}
=_{||a||=||b||=1} 1-\vec{x}\cdot\vec{y} \in [0;2] \]

 This implementation assumes that \(||a||=||b||=1\). If this does not
hold for your data, use {\begingroup\ttfamily{}CosineDistanceFunction\endgroup} instead!

 {\begingroup\ttfamily{}ArcCosineUnitlengthDistanceFunction\endgroup} may sometimes be more
appropriate, but also more computationally expensive.

\classinfo%
{de.lmu.ifi.dbs.elki.distance.distancefunction.}{MahalanobisDistanceFunction}%
{0.7.5}%
{journals/misc/Mahalanobis36}%
{Mahalanobis Distance Function}%

Mahalanobis quadratic form distance for feature vectors.

 For a weight matrix M, this distance is defined as
\[ \text{Mahalanobis}_M(\vec{x},\vec{y}) :=
\sqrt{(\vec{x}-\vec{y})^T \cdot M \cdot (\vec{x}-\vec{y})} \]

\classinfo%
{de.lmu.ifi.dbs.elki.distance.distancefunction.}{RandomStableDistanceFunction}%
{0.4.0}%
{}%
{Random Stable Distance Function}%

This is a dummy distance providing random values (obviously not metrical),
useful mostly for unit tests and baseline evaluations: obviously this
distance provides no benefit whatsoever.

This distance is based on the combined hash codes of the two objects queried,
if they are different. Extra caution is done to ensure symmetry and objects
with the same ID will have a distance of 0. Obviously this distance is not
metrical.

\classinfo%
{de.lmu.ifi.dbs.elki.distance.distancefunction.}{SharedNearestNeighborJaccardDistanceFunction}%
{0.4.0}%
{}%
{Shared Nearest Neighbor Jaccard Distance Function}%

SharedNearestNeighborJaccardDistanceFunction computes the Jaccard
coefficient, which is a proper distance metric.

\classinfo%
{de.lmu.ifi.dbs.elki.distance.distancefunction.}{WeightedCanberraDistanceFunction}%
{0.5.0}%
{}%
{Weighted Canberra Distance Function}%

Weighted Canberra distance function, a variation of Manhattan distance.

\pkginfo{7}%
{de.lmu.ifi.dbs.elki.distance.distancefunction.}{adapter}%
{Adapter}%

Distance functions deriving distances from e.g. similarity measures

\classinfo%
{de.lmu.ifi.dbs.elki.distance.distancefunction.adapter.}{ArccosSimilarityAdapter}%
{0.2}%
{}%
{Arccos Similarity Adapter}%

Adapter from a normalized similarity function to a distance function using
{\begingroup\ttfamily{}arccos(sim)\endgroup}.

\classinfo%
{de.lmu.ifi.dbs.elki.distance.distancefunction.adapter.}{LinearAdapterLinear}%
{0.2}%
{}%
{Linear Adapter Linear}%

Adapter from a normalized similarity function to a distance function using
{\begingroup\ttfamily{}1 - sim\endgroup}.

\classinfo%
{de.lmu.ifi.dbs.elki.distance.distancefunction.adapter.}{LnSimilarityAdapter}%
{0.2}%
{}%
{Ln Similarity Adapter}%

Adapter from a normalized similarity function to a distance function using
{\begingroup\ttfamily{}-log(sim)\endgroup}.

\pkginfo{7}%
{de.lmu.ifi.dbs.elki.distance.distancefunction.}{colorhistogram}%
{Colorhistogram}%

Distance functions using correlations.

\classinfo%
{de.lmu.ifi.dbs.elki.distance.distancefunction.colorhistogram.}{HSBHistogramQuadraticDistanceFunction}%
{0.3}%
{DBLP:conf/mm/SmithC96}%
{HSB Histogram Quadratic Distance Function}%

Distance function for HSB color histograms based on a quadratic form and
color similarity.

 The matrix is filled according to:

 VisualSEEk: a fully automated content-based image query system\\
 J. R. Smith, S. F. Chang\\
 Proc. 4th ACM Int. Conf. on Multimedia 1997

\classinfo%
{de.lmu.ifi.dbs.elki.distance.distancefunction.colorhistogram.}{HistogramIntersectionDistanceFunction}%
{0.3}%
{DBLP:journals/ijcv/SwainB91}%
{Histogram Intersection Distance Function}%

Intersection distance for color histograms.

 Distance function for color histograms that emphasizes 'strong' bins.

\classinfo%
{de.lmu.ifi.dbs.elki.distance.distancefunction.colorhistogram.}{RGBHistogramQuadraticDistanceFunction}%
{0.3}%
{DBLP:journals/pami/HafnerSEFN95}%
{RGB Histogram Quadratic Distance Function}%

Distance function for RGB color histograms based on a quadratic form and
color similarity.

\pkginfo{7}%
{de.lmu.ifi.dbs.elki.distance.distancefunction.}{correlation}%
{Correlation}%

Distance functions using correlations.

\classinfo%
{de.lmu.ifi.dbs.elki.distance.distancefunction.correlation.}{AbsolutePearsonCorrelationDistanceFunction}%
{0.7.0}%
{}%
{Absolute Pearson Correlation Distance Function}%

Absolute Pearson correlation distance function for feature vectors.

The absolute Pearson correlation distance is computed from the Pearson
correlation coefficient {\begingroup\ttfamily{}r\endgroup} as: {\begingroup\ttfamily{}1-abs(r)\endgroup}.

The distance between two vectors will be low (near 0), if their attribute
values are dimension-wise strictly positively or negatively correlated, it
will be high (near 1), if their attribute values are dimension-wise
uncorrelated.

\classinfo%
{de.lmu.ifi.dbs.elki.distance.distancefunction.correlation.}{AbsoluteUncenteredCorrelationDistanceFunction}%
{0.7.0}%
{}%
{Absolute Uncentered Correlation Distance Function}%

Absolute uncentered correlation distance function for feature vectors.

This is highly similar to {\begingroup\ttfamily{}AbsolutePearsonCorrelationDistanceFunction\endgroup},
but uses a fixed mean of 0 instead of the sample mean.

\classinfo%
{de.lmu.ifi.dbs.elki.distance.distancefunction.correlation.}{PearsonCorrelationDistanceFunction}%
{0.3}%
{}%
{Pearson Correlation Distance Function}%

Pearson correlation distance function for feature vectors.

The Pearson correlation distance is computed from the Pearson correlation
coefficient {\begingroup\ttfamily{}r\endgroup} as: {\begingroup\ttfamily{}1-r\endgroup}. Hence, possible values of
this distance are between 0 and 2.

The distance between two vectors will be low (near 0), if their attribute
values are dimension-wise strictly positively correlated, it will be high
(near 2), if their attribute values are dimension-wise strictly negatively
correlated. For Features with uncorrelated attributes, the distance value
will be intermediate (around 1).

\classinfo%
{de.lmu.ifi.dbs.elki.distance.distancefunction.correlation.}{SquaredPearsonCorrelationDistanceFunction}%
{0.3}%
{}%
{Squared Pearson Correlation Distance Function}%

Squared Pearson correlation distance function for feature vectors.

 The squared Pearson correlation distance is computed from the
Pearson correlation coefficient \(r\) as: \(1-r^2\).
Hence, possible values of this distance are between 0 and 1.

 The distance between two vectors will be low (near 0), if their attribute
values are dimension-wise strictly positively or negatively correlated.
For features with uncorrelated attributes, the distance value will be high
(near 1).

\classinfo%
{de.lmu.ifi.dbs.elki.distance.distancefunction.correlation.}{SquaredUncenteredCorrelationDistanceFunction}%
{0.7.0}%
{}%
{Squared Uncentered Correlation Distance Function}%

Squared uncentered correlation distance function for feature vectors.

This is highly similar to {\begingroup\ttfamily{}SquaredPearsonCorrelationDistanceFunction\endgroup},
but uses a fixed mean of 0 instead of the sample mean.

\classinfo%
{de.lmu.ifi.dbs.elki.distance.distancefunction.correlation.}{UncenteredCorrelationDistanceFunction}%
{0.7.0}%
{}%
{Uncentered Correlation Distance Function}%

Uncentered correlation distance.

This is highly similar to {\begingroup\ttfamily{}PearsonCorrelationDistanceFunction\endgroup}, but
uses a fixed mean of 0 instead of the sample mean.

\classinfo%
{de.lmu.ifi.dbs.elki.distance.distancefunction.correlation.}{WeightedPearsonCorrelationDistanceFunction}%
{0.4.0}%
{}%
{Weighted Pearson Correlation Distance Function}%

Pearson correlation distance function for feature vectors.

The Pearson correlation distance is computed from the Pearson correlation
coefficient {\begingroup\ttfamily{}r\endgroup} as: {\begingroup\ttfamily{}1-r\endgroup}. Hence, possible values of
this distance are between 0 and 2.

The distance between two vectors will be low (near 0), if their attribute
values are dimension-wise strictly positively correlated, it will be high
(near 2), if their attribute values are dimension-wise strictly negatively
correlated. For Features with uncorrelated attributes, the distance value
will be intermediate (around 1).

This variation is for weighted dimensions.

\classinfo%
{de.lmu.ifi.dbs.elki.distance.distancefunction.correlation.}{WeightedSquaredPearsonCorrelationDistanceFunction}%
{0.4.0}%
{}%
{Weighted Squared Pearson Correlation Distance Function}%

Weighted squared Pearson correlation distance function for feature vectors.

 The squared Pearson correlation distance is computed from the
Pearson correlation coefficient \(r\) as: \(1-r^2\).
Hence, possible values of this distance are between 0 and 1.

 The distance between two vectors will be low (near 0), if their attribute
values are dimension-wise strictly positively or negatively correlated.
For features with uncorrelated attributes, the distance value will be high
(near 1).

 This variation is for weighted dimensions.

\pkginfo{7}%
{de.lmu.ifi.dbs.elki.distance.distancefunction.}{external}%
{External}%

Distance functions using external data sources.

\classinfo%
{de.lmu.ifi.dbs.elki.distance.distancefunction.external.}{AsciiDistanceParser}%
{0.1}%
{}%
{Ascii Distance Parser}%

Parser for parsing one distance value per line.

 A line must have the following format: {\begingroup\ttfamily{}id1 id2 distanceValue\endgroup}, where
id1 and id2 are integers starting at 0 representing the two ids belonging to
the distance value. Lines starting with "\#" will be ignored.

\classinfo%
{de.lmu.ifi.dbs.elki.distance.distancefunction.external.}{DiskCacheBasedDoubleDistanceFunction}%
{0.2}%
{}%
{Disk Cache Based Double Distance Function}%

Distance function that is based on double distances given by a distance
matrix of an external binary matrix file.

\classinfo%
{de.lmu.ifi.dbs.elki.distance.distancefunction.external.}{DiskCacheBasedFloatDistanceFunction}%
{0.2}%
{}%
{Disk Cache Based Float Distance Function}%

Distance function that is based on float distances given by a distance matrix
of an external binary matrix file.

\classinfo%
{de.lmu.ifi.dbs.elki.distance.distancefunction.external.}{FileBasedSparseDoubleDistanceFunction}%
{0.1}%
{}%
{File Based Sparse Double Distance Function}%

Distance function that is based on double distances given by a distance
matrix of an external ASCII file.

 Note: parsing an ASCII file is rather expensive.

 See {\begingroup\ttfamily{}AsciiDistanceParser\endgroup} for the default input format.

\classinfo%
{de.lmu.ifi.dbs.elki.distance.distancefunction.external.}{FileBasedSparseFloatDistanceFunction}%
{0.1}%
{}%
{File Based Sparse Float Distance Function}%

Distance function that is based on float distances given by a distance matrix
of an external ASCII file.

 Note: parsing an ASCII file is rather expensive.

 See {\begingroup\ttfamily{}AsciiDistanceParser\endgroup} for the default input format.

\pkginfo{7}%
{de.lmu.ifi.dbs.elki.distance.distancefunction.}{geo}%
{Geo}%

Geographic (earth) distance functions.

\classinfo%
{de.lmu.ifi.dbs.elki.distance.distancefunction.geo.}{DimensionSelectingLatLngDistanceFunction}%
{0.4.0}%
{DBLP:conf/ssd/SchubertZK13}%
{Dimension Selecting Lat Lng Distance Function}%

Distance function for 2D vectors in Latitude, Longitude form.

 The input data must be in degrees (not radians), and the output distance will
be in meters (see {\begingroup\ttfamily{}EarthModel.distanceDeg(double, double, double, double)\endgroup}).

 This implementation allows index accelerated queries using R*-trees (by
providing a point-to-rectangle minimum distance).

\classinfo%
{de.lmu.ifi.dbs.elki.distance.distancefunction.geo.}{LatLngDistanceFunction}%
{0.4.0}%
{DBLP:conf/ssd/SchubertZK13}%
{LatLng Distance Function}%

Distance function for 2D vectors in Latitude, Longitude form.

 The input data must be in degrees (not radians), and the output distance will
be in meters (see {\begingroup\ttfamily{}EarthModel.distanceDeg(double, double, double, double)\endgroup}).

 This implementation allows index accelerated queries using R*-trees (by
providing a point-to-rectangle minimum distance).

\classinfo%
{de.lmu.ifi.dbs.elki.distance.distancefunction.geo.}{LngLatDistanceFunction}%
{0.4.0}%
{DBLP:conf/ssd/SchubertZK13}%
{LngLat Distance Function}%

Distance function for 2D vectors in Longitude, Latitude form.

 The input data must be in degrees (not radians), and the output distance will
be in meters (see {\begingroup\ttfamily{}EarthModel.distanceDeg(double, double, double, double)\endgroup}).

 This implementation allows index accelerated queries using R*-trees (by
providing a point-to-rectangle minimum distance).

\pkginfo{7}%
{de.lmu.ifi.dbs.elki.distance.distancefunction.}{histogram}%
{Histogram}%

Distance functions for one-dimensional histograms.

\classinfo%
{de.lmu.ifi.dbs.elki.distance.distancefunction.histogram.}{HistogramMatchDistanceFunction}%
{0.6.0}%
{journals/misc/Vaserstein69}%
{Histogram Match Distance Function}%

Distance function based on histogram matching, i.e. Manhattan distance on the
cumulative density function.

 This distance function assumes there exist a natural order in the vectors,
i.e. they should be some 1-dimensional histogram.

 This is also known as Earth Movers Distance (EMD), 1st Mallows distance or
1st Wasserstein metric (also Vasershtein metric), for the special case of a
one-dimensional histogram, where the cost is linear in the number of bins to
transport.

\classinfo%
{de.lmu.ifi.dbs.elki.distance.distancefunction.histogram.}{KolmogorovSmirnovDistanceFunction}%
{0.6.0}%
{}%
{Kolmogorov Smirnov Distance Function}%

Distance function based on the Kolmogorov-Smirnov goodness of fit test.

This distance function assumes there exist a natural order in the vectors,
i.e. they should be some 1-dimensional histogram.

\pkginfo{7}%
{de.lmu.ifi.dbs.elki.distance.distancefunction.}{minkowski}%
{Minkowski}%

 Minkowski space L$_{p}$ norms such as the popular Euclidean and
Manhattan distances.

\classinfo%
{de.lmu.ifi.dbs.elki.distance.distancefunction.minkowski.}{EuclideanDistanceFunction}%
{0.1}%
{}%
{Euclidean Distance Function}%

Euclidean distance for {\begingroup\ttfamily{}NumberVector\endgroup}s.

 Euclidean distance is defined as:
\[ \text{Euclidean}(\vec{x},\vec{y}) := \sqrt{\sum\nolimits_i (x_i-y_i)^2} \]

\classinfo%
{de.lmu.ifi.dbs.elki.distance.distancefunction.minkowski.}{LPNormDistanceFunction}%
{0.1}%
{}%
{LP Norm Distance Function}%

L$_{p}$-Norm (Minkowski norms) are a family of distances for
{\begingroup\ttfamily{}NumberVector\endgroup}s.

 The L$_{p}$ distance is defined as:
\[ L_p(\vec{x},\vec{y}) := \left(\sum\nolimits_i (x_i-y_i)\right)^{1/p} \]

 For p >= 1 this is a metric. For p=1, this yields the well known
{\begingroup\ttfamily{}ManhattanDistanceFunction\endgroup}, for p = 2 the standard
{\begingroup\ttfamily{}EuclideanDistanceFunction\endgroup}.

\classinfo%
{de.lmu.ifi.dbs.elki.distance.distancefunction.minkowski.}{ManhattanDistanceFunction}%
{0.1}%
{}%
{Manhattan Distance Function}%

Manhattan distance for {\begingroup\ttfamily{}NumberVector\endgroup}s.

 Manhattan distance is defined as:
\[ \text{Manhattan}(\vec{x},\vec{y}) := \sum_i |x_i-y_i| \]

\classinfo%
{de.lmu.ifi.dbs.elki.distance.distancefunction.minkowski.}{MaximumDistanceFunction}%
{0.3}%
{}%
{Maximum Distance Function}%

Maximum distance for {\begingroup\ttfamily{}NumberVector\endgroup}s.

 The maximum distance is defined as:
\[ \text{Maximum}(\vec{x},\vec{y}) := \max_i |x_i-y_i| \]
and can be seen as limiting case of the {\begingroup\ttfamily{}LPNormDistanceFunction\endgroup} for \( p \rightarrow \infty \).

\classinfo%
{de.lmu.ifi.dbs.elki.distance.distancefunction.minkowski.}{MinimumDistanceFunction}%
{0.3}%
{}%
{Minimum Distance Function}%

Minimum distance for {\begingroup\ttfamily{}NumberVector\endgroup}s.

 Minimum distance is defined as:
\[ \text{Minimum}_p(\vec{x},\vec{y}) := \min_i |x_i-y_i| \]

 This is not a metric, but can sometimes be useful as a lower bound.

\classinfo%
{de.lmu.ifi.dbs.elki.distance.distancefunction.minkowski.}{SparseEuclideanDistanceFunction}%
{0.5.0}%
{}%
{Sparse Euclidean Distance Function}%

Euclidean distance function, optimized for {\begingroup\ttfamily{}SparseNumberVector\endgroup}s.

 Euclidean distance is defined as:
\[ \text{Euclidean}(\vec{x},\vec{y}) := \sqrt{\sum\nolimits_i (x_i-y_i)^2} \]

 For sparse vectors, we can skip those i where both vectors are 0.

\classinfo%
{de.lmu.ifi.dbs.elki.distance.distancefunction.minkowski.}{SparseLPNormDistanceFunction}%
{0.5.0}%
{}%
{Sparse LP Norm Distance Function}%

L$_{p}$-Norm, optimized for {\begingroup\ttfamily{}SparseNumberVector\endgroup}s.

 The L$_{p}$ distance is defined as:
\[ L_p(\vec{x},\vec{y}) := \left(\sum\nolimits_i (x_i-y_i)\right)^{1/p} \]

\classinfo%
{de.lmu.ifi.dbs.elki.distance.distancefunction.minkowski.}{SparseManhattanDistanceFunction}%
{0.5.0}%
{}%
{Sparse Manhattan Distance Function}%

Manhattan distance, optimized for {\begingroup\ttfamily{}SparseNumberVector\endgroup}s.

Manhattan distance is defined as:
\[ \text{Manhattan}(\vec{x},\vec{y}) := \sum_i |x_i-y_i| \]

\classinfo%
{de.lmu.ifi.dbs.elki.distance.distancefunction.minkowski.}{SparseMaximumDistanceFunction}%
{0.5.0}%
{}%
{Sparse Maximum Distance Function}%

Maximum distance, optimized for {\begingroup\ttfamily{}SparseNumberVector\endgroup}s.

The maximum distance is defined as:
\[ \text{Maximum}(\vec{x},\vec{y}) := \max_i |x_i-y_i| \]

and can be seen as limiting case of the {\begingroup\ttfamily{}LPNormDistanceFunction\endgroup} for \( p \rightarrow \infty \).

\classinfo%
{de.lmu.ifi.dbs.elki.distance.distancefunction.minkowski.}{SquaredEuclideanDistanceFunction}%
{0.1}%
{}%
{Squared Euclidean Distance Function}%

Squared Euclidean distance, optimized for {\begingroup\ttfamily{}SparseNumberVector\endgroup}s. This
results in the same rankings as regular Euclidean distance, but saves
computing the square root.

 Squared Euclidean is defined as:
\[ \text{Euclidean}^2(\vec{x},\vec{y}) := \sum_i (x_i-y_i)^2 \]

\classinfo%
{de.lmu.ifi.dbs.elki.distance.distancefunction.minkowski.}{WeightedEuclideanDistanceFunction}%
{0.4.0}%
{}%
{Weighted Euclidean Distance Function}%

Weighted Euclidean distance for {\begingroup\ttfamily{}NumberVector\endgroup}s.

 Weighted Euclidean distance is defined as:
\[ \text{Euclidean}_{\vec{w}}(\vec{x},\vec{y}) :=
\sqrt{\sum\nolimits_i w_i (x_i-y_i)^2} \]

\classinfo%
{de.lmu.ifi.dbs.elki.distance.distancefunction.minkowski.}{WeightedLPNormDistanceFunction}%
{0.4.0}%
{}%
{Weighted LP Norm Distance Function}%

Weighted version of the Minkowski L$_{p}$ norm distance for
{\begingroup\ttfamily{}NumberVector\endgroup}.

 Weighted L$_{p}$ Norms are defined as:
\[ L_{p,\vec{w}}(\vec{x},\vec{y}) := \left(\sum\nolimits_i
w_i |x_i-y_i|^p\right)^{1/p} \]

\classinfo%
{de.lmu.ifi.dbs.elki.distance.distancefunction.minkowski.}{WeightedManhattanDistanceFunction}%
{0.4.0}%
{}%
{Weighted Manhattan Distance Function}%

Weighted version of the Manhattan (L$_{1}$) metric.

 Weighted Manhattan distance is defined as:
\[ \text{Manhattan}_{\vec{w}}(\vec{x},\vec{y}) := \sum_i w_i |x_i-y_i| \]

\classinfo%
{de.lmu.ifi.dbs.elki.distance.distancefunction.minkowski.}{WeightedMaximumDistanceFunction}%
{0.6.0}%
{}%
{Weighted Maximum Distance Function}%

Weighted version of the maximum distance function for
{\begingroup\ttfamily{}NumberVector\endgroup}s.

 Weighted maximum distance is defined as:
\[ \text{Maximum}_{\vec{w}}(\vec{x},\vec{y}) := \max_i w_i |x_i-y_i| \]

\classinfo%
{de.lmu.ifi.dbs.elki.distance.distancefunction.minkowski.}{WeightedSquaredEuclideanDistanceFunction}%
{0.4.0}%
{}%
{Weighted Squared Euclidean Distance Function}%

Weighted squared Euclidean distance for {\begingroup\ttfamily{}NumberVector\endgroup}s. This results in the
same rankings as weighted Euclidean distance, but saves computing the square root.

 Weighted squared Euclidean is defined as:
\[ \text{Euclidean}^2_{\vec{w}}(\vec{x},\vec{y}) := \sum_i w_i (x_i-y_i)^2 \]

\pkginfo{7}%
{de.lmu.ifi.dbs.elki.distance.distancefunction.}{probabilistic}%
{Probabilistic}%

Distance from probability theory, mostly divergences such as K-L-divergence,
J-divergence, F-divergence, χ²-divergence, etc.

\classinfo%
{de.lmu.ifi.dbs.elki.distance.distancefunction.probabilistic.}{ChiDistanceFunction}%
{0.7.5}%
{DBLP:journals/tit/EndresS03,DBLP:conf/iccv/PuzichaRTB99}%
{Chi Distance Function}%

χ distance function, symmetric version.
This is the square root of the {\begingroup\ttfamily{}ChiSquaredDistanceFunction\endgroup}, and can
serve as a fast approximation to
{\begingroup\ttfamily{}SqrtJensenShannonDivergenceDistanceFunction\endgroup}.

 This implementation assumes \(\sum_i x_i=\sum_i y_i\), and is defined as:
\[ \chi(\vec{x},\vec{y}):= \sqrt{2 \sum\nolimits_i
\tfrac{(x_i-x_i)^2}{x_i+y_i}} \]

\classinfo%
{de.lmu.ifi.dbs.elki.distance.distancefunction.probabilistic.}{ChiSquaredDistanceFunction}%
{0.6.0}%
{DBLP:conf/iccv/PuzichaRTB99}%
{Chi Squared Distance Function}%

χ² distance function, symmetric version.

 This implementation assumes \(\sum_i x_i=\sum_i y_i\), and is defined as:
\[\chi^2(\vec{x},\vec{y}):=2\sum\nolimits_i \tfrac{(x_i-x_i)^2}{x_i+y_i}\]

\classinfo%
{de.lmu.ifi.dbs.elki.distance.distancefunction.probabilistic.}{FisherRaoDistanceFunction}%
{0.7.5}%
{doi:10.1007/978-3-642-00234-2,journals/bcalms/Rao45}%
{Fisher Rao Distance Function}%

Fisher-Rao riemannian metric for (discrete) probability distributions.
 \[ \text{Fisher-Rao}(\vec{x},\vec{y})
:= 2 \arccos \sum\nolimits_i \sqrt{p_iq_i} \]

\classinfo%
{de.lmu.ifi.dbs.elki.distance.distancefunction.probabilistic.}{HellingerDistanceFunction}%
{0.7.0}%
{doi:10.1007/978-3-642-00234-2,journals/mathematik/Hellinger1909}%
{Hellinger Distance Function}%

Hellinger metric / affinity / kernel, Bhattacharyya coefficient, fidelity
similarity, Matusita distance, Hellinger-Kakutani metric on a probability
distribution.

 We assume vectors represent normalized probability distributions. Then
\[\text{Hellinger}(\vec{x},\vec{y}):=
\sqrt{\tfrac12\sum\nolimits_i \left(\sqrt{x_i}-\sqrt{y_i}\right)^2 } \]

 The corresponding kernel / similarity is
\[ K_{\text{Hellinger}}(\vec{x},\vec{y}) := \sum\nolimits_i \sqrt{x_i y_i} \]

 If we have normalized probability distributions, we have the nice
property that
\( K_{\text{Hellinger}}(\vec{x},\vec{x}) = \sum\nolimits_i x_i = 1\).
and therefore \( K_{\text{Hellinger}}(\vec{x},\vec{y}) \in [0:1] \).

 Furthermore, we have the following relationship between this variant of the
distance and this kernel:
\[ \text{Hellinger}^2(\vec{x},\vec{y})
= \tfrac12\sum\nolimits_i \left(\sqrt{x_i}-\sqrt{y_i}\right)^2
= \tfrac12\sum\nolimits_i x_i + y_i - 2 \sqrt{x_i y_i} \]
\[ \text{Hellinger}^2(\vec{x},\vec{y})
= \tfrac12K_{\text{Hellinger}}(\vec{x},\vec{x})
+ \tfrac12K_{\text{Hellinger}}(\vec{y},\vec{y})
- K_{\text{Hellinger}}(\vec{x},\vec{y})
= 1 - K_{\text{Hellinger}}(\vec{x},\vec{y}) \]
which implies \(\text{Hellinger}(\vec{x},\vec{y}) \in [0;1]\),
and is very similar to the Euclidean distance and the linear kernel.

 From this, it follows trivially that Hellinger distance corresponds
to the kernel transformation
\(\phi:\vec{x}\mapsto(\tfrac12\sqrt{x_1},\ldots,\tfrac12\sqrt{x_d})\).

 Deza and Deza unfortunately also give a second definition, as:
\[\text{Hellinger-Deza}(\vec{x},\vec{y}):=\sqrt{2\sum\nolimits_i
\left(\sqrt{\tfrac{x_i}{\bar{x}}}-\sqrt{\tfrac{y_i}{\bar{y}}}\right)^2}\]
which has a built-in normalization, and a different scaling that is no longer
bound to $[0;1]$. The 2 in this definition likely should be a \(\frac12\).

 This distance is well suited for histograms, but it is then more efficient to
once normalize the histograms, apply the square roots, and then use Euclidean
distance (i.e., use the "kernel trick" in reverse, materializing the
transformation \(\phi\) given above).

\classinfo%
{de.lmu.ifi.dbs.elki.distance.distancefunction.probabilistic.}{JeffreyDivergenceDistanceFunction}%
{0.5.0}%
{DBLP:conf/iccv/PuzichaRTB99,DBLP:journals/tit/EndresS03,doi:10.1098/rspa.1946.0056,DBLP:journals/tit/Topsoe00}%
{Jeffrey Divergence Distance Function}%

Jeffrey Divergence for {\begingroup\ttfamily{}NumberVector\endgroup}s is a symmetric, smoothened
version of the {\begingroup\ttfamily{}KullbackLeiblerDivergenceAsymmetricDistanceFunction\endgroup}.
Topsøe called this "capacitory discrimination".
 \[JD(\vec{x},\vec{y}):= \sum\nolimits_i
x_i\log\tfrac{2x_i}{x_i+y_i}+y_i\log\tfrac{2y_i}{x_i+y_i}
= KL(\vec{x},\tfrac12(\vec{x}+\vec{y}))
+ KL(\vec{y},\tfrac12(\vec{x}+\vec{y}))\]

\classinfo%
{de.lmu.ifi.dbs.elki.distance.distancefunction.probabilistic.}{JensenShannonDivergenceDistanceFunction}%
{0.6.0}%
{doi:10.1007/978-3-642-00234-2,DBLP:journals/tit/EndresS03,DBLP:journals/tit/Lin91}%
{Jensen Shannon Divergence Distance Function}%

Jensen-Shannon Divergence for {\begingroup\ttfamily{}NumberVector\endgroup}s is a symmetric,
smoothened version of the
{\begingroup\ttfamily{}KullbackLeiblerDivergenceAsymmetricDistanceFunction\endgroup}.

 It essentially is the same as {\begingroup\ttfamily{}JeffreyDivergenceDistanceFunction\endgroup}, only
scaled by half. For completeness, we include both.

 \[JS(\vec{x},\vec{y}):=\tfrac12\sum\nolimits_i
x_i\log\tfrac{2x_i}{x_i+y_i}+y_i\log\tfrac{2y_i}{x_i+y_i}
= \tfrac12 KL(\vec{x},\tfrac12(\vec{x}+\vec{y}))
+ \tfrac12 KL(\vec{y},\tfrac12(\vec{x}+\vec{y}))\]

 There exists a variable definition where the two vectors are weighted with
\(\beta\) and \(1-\beta\), which for the common choice of \(\beta=\tfrac12\)
yields this version.

\classinfo%
{de.lmu.ifi.dbs.elki.distance.distancefunction.probabilistic.}{KullbackLeiblerDivergenceAsymmetricDistanceFunction}%
{0.6.0}%
{books/dover/Kullback59}%
{Kullback Leibler Divergence Asymmetric Distance Function}%

Kullback-Leibler divergence, also known as relative entropy,
information deviation, or just KL-distance (albeit asymmetric).
 \[KL(\vec{x},\vec{y}):=\sum\nolimits_i x_i\log\tfrac{x_i}{y_i}\]

 For a version with the arguments reversed, see
{\begingroup\ttfamily{}KullbackLeiblerDivergenceReverseAsymmetricDistanceFunction\endgroup}.

 For a symmetric version, see {\begingroup\ttfamily{}JeffreyDivergenceDistanceFunction\endgroup}.

\classinfo%
{de.lmu.ifi.dbs.elki.distance.distancefunction.probabilistic.}{KullbackLeiblerDivergenceReverseAsymmetricDistanceFunction}%
{0.6.0}%
{books/dover/Kullback59}%
{Kullback Leibler Divergence Reverse Asymmetric Distance Function}%

Kullback-Leibler divergence, also known as relative entropy, information
deviation or just KL-distance (albeit asymmetric).
 \[KL_R(\vec{x},\vec{y}):=\sum\nolimits_i y_i\log\tfrac{y_i}{x_i}
= KL(\vec{y},\vec{x})\]

 This version has the arguments reversed, see
{\begingroup\ttfamily{}KullbackLeiblerDivergenceAsymmetricDistanceFunction\endgroup} for the "forward"
version.

 For a symmetric version, see {\begingroup\ttfamily{}JeffreyDivergenceDistanceFunction\endgroup}.

\classinfo%
{de.lmu.ifi.dbs.elki.distance.distancefunction.probabilistic.}{SqrtJensenShannonDivergenceDistanceFunction}%
{0.6.0}%
{DBLP:journals/tit/EndresS03}%
{Sqrt Jensen Shannon Divergence Distance Function}%

The square root of Jensen-Shannon divergence is a metric.
 \[\sqrt{JS}(\vec{x},\vec{y}):=\sqrt{\tfrac12\sum\nolimits_i
x_i\log\tfrac{2x_i}{x_i+y_i}+y_i\log\tfrac{2y_i}{x_i+y_i}}
= \sqrt{JS(\vec{x},\vec{y})}\]

 A proof of triangle inequality (for "\(D_{PQ}\)") can be found in Endres and
Schindelin.

\classinfo%
{de.lmu.ifi.dbs.elki.distance.distancefunction.probabilistic.}{TriangularDiscriminationDistanceFunction}%
{0.7.5}%
{DBLP:journals/tit/Topsoe00}%
{Triangular Discrimination Distance Function}%

Triangular Discrimination has relatively tight upper and lower bounds to the
Jensen-Shannon divergence, but is much less expensive.
 \[\text{Triangular-Discrimination}(\vec{x},\vec{y}):=
\sum\nolimits_i \tfrac{|x_i-y_i|^2}{x_i+y_i}\]

 This distance function is meant for distribution vectors that sum to 1, and
does not work on negative values.

 See also {\begingroup\ttfamily{}TriangularDistanceFunction\endgroup} for a metric version.

\classinfo%
{de.lmu.ifi.dbs.elki.distance.distancefunction.probabilistic.}{TriangularDistanceFunction}%
{0.7.5}%
{DBLP:journals/corr/ConnorCVR16}%
{Triangular Distance Function}%

Triangular Distance has relatively tight upper and lower bounds to the
(square root of the) Jensen-Shannon divergence, but is much less expensive.
 \[\text{Triangular-Distance}(\vec{x},\vec{y}):=\sqrt{
\sum\nolimits_i \tfrac{|x_i-y_i|^2}{x_i+y_i}}\]

 This distance function is meant for distribution vectors that sum to 1, and
does not work on negative values.

 This differs from {\begingroup\ttfamily{}TriangularDistanceFunction\endgroup} simply by the square
root, which makes it a proper metric and a good approximation for the much
more expensive {\begingroup\ttfamily{}SqrtJensenShannonDivergenceDistanceFunction\endgroup}.

\pkginfo{7}%
{de.lmu.ifi.dbs.elki.distance.distancefunction.}{set}%
{Set}%

Distance functions for binary and set type data.

\classinfo%
{de.lmu.ifi.dbs.elki.distance.distancefunction.set.}{HammingDistanceFunction}%
{0.7.0}%
{doi:10.1002/j.1538-7305.1950.tb00463.x}%
{Hamming Distance Function}%

Computes the Hamming distance of arbitrary vectors - i.e. counting, on how
many places they differ.

\classinfo%
{de.lmu.ifi.dbs.elki.distance.distancefunction.set.}{JaccardSimilarityDistanceFunction}%
{0.6.0}%
{journals/misc/Jaccard1902}%
{Jaccard Similarity Distance Function}%

A flexible extension of Jaccard similarity to non-binary vectors.

 Jaccard coefficient is commonly defined as \(\frac{A\cap B}{A\cup B}\).

 We can extend this definition to non-binary vectors as follows:
\(\tfrac{|\{i\mid a_i = b_i\}|}{|\{i\mid a_i = 0 \wedge b_i = 0\}|}\)

 For binary vectors, this will obviously be the same quantity. However, this
version is more useful for categorical data.

\pkginfo{7}%
{de.lmu.ifi.dbs.elki.distance.distancefunction.}{strings}%
{Strings}%

Distance functions for strings.

\classinfo%
{de.lmu.ifi.dbs.elki.distance.distancefunction.strings.}{LevenshteinDistanceFunction}%
{0.6.0}%
{journals/misc/Levenshtein66}%
{Levenshtein Distance Function}%

Classic Levenshtein distance on strings.

\classinfo%
{de.lmu.ifi.dbs.elki.distance.distancefunction.strings.}{NormalizedLevenshteinDistanceFunction}%
{0.6.0}%
{journals/misc/Levenshtein66}%
{Normalized Levenshtein Distance Function}%

Levenshtein distance on strings, normalized by string length.

 Note: this is no longer a metric, the triangle inequality is violated.
Example: d("ab","bc")=1, d("ab", "abc")+d("abc","bc")=0.4+0.4=0.8

\pkginfo{7}%
{de.lmu.ifi.dbs.elki.distance.distancefunction.}{subspace}%
{Subspace}%

Distance functions based on subspaces.

\classinfo%
{de.lmu.ifi.dbs.elki.distance.distancefunction.subspace.}{OnedimensionalDistanceFunction}%
{0.1}%
{}%
{Onedimensional Distance Function}%

Distance function that computes the distance between feature vectors as the
absolute difference of their values in a specified dimension only.

\classinfo%
{de.lmu.ifi.dbs.elki.distance.distancefunction.subspace.}{SubspaceEuclideanDistanceFunction}%
{0.1}%
{}%
{Subspace Euclidean Distance Function}%

Euclidean distance function between {\begingroup\ttfamily{}NumberVector\endgroup}s only in specified
dimensions.

\classinfo%
{de.lmu.ifi.dbs.elki.distance.distancefunction.subspace.}{SubspaceLPNormDistanceFunction}%
{0.1}%
{}%
{Subspace LP Norm Distance Function}%

L$_{p}$-Norm distance function between {\begingroup\ttfamily{}NumberVector\endgroup}s only in
specified dimensions.

\classinfo%
{de.lmu.ifi.dbs.elki.distance.distancefunction.subspace.}{SubspaceManhattanDistanceFunction}%
{0.1}%
{}%
{Subspace Manhattan Distance Function}%

Manhattan distance function between {\begingroup\ttfamily{}NumberVector\endgroup}s only in specified
dimensions.

\classinfo%
{de.lmu.ifi.dbs.elki.distance.distancefunction.subspace.}{SubspaceMaximumDistanceFunction}%
{0.1}%
{}%
{Subspace Maximum Distance Function}%

Maximum distance function between {\begingroup\ttfamily{}NumberVector\endgroup}s only in specified
dimensions.

\pkginfo{7}%
{de.lmu.ifi.dbs.elki.distance.distancefunction.}{timeseries}%
{Timeseries}%

Distance functions designed for time series.
 
Note that some regular distance functions (e.g. Euclidean) are also used on time series.

\classinfo%
{de.lmu.ifi.dbs.elki.distance.distancefunction.timeseries.}{DTWDistanceFunction}%
{0.2}%
{DBLP:conf/kdd/BerndtC94}%
{Dynamic Time Warping Distance Function}%

Dynamic Time Warping distance (DTW) for numerical vectors.

\classinfo%
{de.lmu.ifi.dbs.elki.distance.distancefunction.timeseries.}{DerivativeDTWDistanceFunction}%
{0.7.0}%
{DBLP:conf/sdm/KeoghP01}%
{Derivative dynamic time warping}%

Derivative Dynamic Time Warping distance for numerical vectors.

\classinfo%
{de.lmu.ifi.dbs.elki.distance.distancefunction.timeseries.}{EDRDistanceFunction}%
{0.2}%
{DBLP:conf/sigmod/ChenOO05}%
{Edit Distance on Real Sequence}%

Edit Distance on Real Sequence distance for numerical vectors.

\classinfo%
{de.lmu.ifi.dbs.elki.distance.distancefunction.timeseries.}{ERPDistanceFunction}%
{0.2}%
{DBLP:conf/vldb/ChenN04}%
{Edit Distance with Real Penalty}%

Edit Distance With Real Penalty distance for numerical vectors.

\classinfo%
{de.lmu.ifi.dbs.elki.distance.distancefunction.timeseries.}{LCSSDistanceFunction}%
{0.2}%
{DBLP:conf/kdd/VlachosHGK03}%
{Longest Common Subsequence distance function}%

Longest Common Subsequence distance for numerical vectors.

 Originally this was based on the Matlab Code by Michalis Vlachos, but we have
since switched to a version that uses less memory.

\pkginfo{6}%
{de.lmu.ifi.dbs.elki.distance.}{similarityfunction}%
{Similarityfunction}%

Similarity functions.

\classinfo%
{de.lmu.ifi.dbs.elki.distance.similarityfunction.}{FractionalSharedNearestNeighborSimilarityFunction}%
{0.2}%
{}%
{Fractional Shared Nearest Neighbor Similarity Function}%

SharedNearestNeighborSimilarityFunction with a pattern defined to accept
Strings that define a non-negative Integer.

\classinfo%
{de.lmu.ifi.dbs.elki.distance.similarityfunction.}{InvertedDistanceSimilarityFunction}%
{0.5.0}%
{}%
{Inverted Distance Similarity Function}%

Adapter to use a primitive number-distance as similarity measure, by computing
1/distance.

\classinfo%
{de.lmu.ifi.dbs.elki.distance.similarityfunction.}{Kulczynski1SimilarityFunction}%
{0.6.0}%
{doi:10.1007/978-3-642-00234-2}%
{Kulczynski 1 Similarity Function}%

Kulczynski similarity 1.
 \[ s_\text{Kulczynski-1}(\vec{x},\vec{y}):=
\tfrac{\sum\nolimits_i\min\{x_i,y_i\}}{\sum\nolimits_i |x_i-y_i|} \]
or in distance form:
\[ d_\text{Kulczynski-1}(\vec{x},\vec{y}):=
\tfrac{\sum\nolimits_i |x_i-y_i|}{\sum\nolimits_i\min\{x_i,y_i\}} \]

\classinfo%
{de.lmu.ifi.dbs.elki.distance.similarityfunction.}{Kulczynski2SimilarityFunction}%
{0.6.0}%
{doi:10.1007/978-3-642-00234-2}%
{Kulczynski 2 Similarity Function}%

Kulczynski similarity 2.
 \[ s_\text{Kulczynski-2}(\vec{x},\vec{y} :=
\tfrac{n}{2}\left(\tfrac{1}{\bar{x}}+\tfrac{1}{\bar{y}}\right)
\sum\nolimits_i\min\{x_i,y_i\} \]

\classinfo%
{de.lmu.ifi.dbs.elki.distance.similarityfunction.}{SharedNearestNeighborSimilarityFunction}%
{0.1}%
{}%
{Shared Nearest Neighbor Similarity Function}%

SharedNearestNeighborSimilarityFunction with a pattern defined to accept
Strings that define a non-negative Integer.

\pkginfo{7}%
{de.lmu.ifi.dbs.elki.distance.similarityfunction.}{cluster}%
{Cluster}%

Similarity measures for comparing clusters.

\classinfo%
{de.lmu.ifi.dbs.elki.distance.similarityfunction.cluster.}{ClusterIntersectionSimilarityFunction}%
{0.7.0}%
{}%
{Cluster Intersection Similarity Function}%

Measure the similarity of clusters via the intersection size.

\classinfo%
{de.lmu.ifi.dbs.elki.distance.similarityfunction.cluster.}{ClusterJaccardSimilarityFunction}%
{0.7.0}%
{journals/misc/Jaccard1902}%
{Cluster Jaccard Similarity Function}%

Measure the similarity of clusters via the Jaccard coefficient.

\classinfo%
{de.lmu.ifi.dbs.elki.distance.similarityfunction.cluster.}{ClusteringAdjustedRandIndexSimilarityFunction}%
{0.7.0}%
{doi:10.1007/BF01908075}%
{Clustering Adjusted Rand Index Similarity Function}%

Measure the similarity of clusters via the Adjusted Rand Index.

\classinfo%
{de.lmu.ifi.dbs.elki.distance.similarityfunction.cluster.}{ClusteringBCubedF1SimilarityFunction}%
{0.7.0}%
{doi:10.3115/980451.980859}%
{Clustering B Cubed F1 Similarity Function}%

Measure the similarity of clusters via the BCubed F1 Index.

\classinfo%
{de.lmu.ifi.dbs.elki.distance.similarityfunction.cluster.}{ClusteringFowlkesMallowsSimilarityFunction}%
{0.7.0}%
{doi:10.2307/2288117}%
{Clustering Fowlkes Mallows Similarity Function}%

Measure the similarity of clusters via the Fowlkes-Mallows Index.

\classinfo%
{de.lmu.ifi.dbs.elki.distance.similarityfunction.cluster.}{ClusteringRandIndexSimilarityFunction}%
{0.7.0}%
{doi:10.2307/2284239}%
{Clustering Rand Index Similarity Function}%

Measure the similarity of clusters via the Rand Index.

\pkginfo{7}%
{de.lmu.ifi.dbs.elki.distance.similarityfunction.}{kernel}%
{Kernel}%

Kernel functions.

\classinfo%
{de.lmu.ifi.dbs.elki.distance.similarityfunction.kernel.}{LaplaceKernelFunction}%
{0.6.0}%
{}%
{Laplace Kernel Function}%

Laplace / exponential radial basis function kernel.

\classinfo%
{de.lmu.ifi.dbs.elki.distance.similarityfunction.kernel.}{LinearKernelFunction}%
{0.1}%
{}%
{Linear Kernel Function}%

Linear Kernel function that computes a similarity between the two feature
vectors x and y defined by \(x^T\cdot y\).

 Note: this is effectively equivalent to using
{\begingroup\ttfamily{}EuclideanDistanceFunction\endgroup}

\classinfo%
{de.lmu.ifi.dbs.elki.distance.similarityfunction.kernel.}{PolynomialKernelFunction}%
{0.1}%
{}%
{Polynomial Kernel Function}%

Polynomial Kernel function that computes a similarity between the two feature
vectors x and y defined by \((x^T\cdot y+b)^{\text{degree}}\).

\classinfo%
{de.lmu.ifi.dbs.elki.distance.similarityfunction.kernel.}{RadialBasisFunctionKernelFunction}%
{0.6.0}%
{}%
{Radial Basis Function Kernel Function}%

Gaussian radial basis function kernel (RBF Kernel).

\classinfo%
{de.lmu.ifi.dbs.elki.distance.similarityfunction.kernel.}{RationalQuadraticKernelFunction}%
{0.6.0}%
{}%
{Rational Quadratic Kernel Function}%

Rational quadratic kernel, a less computational approximation of the Gaussian
RBF kernel ({\begingroup\ttfamily{}RadialBasisFunctionKernelFunction\endgroup}).

\classinfo%
{de.lmu.ifi.dbs.elki.distance.similarityfunction.kernel.}{SigmoidKernelFunction}%
{0.6.0}%
{}%
{Sigmoid Kernel Function}%

Sigmoid kernel function (aka: hyperbolic tangent kernel, multilayer
perceptron MLP kernel).

\pkginfo{5}%
{de.lmu.ifi.dbs.elki.}{evaluation}%
{Evaluation}%

Functionality for the evaluation of algorithms.

\pkginfo{6}%
{de.lmu.ifi.dbs.elki.evaluation.}{classification}%
{Classification}%

Evaluation of classification algorithms.

\pkginfo{7}%
{de.lmu.ifi.dbs.elki.evaluation.classification.}{holdout}%
{Holdout}%

Holdout and cross-validation strategies for evaluating classifiers.

\classinfo%
{de.lmu.ifi.dbs.elki.evaluation.classification.holdout.}{DisjointCrossValidation}%
{0.7.0}%
{}%
{Disjoint Cross Validation}%

DisjointCrossValidationHoldout provides a set of partitions of a database to
perform cross-validation. The test sets are guaranteed to be disjoint.

\classinfo%
{de.lmu.ifi.dbs.elki.evaluation.classification.holdout.}{LeaveOneOut}%
{0.7.0}%
{}%
{Leave One Out}%

A leave-one-out-holdout is to provide a set of partitions of a database where
each instances once hold out as a test instance while the respectively
remaining instances are training instances.

\classinfo%
{de.lmu.ifi.dbs.elki.evaluation.classification.holdout.}{RandomizedCrossValidation}%
{0.7.0}%
{}%
{Randomized Cross Validation}%

RandomizedCrossValidationHoldout provides a set of partitions of a database
to perform cross-validation. The test sets are not guaranteed to be disjoint.

\classinfo%
{de.lmu.ifi.dbs.elki.evaluation.classification.holdout.}{StratifiedCrossValidation}%
{0.7.0}%
{}%
{Stratified Cross Validation}%

A stratified n-fold crossvalidation to distribute the data to n buckets where
each bucket exhibits approximately the same distribution of classes as does
the complete data set. The buckets are disjoint. The distribution is
deterministic.

\pkginfo{6}%
{de.lmu.ifi.dbs.elki.evaluation.}{clustering}%
{Clustering}%

Evaluation of clustering results.

\classinfo%
{de.lmu.ifi.dbs.elki.evaluation.clustering.}{BCubed}%
{0.5.0}%
{doi:10.3115/980451.980859}%
{B Cubed}%

BCubed measures.

\classinfo%
{de.lmu.ifi.dbs.elki.evaluation.clustering.}{EditDistance}%
{0.5.0}%
{DBLP:conf/sigir/PantelL02}%
{Edit Distance}%

Edit distance measures.

 P. Pantel, D. Lin\\
 Document clustering with committees\\
 Proc. 25th ACM SIGIR Conf. on Research and Development in Information
Retrieval

\classinfo%
{de.lmu.ifi.dbs.elki.evaluation.clustering.}{Entropy}%
{0.5.0}%
{DBLP:conf/colt/Meila03,DBLP:conf/icml/NguyenEB09}%
{Entropy}%

Entropy based measures.

\classinfo%
{de.lmu.ifi.dbs.elki.evaluation.clustering.}{EvaluateClustering}%
{0.4.0}%
{}%
{Evaluate Clustering}%

Evaluate a clustering result by comparing it to an existing cluster label.

\classinfo%
{de.lmu.ifi.dbs.elki.evaluation.clustering.}{LogClusterSizes}%
{0.7.0}%
{}%
{Log Cluster Sizes}%

This class will log simple statistics on the clusters detected, such as the
cluster sizes and the number of clusters.

\classinfo%
{de.lmu.ifi.dbs.elki.evaluation.clustering.}{PairCounting}%
{0.5.0}%
{journals/misc/Jaccard1902,doi:10.1007/BF01908075,doi:10.2307/2288117,doi:10.2307/2284239,doi:10.1007/978-1-4613-0457-9}%
{Pair Counting}%

Pair-counting measures.

\classinfo%
{de.lmu.ifi.dbs.elki.evaluation.clustering.}{SetMatchingPurity}%
{0.5.0}%
{conf/kdd/SteinbachKK00,DBLP:journals/ir/AmigoGAV09a,tr/washington/Meila02,tr/umn/ZhaoK01}%
{Set Matching Purity}%

Set matching purity measures.

\pkginfo{7}%
{de.lmu.ifi.dbs.elki.evaluation.clustering.}{extractor}%
{Extractor}%

Classes to extract clusterings from hierarchical clustering.

\classinfo%
{de.lmu.ifi.dbs.elki.evaluation.clustering.extractor.}{CutDendrogramByHeightExtractor}%
{0.7.0}%
{}%
{Cut Dendrogram By Height Extractor}%

Extract clusters from a hierarchical clustering, during the evaluation phase.

Usually, it is more elegant to use {\begingroup\ttfamily{}ExtractFlatClusteringFromHierarchy\endgroup} as primary algorithm. But in order to extract multiple partitionings
from the same clustering, this can be useful.

\classinfo%
{de.lmu.ifi.dbs.elki.evaluation.clustering.extractor.}{CutDendrogramByNumberOfClustersExtractor}%
{0.7.0}%
{}%
{Cut Dendrogram By Number Of Clusters Extractor}%

Extract clusters from a hierarchical clustering, during the evaluation phase.

Usually, it is more elegant to use {\begingroup\ttfamily{}ExtractFlatClusteringFromHierarchy\endgroup} as primary algorithm. But in order to extract multiple partitionings
from the same clustering, this can be useful.

\classinfo%
{de.lmu.ifi.dbs.elki.evaluation.clustering.extractor.}{HDBSCANHierarchyExtractionEvaluator}%
{0.7.0}%
{}%
{HDBSCAN Hierarchy Extraction Evaluator}%

Extract clusters from a hierarchical clustering, during the evaluation phase.

Usually, it is more elegant to use {\begingroup\ttfamily{}HDBSCANHierarchyExtraction\endgroup} as
primary algorithm. But in order to extract multiple partitionings
from the same clustering, this can be useful.

\classinfo%
{de.lmu.ifi.dbs.elki.evaluation.clustering.extractor.}{SimplifiedHierarchyExtractionEvaluator}%
{0.7.0}%
{}%
{Simplified Hierarchy Extraction Evaluator}%

Extract clusters from a hierarchical clustering, during the evaluation phase.

Usually, it is more elegant to use {\begingroup\ttfamily{}SimplifiedHierarchyExtraction\endgroup} as
primary algorithm. But in order to extract multiple partitionings
from the same clustering, this can be useful.

\pkginfo{7}%
{de.lmu.ifi.dbs.elki.evaluation.clustering.}{internal}%
{Internal}%

Internal evaluation measures for clusterings.

\classinfo%
{de.lmu.ifi.dbs.elki.evaluation.clustering.internal.}{EvaluateCIndex}%
{0.7.0}%
{doi:10.1037/0033-2909.83.6.1072}%
{Evaluate C Index}%

Compute the C-index of a data set.

 Note: This requires pairwise distance computations, so it is not recommended
to use this on larger data sets.

\classinfo%
{de.lmu.ifi.dbs.elki.evaluation.clustering.internal.}{EvaluateConcordantPairs}%
{0.7.0}%
{doi:10.1080/01621459.1975.10480256,doi:10.1146/annurev.es.05.110174.000533}%
{Evaluate Concordant Pairs}%

Compute the Gamma Criterion of a data set.

\classinfo%
{de.lmu.ifi.dbs.elki.evaluation.clustering.internal.}{EvaluateDBCV}%
{0.7.5}%
{DBLP:conf/sdm/MoulaviJCZS14}%
{Evaluate DBCV}%

Compute the Density-Based Clustering Validation Index.

\classinfo%
{de.lmu.ifi.dbs.elki.evaluation.clustering.internal.}{EvaluateDaviesBouldin}%
{0.7.0}%
{DBLP:journals/pami/DaviesB79}%
{Evaluate Davies Bouldin}%

Compute the Davies-Bouldin index of a data set.

\classinfo%
{de.lmu.ifi.dbs.elki.evaluation.clustering.internal.}{EvaluatePBMIndex}%
{0.7.0}%
{DBLP:journals/pr/PakhiraBM04}%
{Evaluate PBM Index}%

Compute the PBM index of a clustering

\classinfo%
{de.lmu.ifi.dbs.elki.evaluation.clustering.internal.}{EvaluateSilhouette}%
{0.7.0}%
{doi:10.1016/0377-04278790125-7}%
{Evaluate Silhouette}%

Compute the silhouette of a data set.

\classinfo%
{de.lmu.ifi.dbs.elki.evaluation.clustering.internal.}{EvaluateSimplifiedSilhouette}%
{0.7.0}%
{}%
{Evaluate Simplified Silhouette}%

Compute the simplified silhouette of a data set.

The simplified silhouette does not use pairwise distances, but distances to
centroids only.

\classinfo%
{de.lmu.ifi.dbs.elki.evaluation.clustering.internal.}{EvaluateSquaredErrors}%
{0.7.0}%
{}%
{Evaluate Squared Errors}%

Evaluate a clustering by reporting the squared errors (SSE, SSQ), as used by
k-means. This should be used with {\begingroup\ttfamily{}SquaredEuclideanDistanceFunction\endgroup} only (when used with other distances, it will manually square the values; but
beware that the result is less meaningful with other distance functions).

For clusterings that provide a cluster prototype object (e.g. k-means), the
prototype will be used. For other algorithms, the centroid will be
recomputed.

\classinfo%
{de.lmu.ifi.dbs.elki.evaluation.clustering.internal.}{EvaluateVarianceRatioCriteria}%
{0.7.0}%
{doi:10.1080/03610927408827101}%
{Evaluate Variance Ratio Criteria}%

Compute the Variance Ratio Criteria of a data set, also known as
Calinski-Harabasz index.

\pkginfo{7}%
{de.lmu.ifi.dbs.elki.evaluation.clustering.}{pairsegments}%
{Pairsegments}%

Pair-segment analysis of multiple clusterings.

\classinfo%
{de.lmu.ifi.dbs.elki.evaluation.clustering.pairsegments.}{ClusterPairSegmentAnalysis}%
{0.5.0}%
{DBLP:conf/icde/AchtertGKSZ12}%
{Cluster Pair Segment Analysis}%

Evaluate clustering results by building segments for their pairs: shared
pairs and differences.

\classinfo%
{de.lmu.ifi.dbs.elki.evaluation.clustering.pairsegments.}{Segments}%
{0.5.0}%
{DBLP:conf/icde/AchtertGKSZ12}%
{Segments}%

Creates segments of two or more clusterings.

 Segments are the equally paired database objects of all given (2+)
clusterings. Given a contingency table, an object Segment represents the
table's cells where an intersection of classes and labels are given. Pair
Segments are created by converting an object Segment into its pair
representation. Converting all object Segments into pair Segments results in
a larger number of pair Segments, if any fragmentation (no perfect match of
clusters) within the contingency table has occurred (multiple cells on one
row or column). Thus for ever object Segment exists a corresponding pair
Segment. Additionally pair Segments represent pairs that are only in one
Clustering which occurs for each split of a clusterings cluster by another
clustering. Here, these pair Segments are referenced as fragmented Segments.
Within the visualization they describe (at least two) pair Segments that have
a corresponding object Segment.

\pkginfo{6}%
{de.lmu.ifi.dbs.elki.evaluation.}{outlier}%
{Outlier}%

Evaluate an outlier score using a misclassification based cost model.

\classinfo%
{de.lmu.ifi.dbs.elki.evaluation.outlier.}{ComputeOutlierHistogram}%
{0.3}%
{}%
{Compute Outlier Histogram}%

Compute a Histogram to evaluate a ranking algorithm.

The parameter {\begingroup\ttfamily{}-hist.positive\endgroup} specifies the class label of "positive"
hits.

\classinfo%
{de.lmu.ifi.dbs.elki.evaluation.outlier.}{JudgeOutlierScores}%
{0.4.0}%
{}%
{Judge Outlier Scores}%

Compute a Histogram to evaluate a ranking algorithm.

The parameter {\begingroup\ttfamily{}-hist.positive\endgroup} specifies the class label of "positive"
hits.

\classinfo%
{de.lmu.ifi.dbs.elki.evaluation.outlier.}{OutlierPrecisionAtKCurve}%
{0.5.0}%
{}%
{Outlier Precision At K Curve}%

Compute a curve containing the precision values for an outlier detection
method.

\classinfo%
{de.lmu.ifi.dbs.elki.evaluation.outlier.}{OutlierPrecisionRecallCurve}%
{0.5.0}%
{}%
{Outlier Precision Recall Curve}%

Compute a curve containing the precision values for an outlier detection
method.

\classinfo%
{de.lmu.ifi.dbs.elki.evaluation.outlier.}{OutlierROCCurve}%
{0.2}%
{}%
{Outlier ROC Curve}%

Compute a ROC curve to evaluate a ranking algorithm and compute the
corresponding ROCAUC value.

The parameter {\begingroup\ttfamily{}-rocauc.positive\endgroup} specifies the class label of
"positive" hits.

The nested algorithm {\begingroup\ttfamily{}-algorithm\endgroup} will be run, the result will be
searched for an iterable or ordering result, which then is compared with the
clustering obtained via the given class label.

\classinfo%
{de.lmu.ifi.dbs.elki.evaluation.outlier.}{OutlierRankingEvaluation}%
{0.7.0}%
{}%
{Outlier Ranking Evaluation}%

Evaluate outlier scores by their ranking

\classinfo%
{de.lmu.ifi.dbs.elki.evaluation.outlier.}{OutlierSmROCCurve}%
{0.5.0}%
{DBLP:conf/pkdd/KlementFJM11}%
{Outlier Sm ROC Curve}%

Smooth ROC curves are a variation of classic ROC curves that takes the scores
into account.

\classinfo%
{de.lmu.ifi.dbs.elki.evaluation.outlier.}{OutlierThresholdClustering}%
{0.5.0}%
{}%
{Outlier Threshold Clustering}%

Pseudo clustering algorithm that builds clusters based on their outlier
score. Useful for transforming a numeric outlier score into a 2-class
dataset.

\pkginfo{6}%
{de.lmu.ifi.dbs.elki.evaluation.}{scores}%
{Scores}%

Evaluation of rankings and scorings.

\classinfo%
{de.lmu.ifi.dbs.elki.evaluation.scores.}{AveragePrecisionEvaluation}%
{0.7.0}%
{}%
{Average Precision Evaluation}%

Evaluate using average precision.

\classinfo%
{de.lmu.ifi.dbs.elki.evaluation.scores.}{DCGEvaluation}%
{0.7.5}%
{DBLP:journals/tois/JarvelinK02}%
{DCG Evaluation}%

Discounted Cumulative Gain.

This evaluation metric would be able to use relevance information, but the
current implementation is for binary labels only (it is easy to add, but
requires API additions or changes).

\classinfo%
{de.lmu.ifi.dbs.elki.evaluation.scores.}{MaximumF1Evaluation}%
{0.7.0}%
{}%
{Maximum F1 Evaluation}%

Evaluate using the maximum F1 score.

\classinfo%
{de.lmu.ifi.dbs.elki.evaluation.scores.}{NDCGEvaluation}%
{0.7.5}%
{DBLP:journals/tois/JarvelinK02}%
{NDCG Evaluation}%

Normalized Discounted Cumulative Gain.

This evaluation metric would be able to use relevance information, but the
current implementation is for binary labels only (it is easy to add, but
requires API additions or changes).

\classinfo%
{de.lmu.ifi.dbs.elki.evaluation.scores.}{PrecisionAtKEvaluation}%
{0.7.0}%
{}%
{Precision At K Evaluation}%

Evaluate using Precision@k, or R-precision (when {\begingroup\ttfamily{}k=0\endgroup}).

When {\begingroup\ttfamily{}k=0\endgroup}, then it is set to the number of positive objects, and the
returned value is the R-precision, or the precision-recall break-even-point
(BEP).

\classinfo%
{de.lmu.ifi.dbs.elki.evaluation.scores.}{ROCEvaluation}%
{0.7.0}%
{}%
{ROC Evaluation}%

Compute ROC (Receiver Operating Characteristics) curves.

A ROC curve compares the true positive rate (y-axis) and false positive rate
(x-axis).

It was first used in radio signal detection, but has since found widespread
use in information retrieval, in particular for evaluating binary
classification problems.

ROC curves are particularly useful to evaluate a ranking of objects with
respect to a binary classification problem: a random sampling will
approximately achieve a ROC value of 0.5, while a perfect separation will
achieve 1.0 (all positives first) or 0.0 (all negatives first). In most use
cases, a score significantly below 0.5 indicates that the algorithm result
has been used the wrong way, and should be used backwards.

\pkginfo{6}%
{de.lmu.ifi.dbs.elki.evaluation.}{similaritymatrix}%
{Similaritymatrix}%

Render a distance matrix to visualize a clustering-distance-combination.

\classinfo%
{de.lmu.ifi.dbs.elki.evaluation.similaritymatrix.}{ComputeSimilarityMatrixImage}%
{0.4.0}%
{}%
{Compute Similarity Matrix Image}%

Compute a similarity matrix for a distance function.

\pkginfo{5}%
{de.lmu.ifi.dbs.elki.}{index}%
{Index}%

Index structure implementations

\pkginfo{6}%
{de.lmu.ifi.dbs.elki.index.}{distancematrix}%
{Distancematrix}%

Precomputed distance matrix.

\classinfo%
{de.lmu.ifi.dbs.elki.index.distancematrix.}{PrecomputedDistanceMatrix}%
{0.7.0}%
{}%
{Precomputed Distance Matrix}%

Distance matrix, for precomputing similarity for a small data set.

 This class uses a linear memory layout (not a ragged array), and assumes
symmetry as well as strictness. This way, it only stores the upper triangle
matrix with double precision. It has to store (n-1) * (n-2) distance values
in memory, requiring 8 * (n-1) * (n-2) bytes. Since Java has a size limit of
arrays of 31 bits (signed integer), we can store at most \(2^16\) objects
(precisely, 65536 objects) in a single array, which needs about 16 GB of RAM.

\classinfo%
{de.lmu.ifi.dbs.elki.index.distancematrix.}{PrecomputedSimilarityMatrix}%
{0.7.0}%
{}%
{Precomputed Similarity Matrix}%

Precomputed similarity matrix, for a small data set.

 This class uses a linear memory layout (not a ragged array), and assumes
symmetry as well as strictness. This way, it only stores the upper triangle
matrix with double precision. It has to store (n-1) * (n-2) similarity values
in memory, requiring 8 * (n-1) * (n-2) bytes. Since Java has a size limit of
arrays of 31 bits (signed integer), we can store at most \(2^16\) objects
(precisely, 65536 objects) in a single array, which needs about 16 GB of RAM.

\pkginfo{6}%
{de.lmu.ifi.dbs.elki.index.}{idistance}%
{Idistance}%

iDistance is a distance based indexing technique, using a reference points embedding.

\classinfo%
{de.lmu.ifi.dbs.elki.index.idistance.}{InMemoryIDistanceIndex}%
{0.7.0}%
{DBLP:conf/vldb/OoiYTJ01,DBLP:journals/tods/JagadishOTYZ05}%
{In Memory I Distance Index}%

In-memory iDistance index, a metric indexing method using a reference point
embedding.

 Important note: we are currently using a different query strategy. The
original publication discusses queries based on repeated radius queries. We use a strategy based on shrinking spheres, iteratively refined
starting with the closes reference point. We also do not use a B+-tree as
data structure, but simple in-memory lists. Therefore, we cannot report page
accesses needed.

 Feel free to contribute improved query strategies. All the code is
essentially here, you only need to query every reference point list, not just
the best.

\pkginfo{6}%
{de.lmu.ifi.dbs.elki.index.}{invertedlist}%
{Invertedlist}%

Indexes using inverted lists.

\classinfo%
{de.lmu.ifi.dbs.elki.index.invertedlist.}{InMemoryInvertedIndex}%
{0.7.0}%
{}%
{In Memory Inverted Index}%

Simple index using inverted lists.

\pkginfo{6}%
{de.lmu.ifi.dbs.elki.index.}{lsh}%
{Lsh}%

Locality Sensitive Hashing

\classinfo%
{de.lmu.ifi.dbs.elki.index.lsh.}{InMemoryLSHIndex}%
{0.6.0}%
{}%
{In Memory LSH Index}%

Locality Sensitive Hashing.

\pkginfo{7}%
{de.lmu.ifi.dbs.elki.index.lsh.}{hashfamilies}%
{Hashfamilies}%

Hash function families for LSH.

\classinfo%
{de.lmu.ifi.dbs.elki.index.lsh.hashfamilies.}{CosineHashFunctionFamily}%
{0.7.0}%
{DBLP:conf/sigir/Henzinger06,DBLP:conf/stoc/Charikar02}%
{Cosine Hash Function Family}%

Hash function family to use with Cosine distance, using simplified hash
functions where the projection is only drawn from +-1, instead of Gaussian
distributions.

\classinfo%
{de.lmu.ifi.dbs.elki.index.lsh.hashfamilies.}{EuclideanHashFunctionFamily}%
{0.6.0}%
{DBLP:conf/compgeom/DatarIIM04}%
{Euclidean Hash Function Family}%

2-stable hash function family for Euclidean distances.

\classinfo%
{de.lmu.ifi.dbs.elki.index.lsh.hashfamilies.}{ManhattanHashFunctionFamily}%
{0.6.0}%
{DBLP:conf/compgeom/DatarIIM04}%
{Manhattan Hash Function Family}%

2-stable hash function family for Euclidean distances.

\pkginfo{7}%
{de.lmu.ifi.dbs.elki.index.lsh.}{hashfunctions}%
{Hashfunctions}%

Hash functions for LSH

\classinfo%
{de.lmu.ifi.dbs.elki.index.lsh.hashfunctions.}{CosineLocalitySensitiveHashFunction}%
{0.7.0}%
{DBLP:conf/stoc/Charikar02}%
{Cosine Locality Sensitive Hash Function}%

Random projection family to use with sparse vectors.

\classinfo%
{de.lmu.ifi.dbs.elki.index.lsh.hashfunctions.}{MultipleProjectionsLocalitySensitiveHashFunction}%
{0.6.0}%
{DBLP:conf/compgeom/DatarIIM04}%
{Multiple Projections Locality Sensitive Hash Function}%

LSH hash function for vector space data. Depending on the choice of random
vectors, it can be appropriate for Manhattan and Euclidean distances.

\pkginfo{6}%
{de.lmu.ifi.dbs.elki.index.}{preprocessed}%
{Preprocessed}%

Index structure based on preprocessors

\pkginfo{7}%
{de.lmu.ifi.dbs.elki.index.preprocessed.}{fastoptics}%
{Fastoptics}%

Preprocessed index used by the FastOPTICS algorithm.

\classinfo%
{de.lmu.ifi.dbs.elki.index.preprocessed.fastoptics.}{RandomProjectedNeighborsAndDensities}%
{0.7.0}%
{DBLP:conf/cikm/SchneiderV13}%
{Random Projected Neighbors And Densities}%

Random Projections used for computing neighbors and density estimates.

 This index is specialized for the algorithm
{\begingroup\ttfamily{}FastOPTICS\endgroup}

\pkginfo{7}%
{de.lmu.ifi.dbs.elki.index.preprocessed.}{knn}%
{Knn}%

Indexes providing KNN and rKNN data.

\classinfo%
{de.lmu.ifi.dbs.elki.index.preprocessed.knn.}{NNDescent}%
{0.7.5}%
{DBLP:conf/www/DongCL11}%
{NN Descent}%

NN-desent (also known as KNNGraph) is an approximate nearest neighbor search
algorithm beginning with a random sample, then iteratively refining this
sample until.

\classinfo%
{de.lmu.ifi.dbs.elki.index.preprocessed.knn.}{SpacefillingKNNPreprocessor}%
{0.7.0}%
{DBLP:conf/dasfaa/SchubertZK15}%
{Spacefilling KNN Preprocessor}%

Compute the nearest neighbors approximatively using space filling curves.

 This version computes the data projections and stores, then queries this data
on-demand. This usually needs less memory (except for very small neighborhood
sizes k) than {\begingroup\ttfamily{}SpacefillingMaterializeKNNPreprocessor\endgroup}, but will also
be slower.

\classinfo%
{de.lmu.ifi.dbs.elki.index.preprocessed.knn.}{SpacefillingMaterializeKNNPreprocessor}%
{0.7.0}%
{DBLP:conf/dasfaa/SchubertZK15}%
{Spacefilling Materialize KNN Preprocessor}%

Compute the nearest neighbors approximatively using space filling curves.

 This version does the bulk kNN-join operation, i.e. precomputes the k nearest
neighbors for every object, then discards the curves. This is usually more
memory intensive but faster than {\begingroup\ttfamily{}SpacefillingKNNPreprocessor\endgroup}.

\pkginfo{7}%
{de.lmu.ifi.dbs.elki.index.preprocessed.}{localpca}%
{Localpca}%

Index using a preprocessed local PCA.

\classinfo%
{de.lmu.ifi.dbs.elki.index.preprocessed.localpca.}{KNNQueryFilteredPCAIndex}%
{0.4.0}%
{}%
{Knn Query Based Local PCA Preprocessor}%

Provides the local neighborhood to be considered in the PCA as the k nearest
neighbors of an object.

\pkginfo{7}%
{de.lmu.ifi.dbs.elki.index.preprocessed.}{preference}%
{Preference}%

Indexes storing preference vectors.

\classinfo%
{de.lmu.ifi.dbs.elki.index.preprocessed.preference.}{DiSHPreferenceVectorIndex}%
{0.1}%
{}%
{DiSH Preference Vector Index}%

Preprocessor for DiSH preference vector assignment to objects of a certain
database.

\classinfo%
{de.lmu.ifi.dbs.elki.index.preprocessed.preference.}{HiSCPreferenceVectorIndex}%
{0.1}%
{DBLP:conf/pkdd/AchtertBKKMZ06}%
{HiSC Preprocessor}%

Preprocessor for HiSC preference vector assignment to objects of a certain
database.

\pkginfo{7}%
{de.lmu.ifi.dbs.elki.index.preprocessed.}{snn}%
{Snn}%

Indexes providing nearest neighbor sets

\classinfo%
{de.lmu.ifi.dbs.elki.index.preprocessed.snn.}{SharedNearestNeighborPreprocessor}%
{0.1}%
{}%
{Shared Nearest Neighbor Preprocessor}%

A preprocessor for annotation of the ids of nearest neighbors to each
database object.

 The k nearest neighbors are assigned based on an arbitrary distance function.

 This functionality is similar but not identical to
{\begingroup\ttfamily{}MaterializeKNNPreprocessor\endgroup}: While it also computes the k nearest
neighbors, it does not keep the actual distances, but organizes the NN set in
a TreeSet for fast set operations.

\pkginfo{6}%
{de.lmu.ifi.dbs.elki.index.}{projected}%
{Projected}%

Projected indexes for data.

\classinfo%
{de.lmu.ifi.dbs.elki.index.projected.}{LatLngAsECEFIndex}%
{0.6.0}%
{}%
{LatLngAsECEF Index}%

Index a 2d data set (consisting of Lat/Lng pairs) by using a projection to 3D
coordinates (WGS-86 to ECEF).

 Earth-Centered, Earth-Fixed (ECEF) is a 3D coordinate system, sometimes also
referred to as XYZ, that uses 3 cartesian axes. The center is at the earths
center of mass, the z axis points to the north pole. X axis is to the prime
meridan at the equator (so latitude 0, longitude 0), and the Y axis is
orthogonal going to the east (latitude 0, longitude 90°E).

 The Euclidean distance in this coordinate system is a lower bound for the
great-circle distance, and Euclidean coordinates are supposedly easier to
index.

 Note: this index will only support the distance function
{\begingroup\ttfamily{}LatLngDistanceFunction\endgroup}, as it uses a projection that will map data
according to this great circle distance. If the query hint "exact" is set, it
will not be used.

\classinfo%
{de.lmu.ifi.dbs.elki.index.projected.}{LngLatAsECEFIndex}%
{0.6.0}%
{}%
{LngLatAsECEF Index}%

Index a 2d data set (consisting of Lng/Lat pairs) by using a projection to 3D
coordinates (WGS-86 to ECEF).

 Earth-Centered, Earth-Fixed (ECEF) is a 3D coordinate system, sometimes also
referred to as XYZ, that uses 3 cartesian axes. The center is at the earths
center of mass, the z axis points to the north pole. X axis is to the prime
meridan at the equator (so latitude 0, longitude 0), and the Y axis is
orthogonal going to the east (latitude 0, longitude 90°E).

 The Euclidean distance in this coordinate system is a lower bound for the
great-circle distance, and Euclidean coordinates are supposedly easier to
index.

 Note: this index will only support the distance function
{\begingroup\ttfamily{}LngLatDistanceFunction\endgroup}, as it uses a projection that will map data
according to this great circle distance. If the query hint "exact" is set, it
will not be used.

\classinfo%
{de.lmu.ifi.dbs.elki.index.projected.}{PINN}%
{0.6.0}%
{DBLP:conf/icdm/VriesCH10}%
{PINN: Projection Indexed Nearest Neighbors}%

Projection-Indexed nearest-neighbors (PINN) is an index to retrieve the
nearest neighbors in high dimensional spaces by using a random projection
based index.

\pkginfo{6}%
{de.lmu.ifi.dbs.elki.index.}{tree}%
{Tree}%

Tree-based index structures

\pkginfo{7}%
{de.lmu.ifi.dbs.elki.index.tree.}{metrical}%
{Metrical}%

Tree-based index structures for metrical vector spaces.

\pkginfo{8}%
{de.lmu.ifi.dbs.elki.index.tree.metrical.}{covertree}%
{Metrical Covertree}%

Cover-tree variations.

\classinfo%
{de.lmu.ifi.dbs.elki.index.tree.metrical.covertree.}{CoverTree}%
{0.7.0}%
{DBLP:conf/icml/BeygelzimerKL06}%
{Cover Tree}%

Cover tree data structure (in-memory). This is a metrical data
structure that is similar to the M-tree, but not as balanced and
disk-oriented. However, by not having these requirements it does not require
the expensive splitting procedures of M-tree.

\classinfo%
{de.lmu.ifi.dbs.elki.index.tree.metrical.covertree.}{SimplifiedCoverTree}%
{0.7.0}%
{}%
{Simplified Cover Tree}%

Simplified cover tree data structure (in-memory). This is a metrical data structure that is similar to the M-tree, but not as balanced and
disk-oriented. However, by not having these requirements it does not require
the expensive splitting procedures of M-tree.

This version does not store the distance to the parent, so it needs only
about 40\% of the memory of {\begingroup\ttfamily{}CoverTree\endgroup} but does more distance
computations for search.

\pkginfo{8}%
{de.lmu.ifi.dbs.elki.index.tree.metrical.}{mtreevariants}%
{Metrical Mtreevariants}%

M-Tree and variants.

\pkginfo{9}%
{de.lmu.ifi.dbs.elki.index.tree.metrical.mtreevariants.}{mktrees}%
{Metrical Mtreevariants Mktrees}%

Metrical index structures based on the concepts of the M-Tree
supporting processing of reverse k nearest neighbor queries by
using the k-nn distances of the entries.

\pkginfo{10}%
{de.lmu.ifi.dbs.elki.index.tree.metrical.mtreevariants.mktrees.}{mkapp}%
{Metrical Mtreevariants Mktrees Mkapp}%

{\begingroup\ttfamily{}MkAppTree\endgroup}

\classinfo%
{de.lmu.ifi.dbs.elki.index.tree.metrical.mtreevariants.mktrees.mkapp.}{MkAppTreeFactory}%
{0.4.0}%
{}%
{MkApp Tree Factory}%

Factory for a MkApp-Tree

\pkginfo{10}%
{de.lmu.ifi.dbs.elki.index.tree.metrical.mtreevariants.mktrees.}{mkmax}%
{Metrical Mtreevariants Mktrees Mkmax}%

{\begingroup\ttfamily{}MkMaxTree\endgroup}

\classinfo%
{de.lmu.ifi.dbs.elki.index.tree.metrical.mtreevariants.mktrees.mkmax.}{MkMaxTreeFactory}%
{0.4.0}%
{}%
{MkMax Tree Factory}%

Factory for MkMaxTrees

\pkginfo{10}%
{de.lmu.ifi.dbs.elki.index.tree.metrical.mtreevariants.mktrees.}{mktab}%
{Metrical Mtreevariants Mktrees Mktab}%

{\begingroup\ttfamily{}MkTabTree\endgroup}

\classinfo%
{de.lmu.ifi.dbs.elki.index.tree.metrical.mtreevariants.mktrees.mktab.}{MkTabTreeFactory}%
{0.4.0}%
{}%
{MkTab Tree Factory}%

Factory for MkTabTrees

\pkginfo{9}%
{de.lmu.ifi.dbs.elki.index.tree.metrical.mtreevariants.}{mtree}%
{Metrical Mtreevariants Mtree}%

{\begingroup\ttfamily{}MTree\endgroup}

\classinfo%
{de.lmu.ifi.dbs.elki.index.tree.metrical.mtreevariants.mtree.}{MTree}%
{0.1}%
{DBLP:conf/vldb/CiacciaPZ97}%
{M-Tree}%

MTree is a metrical index structure based on the concepts of the M-Tree.
Apart from organizing the objects it also provides several methods to search
for certain object in the structure. Persistence is not yet ensured.

\classinfo%
{de.lmu.ifi.dbs.elki.index.tree.metrical.mtreevariants.mtree.}{MTreeFactory}%
{0.4.0}%
{}%
{M Tree Factory}%

Factory for a M-Tree

\pkginfo{9}%
{de.lmu.ifi.dbs.elki.index.tree.metrical.mtreevariants.}{strategies}%
{Metrical Mtreevariants Strategies}%

\pkginfo{10}%
{de.lmu.ifi.dbs.elki.index.tree.metrical.mtreevariants.strategies.}{insert}%
{Metrical Mtreevariants Strategies Insert}%

Insertion (choose path) strategies of nodes in an M-Tree (and variants).

\classinfo%
{de.lmu.ifi.dbs.elki.index.tree.metrical.mtreevariants.strategies.insert.}{MinimumEnlargementInsert}%
{0.6.0}%
{DBLP:conf/vldb/CiacciaPZ97}%
{Minimum Enlargement Insert}%

Minimum enlargement insert - default insertion strategy for the M-tree.

\pkginfo{10}%
{de.lmu.ifi.dbs.elki.index.tree.metrical.mtreevariants.strategies.}{split}%
{Metrical Mtreevariants Strategies Split}%

Splitting strategies of nodes in an M-Tree (and variants).

\classinfo%
{de.lmu.ifi.dbs.elki.index.tree.metrical.mtreevariants.strategies.split.}{FarthestPointsSplit}%
{0.7.5}%
{}%
{Farthest Points Split}%

Farthest points split.

\classinfo%
{de.lmu.ifi.dbs.elki.index.tree.metrical.mtreevariants.strategies.split.}{MLBDistSplit}%
{0.6.0}%
{DBLP:conf/vldb/CiacciaPZ97}%
{MLB Dist Split}%

Encapsulates the required methods for a split of a node in an M-Tree.
The routing objects are chosen according to the MLBDIST strategy.

 The benefit of this strategy is that it works with precomputed distances from
the parent, while most other strategies would require O(n²) distance
computations. So if construction time is critical, this is a good choice.

\classinfo%
{de.lmu.ifi.dbs.elki.index.tree.metrical.mtreevariants.strategies.split.}{MMRadSplit}%
{0.2}%
{DBLP:conf/vldb/CiacciaPZ97}%
{MMRad Split}%

Encapsulates the required methods for a split of a node in an M-Tree. The
routing objects are chosen according to the mMrad strategy.

\classinfo%
{de.lmu.ifi.dbs.elki.index.tree.metrical.mtreevariants.strategies.split.}{MRadSplit}%
{0.2}%
{DBLP:conf/vldb/CiacciaPZ97}%
{MRad Split}%

Encapsulates the required methods for a split of a node in an M-Tree. The
routing objects are chosen according to the Mrad strategy.

\classinfo%
{de.lmu.ifi.dbs.elki.index.tree.metrical.mtreevariants.strategies.split.}{MSTSplit}%
{0.7.5}%
{DBLP:conf/edbt/TrainaTSF00}%
{MST Split}%

Splitting algorithm using the minimum spanning tree (MST), as proposed by the
Slim-Tree variant.

Unfortunately, the slim-tree paper does not detail how to choose the "most
appropriate edge from the longest ones" (to find a more balanced split), so
we try to longest 50\%, and keep the choice which yields the most balanced
split. This seems to work quite well.

\classinfo%
{de.lmu.ifi.dbs.elki.index.tree.metrical.mtreevariants.strategies.split.}{RandomSplit}%
{0.2}%
{DBLP:conf/vldb/CiacciaPZ97}%
{Random Split}%

Encapsulates the required methods for a split of a node in an M-Tree. The
routing objects are chosen according to the RANDOM strategy.

Note: only the routing objects are chosen at random, this is not a random
assignment!

\pkginfo{11}%
{de.lmu.ifi.dbs.elki.index.tree.metrical.mtreevariants.strategies.split.}{distribution}%
{Metrical Mtreevariants Strategies Split Distribution}%

 Entry distsribution strategies of nodes in an M-Tree (and variants).

\classinfo%
{de.lmu.ifi.dbs.elki.index.tree.metrical.mtreevariants.strategies.split.distribution.}{BalancedDistribution}%
{0.7.5}%
{DBLP:conf/vldb/CiacciaPZ97}%
{Balanced Distribution}%

Balanced entry distribution strategy of the M-tree.

\classinfo%
{de.lmu.ifi.dbs.elki.index.tree.metrical.mtreevariants.strategies.split.distribution.}{FarthestBalancedDistribution}%
{0.7.5}%
{}%
{Farthest Balanced Distribution}%

Balanced entry distribution strategy of the M-tree, beginning with the most
difficult points first. This should produce smaller covers.

\classinfo%
{de.lmu.ifi.dbs.elki.index.tree.metrical.mtreevariants.strategies.split.distribution.}{GeneralizedHyperplaneDistribution}%
{0.7.5}%
{DBLP:conf/vldb/CiacciaPZ97}%
{Generalized Hyperplane Distribution}%

Generalized hyperplane entry distribution strategy of the M-tree.

This strategy does not produce balanced trees, but often produces faster
access times, according to the original publication.

\pkginfo{7}%
{de.lmu.ifi.dbs.elki.index.tree.}{spatial}%
{Spatial}%

Tree-based index structures for spatial indexing.

\pkginfo{8}%
{de.lmu.ifi.dbs.elki.index.tree.spatial.}{kd}%
{Spatial Kd}%

K-d-tree and variants.

\classinfo%
{de.lmu.ifi.dbs.elki.index.tree.spatial.kd.}{MinimalisticMemoryKDTree}%
{0.6.0}%
{DBLP:journals/cacm/Bentley75}%
{Minimalistic Memory KD Tree}%

Simple implementation of a static in-memory K-D-tree. Does not support
dynamic updates or anything, but also is very simple and memory efficient:
all it uses is one {\begingroup\ttfamily{}ArrayModifiableDBIDs\endgroup} to sort the data in a
serialized tree.

\classinfo%
{de.lmu.ifi.dbs.elki.index.tree.spatial.kd.}{SmallMemoryKDTree}%
{0.7.0}%
{DBLP:journals/cacm/Bentley75}%
{Small Memory KD Tree}%

Simple implementation of a static in-memory K-D-tree. Does not support
dynamic updates or anything, but also is very simple and memory efficient:
all it uses is one {\begingroup\ttfamily{}ModifiableDoubleDBIDList\endgroup} to sort the data in a
serialized tree and store the current attribute value.

 It needs about 3 times as much memory as {\begingroup\ttfamily{}MinimalisticMemoryKDTree\endgroup} but
it is also considerably faster because it does not need to lookup this value
from the vectors.

\pkginfo{8}%
{de.lmu.ifi.dbs.elki.index.tree.spatial.}{rstarvariants}%
{Spatial Rstarvariants}%

R*-Tree and variants.

\pkginfo{9}%
{de.lmu.ifi.dbs.elki.index.tree.spatial.rstarvariants.}{query}%
{Spatial Rstarvariants Query}%

Queries on the R-Tree family of indexes: kNN and range queries.

\classinfo%
{de.lmu.ifi.dbs.elki.index.tree.spatial.rstarvariants.query.}{EuclideanRStarTreeKNNQuery}%
{0.7.0}%
{DBLP:conf/ssd/HjaltasonS95}%
{Euclidean R Star Tree KNN Query}%

Instance of a KNN query for a particular spatial index.

\classinfo%
{de.lmu.ifi.dbs.elki.index.tree.spatial.rstarvariants.query.}{EuclideanRStarTreeRangeQuery}%
{0.4.0}%
{doi:10.1109/ICICS.1997.652114}%
{Euclidean R Star Tree Range Query}%

Instance of a range query for a particular spatial index.

\classinfo%
{de.lmu.ifi.dbs.elki.index.tree.spatial.rstarvariants.query.}{RStarTreeKNNQuery}%
{0.4.0}%
{DBLP:conf/ssd/HjaltasonS95}%
{R Star Tree KNN Query}%

Instance of a KNN query for a particular spatial index.

\classinfo%
{de.lmu.ifi.dbs.elki.index.tree.spatial.rstarvariants.query.}{RStarTreeRangeQuery}%
{0.4.0}%
{doi:10.1109/ICICS.1997.652114}%
{R Star Tree Range Query}%

Instance of a range query for a particular spatial index.

\pkginfo{9}%
{de.lmu.ifi.dbs.elki.index.tree.spatial.rstarvariants.}{rstar}%
{Spatial Rstarvariants Rstar}%

{\begingroup\ttfamily{}RStarTree\endgroup}

\classinfo%
{de.lmu.ifi.dbs.elki.index.tree.spatial.rstarvariants.rstar.}{RStarTree}%
{0.1}%
{DBLP:conf/sigmod/BeckmannKSS90}%
{R*-Tree}%

RStarTree is a spatial index structure based on the concepts of the R*-Tree.
Apart from organizing the objects it also provides several methods to search
for certain object in the structure and ensures persistence.

\classinfo%
{de.lmu.ifi.dbs.elki.index.tree.spatial.rstarvariants.rstar.}{RStarTreeFactory}%
{0.4.0}%
{}%
{R Star Tree Factory}%

Factory for regular R*-Trees.

\pkginfo{9}%
{de.lmu.ifi.dbs.elki.index.tree.spatial.rstarvariants.}{strategies}%
{Spatial Rstarvariants Strategies}%

\pkginfo{10}%
{de.lmu.ifi.dbs.elki.index.tree.spatial.rstarvariants.strategies.}{bulk}%
{Spatial Rstarvariants Strategies Bulk}%

Packages for bulk-loading R*-Trees.

\classinfo%
{de.lmu.ifi.dbs.elki.index.tree.spatial.rstarvariants.strategies.bulk.}{AdaptiveSortTileRecursiveBulkSplit}%
{0.6.0}%
{}%
{Adaptive Sort Tile Recursive Bulk Split}%

This is variation of the original STR bulk load for non-rectangular data
spaces. Instead of iterating through the dimensions and splitting each by
(approximately) the same factor, this variation tries to adjust the factor to
the extends of the data space. I.e. if the data set is twice as wide as high,
this should produce twice as many partitions on the X than on the Y axis.

Whether or not this offers benefits greatly depends on the distance queries
used. But for symmetric distances, the resulting pages should be more
rectangular, which often is beneficial.

See {\begingroup\ttfamily{}SortTileRecursiveBulkSplit\endgroup} for the original STR bulk load.

\classinfo%
{de.lmu.ifi.dbs.elki.index.tree.spatial.rstarvariants.strategies.bulk.}{FileOrderBulkSplit}%
{0.5.0}%
{}%
{File Order Bulk Split}%

Trivial bulk loading - assumes that the file has been appropriately sorted
before.

\classinfo%
{de.lmu.ifi.dbs.elki.index.tree.spatial.rstarvariants.strategies.bulk.}{MaxExtensionBulkSplit}%
{0.4.0}%
{}%
{Max Extension Bulk Split}%

Split strategy for bulk-loading a spatial tree where the split axes are the
dimensions with maximum extension.

\classinfo%
{de.lmu.ifi.dbs.elki.index.tree.spatial.rstarvariants.strategies.bulk.}{MaxExtensionSortTileRecursiveBulkSplit}%
{0.6.0}%
{}%
{Max Extension Sort Tile Recursive Bulk Split}%

This is variation of the {\begingroup\ttfamily{}SortTileRecursiveBulkSplit\endgroup}, incorporating
some ideas from {\begingroup\ttfamily{}MaxExtensionBulkSplit\endgroup}. Instead of iterating through
the axes in order, it always chooses the axis with the largest extend. This
may rarely lead to the data being split on the same axis twice, but most
importantly it varies the splitting order compared to STR.

{\begingroup\ttfamily{}AdaptiveSortTileRecursiveBulkSplit\endgroup} takes these ideas one step
further, by also varying the fan-out degree.

\classinfo%
{de.lmu.ifi.dbs.elki.index.tree.spatial.rstarvariants.strategies.bulk.}{OneDimSortBulkSplit}%
{0.5.0}%
{doi:10.1145/971699.318900}%
{OneDim Sort Bulk Split}%

Simple bulk loading strategy by sorting the data along the first dimension.

 This is also known as Nearest-X, and attributed to:

 N. Roussopoulos, D. Leifker\\
 Direct spatial search on pictorial databases using packed R-trees\\
 ACM SIGMOD Record 14-4

\classinfo%
{de.lmu.ifi.dbs.elki.index.tree.spatial.rstarvariants.strategies.bulk.}{SortTileRecursiveBulkSplit}%
{0.5.0}%
{DBLP:conf/icde/LeuteneggerEL97}%
{Sort Tile Recursive Bulk Split}%

Sort-Tile-Recursive aims at tiling the data space with a grid-like structure
for partitioning the dataset into the required number of buckets.

\classinfo%
{de.lmu.ifi.dbs.elki.index.tree.spatial.rstarvariants.strategies.bulk.}{SpatialSortBulkSplit}%
{0.5.0}%
{DBLP:conf/cikm/KamelF93}%
{Spatial Sort Bulk Split}%

Bulk loading by spatially sorting the objects, then partitioning the sorted
list appropriately.

 Based conceptually on:

 On packing R-trees\\
 I. Kamel, C. Faloutsos\\
 Proc. 2nd Int. Conf. on Information and Knowledge Management (CIKM)

\pkginfo{10}%
{de.lmu.ifi.dbs.elki.index.tree.spatial.rstarvariants.strategies.}{insert}%
{Spatial Rstarvariants Strategies Insert}%

Insertion strategies for R-Trees

\classinfo%
{de.lmu.ifi.dbs.elki.index.tree.spatial.rstarvariants.strategies.insert.}{ApproximativeLeastOverlapInsertionStrategy}%
{0.5.0}%
{DBLP:conf/sigmod/BeckmannKSS90}%
{Approximative Least Overlap Insertion Strategy}%

The choose subtree method proposed by the R*-Tree with slightly better
performance for large leaf sizes (linear approximation).

 Norbert Beckmann, Hans-Peter Kriegel, Ralf Schneider, Bernhard Seeger\\
 The R*-tree: an efficient and robust access method for points and
rectangles\\
 Proc. 1990 ACM SIGMOD Int. Conf. Management of Data

\classinfo%
{de.lmu.ifi.dbs.elki.index.tree.spatial.rstarvariants.strategies.insert.}{CombinedInsertionStrategy}%
{0.5.0}%
{DBLP:conf/sigmod/BeckmannKSS90}%
{Combined Insertion Strategy}%

Use two different insertion strategies for directory and leaf nodes.

 Using two different strategies was likely first suggested in:

 Norbert Beckmann, Hans-Peter Kriegel, Ralf Schneider, Bernhard Seeger\\
 The R*-tree: an efficient and robust access method for points and
rectangles\\
 Proc. 1990 ACM SIGMOD Int. Conf. Management of Data

\classinfo%
{de.lmu.ifi.dbs.elki.index.tree.spatial.rstarvariants.strategies.insert.}{LeastEnlargementInsertionStrategy}%
{0.5.0}%
{doi:10.1145/971697.602266}%
{Least Enlargement Insertion Strategy}%

The default R-Tree insertion strategy: find rectangle with least volume
enlargement.

\classinfo%
{de.lmu.ifi.dbs.elki.index.tree.spatial.rstarvariants.strategies.insert.}{LeastEnlargementWithAreaInsertionStrategy}%
{0.5.0}%
{DBLP:conf/sigmod/BeckmannKSS90}%
{Least Enlargement With Area Insertion Strategy}%

A slight modification of the default R-Tree insertion strategy: find
rectangle with least volume enlargement, but choose least area on ties.

 Proposed for non-leaf entries in:

 Norbert Beckmann, Hans-Peter Kriegel, Ralf Schneider, Bernhard Seeger\\
 The R*-tree: an efficient and robust access method for points and
rectangles\\
 Proc. 1990 ACM SIGMOD Int. Conf. Management of Data

\classinfo%
{de.lmu.ifi.dbs.elki.index.tree.spatial.rstarvariants.strategies.insert.}{LeastOverlapInsertionStrategy}%
{0.5.0}%
{DBLP:conf/sigmod/BeckmannKSS90}%
{Least Overlap Insertion Strategy}%

The choose subtree method proposed by the R*-Tree for leaf nodes.

\pkginfo{10}%
{de.lmu.ifi.dbs.elki.index.tree.spatial.rstarvariants.strategies.}{overflow}%
{Spatial Rstarvariants Strategies Overflow}%

Overflow treatment strategies for R-Trees

\classinfo%
{de.lmu.ifi.dbs.elki.index.tree.spatial.rstarvariants.strategies.overflow.}{LimitedReinsertOverflowTreatment}%
{0.5.0}%
{DBLP:conf/sigmod/BeckmannKSS90}%
{Limited Reinsert Overflow Treatment}%

Limited reinsertions, as proposed by the R*-Tree: For each real insert, allow
reinsertions to happen only once per level.

\classinfo%
{de.lmu.ifi.dbs.elki.index.tree.spatial.rstarvariants.strategies.overflow.}{SplitOnlyOverflowTreatment}%
{0.5.0}%
{}%
{Split Only Overflow Treatment}%

Always split, as in the original R-Tree

\pkginfo{10}%
{de.lmu.ifi.dbs.elki.index.tree.spatial.rstarvariants.strategies.}{reinsert}%
{Spatial Rstarvariants Strategies Reinsert}%

Reinsertion strategies for R-Trees

\classinfo%
{de.lmu.ifi.dbs.elki.index.tree.spatial.rstarvariants.strategies.reinsert.}{CloseReinsert}%
{0.5.0}%
{DBLP:conf/sigmod/BeckmannKSS90}%
{Close Reinsert}%

Reinsert objects on page overflow, starting with close objects first (even
when they will likely be inserted into the same page again!)

 The strategy preferred by the R*-Tree

\classinfo%
{de.lmu.ifi.dbs.elki.index.tree.spatial.rstarvariants.strategies.reinsert.}{FarReinsert}%
{0.5.0}%
{DBLP:conf/sigmod/BeckmannKSS90}%
{Far Reinsert}%

Reinsert objects on page overflow, starting with farther objects first (even
when they will likely be inserted into the same page again!)

 Alternative strategy mentioned in the R*-tree

\pkginfo{10}%
{de.lmu.ifi.dbs.elki.index.tree.spatial.rstarvariants.strategies.}{split}%
{Spatial Rstarvariants Strategies Split}%

Splitting strategies for R-Trees

\classinfo%
{de.lmu.ifi.dbs.elki.index.tree.spatial.rstarvariants.strategies.split.}{AngTanLinearSplit}%
{0.5.0}%
{DBLP:conf/ssd/AngT97}%
{AngTan Linear Split}%

Line-time complexity split proposed by Ang and Tan.

 This split strategy tries to minimize overlap only, which can however
degenerate to "slices".

\classinfo%
{de.lmu.ifi.dbs.elki.index.tree.spatial.rstarvariants.strategies.split.}{GreeneSplit}%
{0.5.0}%
{DBLP:conf/icde/Greene89}%
{Greene Split}%

Quadratic-time complexity split as used by Diane Greene for the R-Tree.

 Seed selection is quadratic, distribution is O(n log n).

 This contains a slight modification to improve performance with point data:
with points as seeds, the normalized separation is always 1, so we choose the
raw separation then.

 D. Greene\\
 An implementation and performance analysis of spatial data access methods\\
 Proceedings of the Fifth International Conference on Data Engineering

\classinfo%
{de.lmu.ifi.dbs.elki.index.tree.spatial.rstarvariants.strategies.split.}{RTreeLinearSplit}%
{0.5.0}%
{doi:10.1145/971697.602266}%
{R Tree Linear Split}%

Linear-time complexity greedy split as used by the original R-Tree.

\classinfo%
{de.lmu.ifi.dbs.elki.index.tree.spatial.rstarvariants.strategies.split.}{RTreeQuadraticSplit}%
{0.5.0}%
{doi:10.1145/971697.602266}%
{R Tree Quadratic Split}%

Quadratic-time complexity greedy split as used by the original R-Tree.

\classinfo%
{de.lmu.ifi.dbs.elki.index.tree.spatial.rstarvariants.strategies.split.}{TopologicalSplitter}%
{0.4.0}%
{DBLP:conf/sigmod/BeckmannKSS90}%
{Topological Splitter}%

Encapsulates the required parameters for a topological split of a R*-Tree.

\pkginfo{6}%
{de.lmu.ifi.dbs.elki.index.}{vafile}%
{Vafile}%

Vector Approximation File

\classinfo%
{de.lmu.ifi.dbs.elki.index.vafile.}{DAFile}%
{0.5.0}%
{DBLP:conf/ssdbm/KriegelKSZ06}%
{DA File}%

Dimension approximation file, a one-dimensional part of the
{\begingroup\ttfamily{}PartialVAFile\endgroup}.

\classinfo%
{de.lmu.ifi.dbs.elki.index.vafile.}{PartialVAFile}%
{0.5.0}%
{DBLP:conf/ssdbm/KriegelKSZ06}%
{Partial VA File}%

PartialVAFile. In-memory only implementation.

\classinfo%
{de.lmu.ifi.dbs.elki.index.vafile.}{VAFile}%
{0.5.0}%
{tr/ethz/WeberS97}%
{An approximation based data structure for similarity search}%

Vector-approximation file (VAFile)

\pkginfo{5}%
{de.lmu.ifi.dbs.elki.}{math}%
{Math}%

Mathematical operations and utilities used throughout the framework.

\classinfo%
{de.lmu.ifi.dbs.elki.math.}{Mean}%
{0.2}%
{doi:10.1145/365719.365958}%
{Mean}%

Compute the mean using a numerically stable online algorithm.

This class can repeatedly be fed with data using the put() methods, the
resulting values for mean can be queried at any time using getMean().

The high-precision function is based on:

 P. M. Neely\\
 Comparison of Several Algorithms for Computation of Means, Standard
Deviations and Correlation Coefficients\\
 Communications of the ACM 9(7), 1966

\classinfo%
{de.lmu.ifi.dbs.elki.math.}{MeanVariance}%
{0.2}%
{doi:10.2307/1266577,DBLP:conf/ssdbm/SchubertG18,doi:10.1080/00401706.1971.10488826,DBLP:journals/cacm/West79}%
{Mean Variance}%

Do some simple statistics (mean, variance) using a numerically stable online
algorithm.

 This class can repeatedly be fed with data using the add() methods, the
resulting values for mean and average can be queried at any time using
{\begingroup\ttfamily{}Mean.getMean()\endgroup} and {\begingroup\ttfamily{}getSampleVariance()\endgroup}.

 Make sure you have understood variance correctly when using
{\begingroup\ttfamily{}getNaiveVariance()\endgroup} - since this class is fed with samples and
estimates the mean from the samples, {\begingroup\ttfamily{}getSampleVariance()\endgroup} is often
the more appropriate version.

 As experimentally studied in

 Erich Schubert, Michael Gertz\\
 Numerically Stable Parallel Computation of (Co-)Variance\\
 Proc. 30th Int. Conf. Scientific and Statistical Database Management
(SSDBM 2018)

 the current approach is based on:

 E. A. Youngs and E. M. Cramer\\
 Some Results Relevant to Choice of Sum and Sum-of-Product Algorithms\\
 Technometrics 13(3), 1971

 We have originally experimented with:

 B. P. Welford\\
 Note on a method for calculating corrected sums of squares and products\\
 Technometrics 4(3), 1962

 D. H. D. West\\
 Updating Mean and Variance Estimates: An Improved Method\\
 Communications of the ACM 22(9)

\classinfo%
{de.lmu.ifi.dbs.elki.math.}{StatisticalMoments}%
{0.6.0}%
{DBLP:conf/ssdbm/SchubertG18,doi:10.1080/00401706.1971.10488826,web/Terriberry07,tr/sandia/Pebay08}%
{Statistical Moments}%

Track various statistical moments, including mean, variance, skewness and
kurtosis.

\pkginfo{6}%
{de.lmu.ifi.dbs.elki.math.}{geodesy}%
{Geodesy}%

Functions for computing on the sphere / earth.

\classinfo%
{de.lmu.ifi.dbs.elki.math.geodesy.}{Clarke1858SpheroidEarthModel}%
{0.6.0}%
{}%
{Clarke 1858 Spheroid Earth Model}%

The Clarke 1858 spheroid earth model.

 Radius: 6378293.645 m

 Flattening: 1 / 294.26068

\classinfo%
{de.lmu.ifi.dbs.elki.math.geodesy.}{Clarke1880SpheroidEarthModel}%
{0.6.0}%
{}%
{Clarke 1880 Spheroid Earth Model}%

The Clarke 1880 spheroid earth model.

 Radius: 6378249.145 m

 Flattening: 1 / 293.465

\classinfo%
{de.lmu.ifi.dbs.elki.math.geodesy.}{GRS67SpheroidEarthModel}%
{0.6.0}%
{}%
{GRS67 Spheroid Earth Model}%

The GRS 67 spheroid earth model.

 Radius: 6378160.0 m

 Flattening: 1 / 298.25

\classinfo%
{de.lmu.ifi.dbs.elki.math.geodesy.}{GRS80SpheroidEarthModel}%
{0.6.0}%
{}%
{GRS80 Spheroid Earth Model}%

The GRS 80 spheroid earth model, without height model (so not a geoid, just a
spheroid!)

 Radius: 6378137.0 m

 Flattening: 1 / 298.257222101

\classinfo%
{de.lmu.ifi.dbs.elki.math.geodesy.}{SphereUtil}%
{0.5.5}%
{DBLP:conf/ssd/SchubertZK13,web/Williams11,doi:10.1179/sre.1975.23.176.88,journals/skytelesc/Sinnott84}%
{Sphere Util}%

Class with utility functions for distance computations on the sphere.

 Note: the formulas are usually implemented for the unit sphere.

 The majority of formulas are adapted from:

 E. Williams\\
 Aviation Formulary\\
 Online: http://www.edwilliams.org/avform.htm

\classinfo%
{de.lmu.ifi.dbs.elki.math.geodesy.}{SphericalCosineEarthModel}%
{0.6.0}%
{}%
{Spherical Cosine Earth Model}%

A simple spherical earth model using radius 6371009 m.

 For distance computations, this variant uses the Cosine formula, which is
faster but less accurate than the Haversince or Vincenty's formula.

\classinfo%
{de.lmu.ifi.dbs.elki.math.geodesy.}{SphericalHaversineEarthModel}%
{0.6.0}%
{}%
{Spherical Haversine Earth Model}%

A simple spherical earth model using radius 6371009 m.

 For distance computations, this variant uses the Haversine formula, which is
faster but less accurate than Vincenty's formula.

\classinfo%
{de.lmu.ifi.dbs.elki.math.geodesy.}{SphericalVincentyEarthModel}%
{0.6.0}%
{}%
{Spherical Vincenty Earth Model}%

A simple spherical earth model using radius 6371009 m.

\classinfo%
{de.lmu.ifi.dbs.elki.math.geodesy.}{WGS72SpheroidEarthModel}%
{0.6.0}%
{}%
{WGS72 Spheroid Earth Model}%

The WGS72 spheroid earth model, without height model.

 Radius: 6378135.0 m

 Flattening: 1 / 298.26

\classinfo%
{de.lmu.ifi.dbs.elki.math.geodesy.}{WGS84SpheroidEarthModel}%
{0.6.0}%
{}%
{WGS84 Spheroid Earth Model}%

The WGS84 spheroid earth model, without height model (so not a geoid, just a
spheroid!)

 Note that EGM96 uses the same spheroid, but what really makes the difference
is its geoid expansion.

 Radius: 6378137.0 m

 Flattening: 1 / 298.257223563

\pkginfo{6}%
{de.lmu.ifi.dbs.elki.math.}{geometry}%
{Geometry}%

Algorithms from computational geometry.

\classinfo%
{de.lmu.ifi.dbs.elki.math.geometry.}{GrahamScanConvexHull2D}%
{0.4.0}%
{DBLP:journals/ipl/Graham72}%
{Graham Scan Convex Hull 2D}%

Classes to compute the convex hull of a set of points in 2D, using the
classic Grahams scan. Also computes a bounding box.

\classinfo%
{de.lmu.ifi.dbs.elki.math.geometry.}{PrimsMinimumSpanningTree}%
{0.5.5}%
{doi:10.1002/j.1538-7305.1957.tb01515.x}%
{Prims Minimum Spanning Tree}%

Prim's algorithm for finding the minimum spanning tree.

 Implementation for dense graphs, represented as distance matrix.

\classinfo%
{de.lmu.ifi.dbs.elki.math.geometry.}{SweepHullDelaunay2D}%
{0.5.0}%
{web/Sinclair16}%
{Sweep Hull Delaunay 2D}%

Compute the Convex Hull and/or Delaunay Triangulation, using the sweep-hull
approach of David Sinclair.

 Note: This implementation does not check or handle duplicate points!

\pkginfo{6}%
{de.lmu.ifi.dbs.elki.math.}{linearalgebra}%
{Linearalgebra}%

The linear algebra package provides classes and computational methods for
operations on matrices and vectors.

 Some content of this package is adapted from the Jama package.

 Five fundamental matrix decompositions, which consist of pairs or triples
of matrices, permutation vectors, and the like, produce results in five
decomposition classes. These decompositions are accessed by the Matrix
class to compute solutions of simultaneous linear equations, determinants,
inverses and other matrix functions. The five decompositions are:

\begin{itemize}
\item Cholesky Decomposition of symmetric, positive definite matrices.
\item LU Decomposition of rectangular matrices.
\item QR Decomposition of rectangular matrices.
\item Singular Value Decomposition of rectangular matrices.
\item Eigenvalue Decomposition of both symmetric and nonsymmetric square
matrices.
\end{itemize}
Example of use:
 Solve a linear system \(Ax=b\) and compute the residual norm, \(||b-Ax||\).

 {\begingroup\ttfamily{}double[][] matrix = { {1.,2.,3}, {4.,5.,6.}, {7.,8.,10.} };\endgroup}\\
{\begingroup\ttfamily{}double[] b = MathUtil.randomDoubleArray(3, new Random());\endgroup}\\
{\begingroup\ttfamily{}double[] x = VMath.solve(matrix, b);\endgroup}\\
{\begingroup\ttfamily{}double[] r = VMath.minusEquals(VMath.times(matrix, x), b);\endgroup}\\
{\begingroup\ttfamily{}double norm = VMath.euclideanLength(r);\endgroup}

 The original Jama-package has been developed by
the \href{http://www.mathworks.com/}{MathWorks} and
\href{http://www.nist.gov/}{NIST} and can be found at
\href{http://math.nist.gov/javanumerics/jama/}{math.nist.gov}.

Here, for the adaption some classes and methods convenient for data mining
applications within ELKI were added.
Furthermore some erroneous comments were corrected and the coding-style was
subtly changed to a more Java-typical style.

\classinfo%
{de.lmu.ifi.dbs.elki.math.linearalgebra.}{VMath}%
{0.5.0}%
{journals/misc/Mahalanobis36}%
{Vector and Matrix Math Library}%

Class providing basic vector mathematics, for low-level vectors stored as
{\begingroup\ttfamily{}double[]\endgroup}. While this is less nice syntactically, it reduces memory
usage and VM overhead.

\pkginfo{7}%
{de.lmu.ifi.dbs.elki.math.linearalgebra.}{pca}%
{Pca}%

Principal Component Analysis (PCA) and Eigenvector processing.

\classinfo%
{de.lmu.ifi.dbs.elki.math.linearalgebra.pca.}{AutotuningPCA}%
{0.5.0}%
{DBLP:conf/ssdbm/KriegelKSZ08}%
{Autotuning PCA}%

Performs a self-tuning local PCA based on the covariance matrices of given
objects. At most the closest 'k' points are used in the calculation and a
weight function is applied.

 The number of points used depends on when the strong eigenvectors exhibit the
clearest correlation.

\classinfo%
{de.lmu.ifi.dbs.elki.math.linearalgebra.pca.}{PCARunner}%
{0.2}%
{}%
{PCA Runner}%

Class to run PCA on given data.

The various methods will start PCA at different places (e.g. with database
IDs, database query results, a precomputed covariance matrix or eigenvalue
decomposition).

The runner can be parameterized by setting a covariance matrix builder (e.g.
to a weighted covariance matrix builder)

\classinfo%
{de.lmu.ifi.dbs.elki.math.linearalgebra.pca.}{RANSACCovarianceMatrixBuilder}%
{0.5.5}%
{DBLP:conf/icdm/KriegelKSZ12,DBLP:journals/cacm/FischlerB81}%
{RANSAC Covariance Matrix Builder}%

RANSAC based approach to a more robust covariance matrix computation.

 This is an experimental adoption of RANSAC to this problem, not a
generic RANSAC implementation!

 While using RANSAC for PCA at first sounds like a good idea, it does not
work very well in high-dimensional spaces. The problem is that PCA has
O(n²) degrees of freedom, so we need to sample very many objects, then
perform an O(n³) matrix operation to compute PCA, for each attempt.

\classinfo%
{de.lmu.ifi.dbs.elki.math.linearalgebra.pca.}{StandardCovarianceMatrixBuilder}%
{0.2}%
{}%
{Standard Covariance Matrix Builder}%

Class for building a "traditional" covariance matrix vai
{\begingroup\ttfamily{}CovarianceMatrix\endgroup}.
Reasonable default choice for a {\begingroup\ttfamily{}CovarianceMatrixBuilder\endgroup}

\classinfo%
{de.lmu.ifi.dbs.elki.math.linearalgebra.pca.}{WeightedCovarianceMatrixBuilder}%
{0.2}%
{DBLP:conf/ssdbm/KriegelKSZ08}%
{Weighted Covariance Matrix / PCA}%

{\begingroup\ttfamily{}CovarianceMatrixBuilder\endgroup} with weights.

 This builder uses a weight function to weight points differently during build
a covariance matrix. Covariance can be canonically extended with weights, as
shown in the article

\pkginfo{8}%
{de.lmu.ifi.dbs.elki.math.linearalgebra.pca.}{filter}%
{Pca Filter}%

Filter eigenvectors based on their eigenvalues.

\classinfo%
{de.lmu.ifi.dbs.elki.math.linearalgebra.pca.filter.}{DropEigenPairFilter}%
{0.5.0}%
{}%
{Drop EigenPair Filter}%

The "drop" filter looks for the largest drop in normalized relative
eigenvalues.

 Let \( s_1 \ldots s_n \) be the eigenvalues.

 Let \( a_k := 1/(n-k) \sum_{i=k..n} s_i \)

 Then \( r_k := s_k / a_k \) is the relative eigenvalue.

 The drop filter searches for \(\operatorname{arg\,max}_k r_k / r_{k+1} \)

\classinfo%
{de.lmu.ifi.dbs.elki.math.linearalgebra.pca.filter.}{FirstNEigenPairFilter}%
{0.1}%
{}%
{First n Eigenpair filter}%

The FirstNEigenPairFilter marks the n highest eigenpairs as strong
eigenpairs, where n is a user specified number.

\classinfo%
{de.lmu.ifi.dbs.elki.math.linearalgebra.pca.filter.}{LimitEigenPairFilter}%
{0.1}%
{}%
{Limit-based Eigenpair Filter}%

The LimitEigenPairFilter marks all eigenpairs having an (absolute) eigenvalue
below the specified threshold (relative or absolute) as weak eigenpairs, the
others are marked as strong eigenpairs.

\classinfo%
{de.lmu.ifi.dbs.elki.math.linearalgebra.pca.filter.}{PercentageEigenPairFilter}%
{0.1}%
{}%
{Percentage based Eigenpair filter}%

The PercentageEigenPairFilter sorts the eigenpairs in descending order of
their eigenvalues and marks the first eigenpairs, whose sum of eigenvalues is
higher than the given percentage of the sum of all eigenvalues as strong
eigenpairs.

\classinfo%
{de.lmu.ifi.dbs.elki.math.linearalgebra.pca.filter.}{ProgressiveEigenPairFilter}%
{0.2}%
{}%
{Progressive Eigenpair Filter}%

The ProgressiveEigenPairFilter sorts the eigenpairs in descending order of
their eigenvalues and marks the first eigenpairs, whose sum of eigenvalues is
higher than the given percentage of the sum of all eigenvalues as strong
eigenpairs. In contrast to the PercentageEigenPairFilter, it will use a
percentage which changes linearly with the subspace dimensionality. This
makes the parameter more consistent for different dimensionalities and often
gives better results when clusters of different dimensionality exist, since
different percentage alpha levels might be appropriate for different
dimensionalities.

Example calculations of alpha levels:

In a 3D space, a progressive alpha value of 0.5 equals:

- 1D subspace: 50 \% + 1/3 of remainder = 0.667

- 2D subspace: 50 \% + 2/3 of remainder = 0.833

In a 4D space, a progressive alpha value of 0.5 equals:

- 1D subspace: 50\% + 1/4 of remainder = 0.625

- 2D subspace: 50\% + 2/4 of remainder = 0.750

- 3D subspace: 50\% + 3/4 of remainder = 0.875

Reasoning why this improves over PercentageEigenPairFilter:

In a 100 dimensional space, a single Eigenvector representing over 85\% of the
total variance is highly significant, whereas the strongest 85 Eigenvectors
together will by definition always represent at least 85\% of the variance.
PercentageEigenPairFilter can thus not be used with these parameters and
detect both dimensionalities correctly.

The second parameter introduced here, walpha, serves a different function: It
prevents the eigenpair filter to use a statistically weak Eigenvalue just to
reach the intended level, e.g. 84\% + 1\% >= 85\% when 1\% is statistically very
weak.

\classinfo%
{de.lmu.ifi.dbs.elki.math.linearalgebra.pca.filter.}{RelativeEigenPairFilter}%
{0.2}%
{}%
{Relative EigenPair Filter}%

The RelativeEigenPairFilter sorts the eigenpairs in descending order of their
eigenvalues and marks the first eigenpairs who are a certain factor above the
average of the remaining eigenvalues.

It is closely related to the WeakEigenPairFilter, and differs mostly by
comparing to the remaining Eigenvalues, not to the total sum.

There are some situations where one or the other is superior, especially when
it comes to handling nested clusters and strong global correlations that are
not too interesting. These benefits usually only make a difference at higher
dimensionalities.

\classinfo%
{de.lmu.ifi.dbs.elki.math.linearalgebra.pca.filter.}{SignificantEigenPairFilter}%
{0.2}%
{}%
{Significant EigenPair Filter}%

The SignificantEigenPairFilter sorts the eigenpairs in descending order of
their eigenvalues and chooses the contrast of an Eigenvalue to the remaining
Eigenvalues is maximal.

It is closely related to the WeakEigenPairFilter and RelativeEigenPairFilter.
But while the RelativeEigenPairFilter chooses the highest dimensionality that
satisfies the relative alpha levels, the SignificantEigenPairFilter will
chose the local dimensionality such that the 'contrast' is maximal.

There are some situations where one or the other is superior, especially when
it comes to handling nested clusters and strong global correlations that are
not too interesting. These benefits usually only make a difference at higher
dimensionalities.

\classinfo%
{de.lmu.ifi.dbs.elki.math.linearalgebra.pca.filter.}{WeakEigenPairFilter}%
{0.2}%
{}%
{Weak Eigenpair Filter}%

The WeakEigenPairFilter sorts the eigenpairs in descending order of their
eigenvalues and returns the first eigenpairs who are above the average mark
as "strong", the others as "weak".

\pkginfo{8}%
{de.lmu.ifi.dbs.elki.math.linearalgebra.pca.}{weightfunctions}%
{Pca Weightfunctions}%

Weight functions used in weighted PCA via {\begingroup\ttfamily{}WeightedCovarianceMatrixBuilder\endgroup}

\classinfo%
{de.lmu.ifi.dbs.elki.math.linearalgebra.pca.weightfunctions.}{ConstantWeight}%
{0.2}%
{}%
{Constant Weight}%

Constant Weight function

The result is always 1.0

\classinfo%
{de.lmu.ifi.dbs.elki.math.linearalgebra.pca.weightfunctions.}{ErfcStddevWeight}%
{0.2}%
{}%
{Erfc Stddev Weight}%

Gaussian Error Function Weight function, scaled using stddev. This probably
is the most statistically sound weight.

erfc(1 / sqrt(2) * distance / stddev)

\classinfo%
{de.lmu.ifi.dbs.elki.math.linearalgebra.pca.weightfunctions.}{ErfcWeight}%
{0.2}%
{}%
{Erfc Weight}%

Gaussian Error Function Weight function, scaled such that the result it 0.1
at distance == max

erfc(1.1630871536766736 * distance / max)

The value of 1.1630871536766736 is erfcinv(0.1), to achieve the intended
scaling.

\classinfo%
{de.lmu.ifi.dbs.elki.math.linearalgebra.pca.weightfunctions.}{ExponentialStddevWeight}%
{0.2}%
{}%
{Exponential Stddev Weight}%

Exponential Weight function, scaled such that the result it 0.1 at distance
== max

stddev * exp(-.5 * distance/stddev)

This is similar to the Gaussian weight function, except distance/stddev is
not squared.

\classinfo%
{de.lmu.ifi.dbs.elki.math.linearalgebra.pca.weightfunctions.}{ExponentialWeight}%
{0.2}%
{}%
{Exponential Weight}%

Exponential Weight function, scaled such that the result it 0.1 at distance
== max

exp(-2.3025850929940455 * distance/max)

This is similar to the Gaussian weight function, except distance/max is not
squared.

-2.3025850929940455 is log(-.1) to achieve the intended range of 1.0 - 0.1

\classinfo%
{de.lmu.ifi.dbs.elki.math.linearalgebra.pca.weightfunctions.}{GaussStddevWeight}%
{0.2}%
{}%
{Gauss Stddev Weight}%

Gaussian weight function, scaled using standard deviation
\( 1/\sqrt(2\pi) \exp(-\frac{\text{dist}^2}{2\sigma^2}) \)

\classinfo%
{de.lmu.ifi.dbs.elki.math.linearalgebra.pca.weightfunctions.}{GaussWeight}%
{0.2}%
{}%
{Gauss Weight}%

Gaussian weight function, scaled such that the result it 0.1 at distance ==
max, using \( \exp(-2.3025850929940455 \frac{\text{dist}^2}{\max^2}) \).

\classinfo%
{de.lmu.ifi.dbs.elki.math.linearalgebra.pca.weightfunctions.}{InverseLinearWeight}%
{0.2}%
{}%
{Inverse Linear Weight}%

Inverse Linear Weight Function.

This weight is not particularly reasonable. Instead it serves the purpose of
testing the effects of a badly chosen weight function.

This function has increasing weight, from 0.1 to 1.0 at distance == max.

\classinfo%
{de.lmu.ifi.dbs.elki.math.linearalgebra.pca.weightfunctions.}{InverseProportionalStddevWeight}%
{0.2}%
{}%
{Inverse Proportional Stddev Weight}%

Inverse proportional weight function, scaled using the standard deviation.

1 / (1 + distance/stddev)

\classinfo%
{de.lmu.ifi.dbs.elki.math.linearalgebra.pca.weightfunctions.}{InverseProportionalWeight}%
{0.2}%
{}%
{Inverse Proportional Weight}%

Inverse proportional weight function, scaled using the maximum.

1 / (1 + distance/max)

\classinfo%
{de.lmu.ifi.dbs.elki.math.linearalgebra.pca.weightfunctions.}{LinearWeight}%
{0.2}%
{}%
{Linear Weight}%

Linear weight function, scaled using the maximum such that it goes from 1.0
to 0.1

1 - 0.9 * (distance/max)

\classinfo%
{de.lmu.ifi.dbs.elki.math.linearalgebra.pca.weightfunctions.}{QuadraticStddevWeight}%
{0.2}%
{}%
{Quadratic Stddev Weight}%

Quadratic weight function, scaled using the standard deviation.

 We needed another scaling here, we chose the cutoff point to be 3*stddev. If
you need another value, you have to reimplement this class.

 \( \max\{0.0, 1.0 - \frac{\text{dist}^2}{3\sigma^2} \}\)

\classinfo%
{de.lmu.ifi.dbs.elki.math.linearalgebra.pca.weightfunctions.}{QuadraticWeight}%
{0.2}%
{}%
{Quadratic Weight}%

Quadratic weight function, scaled using the maximum to reach 0.1 at that
point.

 1.0 - 0.9 * (distance/max)²

\pkginfo{6}%
{de.lmu.ifi.dbs.elki.math.}{spacefillingcurves}%
{Spacefillingcurves}%

Space filling curves.

\classinfo%
{de.lmu.ifi.dbs.elki.math.spacefillingcurves.}{BinarySplitSpatialSorter}%
{0.5.0}%
{DBLP:journals/cacm/Bentley75}%
{Binary Split Spatial Sorter}%

Spatially sort the data set by repetitive binary splitting, circulating
through the dimensions. This is essentially the bulk-loading proposed for the
k-d-tree, as it will produce a perfectly balanced k-d-tree. The resulting
order is the sequence in which objects would then be stored in the k-d-tree.

 Note that when using this for bulk-loading an R-tree, the result will
not be a k-d-tree, not even remotely similar, as the splits are not
preserved.

\classinfo%
{de.lmu.ifi.dbs.elki.math.spacefillingcurves.}{HilbertSpatialSorter}%
{0.5.0}%
{journals/mathann/Hilbert1891}%
{Hilbert Spatial Sorter}%

Sort object along the Hilbert Space Filling curve by mapping them to their
Hilbert numbers and sorting them.

 Objects are mapped using 31 bits per dimension.

\classinfo%
{de.lmu.ifi.dbs.elki.math.spacefillingcurves.}{PeanoSpatialSorter}%
{0.5.0}%
{journals/mathann/Peano1890}%
{Peano Spatial Sorter}%

Bulk-load an R-tree index by presorting the objects with their position on
the Peano curve.

 The basic shape of this space-filling curve looks like this:

\begin{Verbatim}[fontsize=\verbatimsize]
  3---4   9
  |   |   |
  2   5   8
  |   |   |
  1   6---7
\end{Verbatim}
 Which then expands to the next level as:

\begin{Verbatim}[fontsize=\verbatimsize]
  +-+ +-+ +-+ +-+ E
  | | | | | | | | |
  | +-+ +-+ | | +-+
  |         | |
  | +-+ +-+ | | +-+
  | | | | | | | | |
  +-+ | | +-+ +-+ |
      | |         |
  +-+ | | +-+ +-+ |
  | | | | | | | | |
  S +-+ +-+ +-+ +-+
\end{Verbatim}
 
and so on.

\classinfo%
{de.lmu.ifi.dbs.elki.math.spacefillingcurves.}{ZCurveSpatialSorter}%
{0.5.0}%
{}%
{Z Curve Spatial Sorter}%

Class to sort the data set by their Z-index, without doing a full
materialization of the Z indexes.

\pkginfo{6}%
{de.lmu.ifi.dbs.elki.math.}{statistics}%
{Statistics}%

Statistical tests and methods.

\classinfo%
{de.lmu.ifi.dbs.elki.math.statistics.}{ProbabilityWeightedMoments}%
{0.6.0}%
{doi:10.1080/00401706.1985.10488049,tr/ibm/Hosking00}%
{Probability Weighted Moments}%

Estimate the L-Moments of a sample.

\pkginfo{7}%
{de.lmu.ifi.dbs.elki.math.statistics.}{dependence}%
{Dependence}%

Statistical measures of dependence, such as correlation.

\classinfo%
{de.lmu.ifi.dbs.elki.math.statistics.dependence.}{CorrelationDependenceMeasure}%
{0.7.0}%
{}%
{Correlation Dependence Measure}%

Pearson product-moment correlation coefficient.

\classinfo%
{de.lmu.ifi.dbs.elki.math.statistics.dependence.}{DistanceCorrelationDependenceMeasure}%
{0.7.0}%
{doi:10.1214/009053607000000505}%
{Distance Correlation Dependence Measure}%

Distance correlation.

 The value returned is the square root of the dCor² value. This matches the R
implementation by the original authors.

\classinfo%
{de.lmu.ifi.dbs.elki.math.statistics.dependence.}{HSMDependenceMeasure}%
{0.5.5}%
{DBLP:journals/tvcg/TatuAEBTMK11}%
{HSM Dependence Measure}%

Compute the "interestingness" of dimension connections using the hough
transformation. This is a very visual approach, designed to find certain
patterns in parallel coordinates visualizations. The patterns detected
here occur mostly if you have mutliple clusters of linear patterns
as far as we understood the approach (which is not easy to use,
unfortunately).

\classinfo%
{de.lmu.ifi.dbs.elki.math.statistics.dependence.}{HiCSDependenceMeasure}%
{0.5.5}%
{DBLP:conf/icde/KellerMB12,DBLP:conf/sigmod/AchtertKSZ13}%
{HiCS Dependence Measure}%

Use the statistical tests as used by HiCS to measure dependence of variables.

\classinfo%
{de.lmu.ifi.dbs.elki.math.statistics.dependence.}{HoeffdingsDDependenceMeasure}%
{0.7.0}%
{journals/mathstat/Hoeffding48}%
{Hoeffdings D Dependence Measure}%

Calculate Hoeffding's D as a measure of dependence.

\classinfo%
{de.lmu.ifi.dbs.elki.math.statistics.dependence.}{JensenShannonEquiwidthDependenceMeasure}%
{0.7.0}%
{}%
{Jensen Shannon Equiwidth Dependence Measure}%

Jensen-Shannon Divergence is closely related to mutual information.

The output value is normalized, such that an evenly distributed and identical
distribution will yield a value of 1. Independent distributions may still
yield values close to .25, though.

\classinfo%
{de.lmu.ifi.dbs.elki.math.statistics.dependence.}{MCEDependenceMeasure}%
{0.7.0}%
{DBLP:journals/ivs/Guo03}%
{MCE Dependence Measure}%

Compute a mutual information based dependence measure using a nested means
discretization, originally proposed for ordering axes in parallel coordinate
plots.

\classinfo%
{de.lmu.ifi.dbs.elki.math.statistics.dependence.}{MutualInformationEquiwidthDependenceMeasure}%
{0.7.0}%
{}%
{Mutual Information Equiwidth Dependence Measure}%

Mutual Information (MI) dependence measure by dividing each attribute into
equal-width bins. MI can be seen as Kullback–Leibler divergence of the joint
distribution and the product of the marginal distributions.

For normalization, the resulting values are scaled by {\begingroup\ttfamily{}mi/log(nbins)\endgroup}.
This both cancels out the logarithm base, and normalizes for the number of
bins (a uniform distribution will yield a MI with itself of 1).

\classinfo%
{de.lmu.ifi.dbs.elki.math.statistics.dependence.}{SURFINGDependenceMeasure}%
{0.5.5}%
{DBLP:conf/sigmod/AchtertKSZ13,DBLP:conf/icdm/BaumgartnerPKKK04}%
{SURFING Dependence Measure}%

Compute the similarity of dimensions using the SURFING score. The parameter k
for the k nearest neighbors is currently hard-coded to 10\% of the set size.

 Note that the complexity is roughly O(n n k), so this is a rather slow
method, and with k at 10\% of n, is actually cubic: O(0.1 * n²).

 This version cannot use index support, as the API operates without database
attachment. However, it should be possible to implement some trivial
sorted-list indexes to get a reasonable speedup!

\classinfo%
{de.lmu.ifi.dbs.elki.math.statistics.dependence.}{SlopeDependenceMeasure}%
{0.5.5}%
{DBLP:conf/sigmod/AchtertKSZ13}%
{Slope Dependence Measure}%

Arrange dimensions based on the entropy of the slope spectrum.

 This version only accepts positive correlations, see also
{\begingroup\ttfamily{}SlopeInversionDependenceMeasure\endgroup}.

\classinfo%
{de.lmu.ifi.dbs.elki.math.statistics.dependence.}{SlopeInversionDependenceMeasure}%
{0.5.5}%
{DBLP:conf/sigmod/AchtertKSZ13}%
{Slope Inversion Dependence Measure}%

Arrange dimensions based on the entropy of the slope spectrum.

\classinfo%
{de.lmu.ifi.dbs.elki.math.statistics.dependence.}{SpearmanCorrelationDependenceMeasure}%
{0.7.0}%
{}%
{Spearman Correlation Dependence Measure}%

Spearman rank-correlation coefficient, also known as Spearmans Rho.

\pkginfo{7}%
{de.lmu.ifi.dbs.elki.math.statistics.}{distribution}%
{Distribution}%

Standard distributions, with random generation functionalities.

\classinfo%
{de.lmu.ifi.dbs.elki.math.statistics.distribution.}{BetaDistribution}%
{0.5.0}%
{}%
{Beta Distribution}%

Beta Distribution with implementation of the regularized incomplete beta
function

\classinfo%
{de.lmu.ifi.dbs.elki.math.statistics.distribution.}{CauchyDistribution}%
{0.6.0}%
{}%
{Cauchy Distribution}%

Cauchy distribution.

\classinfo%
{de.lmu.ifi.dbs.elki.math.statistics.distribution.}{ChiDistribution}%
{0.5.0}%
{}%
{Chi Distribution}%

Chi distribution.

\classinfo%
{de.lmu.ifi.dbs.elki.math.statistics.distribution.}{ChiSquaredDistribution}%
{0.5.0}%
{doi:10.2307/2347113}%
{Chi Squared Distribution}%

Chi-Squared distribution (a specialization of the Gamma distribution).

\classinfo%
{de.lmu.ifi.dbs.elki.math.statistics.distribution.}{ConstantDistribution}%
{0.5.0}%
{}%
{Constant Distribution}%

Pseudo distribution, that has a unique constant value.

\classinfo%
{de.lmu.ifi.dbs.elki.math.statistics.distribution.}{ExpGammaDistribution}%
{0.6.0}%
{}%
{Exp Gamma Distribution}%

Exp-Gamma Distribution, with random generation and density functions.

This distribution can be outlined as Y ~ log[Gamma] distributed, or
equivalently exp(Y) ~ Gamma.

Note: this matches the loggamma of SciPy, whereas Wolfram calls this the
Exponential Gamma Distribution "at times confused with the
LogGammaDistribution".

\classinfo%
{de.lmu.ifi.dbs.elki.math.statistics.distribution.}{ExponentialDistribution}%
{0.5.5}%
{}%
{Exponential Distribution}%

Exponential distribution.

\classinfo%
{de.lmu.ifi.dbs.elki.math.statistics.distribution.}{ExponentiallyModifiedGaussianDistribution}%
{0.6.0}%
{}%
{Exponentially Modified Gaussian Distribution}%

Exponentially modified Gaussian (EMG) distribution (ExGaussian distribution)
is a combination of a normal distribution and an exponential distribution.

Note that scipy uses a subtly different parameterization.

\classinfo%
{de.lmu.ifi.dbs.elki.math.statistics.distribution.}{GammaDistribution}%
{0.4.0}%
{doi:10.2307/2347113,doi:10.2307/2347257,DBLP:journals/computing/AhrensD74,DBLP:journals/cacm/AhrensD82}%
{Gamma Distribution}%

Gamma Distribution, with random generation and density functions.

\classinfo%
{de.lmu.ifi.dbs.elki.math.statistics.distribution.}{GeneralizedExtremeValueDistribution}%
{0.6.0}%
{}%
{Generalized Extreme Value Distribution}%

Generalized Extreme Value (GEV) distribution, also known as Fisher–Tippett
distribution.

This is a generalization of the Frechnet, Gumbel and (reversed) Weibull
distributions.

Implementation notice: In ELKI 0.8.0, the sign of the shape was negated.

\classinfo%
{de.lmu.ifi.dbs.elki.math.statistics.distribution.}{GeneralizedLogisticAlternateDistribution}%
{0.6.0}%
{}%
{Generalized Logistic Alternate Distribution}%

Generalized logistic distribution.

One of multiple ways of generalizing the logistic distribution.

Where {\begingroup\ttfamily{}shape=0\endgroup} yields the regular logistic distribution.

\classinfo%
{de.lmu.ifi.dbs.elki.math.statistics.distribution.}{GeneralizedLogisticDistribution}%
{0.6.0}%
{}%
{Generalized Logistic Distribution}%

Generalized logistic distribution. (Type I, Skew-logistic distribution)

 One of multiple ways of generalizing the logistic distribution.
\\
{\begingroup\ttfamily{}pdf(x) = shape * exp(-x) / (1 + exp(-x))**(shape+1)\endgroup}\\
{\begingroup\ttfamily{}cdf(x) = pow(1+exp(-x), -shape)\endgroup}\\
 Where {\begingroup\ttfamily{}shape=1\endgroup} yields the regular logistic distribution.

\classinfo%
{de.lmu.ifi.dbs.elki.math.statistics.distribution.}{GeneralizedParetoDistribution}%
{0.7.0}%
{}%
{Generalized Pareto Distribution}%

Generalized Pareto Distribution (GPD), popular for modeling long tail
distributions.

\classinfo%
{de.lmu.ifi.dbs.elki.math.statistics.distribution.}{GumbelDistribution}%
{0.6.0}%
{}%
{Gumbel Distribution}%

Gumbel distribution, also known as Log-Weibull distribution.

\classinfo%
{de.lmu.ifi.dbs.elki.math.statistics.distribution.}{HaltonUniformDistribution}%
{0.5.5}%
{doi:10.1016/S0895-71770000178-3}%
{Halton Uniform Distribution}%

Halton sequences are a pseudo-uniform distribution. The data is actually too
regular for a true uniform distribution, but as such will of course often
appear to be uniform.

 Technically, they are based on Van der Corput sequence and the Von Neumann
Katutani transformation. These produce a series of integers which then are
converted to floating point values.

 To randomize, we just choose a random starting position, as indicated by

\classinfo%
{de.lmu.ifi.dbs.elki.math.statistics.distribution.}{InverseGaussianDistribution}%
{0.6.0}%
{}%
{Inverse Gaussian Distribution}%

Inverse Gaussian distribution aka Wald distribution.

Beware that SciPy uses a different location parameter.

{\begingroup\ttfamily{}InverseGaussian(a, x) ~ scipy.stats.invgauss(a/x, x)\endgroup} 
Our parameter scheme is in line with common literature. SciPy naming scheme
has comparable notion of location and scale across distributions. So both
have their benefits.

\classinfo%
{de.lmu.ifi.dbs.elki.math.statistics.distribution.}{KappaDistribution}%
{0.6.0}%
{}%
{Kappa Distribution}%

Kappa distribution, by Hosking.

\classinfo%
{de.lmu.ifi.dbs.elki.math.statistics.distribution.}{LaplaceDistribution}%
{0.6.0}%
{}%
{Laplace Distribution}%

Laplace distribution also known as double exponential distribution

\classinfo%
{de.lmu.ifi.dbs.elki.math.statistics.distribution.}{LogGammaDistribution}%
{0.6.0}%
{}%
{Log Gamma Distribution}%

Log-Gamma Distribution, with random generation and density functions.

This distribution can be outlined as Y ~ exp(Gamma) or equivalently Log(Y) ~
Gamma.

Note: this is a different loggamma than scipy uses, but corresponds to the
Log Gamma Distribution of Wolfram, who notes that it is "at times confused
with ExpGammaDistribution".

\classinfo%
{de.lmu.ifi.dbs.elki.math.statistics.distribution.}{LogLogisticDistribution}%
{0.6.0}%
{}%
{Log Logistic Distribution}%

Log-Logistic distribution also known as Fisk distribution.

\classinfo%
{de.lmu.ifi.dbs.elki.math.statistics.distribution.}{LogNormalDistribution}%
{0.5.0}%
{}%
{Log Normal Distribution}%

Log-Normal distribution.

The parameterization of this class is somewhere inbetween of GNU R and SciPy.
Similar to GNU R we use the logmean and logstddev. Similar to Scipy, we also
have a location parameter that shifts the distribution.

Our implementation maps to SciPy's as follows:
\texttt{scipy.stats.lognorm(logstddev, shift, FastMath.exp(logmean))}

\classinfo%
{de.lmu.ifi.dbs.elki.math.statistics.distribution.}{LogisticDistribution}%
{0.6.0}%
{}%
{Logistic Distribution}%

Logistic distribution.

\classinfo%
{de.lmu.ifi.dbs.elki.math.statistics.distribution.}{NormalDistribution}%
{0.5.0}%
{doi:10.18637/jss.v011.i04,web/Ooura96}%
{Normal Distribution}%

Gaussian distribution aka normal distribution

\classinfo%
{de.lmu.ifi.dbs.elki.math.statistics.distribution.}{PoissonDistribution}%
{0.5.0}%
{web/Loader00}%
{Poisson Distribution}%

INCOMPLETE implementation of the poisson distribution.

\classinfo%
{de.lmu.ifi.dbs.elki.math.statistics.distribution.}{RayleighDistribution}%
{0.6.0}%
{}%
{Rayleigh Distribution}%

Rayleigh distribution, a special case of the Weibull distribution.

\classinfo%
{de.lmu.ifi.dbs.elki.math.statistics.distribution.}{SkewGeneralizedNormalDistribution}%
{0.6.0}%
{doi:10.1017/CBO9780511529443}%
{Skew Generalized Normal Distribution}%

Generalized normal distribution by adding a skew term, similar to lognormal
distributions.

 This is one kind of generalized normal distributions. Note that there are
multiple that go by the name of a "Generalized Normal Distribution"; this is
what is currently called "version 2" in English Wikipedia.

\classinfo%
{de.lmu.ifi.dbs.elki.math.statistics.distribution.}{StudentsTDistribution}%
{0.5.0}%
{}%
{Students T Distribution}%

Student's t distribution.

\classinfo%
{de.lmu.ifi.dbs.elki.math.statistics.distribution.}{UniformDistribution}%
{0.2}%
{}%
{Uniform Distribution}%

Uniform distribution.

\classinfo%
{de.lmu.ifi.dbs.elki.math.statistics.distribution.}{WeibullDistribution}%
{0.6.0}%
{}%
{Weibull Distribution}%

Weibull distribution.

\pkginfo{8}%
{de.lmu.ifi.dbs.elki.math.statistics.distribution.}{estimator}%
{Distribution Estimator}%

Estimators for statistical distributions.

\classinfo%
{de.lmu.ifi.dbs.elki.math.statistics.distribution.estimator.}{CauchyMADEstimator}%
{0.6.0}%
{books/Olive08}%
{Cauchy MAD Estimator}%

Estimate Cauchy distribution parameters using Median and MAD.

\classinfo%
{de.lmu.ifi.dbs.elki.math.statistics.distribution.estimator.}{EMGOlivierNorbergEstimator}%
{0.6.0}%
{doi:10.21500/20112084.846}%
{EMG Olivier Norberg Estimator}%

Naive distribution estimation using mean and sample variance.

\classinfo%
{de.lmu.ifi.dbs.elki.math.statistics.distribution.estimator.}{ExpGammaExpMOMEstimator}%
{0.7.5}%
{}%
{Exp Gamma Exp MOM Estimator}%

Simple parameter estimation for the ExpGamma distribution.

This is a very naive estimation, based on the mean and variance only,
sometimes referred to as the "Method of Moments" (MOM).

This estimator based on the {\begingroup\ttfamily{}GammaMOMEstimator\endgroup} and a simple exp data
transformation.

\classinfo%
{de.lmu.ifi.dbs.elki.math.statistics.distribution.estimator.}{ExponentialLMMEstimator}%
{0.6.0}%
{tr/ibm/Hosking00}%
{Exponential LMM Estimator}%

Estimate the parameters of a Gamma Distribution, using the methods of
L-Moments (LMM).

\classinfo%
{de.lmu.ifi.dbs.elki.math.statistics.distribution.estimator.}{ExponentialMADEstimator}%
{0.6.0}%
{books/Olive08}%
{Exponential MAD Estimator}%

Estimate Exponential distribution parameters using Median and MAD.

\classinfo%
{de.lmu.ifi.dbs.elki.math.statistics.distribution.estimator.}{ExponentialMOMEstimator}%
{0.6.0}%
{}%
{Exponential MOM Estimator}%

Estimate Exponential distribution parameters using the mean, which is the
maximum-likelihood estimate (MLE), but not very robust.

\classinfo%
{de.lmu.ifi.dbs.elki.math.statistics.distribution.estimator.}{ExponentialMedianEstimator}%
{0.6.0}%
{preprints/Olive06}%
{Exponential Median Estimator}%

Estimate Exponential distribution parameters using Median and MAD.

\classinfo%
{de.lmu.ifi.dbs.elki.math.statistics.distribution.estimator.}{GammaChoiWetteEstimator}%
{0.6.0}%
{doi:10.2307/1266892}%
{Gamma Choi Wette Estimator}%

Estimate distribution parameters using the method by Choi and Wette.

\classinfo%
{de.lmu.ifi.dbs.elki.math.statistics.distribution.estimator.}{GammaLMMEstimator}%
{0.6.0}%
{tr/ibm/Hosking00}%
{Gamma LMM Estimator}%

Estimate the parameters of a Gamma Distribution, using the methods of
L-Moments (LMM).

 J. R. M. Hosking\\
 Fortran routines for use with the method of L-moments Version 3.03\\
 IBM Research.

\classinfo%
{de.lmu.ifi.dbs.elki.math.statistics.distribution.estimator.}{GammaMOMEstimator}%
{0.6.0}%
{books/duxbury/CasellaB90/Ch7}%
{Gamma MOM Estimator}%

Simple parameter estimation for the Gamma distribution.

 This is a very naive estimation, based on the mean and variance only,
sometimes referred to as the "Method of Moments" (MOM).

\classinfo%
{de.lmu.ifi.dbs.elki.math.statistics.distribution.estimator.}{GeneralizedExtremeValueLMMEstimator}%
{0.6.0}%
{doi:10.1080/00401706.1985.10488049}%
{Generalized Extreme Value LMM Estimator}%

Estimate the parameters of a Generalized Extreme Value Distribution, using
the methods of L-Moments (LMM).

\classinfo%
{de.lmu.ifi.dbs.elki.math.statistics.distribution.estimator.}{GeneralizedLogisticAlternateLMMEstimator}%
{0.6.0}%
{tr/ibm/Hosking00}%
{Generalized Logistic Alternate LMM Estimator}%

Estimate the parameters of a Generalized Logistic Distribution, using the
methods of L-Moments (LMM).

\classinfo%
{de.lmu.ifi.dbs.elki.math.statistics.distribution.estimator.}{GeneralizedParetoLMMEstimator}%
{0.7.0}%
{doi:10.1080/00401706.1985.10488049}%
{Generalized Pareto LMM Estimator}%

Estimate the parameters of a Generalized Pareto Distribution (GPD), using the
methods of L-Moments (LMM).

\classinfo%
{de.lmu.ifi.dbs.elki.math.statistics.distribution.estimator.}{GumbelLMMEstimator}%
{0.6.0}%
{tr/ibm/Hosking00}%
{Gumbel LMM Estimator}%

Estimate the parameters of a Gumbel Distribution, using the methods of
L-Moments (LMM).

\classinfo%
{de.lmu.ifi.dbs.elki.math.statistics.distribution.estimator.}{GumbelMADEstimator}%
{0.6.0}%
{books/Olive08}%
{Gumbel MAD Estimator}%

Parameter estimation via median and median absolute deviation from median
(MAD).

\classinfo%
{de.lmu.ifi.dbs.elki.math.statistics.distribution.estimator.}{InverseGaussianMLEstimator}%
{0.6.0}%
{}%
{Inverse Gaussian ML Estimator}%

Estimate parameter of the inverse Gaussian (Wald) distribution.

\classinfo%
{de.lmu.ifi.dbs.elki.math.statistics.distribution.estimator.}{InverseGaussianMOMEstimator}%
{0.6.0}%
{}%
{Inverse Gaussian MOM Estimator}%

Estimate parameter of the inverse Gaussian (Wald) distribution.

\classinfo%
{de.lmu.ifi.dbs.elki.math.statistics.distribution.estimator.}{LaplaceLMMEstimator}%
{0.6.0}%
{}%
{Laplace LMM Estimator}%

Estimate Laplace distribution parameters using the method of L-Moments (LMM).

\classinfo%
{de.lmu.ifi.dbs.elki.math.statistics.distribution.estimator.}{LaplaceMADEstimator}%
{0.6.0}%
{books/Olive08}%
{Laplace MAD Estimator}%

Estimate Laplace distribution parameters using Median and MAD.

\classinfo%
{de.lmu.ifi.dbs.elki.math.statistics.distribution.estimator.}{LaplaceMLEEstimator}%
{0.6.0}%
{doi:10.2307/2683252}%
{Laplace MLE Estimator}%

Estimate Laplace distribution parameters using Median and mean deviation from
median.

\classinfo%
{de.lmu.ifi.dbs.elki.math.statistics.distribution.estimator.}{LogGammaLogMOMEstimator}%
{0.7.5}%
{}%
{Log Gamma Log MOM Estimator}%

Simple parameter estimation for the LogGamma distribution.

This is a very naive estimation, based on the mean and variance only,
sometimes referred to as the "Method of Moments" (MOM).

This estimator based on the {\begingroup\ttfamily{}GammaMOMEstimator\endgroup} and a simple log data
transformation.

\classinfo%
{de.lmu.ifi.dbs.elki.math.statistics.distribution.estimator.}{LogLogisticMADEstimator}%
{0.6.0}%
{books/Olive08}%
{Log Logistic MAD Estimator}%

Estimate Logistic distribution parameters using Median and MAD.

\classinfo%
{de.lmu.ifi.dbs.elki.math.statistics.distribution.estimator.}{LogNormalBilkovaLMMEstimator}%
{0.6.0}%
{journals/naun/Bilkova12}%
{Log Normal Bilkova LMM Estimator}%

Alternate estimate the parameters of a log Gamma Distribution, using the
methods of L-Moments (LMM) for the Generalized Normal Distribution.

\classinfo%
{de.lmu.ifi.dbs.elki.math.statistics.distribution.estimator.}{LogNormalLMMEstimator}%
{0.6.0}%
{tr/ibm/Hosking00}%
{Log Normal LMM Estimator}%

Estimate the parameters of a log Normal Distribution, using the methods of
L-Moments (LMM) for the Generalized Normal Distribution.

\classinfo%
{de.lmu.ifi.dbs.elki.math.statistics.distribution.estimator.}{LogNormalLevenbergMarquardtKDEEstimator}%
{0.6.0}%
{}%
{Log Normal Levenberg Marquardt KDE Estimator}%

Distribution parameter estimation using Levenberg-Marquardt iterative
optimization and a kernel density estimation.

 Note: this estimator is rather expensive, and needs optimization in the KDE
phase, which currently is O(n²)!

 This estimator is primarily attractive when only part of the distribution was
observed.

\classinfo%
{de.lmu.ifi.dbs.elki.math.statistics.distribution.estimator.}{LogNormalLogMADEstimator}%
{0.6.0}%
{doi:10.2307/2285666}%
{Log Normal Log MAD Estimator}%

Estimator using Medians. More robust to outliers, and just slightly more
expensive (needs to copy the data for partial sorting to find the median).

\classinfo%
{de.lmu.ifi.dbs.elki.math.statistics.distribution.estimator.}{LogNormalLogMOMEstimator}%
{0.6.0}%
{}%
{Log Normal Log MOM Estimator}%

Naive distribution estimation using mean and sample variance.

This is a maximum-likelihood-estimator (MLE).

\classinfo%
{de.lmu.ifi.dbs.elki.math.statistics.distribution.estimator.}{LogisticLMMEstimator}%
{0.6.0}%
{tr/ibm/Hosking00}%
{Logistic LMM Estimator}%

Estimate the parameters of a Logistic Distribution, using the methods of
L-Moments (LMM).

\classinfo%
{de.lmu.ifi.dbs.elki.math.statistics.distribution.estimator.}{LogisticMADEstimator}%
{0.6.0}%
{preprints/Olive06}%
{Logistic MAD Estimator}%

Estimate Logistic distribution parameters using Median and MAD.

\classinfo%
{de.lmu.ifi.dbs.elki.math.statistics.distribution.estimator.}{NormalLMMEstimator}%
{0.6.0}%
{tr/ibm/Hosking00}%
{Normal LMM Estimator}%

Estimate the parameters of a normal distribution using the method of
L-Moments (LMM).

\classinfo%
{de.lmu.ifi.dbs.elki.math.statistics.distribution.estimator.}{NormalLevenbergMarquardtKDEEstimator}%
{0.6.0}%
{}%
{Normal Levenberg Marquardt KDE Estimator}%

Distribution parameter estimation using Levenberg-Marquardt iterative
optimization and a kernel density estimation.

 Note: this estimator is rather expensive, and needs optimization in the KDE
phase, which currently is O(n²)!

 This estimator is primarily attractive when only part of the distribution was
observed.

\classinfo%
{de.lmu.ifi.dbs.elki.math.statistics.distribution.estimator.}{NormalMADEstimator}%
{0.6.0}%
{doi:10.2307/2285666}%
{Normal MAD Estimator}%

Estimator using Medians. More robust to outliers, and just slightly more
expensive (needs to copy the data for partial sorting to find the median).

\classinfo%
{de.lmu.ifi.dbs.elki.math.statistics.distribution.estimator.}{NormalMOMEstimator}%
{0.6.0}%
{}%
{Normal MOM Estimator}%

Naive maximum-likelihood estimations for the normal distribution using mean
and sample variance.

While this is the most commonly used estimator, it is not very robust against
extreme values.

\classinfo%
{de.lmu.ifi.dbs.elki.math.statistics.distribution.estimator.}{RayleighLMMEstimator}%
{0.6.0}%
{}%
{Rayleigh LMM Estimator}%

Estimate the scale parameter of a (non-shifted) RayleighDistribution using
the method of L-Moments (LMM).

\classinfo%
{de.lmu.ifi.dbs.elki.math.statistics.distribution.estimator.}{RayleighMADEstimator}%
{0.6.0}%
{books/Olive08}%
{Rayleigh MAD Estimator}%

Estimate the parameters of a RayleighDistribution using the MAD.

\classinfo%
{de.lmu.ifi.dbs.elki.math.statistics.distribution.estimator.}{RayleighMLEEstimator}%
{0.6.0}%
{}%
{Rayleigh MLE Estimator}%

Estimate the scale parameter of a (non-shifted) RayleighDistribution using a
maximum likelihood estimate.

\classinfo%
{de.lmu.ifi.dbs.elki.math.statistics.distribution.estimator.}{SkewGNormalLMMEstimator}%
{0.6.0}%
{tr/ibm/Hosking00}%
{Skew G Normal LMM Estimator}%

Estimate the parameters of a skew Normal Distribution (Hoskin's Generalized
Normal Distribution), using the methods of L-Moments (LMM).

\classinfo%
{de.lmu.ifi.dbs.elki.math.statistics.distribution.estimator.}{UniformEnhancedMinMaxEstimator}%
{0.6.0}%
{}%
{Uniform Enhanced Min Max Estimator}%

Slightly improved estimation, that takes sample size into account and
enhances the interval appropriately.

\classinfo%
{de.lmu.ifi.dbs.elki.math.statistics.distribution.estimator.}{UniformLMMEstimator}%
{0.6.0}%
{}%
{Uniform LMM Estimator}%

Estimate the parameters of a normal distribution using the method of
L-Moments (LMM).

\classinfo%
{de.lmu.ifi.dbs.elki.math.statistics.distribution.estimator.}{UniformMADEstimator}%
{0.6.0}%
{books/Olive08}%
{Uniform MAD Estimator}%

Estimate Uniform distribution parameters using Median and MAD.

\classinfo%
{de.lmu.ifi.dbs.elki.math.statistics.distribution.estimator.}{UniformMinMaxEstimator}%
{0.6.0}%
{}%
{Uniform Min Max Estimator}%

Estimate the uniform distribution by computing min and max.

\classinfo%
{de.lmu.ifi.dbs.elki.math.statistics.distribution.estimator.}{WeibullLMMEstimator}%
{0.6.0}%
{}%
{Weibull LMM Estimator}%

Estimate parameters of the Weibull distribution using the method of L-Moments
(LMM).

\classinfo%
{de.lmu.ifi.dbs.elki.math.statistics.distribution.estimator.}{WeibullLogMADEstimator}%
{0.6.0}%
{books/Olive08}%
{Weibull Log MAD Estimator}%

Parameter estimation via median and median absolute deviation from median
(MAD).

\pkginfo{9}%
{de.lmu.ifi.dbs.elki.math.statistics.distribution.estimator.}{meta}%
{Distribution Estimator Meta}%

Meta estimators: estimators that do not actually estimate themselves, but instead use other estimators, e.g. on a trimmed data set, or as an ensemble.

\classinfo%
{de.lmu.ifi.dbs.elki.math.statistics.distribution.estimator.meta.}{BestFitEstimator}%
{0.6.0}%
{}%
{Best Fit Estimator}%

A meta estimator that will try a number of (inexpensive) estimations, then
choose whichever works best.

\classinfo%
{de.lmu.ifi.dbs.elki.math.statistics.distribution.estimator.meta.}{TrimmedEstimator}%
{0.6.0}%
{}%
{Trimmed Estimator}%

Trimmed wrapper around other estimators. Sorts the data, trims it, then
analyzes it using another estimator.

\classinfo%
{de.lmu.ifi.dbs.elki.math.statistics.distribution.estimator.meta.}{WinsorizingEstimator}%
{0.6.0}%
{doi:10.1214/aoms/1177730388}%
{Winsorizing Estimator}%

Winsorizing or Georgization estimator. Similar to trimming, this is supposed
to be more robust to outliers. However, instead of removing the extreme
values, they are instead replaced with the cutoff value. This keeps the
quantity of the data the same, and will have a lower impact on variance and
similar measures.

\pkginfo{7}%
{de.lmu.ifi.dbs.elki.math.statistics.}{intrinsicdimensionality}%
{Intrinsicdimensionality}%

Methods for estimating the intrinsic dimensionality.

\classinfo%
{de.lmu.ifi.dbs.elki.math.statistics.intrinsicdimensionality.}{ALIDEstimator}%
{0.7.5}%
{tr/nii/ChellyHK16}%
{ALID Estimator}%

ALID estimator of the intrinsic dimensionality (maximum likelihood estimator
for ID using auxiliary distances).

\classinfo%
{de.lmu.ifi.dbs.elki.math.statistics.intrinsicdimensionality.}{AggregatedHillEstimator}%
{0.7.0}%
{doi:10.1198/073500101316970421}%
{Aggregated Hill Estimator}%

Estimator using the weighted average of multiple hill estimators.

\classinfo%
{de.lmu.ifi.dbs.elki.math.statistics.intrinsicdimensionality.}{EnsembleEstimator}%
{0.7.0}%
{}%
{Ensemble Estimator}%

Ensemble estimator taking the median of three of our best estimators.

However, the method-of-moments estimator seems to work best at least on
artificial distances - you don't benefit from always choosing the second
best, so this ensemble approach does not appear to help.

This is an experimental estimator. Please cite ELKI when using.

\classinfo%
{de.lmu.ifi.dbs.elki.math.statistics.intrinsicdimensionality.}{GEDEstimator}%
{0.7.0}%
{DBLP:conf/icdm/HouleKN12}%
{GED Estimator}%

Generalized Expansion Dimension for estimating the intrinsic dimensionality.

\classinfo%
{de.lmu.ifi.dbs.elki.math.statistics.intrinsicdimensionality.}{HillEstimator}%
{0.7.0}%
{doi:10.1214/aos/1176343247}%
{Hill Estimator}%

Hill estimator of the intrinsic dimensionality (maximum likelihood estimator
for ID).

\classinfo%
{de.lmu.ifi.dbs.elki.math.statistics.intrinsicdimensionality.}{LMomentsEstimator}%
{0.7.0}%
{tr/ibm/Hosking00,DBLP:conf/kdd/AmsalegCFGHKN15}%
{L Moments Estimator}%

Probability weighted moments based estimator using L-Moments.

 This is dervied from the PWM estimators of Amsaleg et al. using the L-Moments
estimation for the exponential distribution.

\classinfo%
{de.lmu.ifi.dbs.elki.math.statistics.intrinsicdimensionality.}{MOMEstimator}%
{0.7.0}%
{DBLP:conf/kdd/AmsalegCFGHKN15}%
{MOM Estimator}%

Methods of moments estimator, using the first moment (i.e. average).

 This could be generalized to higher order moments, but the variance increases
with the order, and we need this to work well with small sample sizes.

\classinfo%
{de.lmu.ifi.dbs.elki.math.statistics.intrinsicdimensionality.}{PWM2Estimator}%
{0.7.0}%
{DBLP:conf/kdd/AmsalegCFGHKN15,doi:10.1029/WR015i005p01055}%
{PWM2 Estimator}%

Probability weighted moments based estimator, using the second moment.

 It can be shown theoretically that this estimator is expected to have a
higher variance than the one using the first moment only, it is included for
completeness only.

\classinfo%
{de.lmu.ifi.dbs.elki.math.statistics.intrinsicdimensionality.}{PWMEstimator}%
{0.7.0}%
{DBLP:conf/kdd/AmsalegCFGHKN15,doi:10.1029/WR015i005p01055}%
{PWM Estimator}%

Probability weighted moments based estimator.

\classinfo%
{de.lmu.ifi.dbs.elki.math.statistics.intrinsicdimensionality.}{RVEstimator}%
{0.7.0}%
{DBLP:conf/kdd/AmsalegCFGHKN15}%
{RV Estimator}%

Regularly Varying Functions estimator of the intrinsic dimensionality

\classinfo%
{de.lmu.ifi.dbs.elki.math.statistics.intrinsicdimensionality.}{ZipfEstimator}%
{0.7.0}%
{doi:10.3150/bj/1137421635,doi:10.1524/strm.1996.14.4.353,doi:10.1080/15326349608807407}%
{Zipf Estimator}%

Zipf estimator (qq-estimator) of the intrinsic dimensionality.

 Unfortunately, this estimator appears to have a bias. We have empirically
modified the plot position such that bias is reduced, but could not find the
proper way of removing this bias for small samples.

\pkginfo{7}%
{de.lmu.ifi.dbs.elki.math.statistics.}{kernelfunctions}%
{Kernelfunctions}%

Kernel functions from statistics.

\classinfo%
{de.lmu.ifi.dbs.elki.math.statistics.kernelfunctions.}{BiweightKernelDensityFunction}%
{0.6.0}%
{doi:10.1016/0167-71528890050-8}%
{Biweight Kernel Density Function}%

Biweight (Quartic) kernel density estimator.

\classinfo%
{de.lmu.ifi.dbs.elki.math.statistics.kernelfunctions.}{CosineKernelDensityFunction}%
{0.6.0}%
{}%
{Cosine Kernel Density Function}%

Cosine kernel density estimator.

\classinfo%
{de.lmu.ifi.dbs.elki.math.statistics.kernelfunctions.}{EpanechnikovKernelDensityFunction}%
{0.6.0}%
{doi:10.1016/0167-71528890050-8}%
{Epanechnikov Kernel Density Function}%

Epanechnikov kernel density estimator.

\classinfo%
{de.lmu.ifi.dbs.elki.math.statistics.kernelfunctions.}{GaussianKernelDensityFunction}%
{0.6.0}%
{doi:10.1016/0167-71528890050-8}%
{Gaussian Kernel Density Function}%

Gaussian kernel density estimator.

\classinfo%
{de.lmu.ifi.dbs.elki.math.statistics.kernelfunctions.}{KernelDensityFunction}%
{0.6.0}%
{doi:10.1016/0167-71528890050-8}%
{Kernel Density Function}%

Inner function of a kernel density estimator.

 Note: as of now, this API does not support asymmetric kernels, which would be
difficult in the multivariate case.

\classinfo%
{de.lmu.ifi.dbs.elki.math.statistics.kernelfunctions.}{TriangularKernelDensityFunction}%
{0.6.0}%
{}%
{Triangular Kernel Density Function}%

Triangular kernel density estimator.

\classinfo%
{de.lmu.ifi.dbs.elki.math.statistics.kernelfunctions.}{TricubeKernelDensityFunction}%
{0.6.0}%
{}%
{Tricube Kernel Density Function}%

Tricube kernel density estimator.

\classinfo%
{de.lmu.ifi.dbs.elki.math.statistics.kernelfunctions.}{TriweightKernelDensityFunction}%
{0.6.0}%
{doi:10.1016/0167-71528890050-8}%
{Triweight Kernel Density Function}%

Triweight kernel density estimator.

\classinfo%
{de.lmu.ifi.dbs.elki.math.statistics.kernelfunctions.}{UniformKernelDensityFunction}%
{0.6.0}%
{doi:10.1016/0167-71528890050-8}%
{Uniform Kernel Density Function}%

Uniform / Rectangular kernel density estimator.

\pkginfo{7}%
{de.lmu.ifi.dbs.elki.math.statistics.}{tests}%
{Tests}%

Statistical tests.

\classinfo%
{de.lmu.ifi.dbs.elki.math.statistics.tests.}{AndersonDarlingTest}%
{0.7.0}%
{doi:10.1214/aoms/1177729437,doi:10.1080/01621459.1974.10480196}%
{Anderson Darling Test}%

Perform Anderson-Darling test for a Gaussian distribution.

 This is a test against normality / goodness of fit. I.e. you can use
it to reject the hypothesis that the data is normal distributed.
Such tests are sensitive to data set size: on small samples, even large
deviations could be by-chance and thus not allow rejection. On the other
hand, on large data sets even a slight deviation can be unlikely to happen if
the data were indeed normal distributed. Thus, this test is more likely to
fail to reject small data sets even when they intuitively do not appear to be
normal distributed, while it will reject large data sets that originate from
a distribution only slightly different from the normal distribution.

 Before using, make sure you have understood statistical tests, and the
difference between failure-to-reject and acceptance!

 The data size should be at least 8 before the results start getting somewhat
reliable. For large data sets, the chance of rejecting the normal
distribution hypothesis increases a lot: no real data looks exactly like a
normal distribution.

\classinfo%
{de.lmu.ifi.dbs.elki.math.statistics.tests.}{KolmogorovSmirnovTest}%
{0.5.0}%
{}%
{Kolmogorov Smirnov Test}%

Kolmogorov-Smirnov test.

Class that tests two given real-valued data samples on whether they might
have originated from the same underlying distribution using the
Kolmogorov-Smirnov test statistic that compares the two empirical cumulative
distribution functions. The KS statistic is defined as the maximum absolute
difference of the empirical CDFs.

\classinfo%
{de.lmu.ifi.dbs.elki.math.statistics.tests.}{StandardizedTwoSampleAndersonDarlingTest}%
{0.7.0}%
{doi:10.1093/biomet/63.1.161,doi:10.1080/01621459.1987.10478517,doi:10.1214/aoms/1177706788}%
{Standardized Two Sample Anderson Darling Test}%

Perform a two-sample Anderson-Darling rank test, and standardize the
statistic according to Scholz and Stephens. Ties are handled as discussed in
Equation 7 of Scholz and Stephens.

 To access the non-standardized A2 scores, use the function
{\begingroup\ttfamily{}unstandardized(double[][])\endgroup}.

 Compared to the Cramer-van Mises test, the Anderson-Darling test puts more
weight on the tail of the distribution. This variant only uses the ranks.

\classinfo%
{de.lmu.ifi.dbs.elki.math.statistics.tests.}{WelchTTest}%
{0.5.0}%
{}%
{Welch T Test}%

Calculates a test statistic according to Welch's t test for two samples
Supplies methods for calculating the degrees of freedom according to the
Welch-Satterthwaite Equation. Also directly calculates a two-sided p-value
for the underlying t-distribution

\pkginfo{5}%
{de.lmu.ifi.dbs.elki.}{utilities}%
{Utilities}%

Utility and helper classes - commonly used data structures, output formatting, exceptions, ...

Specialized utility classes (which often collect static utility methods only) can be found
in other places of ELKI as well, as seen below.

Important utility function collections:
\begin{itemize}
\item Basic and low-level:\begin{itemize}
\item {\begingroup\ttfamily{}Util\endgroup}: Miscellaneous utility functions.
\item {\begingroup\ttfamily{}LoggingUtil\endgroup}: simple logging access.
\item {\begingroup\ttfamily{}MathUtil\endgroup}: Mathematics utility functions.
\item {\begingroup\ttfamily{}VectorUtil\endgroup}: Vector and Matrix functions.
\item {\begingroup\ttfamily{}SpatialUtil\endgroup}: Spatial MBR computations (intersection, union etc.).
\item {\begingroup\ttfamily{}ByteArrayUtil\endgroup}: byte array processing (low-level IO via byte arrays).
\item {\begingroup\ttfamily{}FileUtil\endgroup}: File and file name utility functions.
\item {\begingroup\ttfamily{}ClassGenericsUtil\endgroup}: Generic classes (instantiation, arrays of arrays, sets that require safe but unchecked casts).
\end{itemize}

\item Database-related:\begin{itemize}
\item {\begingroup\ttfamily{}TypeUtil\endgroup}: Data type utility functions and common type definitions.
\item {\begingroup\ttfamily{}QueryUtil\endgroup}: Database Query API simplifications.
\item {\begingroup\ttfamily{}DBIDUtil\endgroup}: Database ID DBID handling.
\item {\begingroup\ttfamily{}DataStoreUtil\endgroup}: Data storage layer (like Maps).
\item {\begingroup\ttfamily{}DatabaseUtil\endgroup}: database utility functions (centroid etc.).
\item {\begingroup\ttfamily{}ResultUtil\endgroup}: result processing functions (e.g. extracting sub-results).
\end{itemize}

\item Output-related:\begin{itemize}
\item {\begingroup\ttfamily{}FormatUtil\endgroup}: output formatting.
\item {\begingroup\ttfamily{}SVGUtil\endgroup}: SVG generation (XML DOM based).
\item {\begingroup\ttfamily{}BatikUtil\endgroup}: Apache Batik SVG utilities (coordinate transforms screen to canvas).
\end{itemize}

\item Specialized:\begin{itemize}
\item {\begingroup\ttfamily{}OptionUtil\endgroup}: Managing parameter settings
\item {\begingroup\ttfamily{}ELKIServiceRegistry\endgroup}: class and classpath inspection.
\item {\begingroup\ttfamily{}RStarTreeUtil\endgroup}: reporting page file accesses.
\end{itemize}

\end{itemize}

\pkginfo{6}%
{de.lmu.ifi.dbs.elki.utilities.}{datastructures}%
{Datastructures}%

Basic memory structures such as heaps and object hierarchies.

\pkginfo{7}%
{de.lmu.ifi.dbs.elki.utilities.datastructures.}{arrays}%
{Arrays}%

Utilities for arrays: advanced sorting for primitvie arrays.

\classinfo%
{de.lmu.ifi.dbs.elki.utilities.datastructures.arrays.}{IntegerArrayQuickSort}%
{0.5.5}%
{web/Yaroslavskiy09}%
{Integer Array Quick Sort}%

Class to sort an int array, using a modified quicksort.

\pkginfo{7}%
{de.lmu.ifi.dbs.elki.utilities.datastructures.}{unionfind}%
{Unionfind}%

Union-find data structures.

\classinfo%
{de.lmu.ifi.dbs.elki.utilities.datastructures.unionfind.}{WeightedQuickUnionInteger}%
{0.7.1}%
{DBLP:books/daglib/0004943}%
{Weighted Quick Union Integer}%

Union-find algorithm for primitive integers, with optimizations.

 This is the weighted quick union approach, weighted by count and using
path-halving for optimization.

\classinfo%
{de.lmu.ifi.dbs.elki.utilities.datastructures.unionfind.}{WeightedQuickUnionRangeDBIDs}%
{0.7.0}%
{DBLP:books/daglib/0004943}%
{Weighted Quick Union Range DBI Ds}%

Union-find algorithm for {\begingroup\ttfamily{}DBIDRange\endgroup} only, with optimizations.

 To instantiate, use {\begingroup\ttfamily{}UnionFindUtil.make(de.lmu.ifi.dbs.elki.database.ids.StaticDBIDs)\endgroup}. This version is optimized for
{\begingroup\ttfamily{}DBIDRange\endgroup}s.

 This is the weighted quick union approach, weighted by count and using
path-halving for optimization.

\classinfo%
{de.lmu.ifi.dbs.elki.utilities.datastructures.unionfind.}{WeightedQuickUnionStaticDBIDs}%
{0.7.0}%
{DBLP:books/daglib/0004943}%
{Weighted Quick Union Static DBI Ds}%

Union-find algorithm for {\begingroup\ttfamily{}StaticDBIDs\endgroup}, with optimizations.

 To instantiate, use {\begingroup\ttfamily{}UnionFindUtil.make(de.lmu.ifi.dbs.elki.database.ids.StaticDBIDs)\endgroup}, which will automatically
choose the best implementation available.

 This is the weighted quick union approach, weighted by count and using
path-halving for optimization.

 This version needs more memory than {\begingroup\ttfamily{}WeightedQuickUnionRangeDBIDs\endgroup} but
can work with any unmodifiable DBID set (although {\begingroup\ttfamily{}ArrayDBIDs\endgroup} are
recommended).

\pkginfo{6}%
{de.lmu.ifi.dbs.elki.utilities.}{ensemble}%
{Ensemble}%

Utility classes for simple ensembles.

\classinfo%
{de.lmu.ifi.dbs.elki.utilities.ensemble.}{EnsembleVotingInverseMultiplicative}%
{0.6.0}%
{}%
{Ensemble Voting Inverse Multiplicative}%

Inverse multiplicative voting:
\( 1-\prod_i(1-s_i) \)

\classinfo%
{de.lmu.ifi.dbs.elki.utilities.ensemble.}{EnsembleVotingMax}%
{0.5.5}%
{}%
{Ensemble Voting Max}%

Simple combination rule, by taking the maximum.

\classinfo%
{de.lmu.ifi.dbs.elki.utilities.ensemble.}{EnsembleVotingMean}%
{0.5.5}%
{}%
{Ensemble Voting Mean}%

Simple combination rule, by taking the mean

\classinfo%
{de.lmu.ifi.dbs.elki.utilities.ensemble.}{EnsembleVotingMedian}%
{0.5.5}%
{}%
{Ensemble Voting Median}%

Simple combination rule, by taking the median.

Note: median is very similar to a majority voting!

\classinfo%
{de.lmu.ifi.dbs.elki.utilities.ensemble.}{EnsembleVotingMin}%
{0.5.5}%
{}%
{Ensemble Voting Min}%

Simple combination rule, by taking the minimum.

\classinfo%
{de.lmu.ifi.dbs.elki.utilities.ensemble.}{EnsembleVotingMultiplicative}%
{0.6.0}%
{}%
{Ensemble Voting Multiplicative}%

Inverse multiplicative voting:
\( \prod_i s_i \)

\pkginfo{6}%
{de.lmu.ifi.dbs.elki.utilities.}{random}%
{Random}%

Random number generation.

\classinfo%
{de.lmu.ifi.dbs.elki.utilities.random.}{FastNonThreadsafeRandom}%
{0.6.0}%
{blog/Lemire16}%
{Fast Non Threadsafe Random}%

Drop-in replacement for {\begingroup\ttfamily{}Random\endgroup}, but not using atomic long
seeds. This implementation is no longer thread-safe (but faster)!

 It is still the same Linear Congruential Generator (LCG), with a cycle length
of 2$^{48}$, of which we only use 32 bits at a time. Given the same
seed, it is expected to produce the exact same random sequence as Java's
{\begingroup\ttfamily{}Random\endgroup}.

\classinfo%
{de.lmu.ifi.dbs.elki.utilities.random.}{XorShift1024NonThreadsafeRandom}%
{0.7.0}%
{blog/Lemire16,web/Vigna14}%
{Xor Shift 1024 Non Threadsafe Random}%

Replacement for Java's {\begingroup\ttfamily{}Random\endgroup} class, using a different
random number generation strategy. Java's random generator is optimized for
speed, but may lack the randomness needed for more complex experiments.

 This approach is based on the work on XorShift1024* by Sebastiano Vigna, with
the original copyright statement:

 Written in 2014 by Sebastiano Vigna (vigna@acm.org)

 To the extent possible under law, the author has dedicated all copyright and
related and neighboring rights to this software to the public domain
worldwide. This software is distributed without any warranty.

 See http://creativecommons.org/publicdomain/zero/1.0/

\classinfo%
{de.lmu.ifi.dbs.elki.utilities.random.}{XorShift64NonThreadsafeRandom}%
{0.7.0}%
{blog/Lemire16,web/Vigna14}%
{Xor Shift 64 Non Threadsafe Random}%

Replacement for Java's {\begingroup\ttfamily{}Random\endgroup} class, using a different
random number generation strategy. Java's random generator is optimized for
speed, but may lack the randomness needed for more complex experiments.

 This approach is based on the work on XorShift64* by Sebastiano Vigna, with
the original copyright statement:

 Written in 2014 by Sebastiano Vigna (vigna@acm.org)

 To the extent possible under law, the author has dedicated all copyright and
related and neighboring rights to this software to the public domain
worldwide. This software is distributed without any warranty.

 See http://creativecommons.org/publicdomain/zero/1.0/

\classinfo%
{de.lmu.ifi.dbs.elki.utilities.random.}{Xoroshiro128NonThreadsafeRandom}%
{0.7.5}%
{blog/Lemire16,web/BlackmanV16}%
{Xoroshiro 128 Non Threadsafe Random}%

Replacement for Java's {\begingroup\ttfamily{}Random\endgroup} class, using a different
random number generation strategy. Java's random generator is optimized for
speed, but may lack the randomness needed for more complex experiments.

 This approach is based on the work on Xoroshiro128+ by Sebastiano Vigna,
with the original copyright statement:

 Written in 2016 by David Blackman and Sebastiano Vigna (vigna@acm.org)

 To the extent possible under law, the author has dedicated all copyright and
related and neighboring rights to this software to the public domain
worldwide. This software is distributed without any warranty.

 See http://creativecommons.org/publicdomain/zero/1.0/

\pkginfo{6}%
{de.lmu.ifi.dbs.elki.utilities.}{referencepoints}%
{Referencepoints}%

Package containing strategies to obtain reference points

Shared code for various algorithms that use reference points.

\classinfo%
{de.lmu.ifi.dbs.elki.utilities.referencepoints.}{AxisBasedReferencePoints}%
{0.3}%
{}%
{Axis Based Reference Points}%

Strategy to pick reference points by placing them on the axis ends.

This strategy produces n+2 reference points that lie on the edges of the
surrounding cube.

\classinfo%
{de.lmu.ifi.dbs.elki.utilities.referencepoints.}{FullDatabaseReferencePoints}%
{0.3}%
{}%
{Full Database Reference Points}%

Strategy to use the complete database as reference points.

\classinfo%
{de.lmu.ifi.dbs.elki.utilities.referencepoints.}{GridBasedReferencePoints}%
{0.3}%
{}%
{Grid Based Reference Points}%

Grid-based strategy to pick reference points.

\classinfo%
{de.lmu.ifi.dbs.elki.utilities.referencepoints.}{RandomGeneratedReferencePoints}%
{0.3}%
{}%
{Random Generated Reference Points}%

Reference points generated randomly within the used data space.

\classinfo%
{de.lmu.ifi.dbs.elki.utilities.referencepoints.}{RandomSampleReferencePoints}%
{0.3}%
{}%
{Random Sample Reference Points}%

Random-Sampling strategy for picking reference points.

\classinfo%
{de.lmu.ifi.dbs.elki.utilities.referencepoints.}{StarBasedReferencePoints}%
{0.3}%
{}%
{Star Based Reference Points}%

Star-based strategy to pick reference points.

\pkginfo{6}%
{de.lmu.ifi.dbs.elki.utilities.}{scaling}%
{Scaling}%

Scaling functions: linear, logarithmic, gamma, clipping, ...

\classinfo%
{de.lmu.ifi.dbs.elki.utilities.scaling.}{ClipScaling}%
{0.3}%
{}%
{Clip Scaling}%

Scale implementing a simple clipping. Values less than the specified minimum
will be set to the minimum, values larger than the maximum will be set to the
maximum.

\classinfo%
{de.lmu.ifi.dbs.elki.utilities.scaling.}{GammaScaling}%
{0.3}%
{}%
{Gamma Scaling}%

Non-linear scaling function using a Gamma curve.

\classinfo%
{de.lmu.ifi.dbs.elki.utilities.scaling.}{IdentityScaling}%
{0.3}%
{}%
{Identity Scaling}%

The trivial "identity" scaling function.

\classinfo%
{de.lmu.ifi.dbs.elki.utilities.scaling.}{LinearScaling}%
{0.3}%
{}%
{Linear Scaling}%

Simple linear scaling function.

\classinfo%
{de.lmu.ifi.dbs.elki.utilities.scaling.}{MinusLogScaling}%
{0.3}%
{}%
{Minus Log Scaling}%

Scaling function to invert values by computing -1 * Math.log(x)

\pkginfo{7}%
{de.lmu.ifi.dbs.elki.utilities.scaling.}{outlier}%
{Outlier}%

Scaling of Outlier scores, that require a statistical analysis of the occurring values

\classinfo%
{de.lmu.ifi.dbs.elki.utilities.scaling.outlier.}{COPOutlierScaling}%
{0.6.0}%
{DBLP:conf/icdm/KriegelKSZ12,DBLP:conf/sdm/KriegelKSZ11}%
{COP Outlier Scaling}%

CDF based outlier score scaling.

 Enhanced version of the scaling proposed in:

 Hans-Peter Kriegel, Peer Kröger, Erich Schubert, Arthur Zimek\\
 Interpreting and Unifying Outlier Scores\\
 Proc. 11th SIAM International Conference on Data Mining (SDM 2011)

 See also:

 Hans-Peter Kriegel, Peer Kröger, Erich Schubert, Arthur Zimek\\
 Outlier Detection in Arbitrarily Oriented Subspaces\\
 in: Proc. IEEE Int. Conf. on Data Mining (ICDM 2012)

\classinfo%
{de.lmu.ifi.dbs.elki.utilities.scaling.outlier.}{HeDESNormalizationOutlierScaling}%
{0.4.0}%
{DBLP:conf/dasfaa/VuAG10}%
{HeDES Normalization Outlier Scaling}%

Normalization used by HeDES

\classinfo%
{de.lmu.ifi.dbs.elki.utilities.scaling.outlier.}{LogRankingPseudoOutlierScaling}%
{0.7.0}%
{}%
{Log Ranking Pseudo Outlier Scaling}%

This is a pseudo outlier scoring obtained by only considering the ranks of
the objects. However, the ranks are not mapped linearly to scores, but using
a normal distribution.

\classinfo%
{de.lmu.ifi.dbs.elki.utilities.scaling.outlier.}{MinusLogGammaScaling}%
{0.3}%
{DBLP:conf/sdm/KriegelKSZ11}%
{Minus Log Gamma Scaling}%

Scaling that can map arbitrary values to a probability in the range of [0:1],
by assuming a Gamma distribution on the data and evaluating the Gamma CDF.

\classinfo%
{de.lmu.ifi.dbs.elki.utilities.scaling.outlier.}{MinusLogStandardDeviationScaling}%
{0.3}%
{DBLP:conf/sdm/KriegelKSZ11}%
{Minus Log Standard Deviation Scaling}%

Scaling that can map arbitrary values to a probability in the range of [0:1].

 Transformation is done using the formula
\(\max\{0, \mathrm{erf}(\lambda \frac{x-\mu}{\sigma\sqrt{2}})\}\)

 Where mean can be fixed to a given value, and stddev is then computed against
this mean.

\classinfo%
{de.lmu.ifi.dbs.elki.utilities.scaling.outlier.}{MixtureModelOutlierScaling}%
{0.4.0}%
{DBLP:conf/icdm/GaoT06}%
{Mixture Model Outlier Scaling}%

Tries to fit a mixture model (exponential for inliers and gaussian for
outliers) to the outlier score distribution.

Note: we found this method to often fail, and fit the normal distribution to
the inliers instead of the outliers, yielding reversed results.

\classinfo%
{de.lmu.ifi.dbs.elki.utilities.scaling.outlier.}{MultiplicativeInverseScaling}%
{0.3}%
{DBLP:conf/sdm/KriegelKSZ11}%
{Multiplicative Inverse Scaling}%

Scaling function to invert values by computing 1/x, but in a variation that
maps the values to the [0:1] interval and avoiding division by 0.

 The exact formula can be written as
\[ 1 / (v \cdot \max_{x\neq 0}\frac{1}{|x|}) = \min_{x \neq 0}(|x|) / v \]
with 1 / 0 := 1

\classinfo%
{de.lmu.ifi.dbs.elki.utilities.scaling.outlier.}{OutlierGammaScaling}%
{0.3}%
{DBLP:conf/sdm/KriegelKSZ11}%
{Outlier Gamma Scaling}%

Scaling that can map arbitrary values to a probability in the range of [0:1]
by assuming a Gamma distribution on the values.

\classinfo%
{de.lmu.ifi.dbs.elki.utilities.scaling.outlier.}{OutlierLinearScaling}%
{0.3}%
{}%
{Outlier Linear Scaling}%

Scaling that can map arbitrary values to a value in the range of [0:1].

Transformation is done by linear mapping onto 0:1 using the minimum and
maximum values.

\classinfo%
{de.lmu.ifi.dbs.elki.utilities.scaling.outlier.}{OutlierMinusLogScaling}%
{0.3}%
{DBLP:conf/sdm/KriegelKSZ11}%
{Outlier Minus Log Scaling}%

Scaling function to invert values by computing -log(x)

 Useful for example for scaling
{\begingroup\ttfamily{}ABOD\endgroup}, but see
{\begingroup\ttfamily{}MinusLogStandardDeviationScaling\endgroup} and {\begingroup\ttfamily{}MinusLogGammaScaling\endgroup} for
more advanced scalings for this algorithm.

\classinfo%
{de.lmu.ifi.dbs.elki.utilities.scaling.outlier.}{OutlierSqrtScaling}%
{0.3}%
{}%
{Outlier Sqrt Scaling}%

Scaling that can map arbitrary positive values to a value in the range of
[0:1].

 Transformation is done by taking the square root, then doing a linear linear
mapping onto 0:1 using the minimum values seen.

\classinfo%
{de.lmu.ifi.dbs.elki.utilities.scaling.outlier.}{RankingPseudoOutlierScaling}%
{0.4.0}%
{}%
{Ranking Pseudo Outlier Scaling}%

This is a pseudo outlier scoring obtained by only considering the ranks of
the objects. However, the ranks are not mapped linearly to scores, but using
a normal distribution.

\classinfo%
{de.lmu.ifi.dbs.elki.utilities.scaling.outlier.}{SigmoidOutlierScaling}%
{0.4.0}%
{DBLP:conf/icdm/GaoT06}%
{Sigmoid Outlier Scaling}%

Tries to fit a sigmoid to the outlier scores and use it to convert the values
to probability estimates in the range of 0.0 to 1.0

\classinfo%
{de.lmu.ifi.dbs.elki.utilities.scaling.outlier.}{SqrtStandardDeviationScaling}%
{0.3}%
{DBLP:conf/sdm/KriegelKSZ11}%
{Sqrt Standard Deviation Scaling}%

Scaling that can map arbitrary values to a probability in the range of [0:1].

 Transformation is done using the formulas
\[y = \sqrt{x - \min}\]
\[s = \max\{0, \textrm{erf}(\lambda \frac{y-\mu}{\sigma\sqrt{2}})\}\]

 Where min and mean \(\mu\) can be fixed to a given value, and stddev
\(\sigma\) is then computed against this mean.

\classinfo%
{de.lmu.ifi.dbs.elki.utilities.scaling.outlier.}{StandardDeviationScaling}%
{0.3}%
{DBLP:conf/sdm/KriegelKSZ11}%
{Standard Deviation Scaling}%

Scaling that can map arbitrary values to a probability in the range of [0:1].

 Transformation is done using the formula
\(\max\{0, \mathrm{erf}(\lambda \frac{x-\mu}{\sigma\sqrt{2}})\}\)

 Where mean can be fixed to a given value, and stddev is then computed against
this mean.

\classinfo%
{de.lmu.ifi.dbs.elki.utilities.scaling.outlier.}{TopKOutlierScaling}%
{0.3}%
{}%
{TopK Outlier Scaling}%

Outlier scaling function that only keeps the top k outliers.

\pkginfo{5}%
{de.lmu.ifi.dbs.elki.}{visualization}%
{Visualization}%

Visualization package of ELKI.

\pkginfo{6}%
{de.lmu.ifi.dbs.elki.visualization.}{parallel3d}%
{Parallel3d}%

3DPC: 3D parallel coordinate plot visualization for ELKI.

This is an add-on module. Details were published as:

\classinfo%
{de.lmu.ifi.dbs.elki.visualization.parallel3d.}{OpenGL3DParallelCoordinates}%
{0.6.0}%
{DBLP:conf/sigmod/AchtertKSZ13}%
{Open GL3D Parallel Coordinates}%

Simple JOGL2 based parallel coordinates visualization.

\classinfo%
{de.lmu.ifi.dbs.elki.visualization.parallel3d.}{Parallel3DRenderer}%
{0.6.0}%
{DBLP:conf/sigmod/AchtertKSZ13}%
{Parallel 3D Renderer}%

Renderer for 3D parallel plots.

 The tricky part here is the vertex buffer layout. We are drawing lines, so we
need two vertices for each macro edge (edge between axes in the plot). We
furthermore need the following properties: we need to draw edges sorted by
depth to allow alpha and smoothing to work, and we need to be able to have
different colors for clusters. An efficient batch therefore will consist of
one edge-color combination. The input data comes in color-object ordering, so
we need to seek through the edges when writing the buffer.

 In total, we have 2 * obj.size * edges.size vertices.

 Where obj.size = sum(col.sizes)

\pkginfo{7}%
{de.lmu.ifi.dbs.elki.visualization.parallel3d.}{layout}%
{Layout}%

Layouting algorithms for 3D parallel coordinate plots.

\classinfo%
{de.lmu.ifi.dbs.elki.visualization.parallel3d.layout.}{CompactCircularMSTLayout3DPC}%
{0.6.0}%
{DBLP:conf/sigmod/AchtertKSZ13}%
{Compact Circular MST Layout 3DPC}%

Simple circular layout based on the minimum spanning tree.

\classinfo%
{de.lmu.ifi.dbs.elki.visualization.parallel3d.layout.}{MultidimensionalScalingMSTLayout3DPC}%
{0.6.0}%
{DBLP:conf/sigmod/AchtertKSZ13}%
{Multidimensional Scaling MST Layout 3DPC}%

Layout the axes by multi-dimensional scaling.

\classinfo%
{de.lmu.ifi.dbs.elki.visualization.parallel3d.layout.}{SimpleCircularMSTLayout3DPC}%
{0.6.0}%
{DBLP:conf/sigmod/AchtertKSZ13}%
{Simple Circular MST Layout 3DPC}%

Simple circular layout based on the minimum spanning tree.

\pkginfo{6}%
{de.lmu.ifi.dbs.elki.visualization.}{projector}%
{Projector}%

Projectors are responsible for finding appropriate projections for data relations.

\classinfo%
{de.lmu.ifi.dbs.elki.visualization.projector.}{HistogramFactory}%
{0.4.0}%
{}%
{Histogram Factory}%

Produce one-dimensional projections.

\classinfo%
{de.lmu.ifi.dbs.elki.visualization.projector.}{OPTICSProjectorFactory}%
{0.4.0}%
{}%
{OPTICS Projector Factory}%

Produce OPTICS plot projections

\classinfo%
{de.lmu.ifi.dbs.elki.visualization.projector.}{ParallelPlotFactory}%
{0.5.0}%
{}%
{Parallel Plot Factory}%

Produce parallel axes projections.

\classinfo%
{de.lmu.ifi.dbs.elki.visualization.projector.}{ParallelPlotProjector}%
{0.5.0}%
{doi:10.1007/978-0-387-68628-8}%
{Parallel Plot Projector}%

ParallelPlotProjector is responsible for producing a parallel axes
visualization.

\classinfo%
{de.lmu.ifi.dbs.elki.visualization.projector.}{ScatterPlotFactory}%
{0.4.0}%
{}%
{Scatter Plot Factory}%

Produce scatterplot projections.

\pkginfo{6}%
{de.lmu.ifi.dbs.elki.visualization.}{visualizers}%
{Visualizers}%

Visualizers for various results

\pkginfo{7}%
{de.lmu.ifi.dbs.elki.visualization.visualizers.}{actions}%
{Actions}%

Action-only "visualizers" that only produce menu entries.

\classinfo%
{de.lmu.ifi.dbs.elki.visualization.visualizers.actions.}{ClusterStyleAction}%
{0.7.0}%
{}%
{Cluster Style Action}%

Actions to use clusterings for styling.

\pkginfo{7}%
{de.lmu.ifi.dbs.elki.visualization.visualizers.}{histogram}%
{Histogram}%

Visualizers based on 1D projected histograms.

\classinfo%
{de.lmu.ifi.dbs.elki.visualization.visualizers.histogram.}{ColoredHistogramVisualizer}%
{0.4.0}%
{}%
{Colored Histogram Visualizer}%

Generates a SVG-Element containing a histogram representing the distribution
of the database's objects.

\pkginfo{7}%
{de.lmu.ifi.dbs.elki.visualization.visualizers.}{optics}%
{Optics}%

Visualizers that do work on OPTICS plots

\classinfo%
{de.lmu.ifi.dbs.elki.visualization.visualizers.optics.}{OPTICSClusterVisualization}%
{0.4.0}%
{}%
{OPTICS Cluster Visualization}%

Visualize the clusters and cluster hierarchy found by OPTICS on the OPTICS
Plot.

\classinfo%
{de.lmu.ifi.dbs.elki.visualization.visualizers.optics.}{OPTICSPlotCutVisualization}%
{0.4.0}%
{}%
{OPTICS Plot Cut Visualization}%

Visualizes a cut in an OPTICS Plot to select an Epsilon value and generate a
new clustering result.

\classinfo%
{de.lmu.ifi.dbs.elki.visualization.visualizers.optics.}{OPTICSPlotSelectionVisualization}%
{0.4.0}%
{}%
{OPTICS Plot Selection Visualization}%

Handle the marker in an OPTICS plot.

\classinfo%
{de.lmu.ifi.dbs.elki.visualization.visualizers.optics.}{OPTICSPlotVisualizer}%
{0.3}%
{}%
{OPTICS Plot Visualizer}%

Visualize an OPTICS result by constructing an OPTICS plot for it.

\classinfo%
{de.lmu.ifi.dbs.elki.visualization.visualizers.optics.}{OPTICSSteepAreaVisualization}%
{0.4.0}%
{}%
{OPTICS Steep Area Visualization}%

Visualize the steep areas found in an OPTICS plot

\pkginfo{7}%
{de.lmu.ifi.dbs.elki.visualization.visualizers.}{pairsegments}%
{Pairsegments}%

Visualizers for inspecting cluster differences using pair counting segments.

\classinfo%
{de.lmu.ifi.dbs.elki.visualization.visualizers.pairsegments.}{CircleSegmentsVisualizer}%
{0.5.0}%
{DBLP:conf/icde/AchtertGKSZ12}%
{Circle Segments Visualizer}%

Visualizer to draw circle segments of clusterings and enable interactive
selection of segments. For "empty" segments, all related segments are
selected instead, to visualize the differences.

\pkginfo{7}%
{de.lmu.ifi.dbs.elki.visualization.visualizers.}{parallel}%
{Parallel}%

Visualizers based on parallel coordinates.

\classinfo%
{de.lmu.ifi.dbs.elki.visualization.visualizers.parallel.}{AxisReorderVisualization}%
{0.5.5}%
{}%
{Axis Reorder Visualization}%

Interactive SVG-Elements for reordering the axes.

\classinfo%
{de.lmu.ifi.dbs.elki.visualization.visualizers.parallel.}{AxisVisibilityVisualization}%
{0.5.5}%
{}%
{Axis Visibility Visualization}%

Layer for controlling axis visbility in parallel coordinates.

\classinfo%
{de.lmu.ifi.dbs.elki.visualization.visualizers.parallel.}{BoundingBoxVisualization}%
{0.5.0}%
{}%
{Bounding Box Visualization}%

Draw spatial objects (except vectors!)

\classinfo%
{de.lmu.ifi.dbs.elki.visualization.visualizers.parallel.}{LineVisualization}%
{0.7.0}%
{}%
{Line Visualization}%

Generates data lines.

\classinfo%
{de.lmu.ifi.dbs.elki.visualization.visualizers.parallel.}{ParallelAxisVisualization}%
{0.5.0}%
{}%
{Parallel Axis Visualization}%

Generates a SVG-Element containing axes, including labeling.

\pkginfo{8}%
{de.lmu.ifi.dbs.elki.visualization.visualizers.parallel.}{cluster}%
{Parallel Cluster}%

Visualizers for clustering results based on parallel coordinates.

\classinfo%
{de.lmu.ifi.dbs.elki.visualization.visualizers.parallel.cluster.}{ClusterOutlineVisualization}%
{0.5.0}%
{}%
{Cluster Outline Visualization}%

Generates a SVG-Element that visualizes the area covered by a cluster.

\classinfo%
{de.lmu.ifi.dbs.elki.visualization.visualizers.parallel.cluster.}{ClusterParallelMeanVisualization}%
{0.5.0}%
{}%
{Cluster Parallel Mean Visualization}%

Generates a SVG-Element that visualizes cluster means.

\pkginfo{8}%
{de.lmu.ifi.dbs.elki.visualization.visualizers.parallel.}{index}%
{Parallel Index}%

Visualizers for index structure based on parallel coordinates.

\classinfo%
{de.lmu.ifi.dbs.elki.visualization.visualizers.parallel.index.}{RTreeParallelVisualization}%
{0.5.0}%
{}%
{R Tree Parallel Visualization}%

Visualize the of an R-Tree based index.

\pkginfo{8}%
{de.lmu.ifi.dbs.elki.visualization.visualizers.parallel.}{selection}%
{Parallel Selection}%

Visualizers for object selection based on parallel projections.

\classinfo%
{de.lmu.ifi.dbs.elki.visualization.visualizers.parallel.selection.}{SelectionAxisRangeVisualization}%
{0.5.0}%
{}%
{Selection Axis Range Visualization}%

Visualizer for generating an SVG-Element representing the selected range.

\classinfo%
{de.lmu.ifi.dbs.elki.visualization.visualizers.parallel.selection.}{SelectionLineVisualization}%
{0.5.0}%
{}%
{Selection Line Visualization}%

Visualizer for generating SVG-Elements representing the selected objects

\classinfo%
{de.lmu.ifi.dbs.elki.visualization.visualizers.parallel.selection.}{SelectionToolAxisRangeVisualization}%
{0.5.0}%
{}%
{Selection Tool Axis Range Visualization}%

Tool-Visualization for the tool to select axis ranges

\classinfo%
{de.lmu.ifi.dbs.elki.visualization.visualizers.parallel.selection.}{SelectionToolLineVisualization}%
{0.5.0}%
{}%
{Selection Tool Line Visualization}%

Tool-Visualization for the tool to select objects

\pkginfo{7}%
{de.lmu.ifi.dbs.elki.visualization.visualizers.}{scatterplot}%
{Scatterplot}%

Visualizers based on scatterplots.

\classinfo%
{de.lmu.ifi.dbs.elki.visualization.visualizers.scatterplot.}{AxisVisualization}%
{0.4.0}%
{}%
{Axis Visualization}%

Generates a SVG-Element containing axes, including labeling.

\classinfo%
{de.lmu.ifi.dbs.elki.visualization.visualizers.scatterplot.}{MarkerVisualization}%
{0.5.0}%
{}%
{Marker Visualization}%

Visualize e.g. a clustering using different markers for different clusters.
This visualizer is not constraint to clusters. It can in fact visualize any
kind of result we have a style source for.

\classinfo%
{de.lmu.ifi.dbs.elki.visualization.visualizers.scatterplot.}{PolygonVisualization}%
{0.4.0}%
{}%
{Polygon Visualization}%

Renders PolygonsObject in the data set.

\classinfo%
{de.lmu.ifi.dbs.elki.visualization.visualizers.scatterplot.}{ReferencePointsVisualization}%
{0.4.0}%
{}%
{Reference Points Visualization}%

The actual visualization instance, for a single projection

\classinfo%
{de.lmu.ifi.dbs.elki.visualization.visualizers.scatterplot.}{TooltipScoreVisualization}%
{0.4.0}%
{}%
{Tooltip Score Visualization}%

Generates a SVG-Element containing Tooltips. Tooltips remain invisible until
their corresponding Marker is touched by the cursor and stay visible as long
as the cursor lingers on the marker.

\classinfo%
{de.lmu.ifi.dbs.elki.visualization.visualizers.scatterplot.}{TooltipStringVisualization}%
{0.4.0}%
{}%
{Tooltip String Visualization}%

Generates a SVG-Element containing Tooltips. Tooltips remain invisible until
their corresponding Marker is touched by the cursor and stay visible as long
as the cursor lingers on the marker.

\pkginfo{8}%
{de.lmu.ifi.dbs.elki.visualization.visualizers.scatterplot.}{cluster}%
{Scatterplot Cluster}%

Visualizers for clustering results based on 2D projections.

\classinfo%
{de.lmu.ifi.dbs.elki.visualization.visualizers.scatterplot.cluster.}{ClusterHullVisualization}%
{0.5.0}%
{}%
{Cluster Hull Visualization}%

Visualizer for generating an SVG-Element containing the convex hull / alpha
shape of each cluster.

\classinfo%
{de.lmu.ifi.dbs.elki.visualization.visualizers.scatterplot.cluster.}{ClusterMeanVisualization}%
{0.7.0}%
{}%
{Cluster Mean Visualization}%

Visualize the mean of a KMeans-Clustering

\classinfo%
{de.lmu.ifi.dbs.elki.visualization.visualizers.scatterplot.cluster.}{ClusterOrderVisualization}%
{0.5.0}%
{}%
{Cluster Order Visualization}%

Cluster order visualizer: connect objects via the spanning tree the cluster
order represents.

\classinfo%
{de.lmu.ifi.dbs.elki.visualization.visualizers.scatterplot.cluster.}{ClusterStarVisualization}%
{0.5.0}%
{}%
{Cluster Star Visualization}%

Visualize the mean of a KMeans-Clustering using stars.

\classinfo%
{de.lmu.ifi.dbs.elki.visualization.visualizers.scatterplot.cluster.}{EMClusterVisualization}%
{0.5.0}%
{}%
{EM Cluster Visualization}%

Visualizer for generating SVG-Elements containing ellipses for first, second
and third standard deviation. In more than 2-dimensional data, the class
tries to approximate the cluster extends.

\classinfo%
{de.lmu.ifi.dbs.elki.visualization.visualizers.scatterplot.cluster.}{VoronoiVisualization}%
{0.5.0}%
{}%
{Voronoi Visualization}%

Visualizer drawing Voronoi cells for k-means clusterings.

See also: {\begingroup\ttfamily{}KMeans clustering\endgroup}

\pkginfo{8}%
{de.lmu.ifi.dbs.elki.visualization.visualizers.scatterplot.}{density}%
{Scatterplot Density}%

Visualizers for data set density in a scatterplot projection.

\classinfo%
{de.lmu.ifi.dbs.elki.visualization.visualizers.scatterplot.density.}{DensityEstimationOverlay}%
{0.3}%
{doi:10.1002/9780470316849}%
{Density Estimation Overlay}%

A simple density estimation visualization, based on a simple kernel-density
in the projection, not the actual data!

\pkginfo{8}%
{de.lmu.ifi.dbs.elki.visualization.visualizers.scatterplot.}{index}%
{Scatterplot Index}%

Visualizers for index structures based on 2D projections.

\classinfo%
{de.lmu.ifi.dbs.elki.visualization.visualizers.scatterplot.index.}{TreeMBRVisualization}%
{0.5.0}%
{}%
{Tree MBR Visualization}%

Visualize the bounding rectangles of an R-Tree based index.

\classinfo%
{de.lmu.ifi.dbs.elki.visualization.visualizers.scatterplot.index.}{TreeSphereVisualization}%
{0.5.0}%
{}%
{Tree Sphere Visualization}%

Visualize the bounding sphere of a metric index.

\pkginfo{8}%
{de.lmu.ifi.dbs.elki.visualization.visualizers.scatterplot.}{outlier}%
{Scatterplot Outlier}%

Visualizers for outlier scores based on 2D projections.

\classinfo%
{de.lmu.ifi.dbs.elki.visualization.visualizers.scatterplot.outlier.}{BubbleVisualization}%
{0.5.0}%
{DBLP:conf/dasfaa/AchtertKRSWZ10}%
{Bubble Visualization}%

Generates a SVG-Element containing bubbles. A Bubble is a circle visualizing
an outlierness-score, with its center at the position of the visualized
object and its radius depending on the objects score.

\classinfo%
{de.lmu.ifi.dbs.elki.visualization.visualizers.scatterplot.outlier.}{COPVectorVisualization}%
{0.5.5}%
{DBLP:conf/icdm/KriegelKSZ12}%
{COP: Correlation Outlier Probability}%

Visualize error vectors as produced by COP.

\pkginfo{8}%
{de.lmu.ifi.dbs.elki.visualization.visualizers.scatterplot.}{selection}%
{Scatterplot Selection}%

Visualizers for object selection based on 2D projections.

\classinfo%
{de.lmu.ifi.dbs.elki.visualization.visualizers.scatterplot.selection.}{DistanceFunctionVisualization}%
{0.5.5}%
{}%
{Distance Function Visualization}%

Factory for visualizers to generate an SVG-Element containing dots as markers
representing the kNN of the selected Database objects.

To use this, add a kNN preprocessor index to your database!

\classinfo%
{de.lmu.ifi.dbs.elki.visualization.visualizers.scatterplot.selection.}{MoveObjectsToolVisualization}%
{0.4.0}%
{}%
{Move Objects Tool Visualization}%

Tool to move the currently selected objects.

\classinfo%
{de.lmu.ifi.dbs.elki.visualization.visualizers.scatterplot.selection.}{SelectionConvexHullVisualization}%
{0.4.0}%
{}%
{Selection Convex Hull Visualization}%

Visualizer for generating an SVG-Element containing the convex hull of the
selected points

\classinfo%
{de.lmu.ifi.dbs.elki.visualization.visualizers.scatterplot.selection.}{SelectionCubeVisualization}%
{0.4.0}%
{}%
{Selection Cube Visualization}%

Visualizer for generating an SVG-Element containing a cube as marker
representing the selected range for each dimension

\classinfo%
{de.lmu.ifi.dbs.elki.visualization.visualizers.scatterplot.selection.}{SelectionDotVisualization}%
{0.4.0}%
{}%
{Selection Dot Visualization}%

Visualizer for generating an SVG-Element containing dots as markers
representing the selected Database's objects.

\classinfo%
{de.lmu.ifi.dbs.elki.visualization.visualizers.scatterplot.selection.}{SelectionToolCubeVisualization}%
{0.4.0}%
{}%
{Selection Tool Cube Visualization}%

Tool-Visualization for the tool to select ranges.

\classinfo%
{de.lmu.ifi.dbs.elki.visualization.visualizers.scatterplot.selection.}{SelectionToolDotVisualization}%
{0.4.0}%
{}%
{Selection Tool Dot Visualization}%

Tool-Visualization for the tool to select objects

\pkginfo{8}%
{de.lmu.ifi.dbs.elki.visualization.visualizers.scatterplot.}{uncertain}%
{Scatterplot Uncertain}%

Visualizers for uncertain data.

\classinfo%
{de.lmu.ifi.dbs.elki.visualization.visualizers.scatterplot.uncertain.}{UncertainBoundingBoxVisualization}%
{0.7.0}%
{}%
{Uncertain Bounding Box Visualization}%

Visualize uncertain objects by their bounding box.

Note: this is currently a hack. Our projection only applies to vector field
relations currently, and this visualizer activates if such a relation (e.g. a
sample, or the center of mass) has a parent relation of type UncertainObject.
But it serves the purpose.

\classinfo%
{de.lmu.ifi.dbs.elki.visualization.visualizers.scatterplot.uncertain.}{UncertainInstancesVisualization}%
{0.7.0}%
{}%
{Uncertain Instances Visualization}%

Visualize a single derived sample from an uncertain database.

Note: this is currently a hack. Our projection only applies to vector field
relations currently, and this visualizer activates if such a relation (e.g. a
sample, or the center of mass) has a parent relation of type UncertainObject.
But it serves the purpose.

\classinfo%
{de.lmu.ifi.dbs.elki.visualization.visualizers.scatterplot.uncertain.}{UncertainSamplesVisualization}%
{0.7.0}%
{}%
{Uncertain Samples Visualization}%

Visualize uncertain objects by multiple samples.

Note: this is currently a hack. Our projection only applies to vector field
relations currently, and this visualizer activates if such a relation (e.g. a
sample, or the center of mass) has a parent relation of type UncertainObject.
But it serves the purpose.

\pkginfo{7}%
{de.lmu.ifi.dbs.elki.visualization.visualizers.}{visunproj}%
{Visunproj}%

Visualizers that do not use a particular projection.

\classinfo%
{de.lmu.ifi.dbs.elki.visualization.visualizers.visunproj.}{DendrogramVisualization}%
{0.7.5}%
{}%
{Dendrogram Visualization}%

Dendrogram visualizer.

\classinfo%
{de.lmu.ifi.dbs.elki.visualization.visualizers.visunproj.}{EvaluationVisualization}%
{0.4.0}%
{}%
{Evaluation Visualization}%

Pseudo-Visualizer, that lists the cluster evaluation results found.

\classinfo%
{de.lmu.ifi.dbs.elki.visualization.visualizers.visunproj.}{HistogramVisualization}%
{0.3}%
{}%
{Histogram Visualization}%

Visualizer to draw histograms.

\classinfo%
{de.lmu.ifi.dbs.elki.visualization.visualizers.visunproj.}{KeyVisualization}%
{0.3}%
{}%
{Key Visualization}%

Visualizer, displaying the key for a clustering.

\classinfo%
{de.lmu.ifi.dbs.elki.visualization.visualizers.visunproj.}{LabelVisualization}%
{0.3}%
{}%
{Label Visualization}%

Trivial "visualizer" that displays a static label. The visualizer is meant to
be used for dimension labels in the overview.

\classinfo%
{de.lmu.ifi.dbs.elki.visualization.visualizers.visunproj.}{PixmapVisualizer}%
{0.4.0}%
{}%
{Pixmap Visualizer}%

Visualize an arbitrary pixmap result.

\classinfo%
{de.lmu.ifi.dbs.elki.visualization.visualizers.visunproj.}{SettingsVisualization}%
{0.3}%
{}%
{Settings Visualization}%

Pseudo-Visualizer, that lists the settings of the algorithm-

\classinfo%
{de.lmu.ifi.dbs.elki.visualization.visualizers.visunproj.}{SimilarityMatrixVisualizer}%
{0.4.0}%
{}%
{Similarity Matrix Visualizer}%

Visualize a similarity matrix with object labels

\classinfo%
{de.lmu.ifi.dbs.elki.visualization.visualizers.visunproj.}{XYCurveVisualization}%
{0.3}%
{}%
{XY Curve Visualization}%

Visualizer to render a simple 2D curve such as a ROC curve.

\classinfo%
{de.lmu.ifi.dbs.elki.visualization.visualizers.visunproj.}{XYPlotVisualization}%
{0.7.0}%
{}%
{XY Plot Visualization}%

Visualizer to render a simple 2D curve such as a ROC curve.

\pagebreak
\bibliographystyle{alpha}
\bibliography{elki,wrapper,addlit}

\newcommand{\etalchar}[1]{$^{#1}$}
\begin{thebibliography}{RRdlIRS06}

\bibitem[ABD{\etalchar{+}}08]{DBLP:conf/sdm/AchtertBDKZ08}
Elke Achtert, Christian Böhm, Jörn David, Peer Kröger, and Arthur Zimek.
\newblock Robust clustering in arbitrarily oriented subspaces.
\newblock In {\em SDM}, pages 763--774, 2008.

\bibitem[ABK{\etalchar{+}}06a]{DBLP:conf/pkdd/AchtertBKKMZ06}
Elke Achtert, Christian Böhm, Hans-Peter Kriegel, Peer Kröger, Ina
  Müller-Gorman, and Arthur Zimek.
\newblock Finding hierarchies of subspace clusters.
\newblock In {\em PKDD}, pages 446--453, 2006.

\bibitem[ABK{\etalchar{+}}06b]{DBLP:conf/kdd/AchtertBKKZ06}
Elke Achtert, Christian Böhm, Hans-Peter Kriegel, Peer Kröger, and Arthur
  Zimek.
\newblock Deriving quantitative models for correlation clusters.
\newblock In {\em KDD}, pages 4--13, 2006.

\bibitem[ABK06c]{DBLP:conf/pakdd/AchtertBK06}
Elke Achtert, Christian Böhm, and Peer Kröger.
\newblock Deli-clu: Boosting robustness, completeness, usability, and
  efficiency of hierarchical clustering by a closest pair ranking.
\newblock In {\em PAKDD}, pages 119--128, 2006.

\bibitem[ABK{\etalchar{+}}07a]{DBLP:conf/dasfaa/AchtertBKKMZ07}
Elke Achtert, Christian Böhm, Hans-Peter Kriegel, Peer Kröger, Ina
  Müller-Gorman, and Arthur Zimek.
\newblock Detection and visualization of subspace cluster hierarchies.
\newblock In {\em DASFAA}, pages 152--163, 2007.

\bibitem[ABK{\etalchar{+}}07b]{DBLP:conf/ssdbm/AchtertBKKZ07}
Elke Achtert, Christian Böhm, Hans-Peter Kriegel, Peer Kröger, and Arthur
  Zimek.
\newblock On exploring complex relationships of correlation clusters.
\newblock In {\em SSDBM}, page~7, 2007.

\bibitem[ABK{\etalchar{+}}07c]{DBLP:conf/sdm/AchtertBKKZ07}
Elke Achtert, Christian Böhm, Hans-Peter Kriegel, Peer Kröger, and Arthur
  Zimek.
\newblock Robust, complete, and efficient correlation clustering.
\newblock In {\em SDM}, pages 413--418, 2007.

\bibitem[ABKS99]{DBLP:conf/sigmod/AnkerstBKS99}
Mihael Ankerst, Markus~M. Breunig, Hans-Peter Kriegel, and Jörg Sander.
\newblock Optics: Ordering points to identify the clustering structure.
\newblock In {\em SIGMOD Conference}, pages 49--60, 1999.

\bibitem[ABKZ06]{DBLP:conf/ssdbm/AchtertBKZ06}
Elke Achtert, Christian Böhm, Peer Kröger, and Arthur Zimek.
\newblock Mining hierarchies of correlation clusters.
\newblock In {\em SSDBM}, pages 119--128, 2006.

\bibitem[ACF{\etalchar{+}}15]{DBLP:conf/kdd/AmsalegCFGHKN15}
Laurent Amsaleg, Oussama Chelly, Teddy Furon, Stéphane Girard, Michael~E.
  Houle, Ken{-}ichi Kawarabayashi, and Michael Nett.
\newblock Estimating local intrinsic dimensionality.
\newblock In {\em KDD}, pages 29--38, 2015.

\bibitem[Ach01]{DBLP:conf/pods/Achlioptas01}
Dimitris Achlioptas.
\newblock Database-friendly random projections.
\newblock In {\em PODS}, 2001.

\bibitem[AD52]{doi:10.1214/aoms/1177729437}
T.~W. Anderson and D.~A. Darling.
\newblock Asymptotic theory of certain "goodness of fit" criteria based on
  stochastic processes.
\newblock {\em The Annals of Mathematical Statistics}, 23(2):193--212, 1952.

\bibitem[AD74]{DBLP:journals/computing/AhrensD74}
Joachim~H. Ahrens and Ulrich Dieter.
\newblock Computer methods for sampling from gamma, beta, poisson and bionomial
  distributions.
\newblock {\em Computing}, 12(3):223--246, 1974.

\bibitem[AD82]{DBLP:journals/cacm/AhrensD82}
Joachim~H. Ahrens and Ulrich Dieter.
\newblock Generating gamma variates by a modified rejection technique.
\newblock {\em Commun. ACM}, 25(1):47--54, 1982.

\bibitem[AGAV09]{DBLP:journals/ir/AmigoGAV09a}
Enrique Amigó, Julio Gonzalo, Javier Artiles, and Felisa Verdejo.
\newblock A comparison of extrinsic clustering evaluation metrics based on
  formal constraints.
\newblock {\em Inf. Retr.}, 12(5):613, 2009.

\bibitem[AGGR98]{DBLP:conf/sigmod/AgrawalGGR98}
Rakesh Agrawal, Johannes Gehrke, Dimitrios Gunopulos, and Prabhakar Raghavan.
\newblock Automatic subspace clustering of high dimensional data for data
  mining applications.
\newblock In {\em SIGMOD Conference}, pages 94--105, 1998.

\bibitem[AGK{\etalchar{+}}12]{DBLP:conf/icde/AchtertGKSZ12}
Elke Achtert, Sascha Goldhofer, Hans{-}Peter Kriegel, Erich Schubert, and
  Arthur Zimek.
\newblock Evaluation of clusterings - metrics and visual support.
\newblock In {\em {IEEE} 28th International Conference on Data Engineering
  {(ICDE} 2012), Washington, DC, {USA} (Arlington, Virginia), 1-5 April, 2012},
  pages 1285--1288, 2012.

\bibitem[AIS93]{DBLP:conf/sigmod/AgrawalIS93}
Rakesh Agrawal, Tomasz Imielinski, and Arun~N. Swami.
\newblock Mining association rules between sets of items in large databases.
\newblock In {\em SIGMOD Conference}, pages 207--216, 1993.

\bibitem[Aka73]{conf/isit/Akaike73}
Hirotogu Akaike.
\newblock Information theory and an extension of the maximum likelihood
  principle.
\newblock In {\em Proceedings of the 2nd International Symposium on Information
  Theory}, pages 267--281, 1973.

\bibitem[AKR{\etalchar{+}}10]{DBLP:conf/dasfaa/AchtertKRSWZ10}
Elke Achtert, Hans{-}Peter Kriegel, Lisa Reichert, Erich Schubert, Remigius
  Wojdanowski, and Arthur Zimek.
\newblock Visual evaluation of outlier detection models.
\newblock In {\em Database Systems for Advanced Applications, 15th
  International Conference, {DASFAA} 2010, Tsukuba, Japan, April 1-4, 2010,
  Proceedings, Part {II}}, pages 396--399, 2010.

\bibitem[AKSZ13]{DBLP:conf/sigmod/AchtertKSZ13}
Elke Achtert, Hans{-}Peter Kriegel, Erich Schubert, and Arthur Zimek.
\newblock Interactive data mining with 3d-parallel-coordinate-trees.
\newblock In {\em Proceedings of the {ACM} {SIGMOD} International Conference on
  Management of Data, {SIGMOD} 2013, New York, NY, USA, June 22-27, 2013},
  pages 1009--1012, 2013.

\bibitem[And73a]{books/academic/Anderberg73/Ch6}
Michael~R. Anderberg.
\newblock {\em Cluster Analysis for Applications}, chapter Hierarchical
  Clustering Methods.
\newblock 1973.

\bibitem[And73b]{books/academic/Anderberg73/Ch7}
Michael~R. Anderberg.
\newblock {\em Cluster Analysis for Applications}, chapter Nonhierarchical
  Clustering Methods.
\newblock 1973.

\bibitem[AP02]{DBLP:conf/pkdd/AngiulliP02}
Fabrizio Angiulli and Clara Pizzuti.
\newblock Fast outlier detection in high dimensional spaces.
\newblock In {\em PKDD}, pages 15--26, 2002.

\bibitem[AS94]{DBLP:conf/vldb/AgrawalS94}
Rakesh Agrawal and Ramakrishnan Srikant.
\newblock Fast algorithms for mining association rules in large databases.
\newblock In {\em VLDB}, pages 487--499, 1994.

\bibitem[AT97]{DBLP:conf/ssd/AngT97}
Chuan-Heng Ang and T.~C. Tan.
\newblock New linear node splitting algorithm for r-trees.
\newblock In {\em SSD}, pages 339--349, 1997.

\bibitem[AV07]{DBLP:conf/soda/ArthurV07}
David Arthur and Sergei Vassilvitskii.
\newblock k-means++: the advantages of careful seeding.
\newblock In {\em SODA}, pages 1027--1035, 2007.

\bibitem[AWY{\etalchar{+}}99]{doi:10.1145/304181.304188}
Charu~C. Aggarwal, Joel~L. Wolf, Philip~S. Yu, Cecilia Procopiuc, and Jong~Soo
  Park.
\newblock Fast algorithms for projected clustering.
\newblock {\em ACM SIGMOD Record}, 28(2):61--72, 1999.

\bibitem[AY00]{DBLP:conf/sigmod/AggarwalY00}
Charu~C. Aggarwal and Philip~S. Yu.
\newblock Finding generalized projected clusters in high dimensional spaces.
\newblock In {\em SIGMOD Conference}, pages 70--81, 2000.

\bibitem[AY01]{DBLP:conf/sigmod/AggarwalY01}
Charu~C. Aggarwal and Philip~S. Yu.
\newblock Outlier detection for high dimensional data.
\newblock In {\em SIGMOD Conference}, pages 37--46, 2001.

\bibitem[AYN{\etalchar{+}}05]{DBLP:journals/bioinformatics/AoYNCFMS05}
Sio~Iong Ao, Kevin~Y. Yip, Michael~K. Ng, David Wai-Lok Cheung, Pui-Yee Fong,
  Ian Melhado, and Pak~Chung Sham.
\newblock Clustag: hierarchical clustering and graph methods for selecting tag
  snps.
\newblock {\em Bioinformatics}, 21(8):1735--1736, 2005.

\bibitem[BB98]{doi:10.3115/980451.980859}
Amit Bagga and Breck Baldwin.
\newblock Entity-based cross-document coreferencing using the vector space
  model.
\newblock In {\em Proceedings of the 17th international conference on
  Computational linguistics -}, 1998.

\bibitem[BC57]{doi:10.2307/1942268}
J.~Roger Bray and J.~T. Curtis.
\newblock An ordination of the upland forest communities of southern wisconsin.
\newblock {\em Ecological Monographs}, 27(4):325--349, 1957.

\bibitem[BC94]{DBLP:conf/kdd/BerndtC94}
Donald~J. Berndt and James Clifford.
\newblock Using dynamic time warping to find patterns in time series.
\newblock In {\em KDD Workshop}, pages 359--370, 1994.

\bibitem[BDG05]{doi:10.3150/bj/1137421635}
J.~Beirlant, G.~Dierckx, and A.~Guillou.
\newblock Estimation of the extreme-value index and generalized quantile plots.
\newblock {\em Bernoulli}, 11(6):949--970, 2005.

\bibitem[Ben75]{DBLP:journals/cacm/Bentley75}
Jon~Louis Bentley.
\newblock Multidimensional binary search trees used for associative searching.
\newblock {\em Commun. ACM}, 18(9):509--517, 1975.

\bibitem[Ber76]{doi:10.2307/2347257}
J.~M. Bernardo.
\newblock Algorithm as 103: Psi (digamma) function.
\newblock {\em Applied Statistics}, 25(3):315, 1976.

\bibitem[BF98]{DBLP:conf/icml/BradleyF98}
Paul~S. Bradley and Usama~M. Fayyad.
\newblock Refining initial points for k-means clustering.
\newblock In {\em ICML}, pages 91--99, 1998.

\bibitem[BFOS84]{books/wa/BreimanFOS84}
Leo Breiman, Jerome~H. Friedman, R.~A. Olshen, and Charles~J. Stone.
\newblock {\em Classification and Regression Trees}.
\newblock Wadsworth, 1984.

\bibitem[BH75]{doi:10.1080/01621459.1975.10480256}
Frank~B. Baker and Lawrence~J. Hubert.
\newblock Measuring the power of hierarchical cluster analysis.
\newblock {\em Journal of the American Statistical Association},
  70(349):31--38, 1975.

\bibitem[BKK96]{DBLP:conf/vldb/BerchtoldKK96}
Stefan Berchtold, Daniel~A. Keim, and Hans-Peter Kriegel.
\newblock The x-tree : An index structure for high-dimensional data.
\newblock In {\em VLDB}, pages 28--39, 1996.

\bibitem[BKKK04]{DBLP:conf/icdm/BohmKKK04}
Christian Böhm, Karin Kailing, Hans-Peter Kriegel, and Peer Kröger.
\newblock Density connected clustering with local subspace preferences.
\newblock In {\em ICDM}, pages 27--34, 2004.

\bibitem[BKKZ04]{DBLP:conf/sigmod/BohmKKZ04}
Christian Böhm, Karin Kailing, Peer Kröger, and Arthur Zimek.
\newblock Computing clusters of correlation connected objects.
\newblock In {\em SIGMOD Conference}, pages 455--466, 2004.

\bibitem[BKL06]{DBLP:conf/icml/BeygelzimerKL06}
Alina Beygelzimer, Sham Kakade, and John Langford.
\newblock Cover trees for nearest neighbor.
\newblock In {\em ICML}, pages 97--104, 2006.

\bibitem[BKNS99]{DBLP:conf/pkdd/BreunigKNS99}
Markus~M. Breunig, Hans-Peter Kriegel, Raymond~T. Ng, and Jörg Sander.
\newblock Optics-of: Identifying local outliers.
\newblock In {\em PKDD}, pages 262--270, 1999.

\bibitem[BKNS00]{DBLP:conf/sigmod/BreunigKNS00}
Markus~M. Breunig, Hans-Peter Kriegel, Raymond~T. Ng, and Jörg Sander.
\newblock Lof: Identifying density-based local outliers.
\newblock In {\em SIGMOD Conference}, pages 93--104, 2000.

\bibitem[BKR{\etalchar{+}}09]{DBLP:conf/kdd/BerneckerKRVZ09}
Thomas Bernecker, Hans-Peter Kriegel, Matthias Renz, Florian Verhein, and
  Andreas Züfle.
\newblock Probabilistic frequent itemset mining in uncertain databases.
\newblock In {\em KDD}, pages 119--128, 2009.

\bibitem[BKSS90]{DBLP:conf/sigmod/BeckmannKSS90}
Norbert Beckmann, Hans-Peter Kriegel, Ralf Schneider, and Bernhard Seeger.
\newblock The r*-tree: An efficient and robust access method for points and
  rectangles.
\newblock In {\em SIGMOD Conference}, pages 322--331, 1990.

\bibitem[BMS96]{DBLP:conf/nips/BradleyMS96}
Paul~S. Bradley, Olvi~L. Mangasarian, and W.~Nick Street.
\newblock Clustering via concave minimization.
\newblock In {\em NIPS}, pages 368--374, 1996.

\bibitem[BMS97]{DBLP:conf/sigmod/BrinMS97}
Sergey Brin, Rajeev Motwani, and Craig Silverstein.
\newblock Beyond market baskets: Generalizing association rules to
  correlations.
\newblock In {\em SIGMOD Conference}, pages 265--276, 1997.

\bibitem[BMUT97]{DBLP:conf/sigmod/BrinMUT97}
Sergey Brin, Rajeev Motwani, Jeffrey~D. Ullman, and Shalom Tsur.
\newblock Dynamic itemset counting and implication rules for market basket
  data.
\newblock In {\em SIGMOD Conference}, pages 255--264, 1997.

\bibitem[BN93]{books/prentice/BassevilleN93/C2}
M.~Basseville and I.~V. Nikiforov.
\newblock {\em Detection of Abrupt Changes - Theory and Application}, chapter
  Section 2.6: Off-line Change Detection.
\newblock 1993.

\bibitem[BPK{\etalchar{+}}04]{DBLP:conf/icdm/BaumgartnerPKKK04}
Christian Baumgartner, Claudia Plant, Karin Kailing, Hans-Peter Kriegel, and
  Peer Kröger.
\newblock Subspace selection for clustering high-dimensional data.
\newblock In {\em ICDM}, pages 11--18, 2004.

\bibitem[BR75]{doi:10.2307/2347113}
D.~J. Best and D.~E. Roberts.
\newblock Algorithm as 91: The percentage points of the χ 2 distribution.
\newblock {\em Applied Statistics}, 24(3):385, 1975.

\bibitem[Bre96]{DBLP:journals/ml/Breiman96b}
Leo Breiman.
\newblock Bagging predictors.
\newblock {\em Machine Learning}, 24(2):123--140, 1996.

\bibitem[BSHW06]{DBLP:conf/vldb/BenjellounSHW06}
Omar Benjelloun, Anish~Das Sarma, Alon~Y. Halevy, and Jennifer Widom.
\newblock Uldbs: Databases with uncertainty and lineage.
\newblock In {\em VLDB}, pages 953--964, 2006.

\bibitem[BT11]{doi:10.1198/jasa.2011.tm10183}
Jacob Bien and Robert Tibshirani.
\newblock Hierarchical clustering with prototypes via minimax linkage.
\newblock {\em Journal of the American Statistical Association},
  106(495):1075--1084, 2011.

\bibitem[BV16]{web/BlackmanV16}
D.~Blackman and S.~Vigna.
\newblock xoroshiro+ / xorshift* / xorshift+ generators and the prng shootout,
  2016.

\bibitem[BY07]{DBLP:conf/icannga/BiciciY07}
Ergun Biçici and Deniz Yuret.
\newblock Locally scaled density based clustering.
\newblock In {\em ICANNGA (1)}, pages 739--748, 2007.

\bibitem[Bí12]{journals/naun/Bilkova12}
D.~Bílková.
\newblock Lognormal distribution and using l-moment method for estimating its
  parameters.
\newblock {\em Int. Journal of Mathematical Models and Methods in Applied
  Sciences (NAUN)}, 1(6), 2012.

\bibitem[CB90]{books/duxbury/CasellaB90/Ch7}
George Casella and Roger~L. Berger.
\newblock {\em Statistical Inference}, chapter Point Estimation.
\newblock Duxbury, 1990.

\bibitem[CC00]{DBLP:conf/ismb/ChengC00}
Yizong Cheng and George~M. Church.
\newblock Biclustering of expression data.
\newblock In {\em ISMB}, pages 93--103, 2000.

\bibitem[CCKN06]{DBLP:conf/pakdd/ChauCKN06}
Michael Chau, Reynold Cheng, Ben Kao, and Jackey Ng.
\newblock Uncertain data mining: An example in clustering location data.
\newblock In {\em PAKDD}, pages 199--204, 2006.

\bibitem[CCVR16]{DBLP:journals/corr/ConnorCVR16}
Richard C.~H. Connor, Franco~Alberto Cardillo, Lucia Vadicamo, and Fausto
  Rabitti.
\newblock Hilbert exclusion: Improved metric search through finite isometric
  embeddings.
\newblock {\em CoRR}, abs/1604.08640, 2016.

\bibitem[CG13]{DBLP:conf/sdm/ChawlaG13}
Sanjay Chawla and Aristides Gionis.
\newblock k-means-: A unified approach to clustering and outlier detection.
\newblock In {\em SDM}, pages 189--197, 2013.

\bibitem[CH74]{doi:10.1080/03610927408827101}
T.~Calinski and J.~Harabasz.
\newblock A dendrite method for cluster analysis.
\newblock {\em Communications in Statistics - Theory and Methods}, 3(1):1--27,
  1974.

\bibitem[Cha02]{DBLP:conf/stoc/Charikar02}
Moses Charikar.
\newblock Similarity estimation techniques from rounding algorithms.
\newblock In {\em STOC}, pages 380--388, 2002.

\bibitem[Che95]{DBLP:journals/pami/Cheng95}
Yizong Cheng.
\newblock Mean shift, mode seeking, and clustering.
\newblock {\em IEEE Trans. Pattern Anal. Mach. Intell.}, 17(8):790--799, 1995.

\bibitem[CHK16]{tr/nii/ChellyHK16}
Oussama Chelly, Michael~E. Houle, and Ken{-}ichi Kawarabayashi.
\newblock Enhanced estimation of local intrinsic dimensionality using auxiliary
  distances.
\newblock Technical report, National Institute of Informatics, 2016.

\bibitem[CLB10]{DBLP:conf/kdd/ChenLB10}
Feng Chen, Chang-Tien Lu, and Arnold~P. Boedihardjo.
\newblock Gls-sod: a generalized local statistical approach for spatial outlier
  detection.
\newblock In {\em KDD}, pages 1069--1078, 2010.

\bibitem[CMS13]{DBLP:conf/pakdd/CampelloMS13}
Ricardo J. G.~B. Campello, Davoud Moulavi, and Jörg Sander.
\newblock Density-based clustering based on hierarchical density estimates.
\newblock In {\em PAKDD (2)}, pages 160--172, 2013.

\bibitem[CN04]{DBLP:conf/vldb/ChenN04}
Lei Chen and Raymond~T. Ng.
\newblock On the marriage of lp-norms and edit distance.
\newblock In {\em VLDB}, pages 792--803, 2004.

\bibitem[C{\"{O}}O05]{DBLP:conf/sigmod/ChenOO05}
Lei Chen, M.~Tamer {\"{O}}zsu, and Vincent Oria.
\newblock Robust and fast similarity search for moving object trajectories.
\newblock In {\em SIGMOD Conference}, pages 491--502, 2005.

\bibitem[Cor71]{doi:10.2307/2344237}
R.~M. Cormack.
\newblock A review of classification.
\newblock {\em Journal of the Royal Statistical Society. Series A (General)},
  134(3):321, 1971.

\bibitem[CPZ97]{DBLP:conf/vldb/CiacciaPZ97}
Paolo Ciaccia, Marco Patella, and Pavel Zezula.
\newblock M-tree: An efficient access method for similarity search in metric
  spaces.
\newblock In {\em VLDB}, pages 426--435, 1997.

\bibitem[CS06]{DBLP:journals/kais/ChawlaS06}
Sanjay Chawla and Pei Sun.
\newblock Slom: a new measure for local spatial outliers.
\newblock {\em Knowl. Inf. Syst.}, 9(4):412--429, 2006.

\bibitem[CW69]{doi:10.2307/1266892}
S.~C. Choi and R.~Wette.
\newblock Maximum likelihood estimation of the parameters of the gamma
  distribution and their bias.
\newblock {\em Technometrics}, 11(4):683, 1969.

\bibitem[Dar57]{doi:10.1214/aoms/1177706788}
D.~A. Darling.
\newblock The kolmogorov-smirnov, cramer-von mises tests.
\newblock {\em The Annals of Mathematical Statistics}, 28(4):823--838, 1957.

\bibitem[DB79]{DBLP:journals/pami/DaviesB79}
David~L. Davies and Donald~W. Bouldin.
\newblock A cluster separation measure.
\newblock {\em IEEE Trans. Pattern Anal. Mach. Intell.}, 1(2):224--227, 1979.

\bibitem[DCL11]{DBLP:conf/www/DongCL11}
Wei Dong, Moses Charikar, and Kai Li.
\newblock Efficient k-nearest neighbor graph construction for generic
  similarity measures.
\newblock In {\em WWW}, pages 577--586, 2011.

\bibitem[DD09]{doi:10.1007/978-3-642-00234-2}
Elena Deza and Michel~Marie Deza.
\newblock {\em Encyclopedia of Distances}.
\newblock Springer Berlin Heidelberg, 2009.

\bibitem[Def77]{DBLP:journals/cj/Defays77}
D.~Defays.
\newblock An efficient algorithm for a complete link method.
\newblock {\em Comput. J.}, 20(4):364--366, 1977.

\bibitem[Dic45]{doi:10.2307/1932409}
Lee~R. Dice.
\newblock Measures of the amount of ecologic association between species.
\newblock {\em Ecology}, 26(3):297--302, 1945.

\bibitem[DIIM04]{DBLP:conf/compgeom/DatarIIM04}
Mayur Datar, Nicole Immorlica, Piotr Indyk, and Vahab~S. Mirrokni.
\newblock Locality-sensitive hashing scheme based on p-stable distributions.
\newblock In {\em Symposium on Computational Geometry}, pages 253--262, 2004.

\bibitem[DLPT85]{books/misc/DidayLPT85}
E.~Diday, J.~Lemaire, J.~Pouget, and F.~Testu.
\newblock {\em Elements d'analyse de donnees}.
\newblock Dunod, 1985.

\bibitem[DLR77]{journals/jroyastatsocise2/DempsterLR77}
A.~P. Dempster, N.~M. Laird, and D.~B. Rubin.
\newblock Maximum likelihood from incomplete data via the {EM} algorithm.
\newblock {\em Journal of the Royal Statistical Society, Series B}, 39(1),
  1977.

\bibitem[Dra13]{mathesis/Drake13}
Jonathan Drake.
\newblock Faster k-means clustering.
\newblock Master's thesis, Baylor University, 2013.

\bibitem[DRS09]{DBLP:journals/cacm/DalviRS09}
Nilesh~N. Dalvi, Christopher Ré, and Dan Suciu.
\newblock Probabilistic databases: diamonds in the dirt.
\newblock {\em Commun. ACM}, 52(7):86--94, 2009.

\bibitem[dRTD98]{conf/asci/deRidderTD98}
D.~de~Ridder, D.~M.~J. Tax, and R.~P.~W. Duin.
\newblock An experimental comparison of one-class classification methods.
\newblock In {\em Proc. 4th Ann. Conf. Advanced School for Computing and
  Imaging (ASCI'98)}, 1998.

\bibitem[dVCH10]{DBLP:conf/icdm/VriesCH10}
Timothy de~Vries, Sanjay Chawla, and Michael~E. Houle.
\newblock Finding local anomalies in very high dimensional space.
\newblock In {\em ICDM}, pages 128--137, 2010.

\bibitem[EKSX96]{DBLP:conf/kdd/EsterKSX96}
Martin Ester, Hans-Peter Kriegel, Jörg Sander, and Xiaowei Xu.
\newblock A density-based algorithm for discovering clusters in large spatial
  databases with noise.
\newblock In {\em KDD}, pages 226--231, 1996.

\bibitem[Elk03]{DBLP:conf/icml/Elkan03}
Charles Elkan.
\newblock Using the triangle inequality to accelerate k-means.
\newblock In {\em ICML}, pages 147--153, 2003.

\bibitem[ES03]{DBLP:journals/tit/EndresS03}
Dominik~Maria Endres and Johannes~E. Schindelin.
\newblock A new metric for probability distributions.
\newblock {\em IEEE Trans. Information Theory}, 49(7):1858--1860, 2003.

\bibitem[Esk00]{DBLP:conf/icml/Eskin00}
Eleazar Eskin.
\newblock Anomaly detection over noisy data using learned probability
  distributions.
\newblock In {\em ICML}, pages 255--262, 2000.

\bibitem[ESK03]{DBLP:conf/sdm/ErtozSK03}
Levent Ertöz, Michael Steinbach, and Vipin Kumar.
\newblock Finding clusters of different sizes, shapes, and densities in noisy,
  high dimensional data.
\newblock In {\em SDM}, pages 47--58, 2003.

\bibitem[FB81]{DBLP:journals/cacm/FischlerB81}
Martin~A. Fischler and Robert~C. Bolles.
\newblock Random sample consensus: A paradigm for model fitting with
  applications to image analysis and automated cartography.
\newblock {\em Commun. ACM}, 24(6):381--395, 1981.

\bibitem[FD07]{doi:10.1126/science.1136800}
B.~J. Frey and D.~Dueck.
\newblock Clustering by passing messages between data points.
\newblock {\em Science}, 315(5814):972--976, 2007.

\bibitem[Fis36]{doi:10.1111/j.1469-1809.1936.tb02137.x}
R.~A. Fisher.
\newblock The use of multiple measurements in taxonomic problems.
\newblock {\em Annals of Eugenics}, 7(2):179--188, 1936.

\bibitem[F{\L}P{\etalchar{+}}51]{journals/misc/FlorekLPSZ51}
K.~Florek, J.~{\L}ukaszewicz, J.~Perkal, H.~Steinhaus, and S.~Zubrzycki.
\newblock Sur la liaison et la division des points d'un ensemble fini.
\newblock {\em Colloquium Mathematicae}, 2(3-4):282--285, 1951.

\bibitem[FM83]{doi:10.2307/2288117}
E.~B. Fowlkes and C.~L. Mallows.
\newblock A method for comparing two hierarchical clusterings.
\newblock {\em Journal of the American Statistical Association}, 78(383):553,
  1983.

\bibitem[For65]{journals/biometrics/Forgy65}
E.~W. Forgy.
\newblock Cluster analysis of multivariate data: efficiency versus
  interpretability of classifications.
\newblock {\em Biometrics}, 21(3), 1965.

\bibitem[FR07]{DBLP:journals/classification/FraleyR07}
Chris Fraley and Adrian~E. Raftery.
\newblock Bayesian regularization for normal mixture estimation and model-based
  clustering.
\newblock {\em J. Classification}, 24(2):155--181, 2007.

\bibitem[GBSM02]{DBLP:journals/ida/GalianoBSM02}
Fernando~Berzal Galiano, Ignacio~J. Blanco, Daniel Sánchez, and María
  Amparo~Vila Miranda.
\newblock Measuring the accuracy and interest of association rules: A new
  framework.
\newblock {\em Intell. Data Anal.}, 6(3):221--235, 2002.

\bibitem[Gow67]{doi:10.2307/2528417}
J.~C. Gower.
\newblock A comparison of some methods of cluster analysis.
\newblock {\em Biometrics}, 23(4):623, 1967.

\bibitem[Gra72]{DBLP:journals/ipl/Graham72}
Ronald~L. Graham.
\newblock An efficient algorithm for determining the convex hull of a finite
  planar set.
\newblock {\em Inf. Process. Lett.}, 1(4):132--133, 1972.

\bibitem[Gre89]{DBLP:conf/icde/Greene89}
Diane Greene.
\newblock An implementation and performance analysis of spatial data access
  methods.
\newblock In {\em ICDE}, pages 606--615, 1989.

\bibitem[GT06]{DBLP:conf/icdm/GaoT06}
Jing Gao and Pang-Ning Tan.
\newblock Converting output scores from outlier detection algorithms into
  probability estimates.
\newblock In {\em ICDM}, pages 212--221, 2006.

\bibitem[Guo03]{DBLP:journals/ivs/Guo03}
Diansheng Guo.
\newblock Coordinating computational and visual approaches for interactive
  feature selection and multivariate clustering.
\newblock {\em Information Visualization}, 2(4):232--246, 2003.

\bibitem[Gut84]{doi:10.1145/971697.602266}
Antonin Guttman.
\newblock R-trees.
\newblock {\em ACM SIGMOD Record}, 14(2):47, 1984.

\bibitem[HA85]{doi:10.1007/BF01908075}
Lawrence Hubert and Phipps Arabie.
\newblock Comparing partitions.
\newblock {\em Journal of Classification}, 2(1):193--218, 1985.

\bibitem[Ham50]{doi:10.1002/j.1538-7305.1950.tb00463.x}
R.~W. Hamming.
\newblock Error detecting and error correcting codes.
\newblock {\em Bell System Technical Journal}, 29(2):147--160, 1950.

\bibitem[Ham74]{doi:10.2307/2285666}
Frank~R. Hampel.
\newblock The influence curve and its role in robust estimation.
\newblock {\em Journal of the American Statistical Association}, 69(346):383,
  1974.

\bibitem[Ham10]{DBLP:conf/sdm/Hamerly10}
Greg Hamerly.
\newblock Making k-means even faster.
\newblock In {\em SDM}, pages 130--140, 2010.

\bibitem[Har75]{books/wiley/Hartigan75/C3}
John~A. Hartigan.
\newblock {\em Clustering Algorithms}, chapter Quick Partition Algorithms:
  Leader Algorithm, page~75.
\newblock Wiley, 1975.

\bibitem[HD14]{doi:10.1007/978-3-319-09259-1_2}
Greg Hamerly and Jonathan Drake.
\newblock Accelerating {Lloyd}’s algorithm for k-means clustering.
\newblock In {\em Partitional Clustering Algorithms}, pages 41--78. 2014.

\bibitem[Hel09]{journals/mathematik/Hellinger1909}
E.~Hellinger.
\newblock Neue begründung der theorie quadratischer formen von unendlichvielen
  veränderlichen.
\newblock {\em Journal für die reine und angewandte Mathematik}, 1909.

\bibitem[Hen06]{DBLP:conf/sigir/Henzinger06}
Monika~Rauch Henzinger.
\newblock Finding near-duplicate web pages: a large-scale evaluation of
  algorithms.
\newblock In {\em SIGIR}, pages 284--291, 2006.

\bibitem[HH07]{DBLP:journals/pr/HaralickH07}
Robert~M. Haralick and Rave Harpaz.
\newblock Linear manifold clustering in high dimensional spaces by stochastic
  search.
\newblock {\em Pattern Recognition}, 40(10):2672--2684, 2007.

\bibitem[Hil91]{journals/mathann/Hilbert1891}
D.~Hilbert.
\newblock Ueber die stetige abbildung einer linie auf ein flächenstück.
\newblock {\em Mathematische Annalen}, 38(3):210--271, 1891.

\bibitem[Hil75]{doi:10.1214/aos/1176343247}
Bruce~M. Hill.
\newblock A simple general approach to inference about the tail of a
  distribution.
\newblock {\em The Annals of Statistics}, 3(5):1163--1174, 1975.

\bibitem[HKF04]{DBLP:conf/icpr/HautamakiKF04}
Ville Hautamäki, Ismo Kärkkäinen, and Pasi Fränti.
\newblock Outlier detection using k-nearest neighbour graph.
\newblock In {\em ICPR (3)}, pages 430--433, 2004.

\bibitem[HKKP01]{doi:10.1198/073500101316970421}
Ronald Huisman, Kees~G Koedijk, Clemens J.~M Kool, and Franz Palm.
\newblock Tail-index estimates in small samples.
\newblock {\em Journal of Business \& Economic Statistics}, 19(2):208--216,
  2001.

\bibitem[HKN12]{DBLP:conf/icdm/HouleKN12}
Michael~E. Houle, Hisashi Kashima, and Michael Nett.
\newblock Generalized expansion dimension.
\newblock In {\em ICDM Workshops}, pages 587--594, 2012.

\bibitem[HL76]{doi:10.1037/0033-2909.83.6.1072}
Lawrence~J. Hubert and Joel~R. Levin.
\newblock A general statistical framework for assessing categorical clustering
  in free recall.
\newblock {\em Psychological Bulletin}, 83(6):1072--1080, 1976.

\bibitem[HMTW47]{doi:10.1214/aoms/1177730388}
Cecil Hastings, Frederick Mosteller, John~W. Tukey, and Charles~P. Winsor.
\newblock Low moments for small samples: A comparative study of order
  statistics.
\newblock {\em The Annals of Mathematical Statistics}, 18(3):413--426, 1947.

\bibitem[Hoe48]{journals/mathstat/Hoeffding48}
W.~Hoeffding.
\newblock A non-parametric test of independence.
\newblock {\em The Annals of Mathematical Statistics}, 19, 1948.

\bibitem[Hos00]{tr/ibm/Hosking00}
J.~R.~M. Hosking.
\newblock Fortran routines for use with the method of l-moments version 3.03.
\newblock Technical report, IBM, 2000.

\bibitem[HPY00]{DBLP:conf/sigmod/HanPY00}
Jiawei Han, Jian Pei, and Yiwen Yin.
\newblock Mining frequent patterns without candidate generation.
\newblock In {\em SIGMOD Conference}, pages 1--12, 2000.

\bibitem[HQ04]{DBLP:conf/icig/HuangQ04}
Tianqiang Huang and Xiaolin Qin.
\newblock Detecting outliers in spatial database.
\newblock In {\em ICIG}, pages 556--559, 2004.

\bibitem[HR02]{DBLP:conf/nips/HintonR02}
Geoffrey~E. Hinton and Sam~T. Roweis.
\newblock Stochastic neighbor embedding.
\newblock In {\em NIPS}, pages 833--840, 2002.

\bibitem[HS54]{doi:10.1093/oxfordjournals.aob.a083391}
Brian Hopkins and J.~G. Skellam.
\newblock A new method for determining the type of distribution of plant
  individuals.
\newblock {\em Annals of Botany}, 18(2):213--227, 1954.

\bibitem[HS95]{DBLP:conf/ssd/HjaltasonS95}
Gísli~R. Hjaltason and Hanan Samet.
\newblock Ranking in spatial databases.
\newblock In {\em SSD}, pages 83--95, 1995.

\bibitem[HS03]{DBLP:journals/prl/HuS03}
Tianming Hu and Sam~Yuan Sung.
\newblock Detecting pattern-based outliers.
\newblock {\em Pattern Recognition Letters}, 24(16):3059--3068, 2003.

\bibitem[HS04]{DBLP:journals/ida/HuS04}
Tianming Hu and Sam~Yuan Sung.
\newblock A trimmed mean approach to finding spatial outliers.
\newblock {\em Intell. Data Anal.}, 8(1):79--95, 2004.

\bibitem[HSE{\etalchar{+}}95]{DBLP:journals/pami/HafnerSEFN95}
James~L. Hafner, Harpreet~S. Sawhney, William Equitz, Myron Flickner, and Wayne
  Niblack.
\newblock Efficient color histogram indexing for quadratic form distance
  functions.
\newblock {\em IEEE Trans. Pattern Anal. Mach. Intell.}, 17(7):729--736, 1995.

\bibitem[HSZ18]{DBLP:conf/sisap/HouleSZ18}
Michael~E. Houle, Erich Schubert, and Arthur Zimek.
\newblock On the correlation between local intrinsic dimensionality and
  outlierness.
\newblock In {\em SISAP}, pages 177--191, 2018.

\bibitem[HW97]{doi:10.1017/CBO9780511529443}
J.~R.~M. Hosking and James~R. Wallis.
\newblock {\em Regional Frequency Analysis}.
\newblock Cambridge University Press, 1997.

\bibitem[HWW85]{doi:10.1080/00401706.1985.10488049}
J.~R.~M. Hosking, J.~R. Wallis, and E.~F. Wood.
\newblock Estimation of the generalized extreme-value distribution by the
  method of probability-weighted moments.
\newblock {\em Technometrics}, 27(3):251--261, 1985.

\bibitem[HXD03]{DBLP:journals/prl/HeXD03}
Zengyou He, Xiaofei Xu, and Shengchun Deng.
\newblock Discovering cluster-based local outliers.
\newblock {\em Pattern Recognition Letters}, 24(9-10):1641--1650, 2003.

\bibitem[Ins09]{doi:10.1007/978-0-387-68628-8}
Alfred Inselberg.
\newblock {\em Parallel Coordinates}.
\newblock Springer New York, 2009.

\bibitem[Jac02]{journals/misc/Jaccard1902}
Paul Jaccard.
\newblock Distribution de la florine alpine dans la bassin de dranses et dans
  quelques regiones voisines.
\newblock {\em Bulletin del la Société Vaudoise des Sciences Naturelles},
  1902.

\bibitem[Jan66]{doi:10.1071/BT9660127}
R.~C. Jancey.
\newblock Multidimensional group analysis.
\newblock {\em Australian Journal of Botany}, 14(1):127, 1966.

\bibitem[Jef46]{doi:10.1098/rspa.1946.0056}
H.~Jeffreys.
\newblock An invariant form for the prior probability in estimation problems.
\newblock {\em Proceedings of the Royal Society A: Mathematical, Physical and
  Engineering Sciences}, 186(1007):453--461, 1946.

\bibitem[JHPvdH12]{tr/tilburg/JanssensHPv12}
J.~Janssens, F.~Huszár, E.~Postma, and J.~van~den Herik.
\newblock Stochastic outlier selection.
\newblock Technical report, Tilburg University, 2012.

\bibitem[JK02]{DBLP:journals/tois/JarvelinK02}
Kalervo Järvelin and Jaana Kekäläinen.
\newblock Cumulated gain-based evaluation of ir techniques.
\newblock {\em ACM Trans. Inf. Syst.}, 20(4):422--446, 2002.

\bibitem[Joh67]{journals/psychometrika/Johnson67}
Stephen~C. Johnson.
\newblock Hierarchical clustering schemes.
\newblock {\em Psychometrika}, 32(3):241--254, 1967.

\bibitem[JOT{\etalchar{+}}05]{DBLP:journals/tods/JagadishOTYZ05}
H.~V. Jagadish, Beng~Chin Ooi, Kian-Lee Tan, Cui Yu, and Rui Zhang.
\newblock idistance: An adaptive b.
\newblock {\em ACM Trans. Database Syst.}, 30(2):364--397, 2005.

\bibitem[JTHW06]{DBLP:conf/pakdd/JinTHW06}
Wen Jin, Anthony K.~H. Tung, Jiawei Han, and Wei Wang.
\newblock Ranking outliers using symmetric neighborhood relationship.
\newblock In {\em PAKDD}, pages 577--593, 2006.

\bibitem[JTSF00]{DBLP:conf/edbt/TrainaTSF00}
Caetano~Traina Jr., Agma J.~M. Traina, Bernhard Seeger, and Christos Faloutsos.
\newblock Slim-trees: High performance metric trees minimizing overlap between
  nodes.
\newblock In {\em EDBT}, pages 51--65, 2000.

\bibitem[KF93]{DBLP:conf/cikm/KamelF93}
Ibrahim Kamel and Christos Faloutsos.
\newblock On packing r-trees.
\newblock In {\em CIKM}, pages 490--499, 1993.

\bibitem[KFJM11]{DBLP:conf/pkdd/KlementFJM11}
William Klement, Peter~A. Flach, Nathalie Japkowicz, and Stan Matwin.
\newblock Smooth receiver operating characteristics (smroc) curves.
\newblock In {\em ECML/PKDD (2)}, pages 193--208, 2011.

\bibitem[KKK04]{DBLP:conf/sdm/KroegerKK04}
Peer Kröger, Hans-Peter Kriegel, and Karin Kailing.
\newblock Density-connected subspace clustering for high-dimensional data.
\newblock In {\em SDM}, pages 246--256, 2004.

\bibitem[KKSZ06]{DBLP:conf/ssdbm/KriegelKSZ06}
Hans-Peter Kriegel, Peer Kröger, Matthias Schubert, and Ziyue Zhu.
\newblock Efficient query processing in arbitrary subspaces using vector
  approximations.
\newblock In {\em SSDBM}, pages 184--190, 2006.

\bibitem[KKSZ08]{DBLP:conf/ssdbm/KriegelKSZ08}
Hans-Peter Kriegel, Peer Kröger, Erich Schubert, and Arthur Zimek.
\newblock A general framework for increasing the robustness of pca-based
  correlation clustering algorithms.
\newblock In {\em SSDBM}, pages 418--435, 2008.

\bibitem[KKSZ09a]{DBLP:conf/cikm/KriegelKSZ09}
Hans-Peter Kriegel, Peer Kröger, Erich Schubert, and Arthur Zimek.
\newblock Loop: local outlier probabilities.
\newblock In {\em CIKM}, pages 1649--1652, 2009.

\bibitem[KKSZ09b]{DBLP:conf/pakdd/KriegelKSZ09}
Hans-Peter Kriegel, Peer Kröger, Erich Schubert, and Arthur Zimek.
\newblock Outlier detection in axis-parallel subspaces of high dimensional
  data.
\newblock In {\em PAKDD}, pages 831--838, 2009.

\bibitem[KKSZ11]{DBLP:conf/sdm/KriegelKSZ11}
Hans-Peter Kriegel, Peer Kröger, Erich Schubert, and Arthur Zimek.
\newblock Interpreting and unifying outlier scores.
\newblock In {\em SDM}, pages 13--24, 2011.

\bibitem[KKSZ12]{DBLP:conf/icdm/KriegelKSZ12}
Hans-Peter Kriegel, Peer Kröger, Erich Schubert, and Arthur Zimek.
\newblock Outlier detection in arbitrarily oriented subspaces.
\newblock In {\em ICDM}, pages 379--388, 2012.

\bibitem[KL97]{doi:10.1109/ICICS.1997.652114}
J.~Kuan and P.~Lewis.
\newblock Fast k nearest neighbour search for r-tree family.
\newblock In {\em Proceedings of ICICS, 1997 International Conference on
  Information, Communications and Signal Processing}, 1997.

\bibitem[Kl{\"{o}}96]{DBLP:books/mit/fayyadPSU96/Klosgen96}
Willi Kl{\"{o}}sgen.
\newblock Explora: A multipattern and multistrategy discovery assistant.
\newblock In {\em Advances in Knowledge Discovery and Data Mining}, pages
  249--271. 1996.

\bibitem[KMB12]{DBLP:conf/icde/KellerMB12}
Fabian Keller, Emmanuel Müller, and Klemens Böhm.
\newblock Hics: High contrast subspaces for density-based outlier ranking.
\newblock In {\em ICDE}, pages 1037--1048, 2012.

\bibitem[KN98]{DBLP:conf/vldb/KnorrN98}
Edwin~M. Knorr and Raymond~T. Ng.
\newblock Algorithms for mining distance-based outliers in large datasets.
\newblock In {\em VLDB}, pages 392--403, 1998.

\bibitem[KP01]{DBLP:conf/sdm/KeoghP01}
Eamonn~J. Keogh and Michael~J. Pazzani.
\newblock Derivative dynamic time warping.
\newblock In {\em SDM}, pages 1--11, 2001.

\bibitem[KP05]{DBLP:conf/kdd/KriegelP05}
Hans-Peter Kriegel and Martin Pfeifle.
\newblock Density-based clustering of uncertain data.
\newblock In {\em KDD}, pages 672--677, 2005.

\bibitem[KR86]{doi:10.1016/B978-0-444-87877-9.50039-X}
Leonard Kaufman and Peter~J. Rousseeuw.
\newblock Clustering large data sets.
\newblock In {\em Pattern Recognition in Practice}, pages 425--437. 1986.

\bibitem[KR87]{books/misc/KauRou87}
Leonard Kaufman and Peter~J. Rousseeuw.
\newblock {\em Statistical Data Analysis Based on the L1-Norm and Related
  Methods}, chapter Clustering by means of Medoids, pages 405--416.
\newblock North-Holland, 1987.

\bibitem[KR90a]{doi:10.1002/9780470316801.ch3}
Leonard Kaufman and Peter~J. Rousseeuw.
\newblock {\em Finding Groups in Data}, chapter Clustering Large Applications
  (Program CLARA), pages 126--163.
\newblock John Wiley \& Sons, Inc., 1990.

\bibitem[KR90b]{doi:10.1002/9780470316801.ch5}
Leonard Kaufman and Peter~J. Rousseeuw.
\newblock {\em Finding Groups in Data}, chapter Agglomerative Nesting (Program
  AGNES), pages 199--252.
\newblock John Wiley \& Sons, Inc., 1990.

\bibitem[KR90c]{doi:10.1002/9780470316801.ch2}
Leonard Kaufman and Peter~J. Rousseeuw.
\newblock {\em Finding Groups in Data}, chapter Partitioning Around Medoids
  (Program PAM), pages 68--125.
\newblock John Wiley \& Sons, Inc., 1990.

\bibitem[KR96]{doi:10.1080/15326349608807407}
Marie Kratz and Sidney~I. Resnick.
\newblock The qq-estimator and heavy tails.
\newblock {\em Communications in Statistics. Stochastic Models},
  12(4):699--724, 1996.

\bibitem[KSZ08]{DBLP:conf/kdd/KriegelSZ08}
Hans-Peter Kriegel, Matthias Schubert, and Arthur Zimek.
\newblock Angle-based outlier detection in high-dimensional data.
\newblock In {\em KDD}, pages 444--452, 2008.

\bibitem[KSZ17]{DBLP:journals/kais/KriegelSZ17}
Hans{-}Peter Kriegel, Erich Schubert, and Arthur Zimek.
\newblock The (black) art of runtime evaluation: Are we comparing algorithms or
  implementations?
\newblock {\em Knowl. Inf. Syst.}, 52(2):341--378, 2017.

\bibitem[Kul59]{books/dover/Kullback59}
Solomon Kullback.
\newblock {\em Information Theory and Statistics}.
\newblock Dover, 1959.

\bibitem[LCK03a]{DBLP:conf/icdm/LuCK03}
Chang-Tien Lu, Dechang Chen, and Yufeng Kou.
\newblock Algorithms for spatial outlier detection.
\newblock In {\em ICDM}, pages 597--600, 2003.

\bibitem[LCK03b]{DBLP:conf/ictai/LuCK03}
Chang-Tien Lu, Dechang Chen, and Yufeng Kou.
\newblock Detecting spatial outliers with multiple attributes.
\newblock In {\em ICTAI}, pages 122--128, 2003.

\bibitem[LEL97]{DBLP:conf/icde/LeuteneggerEL97}
Scott~T. Leutenegger, J.~M. Edgington, and Mario~A. López.
\newblock Str: A simple and efficient algorithm for r-tree packing.
\newblock In {\em ICDE}, pages 497--506, 1997.

\bibitem[Lem16]{blog/Lemire16}
D.~Lemire.
\newblock Fast random shuffling.
\newblock Daniel Lemire's blog, 2016.

\bibitem[Lev66]{journals/misc/Levenshtein66}
V.~I. Levenshtein.
\newblock Binary codes capable of correcting deletions, insertions and
  reversals.
\newblock {\em Soviet physics doklady}, 10, 1966.

\bibitem[Lin91]{DBLP:journals/tit/Lin91}
Jianhua Lin.
\newblock Divergence measures based on the shannon entropy.
\newblock {\em IEEE Trans. Information Theory}, 37(1):145--151, 1991.

\bibitem[LK05]{DBLP:conf/kdd/LazarevicK05}
Aleksandar Lazarevic and Vipin Kumar.
\newblock Feature bagging for outlier detection.
\newblock In {\em KDD}, pages 157--166, 2005.

\bibitem[LKC07]{DBLP:conf/icdm/LeeKC07}
Sau~Dan Lee, Ben Kao, and Reynold Cheng.
\newblock Reducing uk-means to k-means.
\newblock In {\em ICDM Workshops}, pages 483--488, 2007.

\bibitem[LLC10]{DBLP:conf/gis/LiuLC10}
Xutong Liu, Chang-Tien Lu, and Feng Chen.
\newblock Spatial outlier detection: random walk based approaches.
\newblock In {\em GIS}, pages 370--379, 2010.

\bibitem[Llo82]{DBLP:journals/tit/Lloyd82}
Stuart~P. Lloyd.
\newblock Least squares quantization in pcm.
\newblock {\em IEEE Trans. Information Theory}, 28(2):129--136, 1982.

\bibitem[LLP07]{DBLP:conf/mldm/LateckiLP07}
Longin~Jan Latecki, Aleksandar Lazarevic, and Dragoljub Pokrajac.
\newblock Outlier detection with kernel density functions.
\newblock In {\em MLDM}, pages 61--75, 2007.

\bibitem[Loa00]{web/Loader00}
C.~Loader.
\newblock Fast and accurate computation of binomial probabilities, 2000.

\bibitem[LW66]{doi:10.1093/comjnl/9.1.60}
G.~N. Lance and W.~T. Williams.
\newblock Computer programs for hierarchical polythetic classification
  ("similarity analyses").
\newblock {\em The Computer Journal}, 9(1):60--64, 1966.

\bibitem[LW67]{doi:10.1093/comjnl/9.4.373}
G.~N. Lance and W.~T. Williams.
\newblock A general theory of classificatory sorting strategies: 1.
  hierarchical systems.
\newblock {\em The Computer Journal}, 9(4):373--380, 1967.

\bibitem[Mac67]{conf/bsmsp/MacQueen67}
J.~MacQueen.
\newblock Some methods for classification and analysis of multivariate
  observations.
\newblock In {\em 5th Berkeley Symp. Math. Statist. Prob.}, 1967.

\bibitem[Mah36]{journals/misc/Mahalanobis36}
P.~C. Mahalanobis.
\newblock On the generalized distance in statistics.
\newblock {\em Proceedings of the National Institute of Sciences of India},
  2(1), 1936.

\bibitem[Mar04]{doi:10.18637/jss.v011.i04}
George Marsaglia.
\newblock Evaluating the normal distribution.
\newblock {\em Journal of Statistical Software}, 11(4), 2004.

\bibitem[MASS08]{DBLP:conf/icde/MullerASS08}
Emmanuel Müller, Ira Assent, Uwe Steinhausen, and Thomas Seidl.
\newblock Outrank: ranking outliers in high dimensional data.
\newblock In {\em ICDE Workshops}, pages 600--603, 2008.

\bibitem[McR71]{journals/misc/McRae71}
D.~J. McRae.
\newblock Mikca: A fortran iv iterative k-means cluster analysis program.
\newblock {\em Behavioral Science}, 16(4), 1971.

\bibitem[Mei02]{tr/washington/Meila02}
Marina Meilă.
\newblock Comparing clusterings.
\newblock Technical Report 418, University of Washington, Seattle, 2002.

\bibitem[Mei03]{DBLP:conf/colt/Meila03}
Marina Meilă.
\newblock Comparing clusterings by the variation of information.
\newblock In {\em COLT}, pages 173--187, 2003.

\bibitem[Mir96]{doi:10.1007/978-1-4613-0457-9}
Boris Mirkin.
\newblock {\em Mathematical Classification and Clustering}.
\newblock Nonconvex Optimization and Its Applications. Springer US, 1996.

\bibitem[MJC{\etalchar{+}}14]{DBLP:conf/sdm/MoulaviJCZS14}
Davoud Moulavi, Pablo~A. Jaskowiak, Ricardo J. G.~B. Campello, Arthur Zimek,
  and Jörg Sander.
\newblock Density-based clustering validation.
\newblock In {\em SDM}, pages 839--847, 2014.

\bibitem[MM08]{DBLP:conf/IEEEcit/MahranM08}
Shaaban Mahran and Khaled Mahar.
\newblock Using grid for accelerating density-based clustering.
\newblock In {\em CIT}, pages 35--40, 2008.

\bibitem[MMG13]{DBLP:conf/ibpria/MomtazMG13}
Rana Momtaz, Nesma Mohssen, and Mohammad~A. Gowayyed.
\newblock Dwof: A robust density-based outlier detection approach.
\newblock In {\em IbPRIA}, pages 517--525, 2013.

\bibitem[MMW79]{doi:10.1029/WR015i005p01055}
J.~{Maciunas Landwehr}, N.~C. Matalas, and J.~R. Wallis.
\newblock Probability weighted moments compared with some traditional
  techniques in estimating gumbel parameters and quantiles.
\newblock {\em Water Resources Research}, 15(5):1055--1064, 1979.

\bibitem[MN88]{doi:10.1016/0167-71528890050-8}
J.S. Marron and D.~Nolan.
\newblock Canonical kernels for density estimation.
\newblock {\em Statistics \& Probability Letters}, 7(3):195--199, 1988.

\bibitem[MNU00]{DBLP:conf/kdd/McCallumNU00}
Andrew McCallum, Kamal Nigam, and Lyle~H. Ungar.
\newblock Efficient clustering of high-dimensional data sets with application
  to reference matching.
\newblock In {\em KDD}, pages 169--178, 2000.

\bibitem[MS65]{journals/misc/MacnaughtonSmith65}
P.~Macnaughton-Smith.
\newblock Some statistical and other numerical techniques for classifying
  individuals.
\newblock Technical Report Home Office Res. Rpt. No. 6, H.M.S.O., London, 1965.

\bibitem[MSE06]{DBLP:conf/icdm/MoiseSE06}
Gabriela Moise, Jörg Sander, and Martin Ester.
\newblock P3c: A robust projected clustering algorithm.
\newblock In {\em ICDM}, pages 414--425, 2006.

\bibitem[MSS10]{DBLP:conf/cikm/MullerSS10}
Emmanuel Müller, Matthias Schiffer, and Thomas Seidl.
\newblock Adaptive outlierness for subspace outlier ranking.
\newblock In {\em CIKM}, pages 1629--1632, 2010.

\bibitem[Mur83]{DBLP:journals/cj/Murtagh83}
Fionn Murtagh.
\newblock A survey of recent advances in hierarchical clustering algorithms.
\newblock {\em Comput. J.}, 26(4):354--359, 1983.

\bibitem[Mü11]{DBLP:journals/corr/abs-1109-2378}
Daniel Müllner.
\newblock Modern hierarchical, agglomerative clustering algorithms.
\newblock {\em CoRR}, abs/1109.2378, 2011.

\bibitem[NAG10]{DBLP:conf/dasfaa/VuAG10}
Hoang~Vu Nguyen, Hock~Hee Ang, and Vivekanand Gopalkrishnan.
\newblock Mining outliers with ensemble of heterogeneous detectors on random
  subspaces.
\newblock In {\em DASFAA (1)}, pages 368--383, 2010.

\bibitem[NEB09]{DBLP:conf/icml/NguyenEB09}
Xuan~Vinh Nguyen, Julien Epps, and James Bailey.
\newblock Information theoretic measures for clusterings comparison: is a
  correction for chance necessary?
\newblock In {\em ICML}, pages 1073--1080, 2009.

\bibitem[Nee66]{doi:10.1145/365719.365958}
Peter~M. Neely.
\newblock Comparison of several algorithms for computation of means, standard
  deviations and correlation coefficients.
\newblock {\em Communications of the ACM}, 9(7):496--499, 1966.

\bibitem[NF16]{DBLP:conf/icml/NewlingF16}
James Newling and François Fleuret.
\newblock Fast k-means with accurate bounds.
\newblock In {\em ICML}, pages 936--944, 2016.

\bibitem[NH02]{DBLP:journals/tkde/NgH02}
Raymond~T. Ng and Jiawei Han.
\newblock Clarans: A method for clustering objects for spatial data mining.
\newblock {\em IEEE Trans. Knowl. Data Eng.}, 14(5):1003--1016, 2002.

\bibitem[Nor84]{doi:10.2307/2683252}
Robert~M. Norton.
\newblock The double exponential distribution: Using calculus to find a maximum
  likelihood estimator.
\newblock {\em The American Statistician}, 38(2):135, 1984.

\bibitem[Oli06]{preprints/Olive06}
D.~J. Olive.
\newblock Robust estimators for transformed location scale families.
\newblock Online, preprint, 2006.

\bibitem[Oli08]{books/Olive08}
D.~J. Olive.
\newblock {\em Applied Robust Statistics}.
\newblock 2008.

\bibitem[ON10]{doi:10.21500/20112084.846}
Jake Olivier and Melissa~M. Norberg.
\newblock Positively skewed data: revisiting the box-cox power transformation.
\newblock {\em International Journal of Psychological Research}, 3(1):68, 2010.

\bibitem[Oou96]{web/Ooura96}
T.~Ooura.
\newblock Gamma / error functions, 1996.

\bibitem[ORSS06]{DBLP:conf/focs/OstrovskyRSS062}
Rafail Ostrovsky, Yuval Rabani, Leonard Schulman, and Chaitanya Swamy.
\newblock The effectiveness of {Lloyd}-type methods for the k-means problem.
\newblock In {\em 2006 47th Annual IEEE Symposium on Foundations of Computer
  Science (FOCS'06)}, 2006.

\bibitem[ORSS12]{DBLP:journals/jacm/OstrovskyRSS12}
Rafail Ostrovsky, Yuval Rabani, Leonard~J. Schulman, and Chaitanya Swamy.
\newblock The effectiveness of lloyd-type methods for the k-means problem.
\newblock {\em J. ACM}, 59(6):28:1--28:22, 2012.

\bibitem[Pag57]{doi:10.2307/2333258}
E.~S. Page.
\newblock On problems in which a change in a parameter occurs at an unknown
  point.
\newblock {\em Biometrika}, 44(1/2):248, 1957.

\bibitem[PBM04]{DBLP:journals/pr/PakhiraBM04}
Malay~Kumar Pakhira, Sanghamitra Bandyopadhyay, and Ujjwal Maulik.
\newblock Validity index for crisp and fuzzy clusters.
\newblock {\em Pattern Recognition}, 37(3):487--501, 2004.

\bibitem[Pea90]{journals/mathann/Peano1890}
G.~Peano.
\newblock Sur une courbe, qui remplit toute une aire plane.
\newblock {\em Mathematische Annalen}, 36(1), 1890.

\bibitem[Pet76]{doi:10.1093/biomet/63.1.161}
A.~N. Pettitt.
\newblock A two-sample anderson-darling rank statistic.
\newblock {\em Biometrika}, 63(1):161--168, 1976.

\bibitem[Phi02]{DBLP:conf/alenex/Phillips02}
Steven~J. Phillips.
\newblock Acceleration of k-means and related clustering algorithms.
\newblock In {\em ALENEX}, pages 166--177, 2002.

\bibitem[Pic85]{doi:10.2307/1427090}
Dominique Picard.
\newblock Testing and estimating change-points in time series.
\newblock {\em Advances in Applied Probability}, 17(04):841--867, 1985.

\bibitem[PJ09]{DBLP:journals/eswa/ParkJ09}
Hae-Sang Park and Chi-Hyuck Jun.
\newblock A simple and fast algorithm for k-medoids clustering.
\newblock {\em Expert Syst. Appl.}, 36(2):3336--3341, 2009.

\bibitem[PJAM02]{DBLP:conf/sigmod/ProcopiucJAM02}
Cecilia~Magdalena Procopiuc, Michael Jones, Pankaj~K. Agarwal, and T.~M.
  Murali.
\newblock A monte carlo algorithm for fast projective clustering.
\newblock In {\em SIGMOD Conference}, pages 418--427, 2002.

\bibitem[PKGF03]{DBLP:conf/icde/PapadimitriouKGF03}
Spiros Papadimitriou, Hiroyuki Kitagawa, Phillip~B. Gibbons, and Christos
  Faloutsos.
\newblock Loci: Fast outlier detection using the local correlation integral.
\newblock In {\em ICDE}, pages 315--326, 2003.

\bibitem[PL02]{DBLP:conf/sigir/PantelL02}
Patrick Pantel and Dekang Lin.
\newblock Document clustering with committees.
\newblock In {\em SIGIR}, pages 199--206, 2002.

\bibitem[PM00]{DBLP:conf/icml/PellegM00}
Dan Pelleg and Andrew~W. Moore.
\newblock X-means: Extending k-means with efficient estimation of the number of
  clusters.
\newblock In {\em ICML}, pages 727--734, 2000.

\bibitem[Pod89]{doi:10.1007/978-94-009-2432-1_5}
János Podani.
\newblock New combinatorial clustering methods.
\newblock In {\em Numerical syntaxonomy}, pages 61--77. 1989.

\bibitem[PPA{\etalchar{+}}12]{DBLP:conf/sc/PatwaryPALMC12}
Md. Mostofa~Ali Patwary, Diana Palsetia, Ankit Agrawal, Wei keng Liao, Fredrik
  Manne, and Alok~N. Choudhary.
\newblock A new scalable parallel {DBSCAN} algorithm using the disjoint-set
  data structure.
\newblock In {\em SC}, page~62, 2012.

\bibitem[Pri57]{doi:10.1002/j.1538-7305.1957.tb01515.x}
R.~C. Prim.
\newblock Shortest connection networks and some generalizations.
\newblock {\em Bell System Technical Journal}, 36(6):1389--1401, 1957.

\bibitem[PRTB99]{DBLP:conf/iccv/PuzichaRTB99}
Jan Puzicha, Yossi Rubner, Carlo Tomasi, and Joachim~M. Buhmann.
\newblock Empirical evaluation of dissimilarity measures for color and texture.
\newblock In {\em ICCV}, pages 1165--1172, 1999.

\bibitem[PS91]{DBLP:books/mit/PF91/Piatetsky91}
Gregory Piatetsky-Shapiro.
\newblock Discovery, analysis, and presentation of strong rules.
\newblock In {\em Knowledge Discovery in Databases}, pages 229--248. AAAI/MIT
  Press, 1991.

\bibitem[PZG06]{DBLP:conf/icdm/PeiZG06}
Yaling Pei, Osmar~R. Zaïane, and Yong Gao.
\newblock An efficient reference-based approach to outlier detection in large
  datasets.
\newblock In {\em ICDM}, pages 478--487, 2006.

\bibitem[Pé08]{tr/sandia/Pebay08}
P.~Pébay.
\newblock Formulas for robust, one-pass parallel computation of covariances and
  arbitrary-order statistical moments.
\newblock Technical Report SAND2008-6212, Sandia National Laboratories, 2008.

\bibitem[Ran71]{doi:10.2307/2284239}
William~M. Rand.
\newblock Objective criteria for the evaluation of clustering methods.
\newblock {\em Journal of the American Statistical Association}, 66(336):846,
  1971.

\bibitem[Rao45]{journals/bcalms/Rao45}
C.~R. Rao.
\newblock Information and the accuracy attainable in the estimation of
  statistical parameters.
\newblock {\em Bulletin of the Calcutta Mathematical Society}, 37(3), 1945.

\bibitem[RL85]{doi:10.1145/971699.318900}
Nick Roussopoulos and Daniel Leifker.
\newblock Direct spatial search on pictorial databases using packed r-trees.
\newblock {\em ACM SIGMOD Record}, 14(4):17--31, 1985.

\bibitem[Roh74]{doi:10.1146/annurev.es.05.110174.000533}
F~J Rohlf.
\newblock Methods of comparing classifications.
\newblock {\em Annual Review of Ecology and Systematics}, 5(1):101--113, 1974.

\bibitem[Rou87]{doi:10.1016/0377-04278790125-7}
Peter~J. Rousseeuw.
\newblock Silhouettes: A graphical aid to the interpretation and validation of
  cluster analysis.
\newblock {\em Journal of Computational and Applied Mathematics}, 20:53--65,
  1987.

\bibitem[RRdlIRS06]{DBLP:journals/jmma/ReynoldsRIR06}
Alan~P. Reynolds, Graeme Richards, Beatriz de~la Iglesia, and Victor~J.
  Rayward-Smith.
\newblock Clustering rules: A comparison of partitioning and hierarchical
  clustering algorithms.
\newblock {\em J. Math. Model. Algorithms}, 5(4):475--504, 2006.

\bibitem[RRS00]{DBLP:conf/sigmod/RamaswamyRS00}
Sridhar Ramaswamy, Rajeev Rastogi, and Kyuseok Shim.
\newblock Efficient algorithms for mining outliers from large data sets.
\newblock In {\em SIGMOD Conference}, pages 427--438, 2000.

\bibitem[SB91]{DBLP:journals/ijcv/SwainB91}
Michael~J. Swain and Dana~H. Ballard.
\newblock Color indexing.
\newblock {\em International Journal of Computer Vision}, 7(1):11--32, 1991.

\bibitem[SC96]{DBLP:conf/mm/SmithC96}
John~R. Smith and Shih-Fu Chang.
\newblock Visualseek: A fully automated content-based image query system.
\newblock In {\em ACM Multimedia}, pages 87--98, 1996.

\bibitem[Sch78]{doi:10.1214/aos/1176344136}
Gideon Schwarz.
\newblock Estimating the dimension of a model.
\newblock {\em The Annals of Statistics}, 6(2):461--464, 1978.

\bibitem[Sco92]{doi:10.1002/9780470316849}
David~W. Scott, editor.
\newblock {\em Multivariate Density Estimation}.
\newblock Wiley Series in Probability and Statistics. John Wiley \& Sons, Inc.,
  1992.

\bibitem[Sed98]{DBLP:books/daglib/0004943}
Robert Sedgewick.
\newblock {\em Algorithms in C - parts 1-4: fundamentals, data structures,
  sorting, searching (3. ed.)}.
\newblock Addison-Wesley-Longman, 1998.

\bibitem[SEKX98]{DBLP:journals/datamine/SanderEKX98}
Jörg Sander, Martin Ester, Hans-Peter Kriegel, and Xiaowei Xu.
\newblock Density-based clustering in spatial databases: The algorithm
  {GDBSCAN} and its applications.
\newblock {\em Data Min. Knowl. Discov.}, 2(2):169--194, 1998.

\bibitem[SG91]{DBLP:books/mit/PF91/SmythG91}
Padhraic Smyth and Rodney~M. Goodman.
\newblock Rule induction using information theory.
\newblock In {\em Knowledge Discovery in Databases}, pages 159--176. 1991.

\bibitem[SG17]{DBLP:conf/sisap/SchubertG17}
Erich Schubert and Michael Gertz.
\newblock Intrinsic t-stochastic neighbor embedding for visualization and
  outlier detection - a remedy against the curse of dimensionality?
\newblock In {\em SISAP}, pages 188--203, 2017.

\bibitem[SG18a]{DBLP:conf/lwa/SchubertG18}
Erich Schubert and Michael Gertz.
\newblock Improving the cluster structure extracted from optics plots.
\newblock In {\em Proc. Lernen, Wissen, Daten, Analysen (LWDA 2018)}, pages
  318--329, 2018.

\bibitem[SG18b]{DBLP:conf/ssdbm/SchubertG18}
Erich Schubert and Michael Gertz.
\newblock Numerically stable parallel computation of (co-)variance.
\newblock In {\em SSDBM}, pages 10:1--10:12, 2018.

\bibitem[Sib73]{DBLP:journals/cj/Sibson73}
R.~Sibson.
\newblock Slink: An optimally efficient algorithm for the single-link cluster
  method.
\newblock {\em Comput. J.}, 16(1):30--34, 1973.

\bibitem[Sin84]{journals/skytelesc/Sinnott84}
R.~W. Sinnott.
\newblock Virtues of the haversine.
\newblock {\em Sky and Telescope}, 68(2), 1984.

\bibitem[Sin16]{web/Sinclair16}
D.~Sinclair.
\newblock S-hull: a fast sweep-hull routine for delaunay triangulation, 2016.

\bibitem[SKE{\etalchar{+}}15]{DBLP:journals/pvldb/SchubertKEZSZ15}
Erich Schubert, Alexander Koos, Tobias Emrich, Andreas Z{\"{u}}fle,
  Klaus~Arthur Schmid, and Arthur Zimek.
\newblock A framework for clustering uncertain data.
\newblock {\em {PVLDB}}, 8(12):1976--1979, 2015.

\bibitem[SKK00]{conf/kdd/SteinbachKK00}
M.~Steinbach, G.~Karypis, and V.~Kumar.
\newblock A comparison of document clustering techniques.
\newblock In {\em KDD workshop on text mining. Vol. 400. No. 1}, 2000.

\bibitem[SLZ03]{DBLP:journals/geoinformatica/ShekharLZ03}
Shashi Shekhar, Chang-Tien Lu, and Pusheng Zhang.
\newblock A unified approach to detecting spatial outliers.
\newblock {\em GeoInformatica}, 7(2):139--166, 2003.

\bibitem[SM02]{journals/kansas/SokalM1902}
R.~R. Sokal and C.~D. Michener.
\newblock A statistical method for evaluating systematic relationship.
\newblock {\em The University of Kansas Science Bulletin}, 38:1409–1438,
  1902.

\bibitem[SM99]{DBLP:conf/dmkdttt/SaharM99}
Sigal Sahar and Yishay Mansour.
\newblock Empirical evaluation of interest-level criteria.
\newblock In {\em Data Mining and Knowledge Discovery: Theory, Tools, and
  Technology}, pages 63--74, 1999.

\bibitem[Sne57]{doi:10.1099/00221287-17-1-201}
P.~H.~A. Sneath.
\newblock The application of computers to taxonomy.
\newblock {\em Microbiology}, 17(1):201--226, 1957.

\bibitem[SPST{\etalchar{+}}01]{DBLP:journals/neco/ScholkopfPSSW01}
Bernhard Schölkopf, John~C. Platt, John Shawe-Taylor, Alexander~J. Smola, and
  Robert~C. Williamson.
\newblock Estimating the support of a high-dimensional distribution.
\newblock {\em Neural Computation}, 13(7):1443--1471, 2001.

\bibitem[SR18]{DBLP:journals/corr/abs-1810-05691}
Erich Schubert and Peter~J. Rousseeuw.
\newblock Faster k-medoids clustering: Improving the pam, clara, and clarans
  algorithms.
\newblock {\em CoRR}, abs/1810.05691, 2018.

\bibitem[SRB07]{doi:10.1214/009053607000000505}
Gábor~J. Székely, Maria~L. Rizzo, and Nail~K. Bakirov.
\newblock Measuring and testing dependence by correlation of distances.
\newblock {\em The Annals of Statistics}, 35(6):2769--2794, 2007.

\bibitem[SS87]{doi:10.1080/01621459.1987.10478517}
F.~W. Scholz and M.~A. Stephens.
\newblock K-sample anderson–darling tests.
\newblock {\em Journal of the American Statistical Association},
  82(399):918--924, 1987.

\bibitem[SS96]{doi:10.1524/strm.1996.14.4.353}
J.~Schultze and J.~Steinebach.
\newblock On least squares estimates of an exponential tail coefficient.
\newblock {\em Statistics \& Risk Modeling}, 14(4), 1996.

\bibitem[SSE{\etalchar{+}}17]{DBLP:journals/tods/SchubertSEKX17}
Erich Schubert, Jörg Sander, Martin Ester, Hans-Peter Kriegel, and Xiaowei Xu.
\newblock {DBSCAN} revisited, revisited: Why and how you should (still) use
  {DBSCAN}.
\newblock {\em ACM Trans. Database Syst.}, 42(3):19:1--19:21, 2017.

\bibitem[SSW{\etalchar{+}}17]{DBLP:journals/corr/abs-1708-03569}
Erich Schubert, Andreas Spitz, Michael Weiler, Johanna Geiß, and Michael
  Gertz.
\newblock Semantic word clouds with background corpus normalization and
  t-distributed stochastic neighbor embedding.
\newblock {\em CoRR}, abs/1708.03569, 2017.

\bibitem[Ste74]{doi:10.1080/01621459.1974.10480196}
M.~A. Stephens.
\newblock Edf statistics for goodness of fit and some comparisons.
\newblock {\em Journal of the American Statistical Association},
  69(347):730--737, 1974.

\bibitem[SV13]{DBLP:conf/cikm/SchneiderV13}
Johannes Schneider and Michail Vlachos.
\newblock Fast parameterless density-based clustering via random projections.
\newblock In {\em CIKM}, pages 861--866, 2013.

\bibitem[SWK14]{DBLP:conf/kdd/SchubertWK14}
Erich Schubert, Michael Weiler, and Hans-Peter Kriegel.
\newblock {SigniTrend}: scalable detection of emerging topics in textual
  streams by hashed significance thresholds.
\newblock In {\em KDD}, pages 871--880, 2014.

\bibitem[SWZK12]{DBLP:conf/sdm/SchubertWZK12}
Erich Schubert, Remigius Wojdanowski, Arthur Zimek, and Hans-Peter Kriegel.
\newblock On evaluation of outlier rankings and outlier scores.
\newblock In {\em SDM}, pages 1047--1058, 2012.

\bibitem[SZK13]{DBLP:conf/ssd/SchubertZK13}
Erich Schubert, Arthur Zimek, and Hans-Peter Kriegel.
\newblock Geodetic distance queries on r-trees for indexing geographic data.
\newblock In {\em SSTD}, pages 146--164, 2013.

\bibitem[SZK14a]{DBLP:conf/sdm/SchubertZK14}
Erich Schubert, Arthur Zimek, and Hans-Peter Kriegel.
\newblock Generalized outlier detection with flexible kernel density estimates.
\newblock In {\em SDM}, pages 542--550, 2014.

\bibitem[SZK14b]{DBLP:journals/datamine/SchubertZK14}
Erich Schubert, Arthur Zimek, and Hans-Peter Kriegel.
\newblock Local outlier detection reconsidered: a generalized view on locality
  with applications to spatial, video, and network outlier detection.
\newblock {\em Data Min. Knowl. Discov.}, 28(1):190--237, 2014.

\bibitem[SZK15]{DBLP:conf/dasfaa/SchubertZK15}
Erich Schubert, Arthur Zimek, and Hans-Peter Kriegel.
\newblock Fast and scalable outlier detection with approximate nearest neighbor
  ensembles.
\newblock In {\em DASFAA (2)}, pages 19--36, 2015.

\bibitem[Sø48]{journals/misc/Sorensen48}
T.~Sørensen.
\newblock A method of establishing groups of equal amplitude in plant sociology
  based on similarity of species and its application to analyses of the
  vegetation on danish commons.
\newblock {\em Kongelige Danske Videnskabernes Selskab}, 5(4), 1948.

\bibitem[TAE{\etalchar{+}}11]{DBLP:journals/tvcg/TatuAEBTMK11}
Andrada Tatu, Georgia Albuquerque, Martin Eisemann, Peter Bak, Holger Theisel,
  Marcus~A. Magnor, and Daniel~A. Keim.
\newblock Automated analytical methods to support visual exploration of
  high-dimensional data.
\newblock {\em IEEE Trans. Vis. Comput. Graph.}, 17(5):584--597, 2011.

\bibitem[TCFC02]{DBLP:conf/pakdd/TangCFC02}
Jian Tang, Zhixiang Chen, Ada Wai-Chee Fu, and David Wai-Lok Cheung.
\newblock Enhancing effectiveness of outlier detections for low density
  patterns.
\newblock In {\em PAKDD}, pages 535--548, 2002.

\bibitem[Ter07]{web/Terriberry07}
T.~B. Terriberry.
\newblock Computing higher-order moments online, 2007.

\bibitem[TK00]{tr/umn/TanK00}
Pang-Ning Tan and Vipin Kumar.
\newblock Interestingness measures for association patterns: A perspective.
\newblock Technical Report 00-036, University of Minnesota, 2000.

\bibitem[TKS04]{DBLP:journals/is/TanKS04}
Pang-Ning Tan, Vipin Kumar, and Jaideep Srivastava.
\newblock Selecting the right objective measure for association analysis.
\newblock {\em Inf. Syst.}, 29(4):293--313, 2004.

\bibitem[Top00]{DBLP:journals/tit/Topsoe00}
Flemming Topsøe.
\newblock Some inequalities for information divergence and related measures of
  discrimination.
\newblock {\em IEEE Trans. Information Theory}, 46(4):1602--1609, 2000.

\bibitem[Vas69]{journals/misc/Vaserstein69}
L.N. Vaserstein.
\newblock Markov processes over denumerable products of spaces describing large
  systems of automata.
\newblock {\em Problemy Peredachi Informatsii / Problems of Information
  Transmission}, 5(3), 1969.

\bibitem[vBHZ15]{tr/nii/BrunkenHZ15}
Jonathan von Brünken, Michael~E. Houle, and Arthur Zimek.
\newblock Intrinsic dimensional outlier detection in high-dimensional data.
\newblock Technical report, National Institute of Informatics, 2015.

\bibitem[vdM14]{DBLP:journals/jmlr/Maaten14}
Laurens van~der Maaten.
\newblock Accelerating {t-SNE} using tree-based algorithms.
\newblock {\em Journal of Machine Learning Research}, 15(1):3221--3245, 2014.

\bibitem[vdMH08]{journals/jmlr/MaatenH08}
Laurens J.~P. van~der Maaten and Geoffrey~E. Hinton.
\newblock Visualizing high-dimensional data using {t-SNE}.
\newblock {\em Journal of Machine Learning Research}, 9:2579--2605, 2008.

\bibitem[VHGK03]{DBLP:conf/kdd/VlachosHGK03}
Michail Vlachos, Marios Hadjieleftheriou, Dimitrios Gunopulos, and Eamonn~J.
  Keogh.
\newblock Indexing multi-dimensional time-series with support for multiple
  distance measures.
\newblock In {\em KDD}, pages 216--225, 2003.

\bibitem[Vig14]{web/Vigna14}
S.~Vigna.
\newblock An experimental exploration of marsaglia's xorshift generators,
  scrambled, 2014.

\bibitem[Vin75]{doi:10.1179/sre.1975.23.176.88}
T.~Vincenty.
\newblock Direct and inverse solutions of geodescis on the ellipsoid with
  application of nested equations.
\newblock {\em Survey Review}, 23(176):88--93, 1975.

\bibitem[vR79]{DBLP:books/bu/Rijsbergen79}
C.~J. van Rijsbergen.
\newblock {\em Information Retrieval}.
\newblock Butterworth, 1979.

\bibitem[War63]{doi:10.1080/01621459.1963.10500845}
Joe~H. Ward.
\newblock Hierarchical grouping to optimize an objective function.
\newblock {\em Journal of the American Statistical Association},
  58(301):236--244, 1963.

\bibitem[WB97]{tr/ethz/WeberS97}
R.~Weber and S.~Blott.
\newblock An approximation based data structure for similarity search.
\newblock Technical Report TR1997b, ETH Zentrum, Zürich, 1997.

\bibitem[Wel62]{doi:10.2307/1266577}
B.~P. Welford.
\newblock Note on a method for calculating corrected sums of squares and
  products.
\newblock {\em Technometrics}, 4(3):419, 1962.

\bibitem[Wes79]{DBLP:journals/cacm/West79}
D.~H.~D. West.
\newblock Updating mean and variance estimates: An improved method.
\newblock {\em Commun. ACM}, 22(9):532--535, 1979.

\bibitem[WH00]{doi:10.1016/S0895-71770000178-3}
X.~Wang and F.J. Hickernell.
\newblock Randomized halton sequences.
\newblock {\em Mathematical and Computer Modelling}, 32(7-8):887--899, 2000.

\bibitem[Wil11]{web/Williams11}
E.~Williams.
\newblock Aviation formulary, 2011.

\bibitem[Wis69]{doi:10.2307/2528688}
David Wishart.
\newblock 256. note: An algorithm for hierarchical classifications.
\newblock {\em Biometrics}, 25(1):165, 1969.

\bibitem[Yar09]{web/Yaroslavskiy09}
V.~Yaroslavskiy.
\newblock Dual-pivot quicksort, 2009.

\bibitem[YC71]{doi:10.1080/00401706.1971.10488826}
Edward~A. Youngs and Elliot~M. Cramer.
\newblock Some results relevant to choice of sum and sum-of-product algorithms.
\newblock {\em Technometrics}, 13(3):657--665, 1971.

\bibitem[YOTJ01]{DBLP:conf/vldb/OoiYTJ01}
Cui Yu, Beng~Chin Ooi, Kian-Lee Tan, and H.~V. Jagadish.
\newblock Indexing the distance: An efficient method to knn processing.
\newblock In {\em VLDB}, pages 421--430, 2001.

\bibitem[YSW09]{doi:10.1109/ITCS.2009.230}
Bo~Yu, Mingqiu Song, and Leilei Wang.
\newblock Local isolation coefficient-based outlier mining algorithm.
\newblock In {\em 2009 International Conference on Information Technology and
  Computer Science}, 2009.

\bibitem[ZCS14]{DBLP:conf/ssdbm/ZimekCS14}
Arthur Zimek, Ricardo J. G.~B. Campello, and Jörg Sander.
\newblock Data perturbation for outlier detection ensembles.
\newblock In {\em SSDBM}, pages 13:1--13:12, 2014.

\bibitem[ZES{\etalchar{+}}14]{DBLP:conf/kdd/ZufleESMZR14}
Andreas Züfle, Tobias Emrich, Klaus~Arthur Schmid, Nikos Mamoulis, Arthur
  Zimek, and Matthias Renz.
\newblock Representative clustering of uncertain data.
\newblock In {\em KDD}, pages 243--252, 2014.

\bibitem[Zha96]{tr/wisc/Zhang97}
T.~Zhang.
\newblock Data clustering for very large datasets plus applications.
\newblock Technical Report 1355, University of Wisconsin Madison, 1996.

\bibitem[ZHJ09]{DBLP:conf/pakdd/ZhangHJ09}
Ke~Zhang, Marcus Hutter, and Huidong Jin.
\newblock A new local distance-based outlier detection approach for scattered
  real-world data.
\newblock In {\em PAKDD}, pages 813--822, 2009.

\bibitem[Zim08]{phd/dnb/Zimek08/Ch18}
Arthur Zimek.
\newblock {\em Correlation Clustering}.
\newblock PhD thesis, Ludwig-Maximilians-Universität München, 2008.
\newblock Application 2: Outlier Detection (Chapter 18).

\bibitem[ZJ14]{DBLP:books/cu/ZM2014}
Mohammed~J. Zaki and Wagner~Meira Jr.
\newblock {\em Data Mining and Analysis: Fundamental Concepts and Algorithms}.
\newblock Cambridge University Press, 2014.

\bibitem[ZK01]{tr/umn/ZhaoK01}
Y.~Zhao and G.~Karypis.
\newblock Criterion functions for document clustering: Experiments and
  analysis.
\newblock Technical Report CS 01-040, University of Minnesota, 2001.

\bibitem[ZPOL97]{DBLP:conf/kdd/ZakiPOL97}
Mohammed~Javeed Zaki, Srinivasan Parthasarathy, Mitsunori Ogihara, and Wei Li.
\newblock New algorithms for fast discovery of association rules.
\newblock In {\em KDD}, pages 283--286, 1997.

\bibitem[ZRL96]{DBLP:conf/sigmod/ZhangRL96}
Tian Zhang, Raghu Ramakrishnan, and Miron Livny.
\newblock Birch: An efficient data clustering method for very large databases.
\newblock In {\em SIGMOD Conference}, pages 103--114, 1996.

\bibitem[ZRL97]{DBLP:journals/datamine/ZhangRL97}
Tian Zhang, Raghu Ramakrishnan, and Miron Livny.
\newblock Birch: A new data clustering algorithm and its applications.
\newblock {\em Data Min. Knowl. Discov.}, 1(2):141--182, 1997.

\bibitem[ZXF08]{DBLP:conf/ictai/ZhaoXF08}
Qinpei Zhao, Mantao Xu, and Pasi Fränti.
\newblock Knee point detection on bayesian information criterion.
\newblock In {\em ICTAI (2)}, pages 431--438, 2008.

\end{thebibliography}

\end{document}